\documentclass[pdflatex,sn-mathphys-num]{sn-jnl}

\usepackage{graphicx}%
\usepackage{multirow}%
\usepackage{amsmath,amssymb,amsfonts}%
\usepackage{amsthm}%
\usepackage{mathrsfs}%
\usepackage[title]{appendix}%
\usepackage{xcolor}%
\usepackage{textcomp}%
\usepackage{manyfoot}%
\usepackage{booktabs}%
\usepackage{algorithm}%
\usepackage{algorithmicx}%
\usepackage{algpseudocode}%
\usepackage{listings}%
\usepackage{makecell}

\usepackage{bm}
\usepackage{color}
\usepackage{xcolor}
\usepackage{caption}
\usepackage{subfigure}
\usepackage[table]{xcolor}
\usepackage{tcolorbox}
\definecolor{lightblue}{HTML}{F0F8FF}
\definecolor{top1}{rgb}{0.68, 0.85, 0.9}
\definecolor{top2}{rgb}{0.78, 0.93, 0.96}
\definecolor{top3}{rgb}{0.88, 0.97, 0.99}

\theoremstyle{thmstyleone}

\newcommand{\re}[1]{{\color{black}#1}}
\newcommand{\rre}[1]{{\color{black}#1}}

\theoremstyle{thmstyletwo}

\theoremstyle{thmstylethree}

\begin{document}

\title[Article Title]{\rre{A Conformation-Centric Generative Foundation Model for Linear Polymer Modeling and Design}}

\author[1,2]{\fnm{Fanmeng} \sur{Wang}}\email{fanmengwang@ruc.edu.cn}
\author[3]{\fnm{\rre{Ruochao}} \sur{\rre{Wang}}}\email{wangruochao.ripp@sinopec.com}
\author[3]{\fnm{Shan} \sur{Mei}}\email{meishan.ripp@sinopec.com}
\author[2]{\fnm{Wentao} \sur{Guo}}\email{guowt1997@gmail.com}
\author[2]{\fnm{Hongshuai} \sur{Wang}}\email{wanghongshuai@dp.tech}
\author*[3]{\fnm{Qi} \sur{Ou}}\email{ouqi.ripp@sinopec.com}
\author*[2]{\fnm{Zhifeng} \sur{Gao}}\email{gaozf@dp.tech}
\author*[1]{\fnm{Hongteng} \sur{Xu}}\email{hongtengxu@ruc.edu.cn}

\affil[1]{\orgdiv{Gaoling School of Artificial Intelligence}, \orgname{Renmin University of China}, \orgaddress{\city{Beijing}, \country{China}}}
\affil[2]{\orgname{DP Technology}, \orgaddress{\city{Beijing}, \country{China}}}
\affil[3]{\orgname{SINOPEC Research Institute of Petroleum Processing Co., Ltd.}, \orgaddress{\city{Beijing}, \country{China}}}

\abstract{
\rre{Linear polymers, macromolecules formed from monomers covalently bonded into continuous chains}, underpin countless technologies and are indispensable to modern life.
While deep learning is advancing polymer science, existing methods typically represent the whole linear polymer solely through monomer-level descriptors, overlooking the global structural information inherent in polymer conformations, which ultimately limits their practical performance. 
Moreover, this important field still lacks a dedicated foundation model that can effectively support diverse downstream tasks, thereby severely constraining progress. 
To address these challenges, we introduce PolyConFM, a foundation model tailored for modeling and designing linear polymers through conformation-centric generative pretraining.
Recognizing that each linear polymer is essentially a continuous chain whose conformation can be naturally decomposed into a sequence of local conformations (i.e., those of its repeating units), we pretrain PolyConFM under the conditional generation paradigm, reconstructing these local conformations via masked autoregressive (MAR) modeling and further generating their orientation transformations to recover the corresponding polymer conformation.
Meanwhile, we construct a linear polymer conformation dataset via molecular dynamics simulations to mitigate data sparsity, thereby enabling conformation-centric pretraining.
Experiments demonstrate that PolyConFM consistently outperforms representative task-specific methods across diverse downstream tasks, thereby equipping polymer science with a powerful tool targeting linear polymers.
}


\maketitle

\section{Introduction}\label{sec: introduction}
In recent decades, linear polymers have become the cornerstone of modern life, underpinning innovations as diverse as lightweight structural materials~\cite{naskar2016polymer}, flexible electronics~\cite{kang2015polymer}, energy storage~\cite{chen2023ladderphane}, catalysis~\cite{taylor2015catalysts}, and biomedicine~\cite{chen2019facile}. 
\rre{As macromolecules formed through the covalent bonding of numerous monomers into continuous chains}, they embody the art of molecular condensation that transforms simple building blocks into functional materials, offering exceptional tunability while complicating experimentation~\cite{audus2017polymer}.
Traditionally, researchers depend on wet-lab experiments and computational methods (e.g., molecular dynamics simulations and polymer informatics tools), complemented by analytical characterization, to investigate these linear systems.
However, these approaches are expensive, time-consuming, and dependent on substantial domain expertise, thereby struggling to meet the rapidly increasing demands~\cite{chen2021polymer, zhang2023transferring, zhang2024multi}. 
In this context, driven by the remarkable success of artificial intelligence across scientific fields~\cite{wang2023scientific, xia2023systematic, van2023ai, han2025survey}, learning-based methods are emerging as a promising avenue for advancing polymer science~\cite{mcdonald2023applied, dobrynin2023forensics, liu2025harnessing}.

Among these methods, polymer pretraining methods have stood out by learning inherent patterns from large-scale unlabeled data to enhance downstream performance while reducing reliance on labeled data~\cite{tran2024design}.
In particular, existing polymer pretraining methods typically leverage various monomer-level descriptors to represent the whole linear polymer~\cite{ge2025machine} and directly borrow small‑molecule pretraining frameworks~\cite{ross2022large, wang2022molecular, zhu2022unified, li2023knowledge} to the polymer field.
For example, some sequence-based methods~\cite{kuenneth2023polybert, xu2023transpolymer, qiu2024polync} directly pretrain language models on millions of polymer SMILES strings~\footnote{The polymer SMILES (P-SMILES) string is a modified SMILES representation, formed through combining the corresponding monomer’s SMILES string with two ``*'' symbols indicating polymerization sites.} using masked prediction or autoregression.
Those methods in~\cite{gao2024self, han2024multimodal, huang2025unified} further incorporate 2D topological information of monomers through contrastive pretraining. 
However, representing polymers with monomer‑level descriptors is fundamentally inappropriate, as these descriptors fail to capture global structural features inherent in polymer conformations, including chain length, tacticity, and long‑range intrachain interactions, which are essential for accurate polymer modeling~\cite{wang2024mmpolymer}.
For instance, atactic and isotactic polypropylene, though derived from the same monomer, differ in chain length and tacticity, and exhibit markedly different glass transition temperatures. 
Monomer‑level descriptors are entirely unable to distinguish these distinctions~\cite{gitsas2008pressure} and thus limit model performance.
\re{Meanwhile, whereas small-molecule pretraining has shifted toward foundation models~\cite{qiang2023bridging, feng2025unigem} that bridge the gap between representation learning and generation within a unified framework to support various downstream tasks}, existing polymer pretraining methods~\cite{kuenneth2023polybert, xu2023transpolymer, qiu2024polync, gao2024self, han2024multimodal, huang2025unified, wang2024mmpolymer} remain focused solely on representation learning for downstream property prediction, providing limited support for various generative tasks (e.g., polymer design), thereby constraining progress in this field.

\begin{figure}[!htbp]
    \centering
    \includegraphics[width=0.95\linewidth]{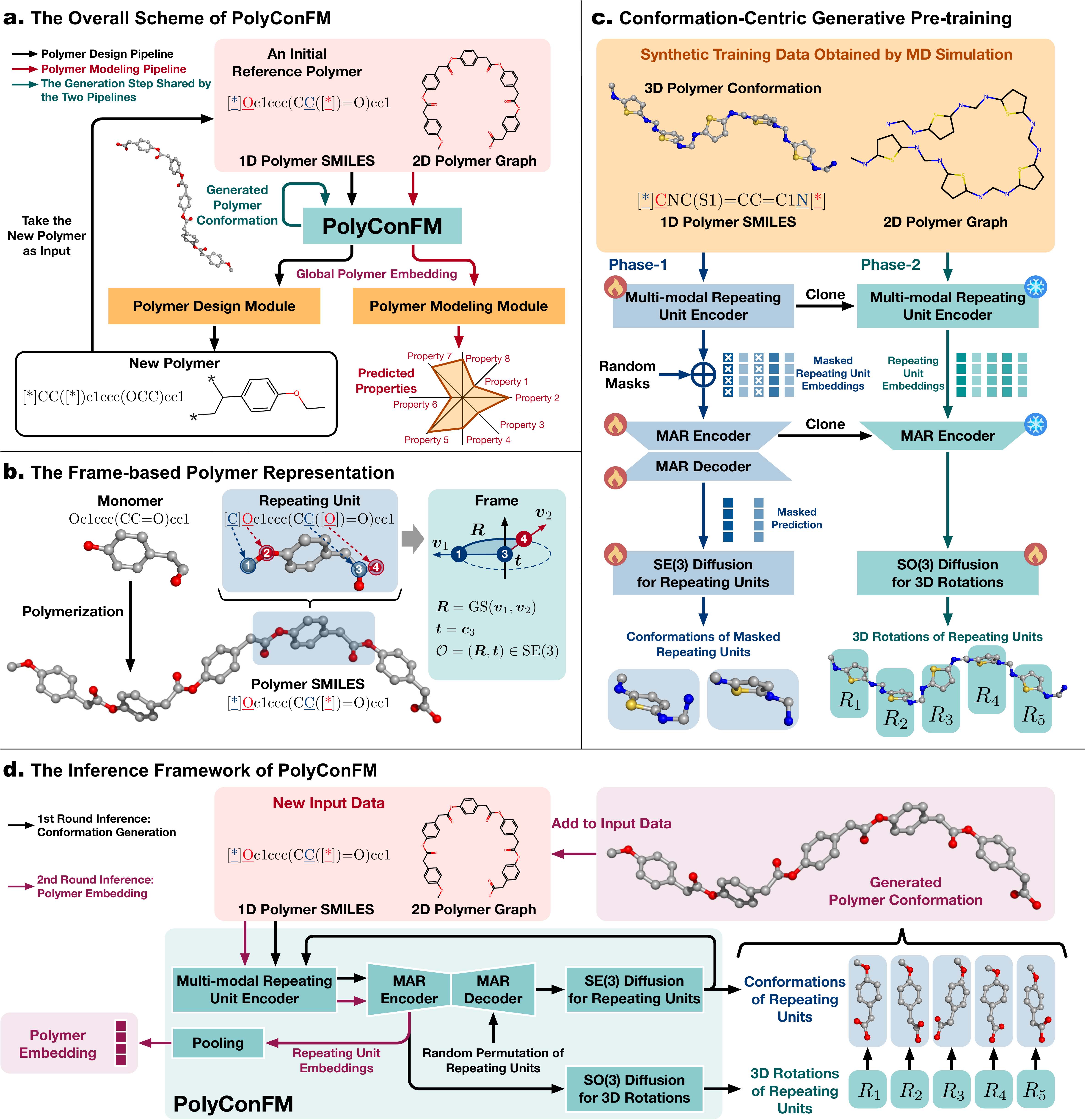}
    \caption{\textbf{Overview of the proposed PolyConFM.}
    \textbf{a.}~The overall scheme of PolyConFM: PolyConFM employs polymer conformation generated by itself as input to provide global structural information for downstream tasks, while the modeling module can also assist the design module via virtual screening to prioritize candidates, thereby positioning PolyConFM as a powerful backbone that seamlessly bridges polymer structure, property, and design.
    \textbf{b.}~The frame-based polymer representation: Each linear polymer is essentially a continuous chain whose conformation can be decomposed into a sequence of repeating-unit conformations with identical SMILES strings and distinct 3D structures, overlapping at those key atoms (e.g., atom-1 of the current repeating unit aligns with atom-3 of the preceding repeating unit). 
    Here, the orientation transformation corresponding to each frame is denoted as $\mathcal{O} = (\bm R, \bm t)$, where rotation transformation $\bm R \in \mathbb{R}^{3 \times 3}$ is calculated through the Gram-Schmidt (GS) procedure~\cite{leon2013gram} on vectors $\bm v_1$ and $\bm v_2$ (i.e., $\bm R$ = GS($\bm v_1$, $\bm v_2$)), and translation transformation $\bm t \in \mathbb{R}^{3}$ is the 3D coordinate of atom-3 \re{(i.e., $\bm t$ = $\bm{c}_3$)}.
    \textbf{c.}~Conformation-centric generative pretraining: PolyConFM is first trained to reconstruct repeating-unit conformations via masked autoregressive (MAR) modeling and then trained to generate the required orientation transformations for assembling them to recover the corresponding polymer conformation, thereby enabling it to capture complex dependencies among repeating units for global structure modeling and simultaneously unlock conformation-generation capabilities for downstream tasks.
    \re{Notably, the teacher-forcing strategy is adopted during pretraining, which directly utilizes ground-truth repeating-unit conformations (rather than the generation from Phase-1) obtained by MD simulation to condition Phase-2, thereby preventing error propagation.}
    \textbf{d.}~The inference framework of PolyConFM: The pretrained PolyConFM generates polymer conformation based on the corresponding 1D and 2D representations, and then derives polymer embedding for downstream tasks by passing the generated conformation back through itself.
    Notably, for the sake of visualization simplicity, only a small fragment of the complete polymer conformation is presented in this figure.
    }
    \label{fig: overview}
\end{figure}

Inspired by recent progress in small-molecule pretraining~\cite{liu2022pretraining, wang2023automated, zhou2023unimol, wang2025token, ni2024pre, qiao2025self}, which firmly establishes the significant value of incorporating molecular conformations (i.e., the stable 3D structures), it is imperative to develop polymer conformation–centric pretraining methods to ensure polymer modeling remains faithful to the underlying chemical and physical principles.
However, polymers are intrinsically more complex than small molecules (e.g., much higher molecular weights and far greater structural flexibility), and a pronounced scarcity of high-quality pretraining data — especially for critical structural data such as polymer conformations~\cite{martin2023emerging}, meaning that simply transplanting small‑molecule paradigms is no longer tenable.
In this context, designing the generative pretraining paradigm around polymer conformation becomes a natural and effective choice.
In particular, polymer conformations directly reflect global structural features, making structure–property relationships explicit and yielding more informative representations for polymer modeling~\cite{wang2023mperformer, janson2023direct, cen2024high, li2025size}.
Meanwhile, generative pretraining captures the underlying data distribution, aligning representation with structurally informed generation and downstream design, which has already demonstrated strong advantages in other scientific domains~\cite{ferruz2022protgpt2, feng2023may, cui2024scgpt, chen2024towards, hao2024large, xiang2024molecular, chen2025xtrimopglm}.

Therefore, we introduce PolyConFM, a foundation model tailored for modeling and designing linear polymers, which effectively overcomes the above challenges through conformation-centric generative pretraining.
In particular, considering that each linear polymer is essentially a continuous chain whose conformation can be naturally decomposed into a sequence of local conformations (i.e., the corresponding conformation of each repeating unit within the linear polymer), we employ these local conformations as token-like structural units for model input.
During pretraining, as illustrated in Figure~\ref{fig: overview}c, we first train PolyConFM to reconstruct these local conformations via masked autoregressive (MAR) modeling and then train it to generate the required orientation transformations~\footnote{As shown in Figure~\ref{fig: overview}b, the bonding atoms between adjacent repeating units naturally overlap (e.g., atom-1 of the current repeating unit aligns with atom-3 of the preceding repeating unit). Therefore, we only need to generate rotational transformations, as translation transformations can be directly derived from the 3D coordinates of those overlapping atoms when assembling.} for assembling them, thereby recovering the corresponding polymer conformation.
This pretraining strategy enables PolyConFM to capture complex dependencies among repeating units for global structure modeling and simultaneously unlock conformation-generation capabilities for diverse downstream tasks, leading to a systematic framework that bridges the gap between representation learning and generation.
Meanwhile, to mitigate the severe scarcity of conformation datasets, we devote considerable time and resources to constructing a dataset of over 50,000 linear polymers with conformations through molecular dynamics simulations.
This dataset not only enables our conformation-centric generative pretraining but also provides strong momentum for future research in this important field.

To comprehensively evaluate PolyConFM, we conduct extensive experiments across diverse tasks and settings, demonstrating its superior performance over task-specific baselines.
In particular, owing to conformation-centric generative pretraining, PolyConFM unlocks the capability to generate polymer conformations that are subsequently used as inputs for downstream tasks, thereby providing crucial global structural information.
On this basis, PolyConFM achieves state-of-the-art performance on the downstream polymer property prediction task by deriving structure-aware polymer embeddings from its self-generated conformations, highlighting its accurate structure–property relationship modeling capability.
Moreover, equipped with these two capabilities, PolyConFM operates with the clear design objective and reliable search guidance, significantly outperforming various baselines on the downstream polymer design task.
Taken together, these promising results establish PolyConFM as an effective and powerful foundation model for linear polymers, seamlessly bridging structure, property, and design.

\section{Results}\label{sec2}
\subsection{PolyConFM Framework}
We introduce PolyConFM, a foundation model tailored for the modeling and design of linear polymers via conformation-centric generative pretraining, thereby capturing global structural features and supporting diverse downstream tasks.
The complete framework, comprising the model architectures and learning paradigms, has been illustrated in Figure~\ref{fig: overview}. 

On the whole, as shown in Figure~\ref{fig: overview}a, PolyConFM employs the polymer conformation generated by itself as input to provide global structural information for downstream tasks, while the polymer modeling module can also assist the polymer design module via virtual screening to prioritize candidates, thereby positioning it as a powerful backbone that seamlessly bridges structure, property, and design. 
In particular, as shown in Figure~\ref{fig: overview}b, under the frame-based polymer representation, each polymer conformation can be specified by a set of repeating-unit conformations and their orientation transformations, thereby enabling the model to accommodate the vast chemical space of linear polymers. 
Further details on the frame-based polymer representation are provided in Section~\ref{sec: rep}.

For conformation-centric generative pretraining, as shown in Figure~\ref{fig: overview}c, we pretrain PolyConFM under the conditional generation paradigm. 
Here, it first learns to generate repeating-unit conformations via masked autoregressive modeling and then learns their orientation transformations to assemble them into the corresponding polymer conformation, thereby capturing inter-unit dependencies for global structure modeling while simultaneously unlocking conformation generation capability for downstream tasks.
Please note that since adjacent repeating-unit conformations are naturally overlapping at those key atoms, only their rotational transformations are required, as corresponding translation transformations can be directly derived from the 3D coordinates of overlapping atoms when assembling.
Further details on conformation-centric generative pretraining are provided in Section~\ref{sec: pretraining}.

For finetuning on downstream tasks, as shown in Figure~\ref{fig: overview}d, we first run inference with the pretrained PolyConFM to generate repeating-unit conformations and their rotation transformations, followed by assembling them into the complete polymer conformation, and add this generated polymer conformation to the input to derive the corresponding global polymer embedding.
Furthermore, as shown in Figure~\ref{fig: overview}a, we employ a simple multi-layer perceptron (MLP) layer as the polymer modeling module, which takes the global polymer embedding as input for downstream property prediction, and a diffusion model as the polymer design module, which takes the global polymer embedding as an additional condition for downstream design.
Further details on finetuning are provided in Section~\ref{sec: finetuning}.

Finally, details on the experimental setup, including datasets, baselines, and metrics, are provided in Section~\ref{sec: exp_setup}. 
The outcomes and observations, including results, analyses, and ablation studies, are provided in the following subsections and Supplementary Information~\ref{SI-sec: exp}.

\subsection{Unlocking Polymer Conformation Generation with PolyConFM}
As illustrated in Figure~\ref{fig: overview} and Section~\ref{sec: pretraining}, conformation-centric generative pretraining has enabled PolyConFM to generate polymer conformations that serve as inputs for downstream tasks.
Here, given the lack of specialized polymer conformation generation methods, we compare PolyConFM's conformation generation capability with various representative molecular conformation generation methods trained on the linear polymer conformation dataset (construction pipeline and statistics of this dataset are provided in Supplementary Information~\ref{SI-sec: datasets}), and evaluate performance using both structure-matching and energy-matching metrics.
More information regarding baselines and metrics is provided in Section~\ref{sec: exp_setup}.

\begin{table*}[tbp]
    \setlength{\tabcolsep}{0.49em}
    \caption{The performance comparison of different methods on the polymer conformation generation task, and the best result for each metric has been bolded. \re{Here, we evaluate generated conformations using structure-matching (S-MAT) and energy-matching (E-MAT) metrics, where Recall-based (-R) metrics measure the coverage of the reference space and Precision-based (-P) metrics assess generation fidelity.}
    In particular, for the scalability evaluation, we double the number of repeating units per polymer in the test set during inference.
    \rre{Here, all reported results are evaluated on the test set.}}
    \centering
        \begin{footnotesize}
        \scalebox{1}{
        \begin{tabular}{llccccccccc}
        \toprule
        & \multirow{3}[2]{*}{Method} & \multicolumn{4}{c}{{Structure}} & \multicolumn{4}{c}{{Energy}} & \multirow{3}[2]{*}{\shortstack{Inference Time$^{*}$\\(min/conf)}}\\
        & & \multicolumn{2}{c}{S-MAT-R $\downarrow$} & \multicolumn{2}{c}{S-MAT-P $\downarrow$} & \multicolumn{2}{c}{E-MAT-R $\downarrow$} & \multicolumn{2}{c}{E-MAT-P $\downarrow$} \\
        \cmidrule(lr){3-4} \cmidrule(lr){5-6} \cmidrule(lr){7-8} \cmidrule(lr){9-10} 
        & & Mean  & Median & Mean  & Median & Mean  & Median & Mean  & Median \\
        \midrule
        \multirow{5}{*}{\begin{sideways}\shortstack{Standard\\Evaluation}\end{sideways}}
            & GeoDiff~\cite{xu2022geodiff} & 93.119  & 89.767  & 95.259  & 91.869  & 21.249  & 18.106  & 64.871  & 58.711 & 3.540 \\
            & TorsionalDiff~\cite{jing2022torsional} & 53.210  & 38.710  & 70.679  & 60.744  & 2.605  & 1.034  & 8.402  & 6.851  & 0.452 \\
            & MCF~\cite{wang2024swallowing}   & 248.432  & 242.866  & 258.891  & 253.239  &  \multicolumn{4}{c}{$> 10^{10}$} & 1.123 \\
            & ET-Flow~\cite{hassan2024flow} & 94.057  & 90.475  & 96.896  & 92.877  & 6.733  & 5.186  & 53.528  & 30.125  & 0.401 \\
            & \cellcolor{lightblue}PolyConFM & \cellcolor{lightblue}\textbf{\kern1pt 35.021} & \cellcolor{lightblue}\textbf{\kern1pt 24.279} & \cellcolor{lightblue}\textbf{\kern1pt 46.861} & \cellcolor{lightblue}\textbf{\kern1pt 37.996} & \cellcolor{lightblue}\textbf{\kern1pt 0.933} & \cellcolor{lightblue}\textbf{\kern1pt 0.359} & \cellcolor{lightblue}\textbf{\kern1pt 6.191} & \cellcolor{lightblue}\textbf{\kern1pt 4.122} & \cellcolor{lightblue}\textbf{\kern1pt 0.397} \\
        \midrule
        \multirow{5}{*}{\begin{sideways}\shortstack{Scalability\\Evaluation}\end{sideways}}
            & GeoDiff~\cite{xu2022geodiff} & 184.668  & 175.607  & 186.861  & 177.645  & 52.614  & 47.872  & 112.883  & 105.197  & 4.979\\
            & TorsionalDiff~\cite{jing2022torsional} & 119.289  & 94.075  & 146.816  & 126.932  & 5.219  & 2.216  & 11.692  & 9.227 & 1.384 \\
            & MCF~\cite{wang2024swallowing}   & 227.691  & 252.796  & 280.805  & 260.882  &  \multicolumn{4}{c}{$> 10^{10}$} & 1.488 \\
            & ET-Flow~\cite{hassan2024flow} & 186.132  & 176.370  & 188.725  & 178.977  & 15.331  & 12.465  & 65.116  & 41.642 & 0.744 \\
            & \cellcolor{lightblue}PolyConFM & \cellcolor{lightblue}\textbf{\kern1pt 65.040} & \cellcolor{lightblue}\textbf{\kern1pt 41.992} & \cellcolor{lightblue}\textbf{\kern1pt 84.626} & \cellcolor{lightblue}\textbf{\kern1pt 64.445} & \cellcolor{lightblue}\textbf{\kern1pt 1.259} & \cellcolor{lightblue}\textbf{\kern1pt 0.609} & \cellcolor{lightblue}\textbf{\kern1pt 5.785} & \cellcolor{lightblue}\textbf{\kern1pt 4.434} & \cellcolor{lightblue}\textbf{\kern1pt 0.637} \\
        \bottomrule
        \end{tabular}}
        \parbox{\linewidth}{\hspace{0.2em} \footnotesize $^{*}$ It represents the average time required to generate polymer conformations during inference.}
        \end{footnotesize}
    \label{tab: exp_conf}
\end{table*}%

Table~\ref{tab: exp_conf} summarizes the performance of various methods on the polymer conformation generation task, covering both the standard evaluation and the scalability evaluation. 
For the standard evaluation, we perform inference to generate conformations at a scale matching the training set (approximately 2,000 atoms per conformation), thereby evaluating their conformation generation capability under in-distribution conditions.
Here, we evaluate generated conformations using structure-matching (S-MAT) and energy-matching (E-MAT) metrics, where Recall-based (-R) metrics measure the coverage of the reference space and Precision-based (-P) metrics assess generation fidelity.
As presented in Table~\ref{tab: exp_conf}(top), PolyConFM achieves state-of-the-art performance across all evaluation metrics while requiring the least inference time, ensuring both effective and efficient polymer conformation generation.
In particular, compared with TorsionalDiff (i.e., the best baseline), PolyConFM improves all evaluation metrics by at least 25\% while maintaining comparable inference efficiency and eliminating the need for predetermined initial structures, highlighting its practicality for generating polymer conformations that are both structurally accurate and energetically realistic.

Moreover, as linear polymers are macromolecules formed through the covalent bonding of monomers into continuous chains, their conformations exhibit multiscale characteristics arising from variations in the number of repeating units incorporated during polymerization.
In this context, we conduct another dedicated evaluation to compare the scalability of various methods when generating polymer conformations at larger scales (i.e., more repeating units).
Here, since models are all trained on conformations with approximately 2,000 atoms, we further perform inference to generate conformations with approximately 4,000 atoms by simply doubling the number of repeating units per polymer.
Meanwhile, we apply the construction pipeline described in Supplementary Information~\ref{SI-sec: datasets} to generate ground‑truth enlarged conformations for performance evaluation.
As presented in Table~\ref{tab: exp_conf} (bottom), the natural advantages of masked autoregressive modeling within its conformation-centric generative pretraining indeed enable PolyConFM to scale effectively, yielding significant improvements over all baselines across all evaluation metrics, thereby setting it apart as the most promising method for multiscale conformation generation.

In addition, we present visualization examples of polymer conformations generated by the best baseline and PolyConFM in~\ref{fig: conf_case}, along with expansion experiments and analyses in Supplementary Information~\ref{SI-sec: exp_conf}, to furnish further insights.
Overall, through conformation-centric generative pretraining, PolyConFM successfully unlocks its conformation generation capability, thereby providing crucial global structural information for downstream tasks.

\subsection{Improved Polymer Property Prediction with PolyConFM}\label{sec: results_property}
As illustrated in Figure~\ref{fig: overview} and Section~\ref{sec: finetuning}, PolyConFM directly employs conformations generated by itself to derive structure-aware polymer embeddings, which are then fed into the polymer modeling module for the downstream polymer property prediction task. 
Here, we instantiate this modeling module as a simple multi-layer perceptron (MLP) layer and compare PolyConFM’s property prediction capability against state‑of‑the‑art baselines across diverse polymer property datasets.
More information regarding baselines and datasets is provided in Section~\ref{sec: exp_setup}.

Table~\ref{tab: exp_proprety} summarizes the performance of various methods on the downstream polymer property prediction task, covering eight typical polymer property datasets.
Here, since these property datasets are all formulated as regression problems, both root mean squared error (RMSE) and coefficient of determination ($R^2$) are used as evaluation metrics, with results reported as mean $\pm$ standard deviation.
As presented in Table~\ref{tab: exp_proprety}, PolyConFM consistently outperforms all baselines across all evaluation metrics on all property datasets, demonstrating superior generalization and robustness on the polymer property prediction task.
In particular, compared with MMPolymer (i.e., the best baseline), PolyConFM achieves tangible improvement on representative datasets.
For example, the RMSE metric decreases by more than 20$\%$ on Eat and by nearly 8\% on Eea, highlighting improved fidelity in modeling structure–property relationships.
Furthermore, the comparison with molecular baselines (i.e., MolCLR, 3D Infomax, and Uni-Mol) also reveals several noteworthy insights.
On the one hand, baselines (i.e., 3D Infomax and Uni-Mol) that incorporate 3D structural information consistently outperform those that do not (i.e., MolCLR), underscoring the critical value of structural features.
On the other hand, these molecular baselines perform substantially worse than the best polymer baseline (i.e., MMPolymer) and fall even further behind PolyConFM, underscoring the inherent limitations of directly transplanting molecular methods to polymer-specific tasks.

To further substantiate the superiority of PolyConFM, we extend property prediction experiments to two critical thermal properties: the glass transition temperature ($T_g$) and melting temperature ($T_m$).
Specifically, aggregated from established literature~\cite{giro2023ai, qiu2024polync} and molecular dynamics simulations following the pipeline in~\cite{liu2025open}, these datasets for $T_{g}$ and $T_m$ depend more strongly on long-range conformations, thus rendering them uniquely suited to highlight the advanced modeling capacity of PolyConFM.
Here, we directly compare PolyConFM against three most competitive baselines (i.e., Uni-Mol, Transpolymer, and MMPolymer) identified in Table~\ref{tab: exp_proprety} under two rigorous dataset splitting strategies.
As demonstrated in Table~\ref{tab:main_tg_tm_results}, PolyConFM consistently achieves state-of-the-art performance for both thermal properties across the two splitting strategies. 
For example, under the scaffold-based splitting strategy, PolyConFM reduces the RMSE by almost 12\% for $T_g$ while improving the $R^2$ by almost 15\% for $T_m$ compared to the best baseline.
Notably, while traditional models mentioned in existing works~\cite{averochkin2026machine, he2026bridging} could report better results (e.g., higher $R^2$ or lower RMSE) for $T_g$ and $T_m$, such outcomes are typically obtained under random-based dataset splitting, which fails to prevent significant structural similarities between training and test sets, thereby allowing models to perform simple interpolation within the familiar chemical space.
In contrast, this work compares PolyConFM and baselines under more rigorous dataset splitting (i.e., cluster-based and scaffold-based), which ensures that training and test sets inhabit distinct chemical regions, thereby forcing models to generalize to entirely unseen chemical families rather than performing simple interpolation.
Given that $T_g$ and $T_m$ are fundamentally governed by long-range conformations rather than local motifs, the results under these rigorous out-of-distribution (OOD) settings more faithfully reflect PolyConFM's predictive power.

In addition, we present t-SNE visualization in~\ref{fig: exp_property_tsne} and predicted–versus–true scatter plots in~\ref{fig: exp_property_scatter} to complement the above numerical results, along with expansion experiments and analyses in Supplementary Information~\ref{SI-sec: exp_property} to furnish further insights.
Overall, PolyConFM significantly improves property prediction, thereby enhancing the practical applicability and robustness of structure–property relationship modeling.

\begin{table*}[t]
    \setlength{\tabcolsep}{0.15em}
    \caption{The performance comparison of different methods on the downstream polymer property prediction task, and the best result for each polymer property dataset has been bolded.
    \rre{Here, all reported results are evaluated on the test set.}} 
    \centering
    \begin{footnotesize}
    \scalebox{1}{
    \begin{tabular}{llcccccccc}
    \toprule
     & Method & Egc   & Egb   & Eea   & Ei    & Xc    & EPS   & Nc   & Eat \\
    \midrule
    \multirow{7}{*}{\begin{sideways}RMSE ($\downarrow$)\end{sideways}}
        & MolCLR~\cite{wang2022molecular} & 0.587$_{\pm\text{0.024}}$ & 0.644$_{\pm\text{0.072}}$ & 0.404$_{\pm\text{0.017}}$ & 0.533$_{\pm\text{0.053}}$ & 21.719$_{\pm\text{1.631}}$ & 0.631$_{\pm\text{0.045}}$ &
        0.117$_{\pm\text{0.015}}$ & 0.094$_{\pm\text{0.033}}$ \\

        & 3D Infomax~\cite{stark20223d} & 0.494$_{\pm\text{0.039}}$ & 0.553$_{\pm\text{0.032}}$ & 0.335$_{\pm\text{0.055}}$ & 0.449$_{\pm\text{0.086}}$ & 19.483$_{\pm\text{2.491}}$ & 0.582$_{\pm\text{0.054}}$ &
        0.101$_{\pm\text{0.018}}$ & 0.094$_{\pm\text{0.039}}$ \\

        & Uni-Mol~\cite{zhou2023unimol} & 0.489$_{\pm\text{0.028}}$ & 0.531$_{\pm\text{0.055}}$  & 0.332$_{\pm\text{0.027}}$ & 0.407$_{\pm\text{0.080}}$ & {17.414}$_{\pm\text{1.581}}$ & {0.536}$_{\pm\text{0.053}}$ & 0.095$_{\pm\text{0.013}}$ & 0.084$_{\pm\text{0.034}}$  \\
        
        & polyBERT~\cite{kuenneth2023polybert} & 0.553$_{\pm\text{0.011}}$ & 0.759$_{\pm\text{0.042}}$ & 0.363$_{\pm\text{0.037}}$ & 0.526$_{\pm\text{0.068}}$ & 18.437$_{\pm\text{0.560}}$ & 0.618$_{\pm\text{0.049}}$ & 0.113$_{\pm\text{0.003}}$ & 0.172$_{\pm\text{0.016}}$ \\

        & Transpolymer~\cite{xu2023transpolymer} & {0.453}$_{\pm\text{0.007}}$ & 0.576$_{\pm\text{0.021}}$ & 0.326$_{\pm\text{0.040}}$ & {0.397}$_{\pm\text{0.061}}$ & {17.740}$_{\pm\text{0.732}}$ & {0.547}$_{\pm\text{0.051}}$ & {0.096}$_{\pm\text{0.016}}$ & 0.147$_{\pm\text{0.093}}$ \\
        
        & MMPolymer~\cite{wang2024mmpolymer} & {{0.431}$_{\pm\text{0.017}}$} & 
        {{0.496}$_{\pm\text{0.031}}$}  & {{0.286}$_{\pm\text{0.029}}$}  & {{0.390}$_{\pm\text{0.057}}$} & {{16.814}$_{\pm\text{0.867}}$} & {{0.511}$_{\pm\text{0.035}}$} & 
        {{0.087}$_{\pm\text{0.010}}$} & {0.061}$_{\pm\text{0.016}}$ \\

        & \cellcolor{lightblue} \kern-2.5pt PolyConFM & 
        {\cellcolor{lightblue}\textbf{0.429}$_{\pm\text{0.016}}$} & {\cellcolor{lightblue}\textbf{0.473}$_{\pm\text{0.052}}$} & {\cellcolor{lightblue}\textbf{0.265}$_{\pm\text{0.032}}$} & {\cellcolor{lightblue}\textbf{0.384}$_{\pm\text{0.072}}$} & {\cellcolor{lightblue}\textbf{16.737}$_{\pm\text{1.136}}$} & {\cellcolor{lightblue}\textbf{0.477}$_{\pm\text{0.028}}$} & 
        {\cellcolor{lightblue}\textbf{0.082}$_{\pm\text{0.009}}$} & {\cellcolor{lightblue}\textbf{0.048}}$_{\pm\text{0.026}}$ \\
    \midrule
    \multirow{7}{*}{\begin{sideways}$R^2$ ($\uparrow$)\end{sideways}}
          & MolCLR~\cite{wang2022molecular} & 0.858$_{\pm\text{0.010}}$ & 0.882$_{\pm\text{0.027}}$ & 0.854$_{\pm\text{0.038}}$ & 0.689$_{\pm\text{0.037}}$ & 0.176$_{\pm\text{0.026}}$ & 0.683$_{\pm\text{0.020}}$ &
          0.764$_{\pm\text{0.037}}$ & 0.885$_{\pm\text{0.104}}$ \\

          & 3D Infomax~\cite{stark20223d} & 0.900$_{\pm\text{0.016}}$ & 0.898$_{\pm\text{0.018}}$ & 0.891$_{\pm\text{0.045}}$ & 0.766$_{\pm\text{0.086}}$ & 0.274$_{\pm\text{0.122}}$ & 0.690$_{\pm\text{0.063}}$ &
          0.797$_{\pm\text{0.086}}$ & 0.869$_{\pm\text{0.097}}$ \\

          & Uni-Mol~\cite{zhou2023unimol} & 0.901$_{\pm\text{0.013}}$ & 0.925$_{\pm\text{0.011}}$  & 0.901$_{\pm\text{0.027}}$ & 0.820$_{\pm\text{0.075}}$ & 0.454$_{\pm\text{0.079}}$ & 0.751$_{\pm\text{0.085}}$ & 0.828$_{\pm\text{0.072}}$ & 0.937$_{\pm\text{0.032}}$  \\
          
          & polyBERT~\cite{kuenneth2023polybert} & 0.875$_{\pm\text{0.006}}$ & 0.844$_{\pm\text{0.034}}$ & 0.880$_{\pm\text{0.035}}$ & 0.705$_{\pm\text{0.085}}$ & 0.384$_{\pm\text{0.066}}$ & 0.681$_{\pm\text{0.058}}$ & 0.769$_{\pm\text{0.034}}$ & 0.672$_{\pm\text{0.119}}$ \\

          & Transpolymer~\cite{xu2023transpolymer} &
         {0.916}$_{\pm\text{0.002}}$ & 0.911$_{\pm\text{0.008}}$ & 0.902$_{\pm\text{0.036}}$ & {0.830}$_{\pm\text{0.059}}$ & {0.430}$_{\pm\text{0.058}}$ & {0.744}$_{\pm\text{0.075}}$ & {0.826}$_{\pm\text{0.071}}$ & 0.800$_{\pm\text{0.172}}$ \\
          
          & MMPolymer~\cite{wang2024mmpolymer} & 
          {{0.924}}$_{\pm\text{0.006}}$ & {{0.934}}$_{\pm\text{0.008}}$ & {{0.925}}$_{\pm\text{0.025}}$ & {{0.836}}$_{\pm\text{0.053}}$ & 
          {{0.488}}$_{\pm\text{0.072}}$ & {{0.779}}$_{\pm\text{0.052}}$ & {{0.864}}$_{\pm\text{0.036}}$ & 0.961$_{\pm\text{0.018}}$ \\

          & \cellcolor{lightblue} \kern-2.5pt PolyConFM & 
          {\cellcolor{lightblue}\textbf{0.925}}$_{\pm\text{0.007}}$ & {\cellcolor{lightblue}\textbf{0.940}}$_{\pm\text{0.009}}$ & {\cellcolor{lightblue}\textbf{0.935}}$_{\pm\text{0.024}}$ & {\cellcolor{lightblue}\textbf{0.839}}$_{\pm\text{0.061}}$ & 
          {\cellcolor{lightblue}\textbf{0.492}}$_{\pm\text{0.088}}$ & {\cellcolor{lightblue}\textbf{0.806}}$_{\pm\text{0.049}}$ & {\cellcolor{lightblue}\textbf{0.875}}$_{\pm\text{0.037}}$ & {\cellcolor{lightblue}\textbf{0.979}}$_{\pm\text{0.016}}$ \\
    \bottomrule
    \end{tabular}}
    \end{footnotesize}
  \label{tab: exp_proprety}
\end{table*}%

\begin{table*}[t]
\setlength{\tabcolsep}{5.9pt}
\centering
\begin{footnotesize}
\caption{\rre{The property prediction results for glass transition temperature ($T_g$) and melting temperature ($T_m$) under two rigorous dataset splitting strategies. All reported results are evaluated on the test set.}}
\label{tab:main_tg_tm_results}
\begin{tabular}{cccccccc}
\toprule
\multirow{2}[2]{*}{\shortstack{Property}} & \multirow{2}[2]{*}{\shortstack{\# Sample}} & \multirow{2}[2]{*}{\shortstack{Data Range}} & \multirow{2}[2]{*}{Method} & \multicolumn{2}{c}{Cluster-based Splitting} & \multicolumn{2}{c}{Scaffold-based Splitting} \\
\cmidrule(lr){5-6} \cmidrule(lr){7-8}
 & & & & RMSE $\downarrow$ & $R^2$ $\uparrow$ & RMSE $\downarrow$ & $R^2$ $\uparrow$ \\
\midrule
\multirow{4}{*}{\shortstack{$T_g$}} 
 & \multirow{4}{*}{3,844} 
 & \multirow{4}{*}{[-118, 500]} 
 & Uni-Mol~\cite{zhou2023unimol}          & 63.855$_{\pm\text{1.688}}$ & 0.614$_{\pm\text{0.020}}$ & 51.177$_{\pm\text{0.791}}$ & 0.656$_{\pm\text{0.011}}$ \\
 & & & Transpolymer~\cite{xu2023transpolymer} & 57.177$_{\pm\text{1.172}}$ & 0.690$_{\pm\text{0.013}}$ & 44.581$_{\pm\text{0.647}}$ & 0.739$_{\pm\text{0.008}}$ \\
 & & & MMPolymer~\cite{wang2024mmpolymer}     & 55.908$_{\pm\text{0.474}}$ & 0.704$_{\pm\text{0.005}}$ & 44.346$_{\pm\text{1.167}}$ & 0.741$_{\pm\text{0.014}}$ \\
 & & & \cellcolor{lightblue}\kern-3pt PolyConFM 
 & {\cellcolor{lightblue}\textbf{52.791}$_{\pm\text{0.821}}$} & {\cellcolor{lightblue}\textbf{0.736}$_{\pm\text{0.008}}$} & {\cellcolor{lightblue}\textbf{38.766}$_{\pm\text{0.864}}$} & {\cellcolor{lightblue}\textbf{0.802}$_{\pm\text{0.009}}$} \\
\midrule
\multirow{4}{*}{\shortstack{$T_m$}} 
 & \multirow{4}{*}{3,179} 
 & \multirow{4}{*}{[-55, 580]} 
 & Uni-Mol~\cite{zhou2023unimol}          & 83.548$_{\pm\text{2.082}}$ & 0.333$_{\pm\text{0.034}}$ & 91.721$_{\pm\text{4.466}}$ & 0.382$_{\pm\text{0.058}}$ \\
 & & & Transpolymer~\cite{xu2023transpolymer} & 79.189$_{\pm\text{1.359}}$ & 0.401$_{\pm\text{0.020}}$ & 82.960$_{\pm\text{1.987}}$ & 0.495$_{\pm\text{0.024}}$ \\
 & & & MMPolymer~\cite{wang2024mmpolymer}     & 79.318$_{\pm\text{2.626}}$ & 0.399$_{\pm\text{0.040}}$ & 82.274$_{\pm\text{1.973}}$ & 0.504$_{\pm\text{0.024}}$ \\  
 & & & \cellcolor{lightblue}\kern-3pt PolyConFM 
 & {\cellcolor{lightblue}\textbf{73.486}$_{\pm\text{0.635}}$} & {\cellcolor{lightblue}\textbf{0.484}$_{\pm\text{0.009}}$} & {\cellcolor{lightblue}\textbf{75.824}$_{\pm\text{4.465}}$} & {\cellcolor{lightblue}\textbf{0.578}$_{\pm\text{0.050}}$} \\
\bottomrule
\end{tabular}
\end{footnotesize}
\end{table*}

\begin{table*}[t]
    \setlength{\tabcolsep}{0.28em}
    \caption{\rre{The performance comparison of different methods on the downstream polymer design task, and the best result for each metric has been bolded. In particular, the conditioning set comprises the synthetic score (Synth.) together with the gas permeabilities of $\text{O}_2$, $\text{N}_2$, and $\text{CO}_2$. Here, all reported results are evaluated on the test set.}} 
    \centering
    \begin{footnotesize}
    \scalebox{1}{
    \begin{tabular}{lccccccccc}
    \toprule
    \multirow{2}[2]{*}{Method} & \multicolumn{4}{c}{Distribution Learning} & \multicolumn{4}{c}{Condition Control} \\
    \cmidrule(lr){2-5} \cmidrule(lr){6-9} 
    & Coverage $\uparrow$ & Diversity $\uparrow$ & Similarity $\uparrow$ & Distance $\downarrow$ & Synth. $\downarrow$ & O2Perm $\downarrow$ & N2Perm $\downarrow$ & CO2Perm $\downarrow$ \\ 
    \midrule
    MolGPT~\cite{bagal2021molgpt}  & $1.000_{\pm0.000}$ & $0.792_{\pm0.001}$ & $0.954_{\pm0.003}$ & $\phantom{0}7.941_{\pm0.016}$  & $1.768_{\pm0.008}$ & $1.075_{\pm0.003}$ & $0.993_{\pm0.001}$ & $1.109_{\pm0.002}$ \\
    GraphGA~\cite{jensen2019graph} & $1.000_{\pm0.000}$ & $0.828_{\pm0.002}$ & $0.959_{\pm0.001}$ & $\phantom{0}8.654_{\pm0.023}$  & $1.583_{\pm0.006}$ & $1.146_{\pm0.005}$ & $1.071_{\pm0.004}$ & $0.979_{\pm0.003}$ \\
    DiGress~\cite{vignac2023digress} & $1.000_{\pm0.000}$ & $\textbf{0.897}_{\pm0.002}$ & $0.385_{\pm0.003}$ & $21.206_{\pm0.021}$ & $2.530_{\pm0.005}$ & $1.796_{\pm0.002}$ & $2.193_{\pm0.002}$ & $1.720_{\pm0.002}$ \\
    GDSS~\cite{jo2022score}   & $0.667_{\pm0.000}$ & $0.826_{\pm0.001}$ & $0.001_{\pm0.000}$ & $35.557_{\pm0.018}$ & $1.361_{\pm0.003}$ & $0.927_{\pm0.005}$ & $1.009_{\pm0.005}$ & $1.268_{\pm0.002}$ \\
    MOOD~\cite{lee2023exploring}   & $0.833_{\pm0.000}$  & $0.843_{\pm0.002}$ & $0.005_{\pm0.002}$ & $40.790_{\pm0.009}$ & $1.532_{\pm0.006}$ & $1.256_{\pm0.005}$ & $1.072_{\pm0.003}$ & $1.402_{\pm0.001}$ \\
    GraphDiT~\cite{liu2024graph}  & $1.000_{\pm0.000}$ & $0.858_{\pm0.001}$ & $0.974_{\pm0.001}$ & $\phantom{0}7.209_{\pm0.035}$  & $1.239_{\pm0.013}$ & $0.871_{\pm0.011}$ & $1.063_{\pm0.016}$ & $0.867_{\pm0.011}$ \\
    \rowcolor{lightblue}
    PolyConFM & $1.000_{\pm0.000}$ & $0.845_{\pm0.001}$ & $\textbf{0.981}_{\pm0.000}$ &  $\phantom{0}\textbf{6.518}_{\pm0.023}$  & $\textbf{0.860}_{\pm0.003}$ &  $\textbf{0.834}_{\pm0.004}$ &  $\textbf{0.987}_{\pm0.005}$ &  $\textbf{0.820}_{\pm0.003}$ \\
    \bottomrule
    \end{tabular}}
    \end{footnotesize}
  \label{tab: exp_design}
\end{table*}

\subsection{Enhanced Polymer Design with PolyConFM}\label{sec: results_design}
As illustrated in Figure~\ref{fig: overview} and Section~\ref{sec: finetuning}, PolyConFM treats polymer design as a typical conditional generation task and leverages the learned global embedding of the reference polymer as an additional conditioning signal, thereby enhancing guidance through effective structure modeling.
Besides, due to the vast chemical space and practical manufacturing constraints, we also formulate this task as generating suitable 2D graph structures that satisfy specific conditions, consistent with previous works.
Here, we employ a graph-based diffusion model as the polymer design module and compare PolyConFM’s design capability with state-of-the-art baselines across diverse evaluation metrics. 
More information regarding datasets, baselines, and metrics is provided in Section~\ref{sec: exp_setup}.

Table~\ref{tab: exp_design} summarizes the performance of various methods on the downstream polymer design task, covering comprehensive evaluation metrics for both distribution learning and condition control.
Here, the conditioning set consists of the synthetic score (Synth.) and three numerical properties (gas permeabilities of $\text{O}_2$, $\text{N}_2$, and $\text{CO}_2$), and our goal is to generate polymers that satisfy these conditions while maintaining distributional consistency.
As presented in Table~\ref{tab: exp_design}, PolyConFM exhibits a favorable balance between distributional fidelity and conditional satisfaction, achieving state-of-the-art performance.
In particular, with respect to distributional fidelity, PolyConFM reaches perfect heavy‑atom type coverage, the highest fragment‑based similarity, and the lowest Fréchet ChemNet Distance, while maintaining competitive diversity.
On the condition‑control side, PolyConFM consistently enhances control across all conditions, reducing MAE (i.e., mean absolute error) on the synthetic score by over 30\% relative to the best baseline.
Taken together, these results indicate that PolyConFM not only accurately captures the reference distribution but also closely adheres to the conditioning signals, underscoring its effectiveness in designing the desired polymers.

In addition, we present the performance of various methods conditioned on a single gas permeability in~\ref{tab: exp_design_add}, along with expansion experiments and analyses in Supplementary Information~\ref{SI-sec: exp_design}, to furnish further insights.
Overall, PolyConFM effectively enhances polymer design, thereby accelerating polymer discovery by generating numerous candidates that satisfy the required conditions.

\subsection{Ablation Experiments}\label{sec: ablation}

\begin{figure*}[t]
    \centering 
        \includegraphics[width=0.97\textwidth]{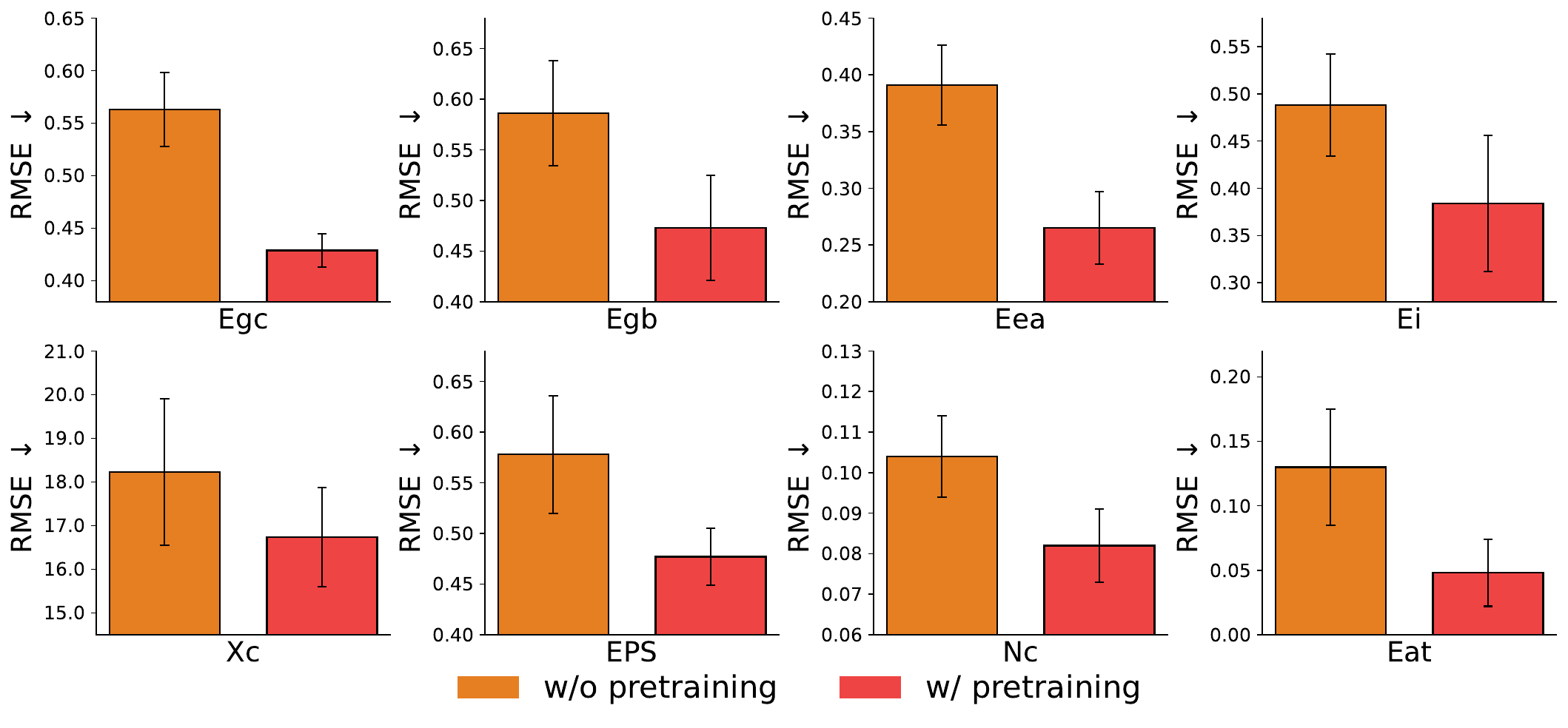} 
    \caption{\rre{The ablation study on the necessity of pretraining, where both variants are implemented with the same model architecture and data pipeline.
    Here, the only independent variable is the training strategy: training from scratch (w/o pretraining) versus pretraining followed by fine-tuning (w/ pretraining), thereby ensuring any observed performance differences can be solely attributed to the pretraining phase. Here, all reported results are evaluated on the test set.}}
    \label{fig: pretraining_abl}
\end{figure*}

\begin{figure*}[t]
    \centering 
        \includegraphics[width=0.97\textwidth]{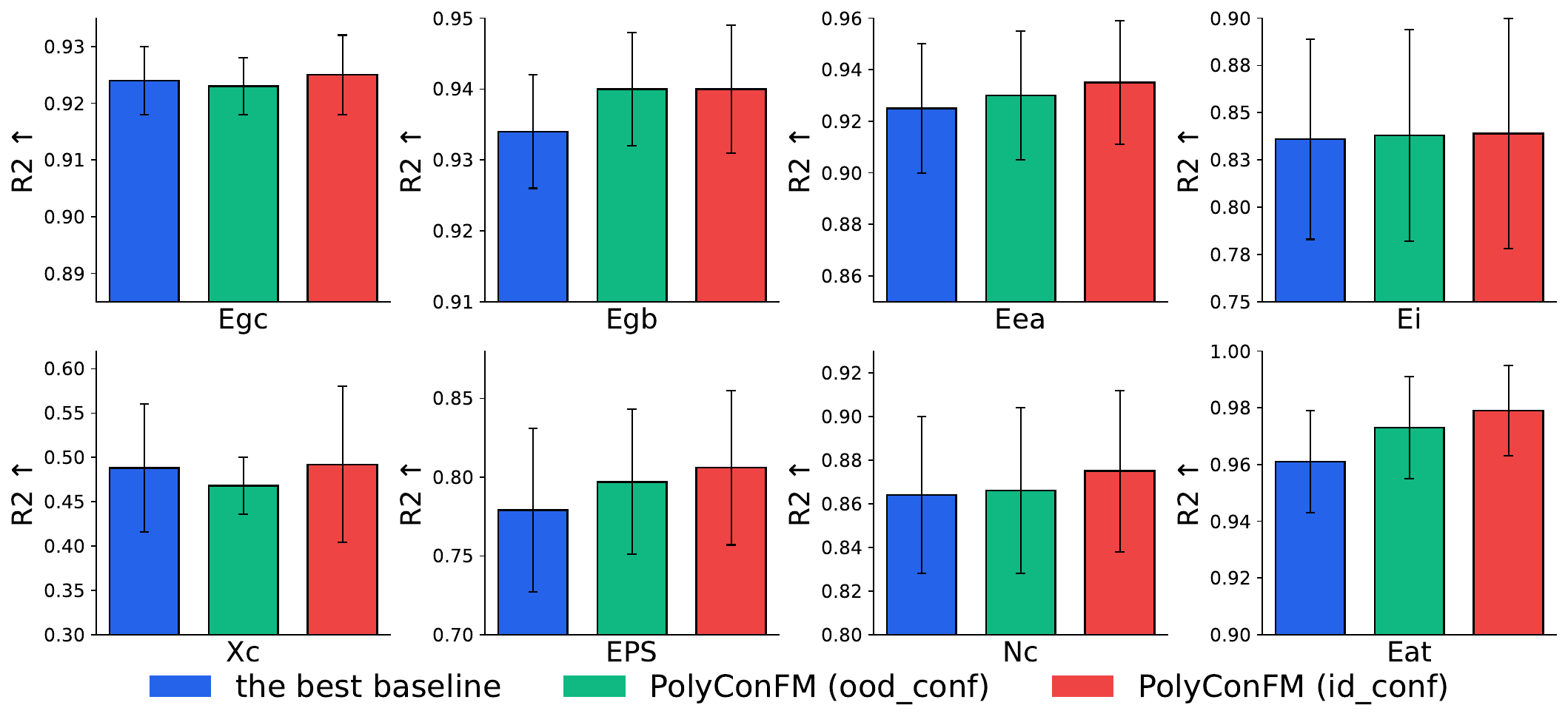} 
    \caption{The ablation study on the downstream polymer property prediction task, where PolyConFM takes externally provided conformations (i.e., ood$\_$conf) and self-generated conformations (i.e., id$\_$conf) as inputs, respectively. The best baseline (i.e., MMPolymer) is also included here for comparison. 
    \rre{Here, all reported results are evaluated on the test set.}}
    \label{fig: property_abl}
\end{figure*}

\begin{figure*}[t]
    \centering 
        \includegraphics[width=0.97\textwidth]{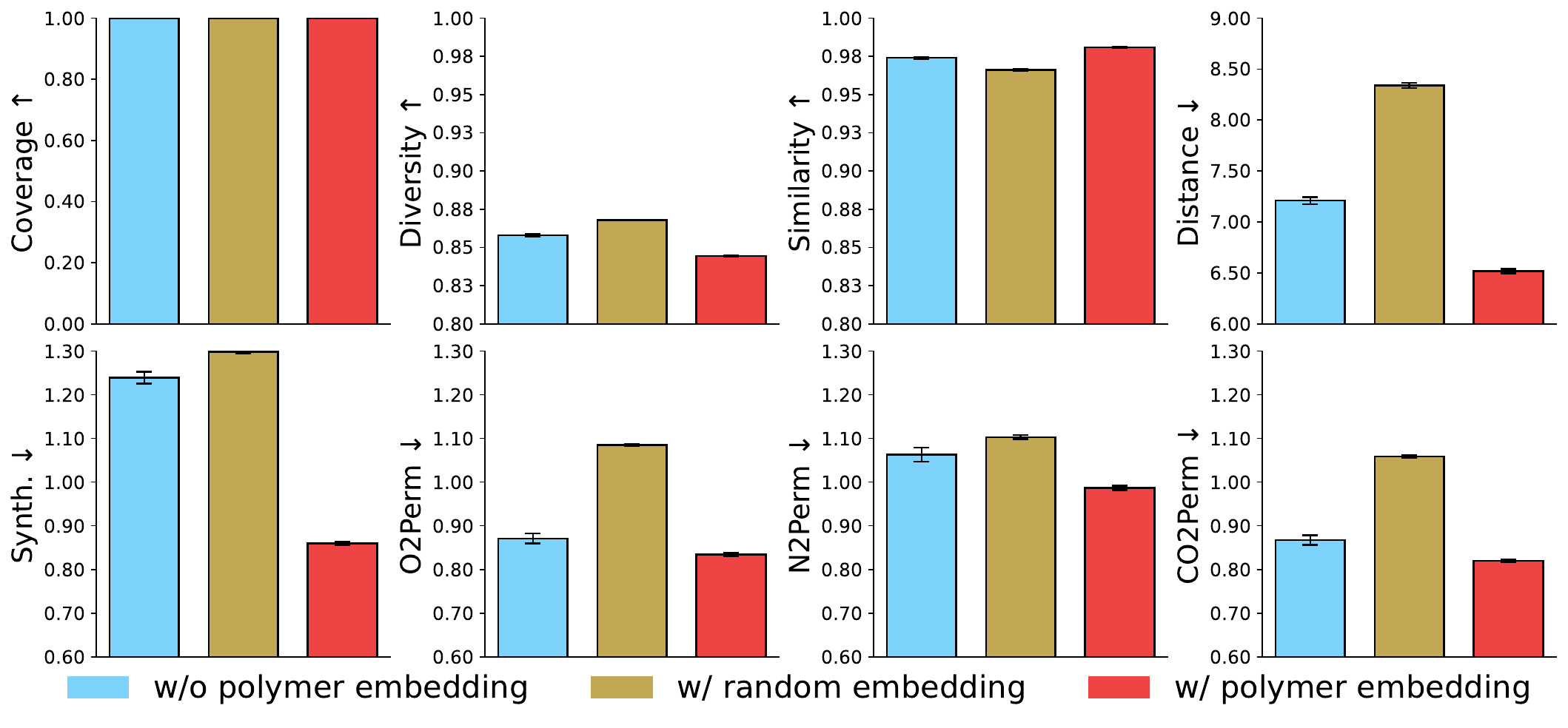} 
    \caption{The ablation study on the downstream polymer design task, where the polymer embedding is either removed or replaced with the random embedding of the same dimension.
    \rre{Here, all reported results are evaluated on the test set.}}
    \label{fig: design_abl}
\end{figure*}

To rigorously validate the necessity of pretraining and demonstrate how much of the downstream gains originate specifically from the pretraining phase, ablation experiments are conducted on the downstream polymer property prediction task.
As presented in Figure~\ref{fig: pretraining_abl}, we compare the standard PolyConFM (i.e., w/ pretraining) against the variant trained from scratch (i.e., w/o pretraining) while keeping all other configurations (e.g., model architecture and data pipeline) identical, thereby isolating the performance gains attributable to the pretraining phase.
Here, these results clearly demonstrate that the standard PolyConFM consistently outperforms the ``w/o pretraining'' variant across all eight property datasets.
This substantial margin confirms that the downstream gains are fundamentally driven by the structural priors captured during our conformation-centric generative pretraining, highlighting its indispensable role as the foundation for accurate polymer modeling.

As illustrated in Section~\ref{sec: results_property}, PolyConFM establishes state-of-the-art performance on the downstream polymer property prediction task through directly leveraging conformations generated by itself to derive structure-aware polymer embeddings.
Here, we further replace these self-generated conformations with those initial structures from the construction pipeline in Supplementary Information~\ref{SI-sec: datasets} to provide noteworthy insight into performance gains.
As presented in Figure~\ref{fig: property_abl}, even taking those externally provided conformations as input, PolyConFM still significantly outperforms the best baseline (i.e., MMPolymer), indicating that conformation-centric generative pretraining equips it with transferable structure-aware priors that remain effective irrespective of the conformation source. 
Meanwhile, using conformations generated by itself indeed improves performance on all property datasets, suggesting that self-generated conformations are better aligned with PolyConFM’s representation space.
Collectively, these results highlight the critical role of conformational information in property prediction and demonstrate that aligning representation and generation yields additional gains, thereby validating both the rationale and the necessity of our conformation-centric generative pretraining paradigm.

In addition, as illustrated in Section~\ref{sec: results_design}, PolyConFM leverages the learned global embedding of the reference polymer as an additional conditioning signal to better guide the polymer design module.
To provide noteworthy insight into performance gain, ablation experiments are conducted through either removing polymer embeddings or replacing them with random embeddings of equal dimension.
As presented in Figure~\ref{fig: design_abl}, compared with the other two variants, incorporating polymer embeddings yields consistent improvements across most evaluation metrics, particularly for those condition-control metrics.
Meanwhile, the random-embedding setting variant performs worse than the no-embedding setting, indicating that arbitrary embeddings provide no benefit. 
Taken together, these results demonstrate that PolyConFM indeed effectively captures structural priors via conformation-centric generative pretraining, thereby enhancing guidance of the downstream polymer design process.

\section{Discussion}
In this work, we introduce PolyConFM, a conformation-centric generative foundation model tailored for linear polymers, to address two fundamental limitations of existing methods: their reliance on monomer-level descriptors that ignore global structural information and their inability to simultaneously support both modeling and design.
Recognizing that each linear polymer is essentially a continuous chain whose conformation can be naturally decomposed into a sequence of repeating-unit conformations, we pretrain PolyConFM under the conditional generation paradigm, reconstructing these repeating-unit conformations via masked autoregressive (MAR) modeling and then generating their orientation transformations to recover the complete polymer conformation, thereby capturing complex dependencies among repeating units for global structure modeling while unlocking its conformation generation capability for diverse downstream tasks. 
Meanwhile, to mitigate the severe scarcity of conformation datasets, we devote considerable time and resources to constructing a dataset of over 50,000 linear polymers with conformations via molecular dynamics simulations, thereby enabling our conformation-centric generative pretraining and facilitating future research in this important field.
Extensive experiments consistently demonstrate that PolyConFM significantly outperforms task-specific methods across diverse tasks, establishing it as a foundation model that seamlessly bridges the structure, property, and design of linear polymers.

Despite these promising results, PolyConFM represents essential foundational progress rather than the ultimate solution for all polymeric systems.
Here, to fully realize the broad potential of conformation-centric foundation models, several open challenges remain to be explored for future research:

First, regarding structural and environmental complexity, the current scope of PolyConFM is primarily limited to linear polymers simulated as isolated single chains.
Building upon this solid foundation, a natural progression for future research is to expand this paradigm for broader structural diversity, particularly non-linear topologies (e.g., branched or cross-linked polymers).
Besides, scaling this paradigm to complex environments, particularly bulk melts and multi-component mixtures (e.g., polymer blends and solutions), remains the compelling avenue for future exploration in this important field.

Second, from the methodological perspective, PolyConFM currently utilizes conformations generated through molecular dynamics (MD) simulations for pretraining.
Although the consistently superior performance of PolyConFM across diverse downstream tasks firmly verifies the sufficiency and effectiveness of these MD-generated conformations in capturing critical structural priors, they inevitably inherit the biases and resolution limits of the empirical force fields employed. 
To capture the underlying physical interactions with higher fidelity, future iterations would benefit from exploring multiscale simulation strategies or from incorporating quantum-mechanical (QM) data into the pretraining pipeline.

Finally, from the practical application perspective, PolyConFM predominantly relies on surrogate models to evaluate generated candidates during polymer design.
While the employed surrogate models (i.e., the random-forest oracle) are widely recognized as standard practice in this field, and we further fortify this evaluation with high-fidelity cross-checks and uncertainty analyses to ensure reliability and prevent reward hacking, these rigorous in silico assessments ultimately lack wet-lab verification.
Since computational generation cannot replace physical realization, bridging the gap between in silico design and wet-lab synthesis remains the critical next step for future work.

\section{Methods}
\subsection{Frame-based Polymer Representation}\label{sec: rep}
For each linear polymer with $N$ atoms, we represent it as a graph ${\mathcal{G}} = (\mathcal{V}, \mathcal{E})$, where $\mathcal{V} = \{v_i\}_{i=1}^{N}$ is the set of atoms and $\mathcal{E} = \{e_{ij}\}_{i,j=1}^{N}$ is the set of bonds. 
Meanwhile, since each atom in $\mathcal{V}$ corresponds to a 3D coordinate vector $\bm c \in \mathbb{R}^{3}$, the corresponding polymer conformation can be represented as $\mathcal{C}=\{\bm{c}_{i}\}_{i=1}^{N}$.

Moreover, considering that linear polymers are macromolecules formed through the covalent bonding of numerous monomers into continuous chains, we can naturally decompose the complete conformation of each linear polymer into a sequence of local conformations (i.e., the corresponding conformations of those repeating units within this linear polymer).
Here, as illustrated in Figure~\ref{fig: overview}b, we extend the standard definition of repeating units in polymer science, incorporating one key atom from the preceding repeating unit and one key atom from the following repeating unit into the current repeating unit (i.e., atom-1 and atom-4).
In this context, adjacent repeating-unit conformations naturally overlap at those key atoms, where atom-1 of the current repeating unit aligns with atom-3 of the preceding repeating unit and atom-4 of the current repeating unit aligns with atom-2 of the following repeating unit, thereby simplifying polymer structure modeling.
Besides, following the modeling strategy widely adopted by protein residue~\cite{jumper2021highly, yim2023se, huguet2024sequence}, the frame can also be extracted based on those key atoms (e.g., atom-1, atom-3, and atom-4) within the corresponding repeating-unit conformation.
In particular, for the $i$-th repeating unit, the corresponding frame contains its orientation transformation~\footnote{The orientation transformation is relative to the standard coordinate system.} $\mathcal{O}_{i} = (\bm R_i, \bm t_i)$, where $\bm R_i \in \mathbb{R}^{3 \times 3}$ denotes the rotation transformation and $\bm t_i \in \mathbb{R}^{3}$ denotes the translation transformation.
Therefore, the polymer conformation can be further represented as follows:
\begin{equation}\label{eq: decompose}  
    \begin{aligned}  
        \mathcal{C} = \{{\mathcal{C}}^u, {\mathcal{O}}\} = \{\{\bm{C}_{i}^{u}\}_{i=1}^{N_u}, \{\mathcal{O}_{i}\}_{i=1}^{N_u}\} ,
    \end{aligned}  
\end{equation} 
where $N_u$ is the number of repeating units within this linear polymer, \re{${\mathcal{C}}^{u}$ is the set of repeating-unit conformations, ${\mathcal{O}}$ is the set of repeating-unit orientation transformations,}
$\bm{C}_{i}^{u} = [\bm c_{i,j}^{u}] \in \mathbb{R}^{(\frac{N}{N_u} + 2)\times 3}$ is the corresponding conformation of the $i$-th repeating unit and $\mathcal{O}_{i} = (\bm R_i, \bm t_i)$ is the corresponding orientation transformation of the $i$-th repeating unit.
For the $j$-th atom in the $i$-th repeating unit's conformation~\footnote{The repeating-unit conformation is also placed in the standard coordinate system through applying the inverse orientation transformation to the corresponding sub-structure within the polymer conformation.}, its corresponding 3D coordinates within the polymer conformation is expressed as $\bm {R}_i \bm c_{i,j}^{u}  + \bm {t}_i$.
Notably, since each repeating unit involves 2 overlapping atoms from the preceding and following repeating units, the number of atoms in each repeating unit is not $\frac{N}{N_u}$ but rather $\frac{N}{N_u} + 2$.

\subsection{Conformation-centric Generative Pretraining}\label{sec: pretraining}
As discussed in Section~\ref{sec: introduction}, designing generative pretraining around polymer conformation is a natural and effective choice for enabling PolyConFM to accurately capture global structural features and effectively support a wide range of downstream tasks.
In particular, we pretrain PolyConFM under the conditional generation paradigm, learning a generative model $p(\mathcal{C}|\mathcal{G})$ to model the empirical distribution of polymer conformation $\mathcal{C}$ conditioned on the corresponding polymer graph $\mathcal{G}$. 
Combined with Equation~\eqref{eq: decompose}, this pretraining objective can be further expressed as follows:
\begin{eqnarray}\label{eq: define}  
\begin{aligned}  
p(\mathcal{C}|\mathcal{G}) = p({\mathcal{C}}^{u}, {\mathcal{O}}|\mathcal{G}) 
= p({\mathcal{C}}^{u}|\mathcal{G}) \cdot p({\mathcal{O}}|\mathcal{G},{\mathcal{C}}^{u}),  
\end{aligned}  
\end{eqnarray}  
where ${\mathcal{C}}^{u}$ is the set of repeating-unit conformations, and ${\mathcal{O}}$ is the set of their orientation transformations.

Therefore, this pretraining objective can be naturally formulated as a two-phase learning process: 
(1) We first train PolyConFM to generate repeating-unit conformations ${\mathcal{C}}^{u}$ conditioned on the polymer graph $\mathcal{G}$, i.e., $p({\mathcal{C}}^{u}|\mathcal{G})$, and (2) then train it to generate their orientation transformations ${\mathcal{O}}$ conditioned on the polymer graph $\mathcal{G}$ and repeating-unit conformations ${\mathcal{C}}^{u}$, i.e., $p({\mathcal{O}}|\mathcal{G},{\mathcal{C}}^{u})$.
\re{Notably, we employ the teacher-forcing strategy to facilitate efficient optimization. Specifically, Phase-2 directly uses ground-truth repeating-unit conformations as conditional inputs during training to avoid error propagation, while the sequential dependency (Phase-1 $\rightarrow$ Phase-2) is strictly enforced during inference.}

In the following subsections, we illustrate this two‑phase learning process in turn.

\subsubsection{Phase-1: Repeating Unit Conformation Generation}
As illustrated in Figure~\ref{fig: overview}c, since the complete polymer conformation can be decomposed into a sequence of repeating-unit conformations \re{and each repeating-unit conformation is jointly shaped by complex spatial interactions from all other units across the entire polymer chain}, we treat these repeating-unit conformations as token-like structural units for model input, and then integrate the masked autoregressive modeling (MAR) with the SE(3) diffusion designed for repeating units to reconstruct them in random oreder, thereby capturing complex dependencies among repeating units for accurate global structure modeling.
The corresponding learning objective $p({\mathcal{C}}^{u}|\mathcal{G})$ in Equation~\eqref{eq: define} is therefore rewritten as follows:
\begin{equation}\label{eq: phase_1_define}  
    p({\mathcal{C}}^{u}|\mathcal{G}) = p(\{\bm{C}_{i}^{u}\}_{i=1}^{N_u}|\mathcal{G}) 
    = p(\{{\mathcal{C}}_{k}^{u}\}_{k=1}^{K}|\mathcal{G}) 
    = \sideset{}{_{k=1}^{K}}\prod p({\mathcal{C}}_{k}^{u}|\mathcal{G}, \{{\mathcal{C}}_{i}^{u}\}_{i=1}^{k-1}),  
\end{equation}  
where ${\mathcal{C}}^{u}=\{\bm{C}_{i}^{u}\}_{i=1}^{N_u}$ is the set of repeating-unit conformations, $\bm{C}_{i}^{u}$ is the corresponding conformation of the $i$-th repeating unit, and \rre{${\mathcal{C}}_{k}^{u}$ is the corresponding subset of ${\mathcal{C}}^{u}$ that contains $\frac{N_u}{K}$ repeating-unit conformations to be generated at the $k$-th autoregressive step}.
Besides, we define a random permutation $\pi$ over $\{1, ..., N_u\}$ to model the sampling order, thereby ${\mathcal{C}}_{k}^{u}$ can be further represented as follows: 
\begin{equation}\label{eq: pretraining_subset_definition}  
    {\mathcal{C}}_{k}^{u} = \{\bm{C}_{\pi(i)}^{u} \mid i \in \{(k-1)m+1, \dots, km\}\},  
\end{equation}  
where $\bm{C}_{\pi(i)}^{u}$ is the corresponding conformation of the $\pi(i)$-th repeating unit, $m = \frac{N_u}{K}$ is the size of the subset, and $\pi$ ensures a random sampling order.  

Here, the key modules used in this phase are introduced below.

\textbf{Multi-modal Repeating Unit Encoder.}
The multi-modal repeating unit encoder $\mathcal{M}$ comprises two important components: a 2D encoder $\mathcal{M}^{2d}$ designed for the polymer graph $\mathcal{G}$ and a 3D encoder $\mathcal{M}^{3d}$ designed for each repeating-unit conformation $\bm{C}_{i}^{u}$ within the polymer conformation.
In this context, the embedding extraction process can be expressed as follows:
\begin{eqnarray}\label{eq: multi_model_embedding}
\begin{aligned}  
    \bm{X}^{u} = \text{Concat}_1(\mathcal{M}^{2d}(\mathcal{G}), \text{Concat}_0(\{\mathcal{M}^{3d} (\bm{C}_{i}^{u})\}_{i=1}^{N_u})) 
    = \text{Concat}_1(\bm{X}^{2d},  \text{Concat}_0(\{\bm{x}^{3d}_i\}_{i=1}^{N_u})),
\end{aligned}  
\end{eqnarray} 
where $\bm{X}^{u} \in \mathbb{R}^{N_u \times D_{u}}$ is the multi-modal repeating unit embedding, 
$\bm{X}^{2d} \in \mathbb{R}^{N_u \times D_{2d}}$ is the 2D embedding of the polymer graph $\mathcal{G}$,
$\bm{x}^{3d}_i \in \mathbb{R}^{1 \times D_{3d}}$ is the 3D embedding of the $i$-th repeating-unit conformation $\bm{C}_{i}^{u}$, and $\text{Concat}_i(\cdot)$ represents the corresponding concatenation operator in the $i$-th dimension.
Here, we employ the encoder architecture from MolCLR~\cite{wang2022molecular} as the 2D encoder $\mathcal{M}^{2d}$ and the encoder architecture from Uni-Mol~\cite{zhou2023unimol} as the 3D encoder $\mathcal{M}^{3d}$. 

\textbf{Masked Autoregressive Modeling.}
To iteratively generate a subset of unknown repeating-unit conformations based on known/predicted repeating-unit conformations (i.e., Equation~\eqref{eq: phase_1_define}), we leverage the masked autoregressive modeling (MAR)~\cite{li2024autoregressive} in the latent space of the multi-modal repeating unit encoder to learn and model their complex dependencies.
\rre{During pretraining, given the multi-modal repeating unit embedding $\bm{X}^u \in \mathbb{R}^{N_u\times D_u}$, we firstly randomly select a subset of repeating units and then mask their corresponding embeddings}, i.e., 
\begin{equation}
\widetilde{\bm{X}}^{u} = \text{Concat}_0(\{\,\bigl(\mathbf{1}[\,i\notin\mathcal{S}_{\text{mask}}\,]\bigr)\,\bm{x}^{u}_{i}\,\}_{i=1}^{N_u}),
\end{equation}
where $\mathcal{S}_{\text{mask}}$ is the index set of those masked repeating units, and $\bm{x}_{i}^{u} \in \mathbb{R}^{1 \times D_u}$ is the $i$-th row of the multi-modal repeating unit embedding $\bm{X}^u$, corresponding to the $i$-th repeating unit.

Furthermore, as shown in Figure~\ref{fig: overview}c, we use $\widetilde{\bm{X}}^{u}$ as the input of the MAR encoder $\Phi$ and then use the MAR decoder $\Psi$ to obtain the corresponding decoded embedding $\bm{Z}^u$ of these repeating units, i.e., 
\begin{equation}\label{eq: mar}  
    \bm{Z}^u = \Psi(\Phi(\widetilde{\bm{X}}^{u})),  
\end{equation}  
where $\bm{Z}^u \in \mathbb{R}^{{N_u} \times D_u}$ is the decoded embedding of these repeating units, serving as the condition of the subsequent SE(3) diffusion designed for repeating units.
Here, we employ the standard Transformer architecture with bidirectional attention~\cite{devlin2019bert} as the MAR encoder $\Phi$ and MAR decoder $\Psi$.

\textbf{SE(3) Diffusion for Repeating Units.}
The goal of masked autoregressive modeling (MAR) is to reconstruct conformations of those masked repeating units by learning their probability distribution conditioned on the corresponding decoded embeddings.
As shown in Figure~\ref{fig: overview}b, repeating-unit conformations are just a specialized form of molecular conformations, characterized by the added complexity of interactions between repeating units. 
Given the remarkable success in generating molecular conformations, diffusion models are well-suited for learning this conditional probability distribution.
Following previous works~\cite{xu2022geodiff, yang2023diffusion, cao2024survey}, the corresponding loss can be formulated as a denoising criterion, i.e.,
\begin{eqnarray}
\begin{aligned}
    \mathcal{L}_\text{phase-1} = \mathbb{E}_{\varepsilon, t}\left[\left\| \varepsilon - \varepsilon_{\theta}(\bm{C}^u_t \vert t, \bm{z}^u) \right\|^2 \right],\quad
    \text{with~}\bm{C}^u_t = \sqrt{\bar{\alpha}_t}\bm{C}^u + \sqrt{1 - \bar{\alpha}_t}\varepsilon,
\end{aligned}
\end{eqnarray}  
where $\bm{C}^u \in \mathbb{R}^{(\frac{N}{N_u} + 2)\times 3}$ is the conformation of one masked repeating unit whose index is in $\mathcal{S}_{\text{mask}}$, \rre{$\bm{z}^u \in \mathbb{R}^{1 \times D_u}$ is the corresponding decoded embedding of this masked repeating unit (i.e., the corresponding row of $\bm{Z}^u$ obtained by Equation~\eqref{eq: mar})}, $\bar{\alpha}_t$ is the predefined noise schedule, $t$ is the time step of this predefined noise schedule, $\varepsilon$ is the noise sampled from the predefined prior distribution, and $\varepsilon_{\theta}$ is the parameterized denoising network for noise estimator.
Here, we employ the diffusion process defined on the torsion angle space to model this conditional probability distribution effectively, and adopt the corresponding torsional diffusion model architecture~\cite{jing2022torsional} as the denoising network $\varepsilon_{\theta}$.

\subsubsection{Phase-2: Orientation Transformation Generation}\label{sec: pretraining_phase2}
As expressed in Equation~\eqref{eq: define}, after training PolyConFM to generate repeating-unit conformations ${\mathcal{C}}^{u}$ conditioned on the polymer graph $\mathcal{G}$, i.e., $p({\mathcal{C}}^{u}|\mathcal{G})$, we still need to train it to generate their orientation transformations ${\mathcal{O}}$ conditioned on the polymer graph $\mathcal{G}$ and repeating-unit conformations ${\mathcal{C}}^{u}$, i.e.,  $p({\mathcal{O}}|\mathcal{G},{\mathcal{C}}^{u})$, thereby assembling them to recover the corresponding polymer conformation.
In particular, as shown in Figure~\ref{fig: overview}b, adjacent repeating-unit conformations are naturally overlapping at those key atoms (e.g., atom-1 of the current repeating unit aligns with atom-3 of the preceding repeating unit).
Therefore, for each repeating unit's orientation transformation, i.e., $\mathcal{O}_{i} = (\bm R_i, \bm t_i)$, we only need to consider the generation of its rotation transformation $\bm{R}_i$ as the corresponding translation transformation $\bm{t}_i$ can be directly derived by aligning the 3D coordinates of those overlapping atoms with the preceding repeating unit after applying rotation transformation $\bm{R}_i$. 

\textbf{SO(3) Diffusion for 3D Rotations.} According to the above analysis, the corresponding learning objective $p({\mathcal{O}}|\mathcal{G},{\mathcal{C}}^{u})$ in Equation~\eqref{eq: define} can be further simplified as $p(\mathbf{R}|\mathcal{G}, {\mathcal{C}}^{u})$, where $\mathbf{R}=[\bm{R}_i]\in\mathbb{R}^{N_u\times 3\times 3}$ is rotation transformations of all repeating units within this linear polymer.
As illustrated in Figure~\ref{fig: overview}c, an SO(3) diffusion model designed for 3D rotations is thus utilized to learn this conditional probability distribution associated with $\mathbf{R}$, i.e.,
\begin{equation}\label{eq: phase_2} 
\widehat{{\mathbf{R}}}^{(0)} = \varphi({\mathcal{O}}^{(t)}, t, \bm{E}^u), ~\text{with~}   
{{\mathcal{O}}}^{(t)} = ({\mathbf{R}}^{(t)}, {\mathbf{T}}^{(t)}).
\end{equation}
where $\varphi$ denotes the denoising network, whose architecture is the same as the one used in~\cite{yim2023se}. 
\rre{$\bm{E}^u \in \mathbb{R}^{N_u \times D_u}$ is the output of the MAR encoder, representing the condition about \(\mathcal{G}\) and \({\mathcal{C}}^{u}\) in Equation~\eqref{eq: define}. 
$\mathbf{R}^{(t)}=[\bm{R}_i^{(t)}]\in \mathbb{R}^{N_u \times 3 \times 3}$ is the rotation transformations obtained at the time step $t$ during the forward diffusion process defined on SO(3)$^{N_u}$. 
${\mathbf{T}}^{(t)}=[\bm{t}_i^{(t)}]\in \mathbb{R}^{N_u \times 3}$ is the translation transformations calculated by aligning the 3D coordinates of those overlapping atoms (i.e., Equation~\eqref{eq: trans}) after applying the corresponding rotation transformations $\mathbf{R}^{(t)}$ to all repeating units within this linear polymer.}
Accordingly, we can learn this SO(3) diffusion model by minimizing the following loss function:
\begin{equation}
  \mathcal{L}_\text{phase-2} = \frac{1}{N_u} \sideset{}{_{i=1}^{N_u}}\sum \| \widehat{\bm R}_i^{(0)} - {\bm R}_i \|^2,
\end{equation}  
where ${\bm R}_i \in \mathbb{R}^{3 \times 3}$ is the ground-truth rotation transformation of the $i$-th repeating unit.

\subsection{Finetuning PolyConFM for Downstream Tasks}\label{sec: finetuning}
After conformation-centric generative pretraining, PolyConFM not only captures complex dependencies among repeating units for global structure modeling but also simultaneously unlocks its conformation generation capability for diverse downstream tasks.
Here, we further finetune it for two core downstream tasks: polymer property prediction and polymer design, which together encompass the principal use cases in polymer science.
In particular, as illustrated in Figure~\ref{fig: overview}a, PolyConFM employs polymer conformation generated by itself as input to provide global structural information for downstream tasks, while the polymer modeling module can also assist the polymer design module via virtual screening to prioritize suitable candidates for wet-lab validation, thereby \rre{positioning PolyConFM as a powerful backbone that seamlessly bridges structure, property, and design of linear polymers}.

Therefore, in this subsection, we will first illustrate how to leverage the pretrained PolyConFM to generate polymer conformations, which serve as input for downstream tasks, before moving on to its finetuning for downstream property prediction and design tasks.

\textbf{Polymer Conformation Generation.}
As analyzed in Section~\ref{sec: pretraining}, conformation-centric generative pretraining has enabled PolyConFM to learn a generative model $p(\mathcal{C}|\mathcal{G})$, which models the empirical distribution of polymer conformation $\mathcal{C}$ conditioned on the corresponding polymer graph $\mathcal{G}$. 
As illustrated in Figure~\ref{fig: overview}d, \rre{for the new input data (i.e., the corresponding 1D polymer SMILES and 2D polymer graph of one new linear polymer), we can directly run inference with the pretrained PolyConFM to generate repeating-unit conformations $\{\widehat{\bm{C}}_{i}^{u}\}_{i=1}^{N_u}$ and then generate their rotation transformations $\{\widehat{\bm R}_{i}\}_{i=1}^{N_u}$.}
In this context, we only need to assemble these generated repeating-unit conformations into the complete polymer conformation and then add it to the input to derive polymer embedding for downstream tasks.
\re{Notably, this architectural dependency indicates that the conformation generation capability of PolyConFM is exclusively unlocked by its pretraining process. 
Without pretraining, PolyConFM would have no generation capabilities, rendering it fundamentally unable to provide the required inputs for various downstream tasks. 
Therefore, pretraining is not merely an optional performance booster, but the fundamental prerequisite that drives the entire PolyConFM pipeline.}

In particular, as mentioned in Section~\ref{sec: rep}, the orientation transformation is relative to the standard coordinate system, meaning that the generated rotation transformations $\{\widehat{\bm R}_{i}\}_{i=1}^{N_u}$ are also relative to the standard coordinate system.
Therefore, we first transform each generated repeating-unit conformation $\widehat{\bm{C}}_{i}^{u}$ back to the standard coordinate system, i.e.,
\begin{equation}  
\widehat{\bm{C}}_{i}^{u,\text{std}} = (\mathcal{O}_{i}^{c})^{-1} \cdot \widehat{\bm{C}}_{i}^{u} = (\widehat{\bm{C}}_{i}^{u} - \bm t_i^c) \cdot (\bm R_i^c)^{-1}, 
\end{equation}  
where $\mathcal{O}_{i}^{c} = (\bm R_i^c, \bm t_i^c)$ is the current orientation transformation calculated based on the 3D coordinates of those key atoms within $\widehat{\bm{C}}_{i}^{u}$, and the corresponding calculation process is illustrated in Figure~\ref{fig: overview}b.

Then we apply the generated rotation transformation $\widehat{\bm R}_{i}$ to the corresponding $\widehat{\bm{C}}_{i}^{u,\text{std}}$, i.e.,
\begin{equation}  
\widehat{\bm{C}}_{i}^{u,\text{rot}} = \widehat{\bm{C}}_{i}^{u,\text{std}} \cdot \widehat{\bm R}_{i}.  
\end{equation}  

Furthermore, as mentioned in Section~\ref{sec: pretraining_phase2}, the corresponding translation transformation of $\widehat{\bm{C}}_{i}^{u,\text{rot}}$ can be directly derived through aligning the 3D coordinates of those overlapping atoms, i.e.,
\begin{equation}\label{eq: trans}  
\hat{\bm t}_{i} =   
\begin{cases}   
\bm{0}, & \text{if } i = 1, \\  
\sideset{}{_{j=1}^{i-1}}\sum (\hat{\bm c}_{j, 3}^{u,\text{rot}} - \hat{\bm c}_{j+1, 1}^{u,\text{rot}}), & \text{if } i > 1.  
\end{cases}  
\end{equation}  
where $\hat{\bm t}_{i} \in \mathbb{R}^{3}$ represents the corresponding translation transformation of $\widehat{\bm{C}}_{i}^{u,\text{rot}}$, $\hat{\bm c}_{j, 3}^{u,\text{rot}} \in \mathbb{R}^{3}$ represents the 3D coordinate of atom-3 in $\widehat{\bm{C}}_{j}^{u,\text{rot}}$, and $\hat{\bm c}_{j+1, 1}^{u,\text{rot}} \in \mathbb{R}^{3}$ represents the 3D coordinate of atom-1 in $\widehat{\bm{C}}_{j+1}^{u,\text{rot}}$.

Finally, we obtain the complete polymer conformation $\widehat{\bm{C}} \in \mathbb{R}^{N \times 3}$ through applying the corresponding translation transformation $\hat{\bm t}_{i}$ to $\widehat{\bm{C}}_{i}^{u,\text{rot}}$, i.e.,
\begin{eqnarray}  
\begin{aligned}
    \widehat{\bm{C}}_{i}^{u,\text{final}} = \widehat{\bm{C}}_{i}^{u,\text{rot}} + \hat{\bm t}_{i},\quad  
    \widehat{\bm{C}} = \{\widehat{\bm{C}}_{i}^{u,\text{final}} \setminus \{\hat{\bm{c}}_{i, 1}^{u,\text{final}}, \hat{\bm{c}}_{i, 4}^{u,\text{final}}\} \}_{i=1}^{N_u},  
\end{aligned}
\end{eqnarray}
where $\widehat{\bm{C}} \in \mathbb{R}^{N\times3}$ is the complete polymer conformation, and $\setminus$ represents the set difference operation for removing those overlapping atoms (i.e., atom-1 and atom-4 within each repeating unit).

Additionally, as illustrated in Figure~\ref{fig: overview}d, after assembling these generated repeating-unit conformations into the complete polymer conformation, we add this generated polymer conformation $\widehat{\bm{C}}$ to the input and further derive the corresponding global polymer embedding for downstream tasks, i.e., 
\begin{equation}\label{eq: pooling}  
  \bm{e}_\text{global} = \frac{1}{N_u}\,\mathbf{1}_{N_u}^{\top} \bm{E}^u
\end{equation}  
where \rre{$N_u$ is the number of repeating units within this linear polymer}, $\bm{E}^u \in \mathbb{R}^{N_u \times D_u}$ is the output of the MAR encoder, and we use mean pooling to obtain the global polymer embedding $\bm{e}_\text{global}$.

\textbf{Polymer Property Prediction.}
Polymer property prediction is a typical representation-learning task that aims to learn informative polymer embeddings and map them to the corresponding property values through supervised learning.
As illustrated in Section~\ref{sec: pretraining}, conformation-centric generative pretraining has enabled PolyConFM to accurately model polymer structures, yielding high-quality and structure-aware polymer embeddings.
Therefore, we directly employ a simple multi-layer perceptron (MLP) layer as the polymer modeling module, and finetune PolyConFM with it to perform this task.

In particular, as illustrated in Figure~\ref{fig: overview}d, we first employ PolyConFM to generate each polymer's conformation, from which we then derive the corresponding global polymer embedding to serve as input for the polymer modeling module.
During finetuning, since public polymer property datasets are all formulated as regression, we learn the model by minimizing the mean squared error (MSE) loss, i.e.,
\begin{equation}
  \mathcal{L}_\text{pred} = (\text{MLP}(\bm{e}_\text{global}) - y)^2
\end{equation}  
where $\bm{e}_\text{global} \in \mathbb{R}^{1 \times D_u}$ is the global polymer embedding obtained through Equation~\eqref{eq: pooling}, and \rre{$y \in \mathbb{R}$ is the corresponding ground-truth property value of this linear polymer}.

\textbf{Polymer Design.}
Polymer design is a typical conditional generation task that focuses on generating polymers satisfying specific conditions, such as desired properties and structural requirements.
As illustrated in Figure~\ref{fig: overview}a, we leverage the global polymer embedding of the reference polymer, learned by PolyConFM,  as an additional conditioning signal to better guide polymer design, thereby thoroughly validating the effectiveness of PolyConFM in modeling polymer structures.
Meanwhile, due to the vast chemical space and practical manufacturing constraints, we further simplify this task into designing suitable 2D graph structures.
Here, we employ a graph-based diffusion model as the polymer design module and finetune it by minimizing the following negative log-likelihood function, i.e., 
\begin{equation}
    \mathcal{L}_\text{design}= \mathbb{E}_{q(G^0)} \mathbb{E}_{q(G^t|G^0)} \left[- \mathbb{E}_{\mathbf{x} \in G^0} \log p_\theta \left( \mathbf{x} \mid G^t , \bm{e}_\text{global},  \mathcal{C}\right) \right]
\end{equation}
where $p_\theta$ is the denoising network for conditional generation, whose architecture is the same as the one used in~\cite{liu2024graph}, $G^t$ is obtained at the time step $t$ during the forward diffusion process defined on the space of 2D graphs, $\bm{e}_\text{global} \in \mathbb{R}^{1 \times D_u}$ is the global polymer embedding obtained through Equation~\eqref{eq: pooling} and $\mathcal{C}={\{c_1, c_2, \dots, c_M\}}$ represents the conditioning set.

\subsection{Experimental Setup}\label{sec: exp_setup}
\subsubsection{Datasets}\label{sec: exp_setup_datasets}
To mitigate the severe scarcity of polymer conformation datasets, a major factor that limits the development of this important field, we devote considerable time and resources to constructing a dataset of over 50,000 linear polymers with conformations (about 2,000 atoms per conformation) through molecular dynamics simulations.
Here, under the guidance of experienced domain experts, we \re{design our molecular dynamics simulations using standard pipelines widely adopted in previous works~\cite{afzal2020high} and assess equilibration
and the charge-model sensitivity of key conformational statistics, thereby ensuring the reliability of this dataset.}
In particular, the initial polymer structures of molecular dynamics simulations are generated using RDKit~\cite{landrum2013rdkit} and AmberTools~\cite{salomon2013overview}, followed by energy minimization, \re{equilibration and sampling} in the canonical ensemble (NVT) with a 1 fs time step for a total duration of 5 ns (5,000,000 steps). 
Besides, each simulation trajectory is obtained using the General AMBER Force Field with the GROMACS package~\cite{van2005gromacs}. 
\re{To ensure the reference conformations accurately represent the equilibrium state, we perform sampling from the final 1 ns of each MD trajectory, during which the system energy has fully converged. \rre{Specifically, five conformations are uniformly sampled from this stable interval to serve as the reference ensemble (also known as “ground truth”) for each polymer, thereby providing the sample-based representation of equilibrium conformational statistics under the employed MD protocol.}}
With this conformation dataset, we not only enable conformation-centric pretraining but also facilitate future research.
\rre{Notably, for computational efficiency, the pretraining conformations are obtained from single-chain vacuum simulations, without solvent or multi-chain effects. These simulations are designed to provide global geometric and topological priors under basic physical constraints during pretraining, rather than to reproduce condensed-phase properties quantitatively. 
Nevertheless, the ablation experiments in Section \ref{sec: ablation} demonstrate that even such basic global structural priors indeed substantially improve the downstream performance of PolyConFM.}

Then, for the downstream polymer property prediction task, we utilize \re{diverse polymer property datasets (denoted as Egc, Egb, Eea, Ei, Xc, EPS, Nc, and Eat, respectively) provided in~\cite{kuenneth2021polymer}}, consistent with previous works~\cite{wang2024mmpolymer}.
In particular, these datasets are derived from density functional theory (DFT) calculations and span typical properties, thus ensuring a reliable and comprehensive assessment.

Finally, for the downstream polymer design task, we utilize the polymer design dataset provided in~\cite{liu2024graph}, after removing invalid polymers (e.g., lacking polymerization sites) to ensure data validity and reliable evaluation.
In particular, this dataset's condition set balances performance (three gas permeabilities: O2Perm, CO2Perm, and N2Perm) with practical feasibility (synthesizability), ensuring a realistic and application-oriented experimental setting.

More information about the above datasets is provided in Supplementary Information~\ref{SI-sec: datasets}.

\subsubsection{Baselines}\label{sec: exp_setup_baselines}
To demonstrate PolyConFM’s superior performance across diverse polymer-related tasks, we compare it with various representative task-specific methods.

For the polymer conformation generation task, in light of the lack of specialized polymer conformation generation methods, we have to utilize various representative molecular conformation generation methods, including GeoDiff~\cite{xu2022geodiff}, TorsionalDiff~\cite{jing2022torsional}, MCF~\cite{wang2024swallowing}, and ET-Flow~\cite{hassan2024flow} as our baselines.
Here, we adapt these baselines for polymer conformation generation by modeling polymers as large molecules with many more atoms. 
In particular, since TorsionalDiff requires an initial polymer structure as input, which cannot be directly generated like small molecules using RDKit~\cite{landrum2013rdkit}, we have to employ the initial polymer structure of the corresponding simulation trajectory as its input, thus unintentionally giving TorsionalDiff a biased advantage over other methods.

For the downstream polymer property prediction task, we utilize various state-of-the-art methods, including polyBERT~\cite{kuenneth2023polybert}, Transpolymer~\cite{xu2023transpolymer}, and MMPolymer~\cite{wang2024mmpolymer} as our baselines.
In particular, these baselines are all polymer pretraining methods designed for property prediction, making them well-suited to demonstrate the superiority of our conformation-centric generative pretraining.
Meanwhile, various representative molecular pretraining methods, including MolCLR~\cite{wang2022molecular}, 3D Infomax~\cite{stark20223d}, and Uni-Mol~\cite{zhou2023unimol}, are also utilized as our baselines to reveal the limitations of directly transplanting molecular methods to polymer-specific tasks, thereby emphasizing the critical need to develop tailored polymer methods that can accommodate their unique characteristics.

For the downstream polymer design task, we utilize the latest GraphDiT~\cite{liu2024graph}, along with its various baselines, including MolGPT~\cite{bagal2021molgpt}, GraphGA~\cite{jensen2019graph}, DiGress~\cite{vignac2023digress}, GDSS~\cite{jo2022score}, and MOOD~\cite{lee2023exploring}, as our baselines.
In particular, these baselines can be divided into two categories: 1) optimization methods that 
treat the conditioning set as a combined objective and minimize the corresponding summed
normalized error; 2) generative methods that directly combine various generative models (e.g., diffusion models) with either predictor-guided or predictor-free conditional strategies.

Notably, \rre{every method (including PolyConFM and all compared baselines) in our work is trained and evaluated using the same data splits (i.e, the identical training, validation, and test sets), thereby ensuring that all comparisons are perfectly aligned and strictly fair.}
More information about the above baselines is provided in Supplementary Information~\ref{SI-sec: baselines}.

\subsubsection{Metrics}\label{sec: exp_setup_metrics}
To ensure fair and transparent comparisons across all tasks, we employ established evaluation metrics from previous works, with minor but essential adjustments tailored to polymers.

For the polymer conformation generation task, where no established polymer-specific metrics exist, we design our evaluation metrics that consider both structure-matching and energy-matching.
Here, we denote the sets of generated and reference conformations \re{corresponding to the same polymer} as $S_g$ and $S_r$, respectively.
In this context, the corresponding structure-matching metrics are defined as follows:  
\begin{equation}\label{eq: structure}
\begin{aligned}
\text{S-MAT-R} = \frac{1}{|S_r|} \sideset{}{_{\bm C \in S_r}}\sum \min_{\widehat{\bm{C}} \in S_g} \text{RMSD}(\bm{C}, \widehat{\bm{C}}), \quad 
\text{S-MAT-P} = \frac{1}{|S_g|} \sideset{}{_{\widehat{\bm{C}} \in S_g}}\sum \min_{\bm{C} \in S_r} \text{RMSD}(\bm{C}, \widehat{\bm{C}}), 
\end{aligned}  
\end{equation}  
where the generated conformation $\widehat{\bm{C}}$ and reference conformation $\bm C $ \re{have been optimally \rre{aligned using the Kabsch algorithm in PyMOL~\cite{delano2002pymol} before computing root-mean-square deviation (RMSD), thereby minimizing structural discrepancy arising from rigid-body rotation and translation.}
In addition, $|S_r|$ and $|S_g|$ denote the total number of conformations in the reference and generated sets, respectively.
Specifically, S-MAT-R (Recall-based) measures the coverage of the reference conformational space by averaging the distance from each reference conformation to its nearest neighbor in the generated set, while S-MAT-P (Precision-based) evaluates the fidelity of the generated samples by averaging the distance from each generated conformation to the closest reference one.
In fact, both S-MAT-R and S-MAT-P have already been widely adopted in the molecular conformation generation field~\cite{xulearning, xu2021end, shi2021learning, luo2021predicting} for quantifying structural discrepancy.
Here, they are also reported in \text{\AA}ngstr\"oms (\text{\AA}), consistent with the unit of atomic coordinates.
Since RMSD itself is an average over the number of atoms, both S-MAT-R and S-MAT-P are not further normalized, with a theoretical range of $[0, +\infty)$. 
Besides, lower values for both metrics indicate superior performance in capturing the target conformational ensemble.}
Meanwhile, the corresponding energy-matching metrics are defined similarly by replacing the structural difference $\text{RMSD}(\bm{C}, \widehat{\bm{C}})$ in Eq.~\eqref{eq: structure} with potential energy difference $| E(\bm{C}) - E(\widehat{\bm{C}}) |$. 
Please note that the Coverage metric, which relies on a fixed RMSD threshold $\delta$ for structural comparison in the small molecule domain~\cite{xu2022geodiff}, is unsuitable for polymer conformation generation as polymers exhibit a much larger conformational space with significant diversity arising from their chain length, flexibility, and repeating units~\cite{chen2021polymer}. Thus, we exclude it from our evaluation metrics but still report the corresponding performance under this metric in Supplementary Information~\ref{SI-sec: exp_conf} for reference.

For the downstream polymer property prediction task, where all public datasets are formulated as regression, we choose the widely adopted root mean squared error (RMSE) and coefficient of determination ($R^2$) as our evaluation metrics, thereby providing complementary insights into model performance while ensuring alignment with previous works~\cite{xu2023transpolymer, wang2024mmpolymer}. \rre{Specifically,
\begin{equation}\label{eq: rmse_r2}
    \text{RMSE} = \sqrt{\frac{\sum_{i=1}^{n} (y_i - \hat{y}_i)^2}{n}}, \quad R^2 = 1 - \frac{\sum_{i=1}^{n} (y_i - \hat{y}_i)^2}{\sum_{i=1}^{n} (y_i - \bar{y})^2} = 1 - \frac{\text{MSE}}{Var(y)} = 1 - \frac{\text{(RMSE)}^2}{Var(y)}
    \end{equation}
where $n$ is the total number of samples in the test set; $y_i$ and $\hat{y}_i$ denote the ground-truth and predicted values for the $i$-th sample, respectively; $\bar{y}$ is the mean of the ground-truth values; and $Var(y)$ represents the variance of the ground-truth labels, which scales the $R^2$ metric.
Notably, since $R^2$ is jointly determined by both the RMSE and  $Var(y)$ of the corresponding test set, comparing $R^2$ across different splitting strategies (i.e., test sets with different $Var(y)$) is statistically meaningless.}
    
For the downstream polymer design task, we adopt those well-designed evaluation metrics already established in previous works~\cite{liu2024graph}, including (1) coverage of heavy atom types relative to the reference set (Coverage); (2) internal diversity among generated samples (Diversity); (3) fragment‑based similarity to the reference set (Similarity); (4) Fréchet ChemNet Distance to the reference distribution (Distance); together with MAE (i.e., mean absolute error) between the generated and conditioned (5) synthetic accessibility score (Synth.) and (6)$\sim$(8) MAE for the numerical conditions (Property), thereby providing a balanced evaluation that jointly considers distribution learning and condition control for both distributional fidelity and condition satisfaction.
In addition, following previous works~\cite{gao2022sample, liu2024graph}, we also employ random forests trained on task-related polymers as the evaluation oracle. 

\rre{Notably, all reported metrics throughout the whole work are based on the test-set
performance.}

\section*{Data Availability}
All datasets used in this work are available at \url{https://zenodo.org/records/17568899}. 
\re{Detailed information regarding their references and links is listed below:

\begin{itemize}
    \item Conformation Dataset: Constructed here and available at \url{https://zenodo.org/records/17568899}.
    
    \item Property Dataset: Obtained from~\cite{kuenneth2021polymer} and available at \url{https://khazana.gatech.edu/dataset}.
    
    \item Design Dataset: Obtained from~\cite{liu2024graph} and available at \url{https://github.com/liugangcode/Graph-DiT}.
\end{itemize}
}

\section*{Code Availability} 
Codes are available at \url{https://github.com/FanmengWang/PolyConFM}, and the corresponding model weights are available at \url{https://zenodo.org/records/17577742}.
\re{Regarding computational resources, the conformation-centric generative pretraining is conducted on eight A100 (80GB) GPUs, taking approximately 120 wall-clock hours.
Furthermore, finetuning for downstream tasks is performed on a single A100 (80GB) GPU, requiring no more than 20 wall-clock hours depending on the specific dataset size.
}

\section*{Acknowledgements}
This work was supported by the National Natural Science Foundation of China (92270110), the Fundamental Research Funds for the Central Universities, and the Research Funds of Renmin University of China. 
We also acknowledge the support provided by the fund for building world-class universities (disciplines) of Renmin University of China and by the funds from Beijing Key Laboratory of Research on Large Models and Intelligent Governance, Engineering Research Center of Next-Generation Intelligent Search and Recommendation, Ministry of Education, and from Intelligent Social Governance Interdisciplinary Platform, Major Innovation \& Planning Interdisciplinary Platform for the ``Double-First Class'' Initiative, Renmin University of China.

\section*{Author Contributions}
H.X., Z.G., and Q.O. conceived and supervised the project.
F.W. implemented the PolyConFM methodology and drafted the initial manuscript.
F.W., R.W., and S.M. performed the experiments and analyzed the results.
R.W. guided and designed the reliability validation for the polymer simulations.
W.G., H.W., and Q.O. provided scientific guidance and domain expertise for polymer modeling and simulations.
All authors participated in the discussion and contributed to the final manuscript.

\section*{Competing Interests}
The authors declare no competing interests.

\bibliography{reference}

\newpage
\begin{appendix}

\captionsetup[figure]{labelformat=empty}
\captionsetup[table]{labelformat=empty}
\renewcommand{\thefigure}{Supplementary Figure \arabic{figure}}
\renewcommand{\thetable}{Supplementary Table \arabic{table}}
\setcounter{figure}{0}
\setcounter{table}{0}

\begin{center}
    \section*{\Large Supplementary Information}
\end{center}
\vspace{2pt}

\section{Details on Datasets}\label{SI-sec: datasets}
\subsection{Polymer Conformation Dataset}
The vast chemical space of polymers makes conformational computation expensive, which in turn leads to the severe scarcity of polymer conformation datasets. 
While some datasets~\cite{feng2023may} indeed exist, they are limited to polymers with less than six repeating units, which is far from realistic scenarios where polymers comprise thousands of atoms.
In this context, under the guidance of experienced domain experts, we devote considerable time and resources to constructing a dataset of over 50,000 linear polymers with conformations (approximately 2,000 atoms per conformation) through molecular dynamics (MD) simulations, which not only enables conformation-centric generative pretraining of PolyConFM but also accelerates subsequent research in this important field.

\textbf{Construction Pipeline.}
Firstly, polymers are constructed and prepared for MD simulations by combining various molecular modeling tools and custom Python scripts.
In particular, monomer conformations are generated by RDKit~\cite{landrum2013rdkit} based on the corresponding SMILES strings and processed to define chain termini and repeating units. 
For each linear polymer, we define a polymeric unit template (PUT) along with the head polymeric terminus (HPT) and tail polymeric terminus (TPT) to represent it. 
RDKit is utilized to analyze atom connectivity and identify those key atoms for polymerization. 
Hydrogen atoms at the polymerization sites are omitted, and chain termini are explicitly parameterized. 
Topology and force field parameters files for the corresponding monomer are generated using the Antechamber and prepgen tools~\cite{salomon2013overview} under the General AMBER Force Field (GAFF)~\cite{wang2004development}.
TLeap is then used to build AMBER-compatible topology and coordinate files for both individual monomers and polymer chains. 
Please note that this MD pipeline is intended to generate a large-scale, physically plausible conformation dataset for pretraining, rather than to establish a universally accurate polymer force-field benchmark. 
Given the challenges and chemistry dependence of polymer parameterization workflows~\cite{turney2025atomistic,gartner2019modeling}, AM1-BCC charges are adopted here as a practical and scalable choice that balances computational efficiency and consistency across diverse polymers~\cite{jakalian2002fast,he2020fast}.
Polymer chains with a degree of polymerization $N_u$ are constructed by repeating the PUT ${N_u}-2$ times and capping the chain with HPT and TPT at the termini. 
Chain lengths are chosen to achieve approximately 2,000 atoms per system. 
AMBER topology and coordinate files are converted to GROMACS-compatible formats using ACPYPE, which are required for subsequent MD simulations. 

Furthermore, the optimization and MD simulations are performed with GROMACS~\cite{van2005gromacs}, which has long been used for polymer simulations~\cite{grunewald2022polyply}.
\re{For each system, a single polymer chain is placed in a vacuum box ($5.0 \times 5.0 \times 5.0$ nm$^3$) with periodic boundary conditions applied in all three directions. 
The box size is chosen as a pragmatic trade-off between computational cost and robustness, enabling high-throughput MD generation of the dataset under a uniform simulation protocol.
All MD simulations are performed using a leap-frog integrator with a time step of 1 fs. The Van der Waals interactions are described by the Lennard-Jones potential with a cutoff distance of 1.2 nm. Electrostatic interactions are treated by the particle mesh Ewald method, with a real-space cutoff of 1.2 nm and an Ewald error tolerance of $1 \times 10^{-5}$. In addition, the LINCS algorithm is employed to constrain all bonds involving hydrogen atoms. 
The initial polymer conformation is relaxed by energy minimization using the steepest descent method, followed by a 5 ns NVT simulation at 298 K with temperature maintained by the V-rescale thermostat. The first 4 ns is used for equilibration, and five conformations are extracted at 200 ps intervals from each polymer during the final 1 ns for dataset construction.
}
Please note that while MD simulations may have some limitations in accurately predicting properties, they are well-suited for efficiently exploring the conformational space of large polymer systems, thereby ensuring the generated conformations are realistic representations of polymer structures without compromising our focus.

\textbf{Dataset Validation.}
To assess the reliability of the polymer conformation dataset, we randomly select eight representative polymers from the full collection and evaluate (\romannumeral1) conformational equilibration and (\romannumeral2) the sensitivity of key conformational statistics to the charge-assignment scheme. The selected polymers span diverse repeat-unit chemistries, while their molecular structures and corresponding P-SMILES strings are listed in~\ref{tab: selected_polymer}.

\begin{table}[tbp]
  \centering
  \setlength{\tabcolsep}{5.5pt}
  \begin{footnotesize}
  \caption{\textbf{\thetable:} \re{Representative polymers used for MD dataset validation.}}
    \begin{tabular}{lccc}
        \toprule
        Polymer Label & P-SMILES & 2D Chemical Structure & Polymerization Degree ($N_u$) \\
        \midrule
        Poly1  &  [*]CC(C=C1)=CC=C1CC(=S)[*] & \raisebox{-0.48\height}{\includegraphics[width=0.12\textwidth, trim=50 50 50 100,clip]{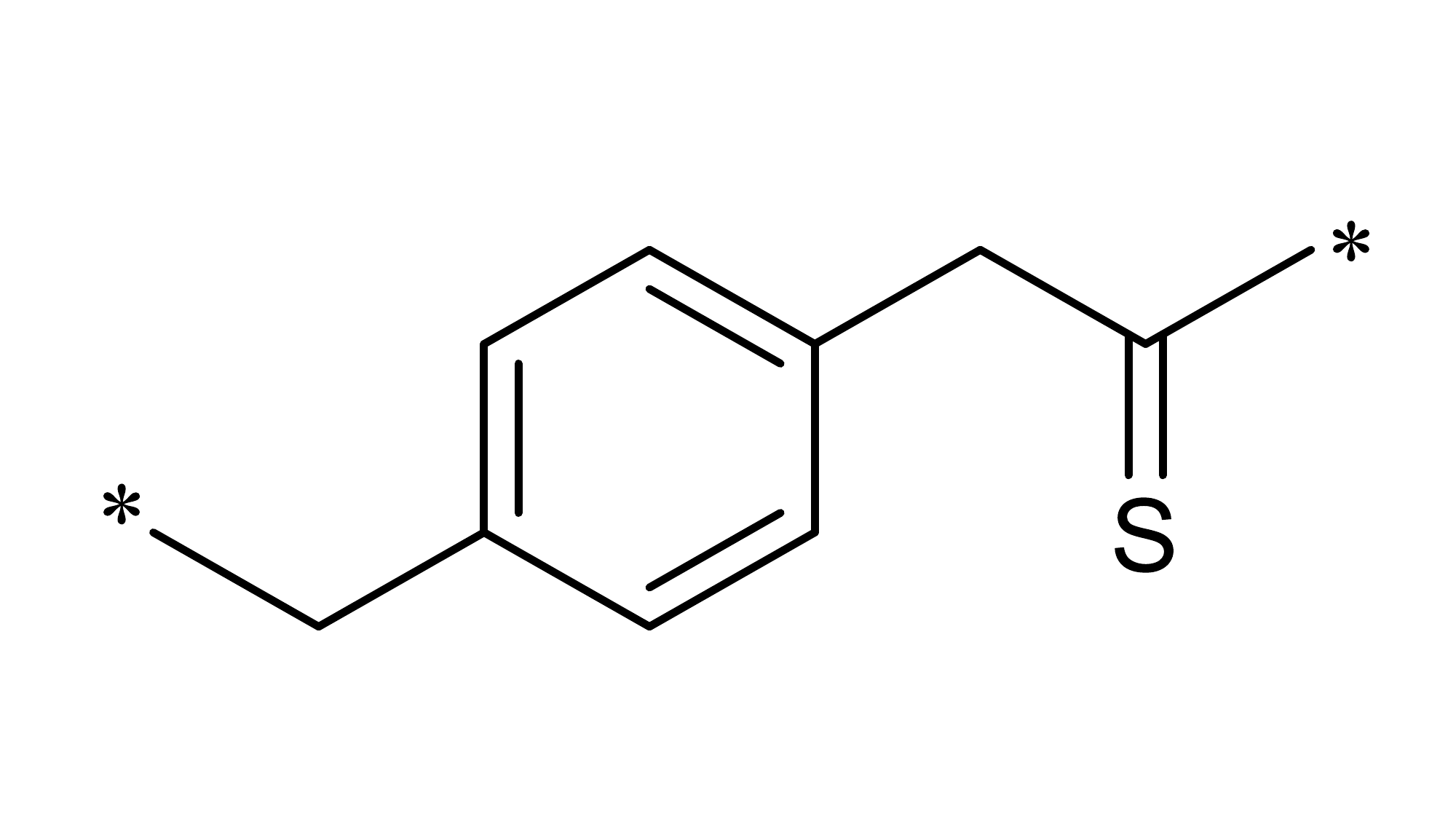}} & 76 \\
        Poly2  &  [*]OC(C)C([*])C & \raisebox{-0.48\height}{\includegraphics[width=0.12\textwidth, trim=80 50 100 100,clip]{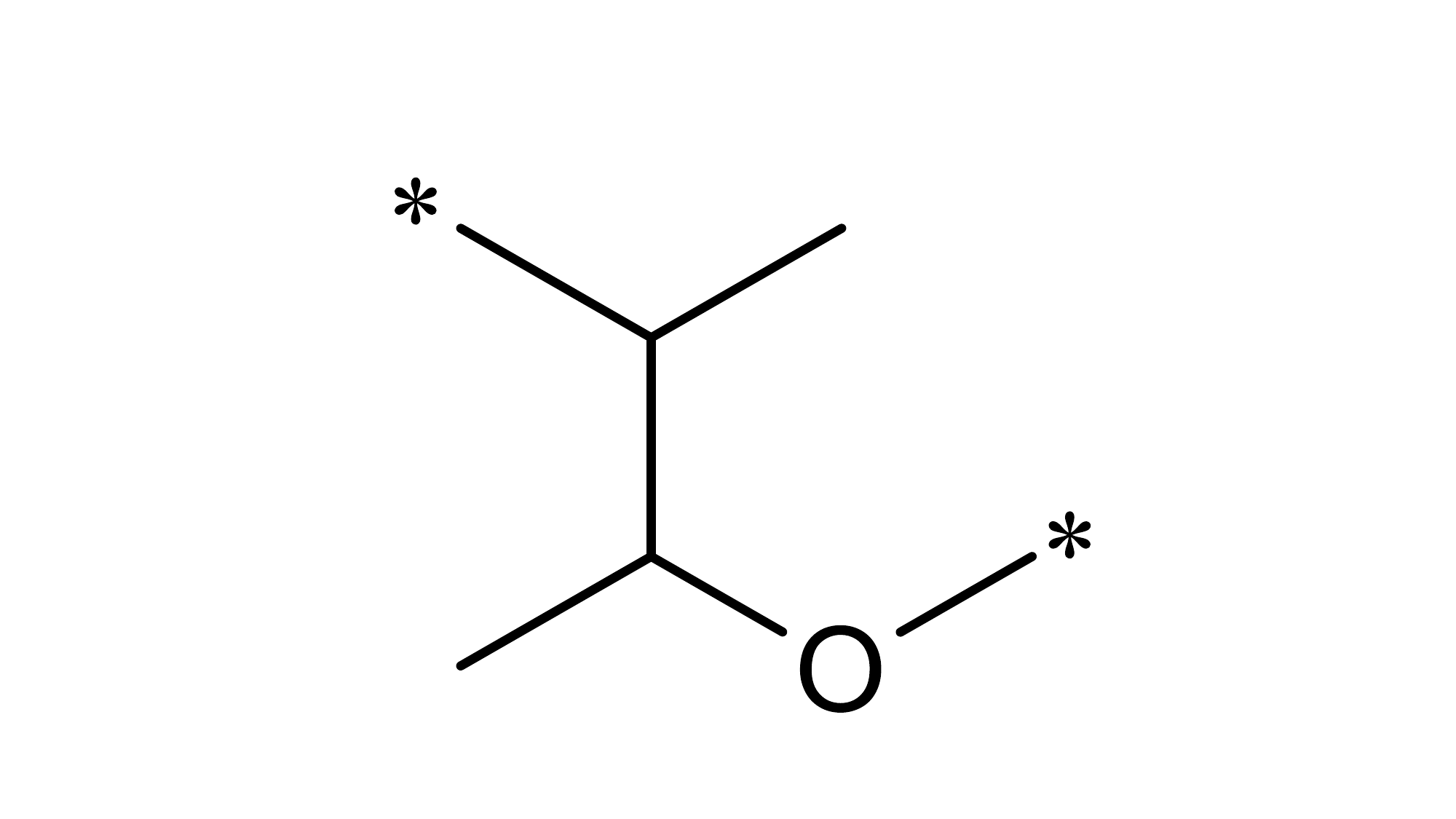}} & 105 \\
        Poly3 &  \makecell[l]{[*]CCCCCCCCCCC(=O)NCCCC\\
        COCCCCCNC(=O)CCCO[*]} & \raisebox{-0.48\height}{\includegraphics[width=0.3\textwidth, trim=0 170 0 170,clip]{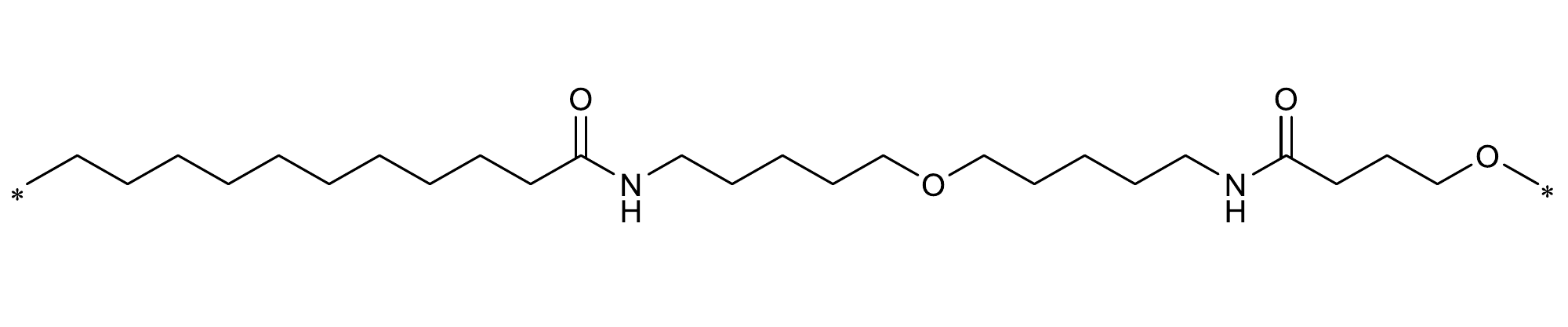}} & 23 \\
        Poly4  &  [*]CC1CC([*])OC(C)O1 & \raisebox{-0.48\height}{\includegraphics[width=0.15\textwidth, trim=80 60 100 130,clip]{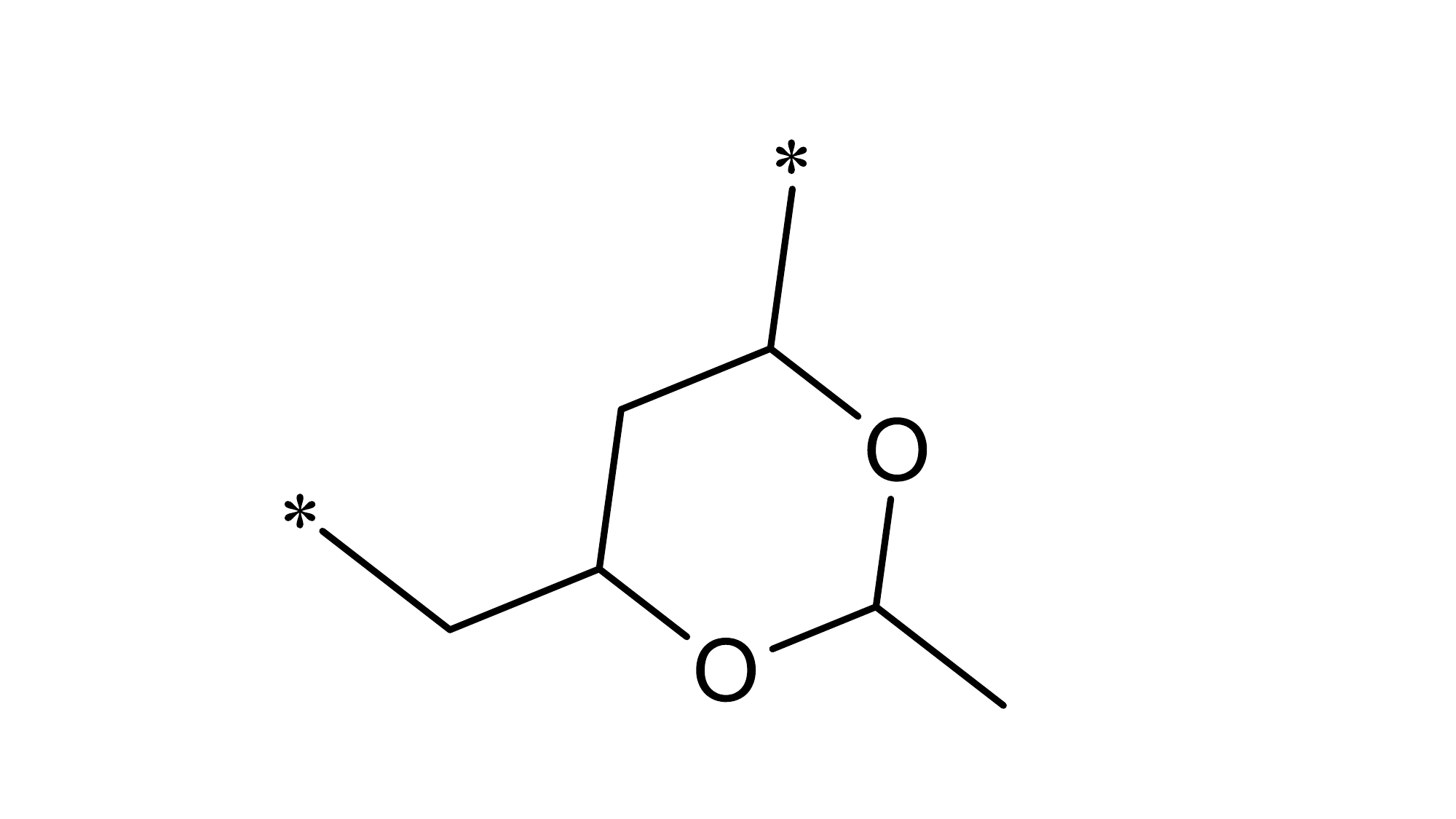}} & 76 \\
        Poly5  &  \makecell[l]{[*]CC1CCC(CNC(=O)CCCCC\\
        CCC(=O)N[*])CC1} & \raisebox{-0.48\height}{\includegraphics[width=0.24\textwidth, trim=0 150 0 150,clip]{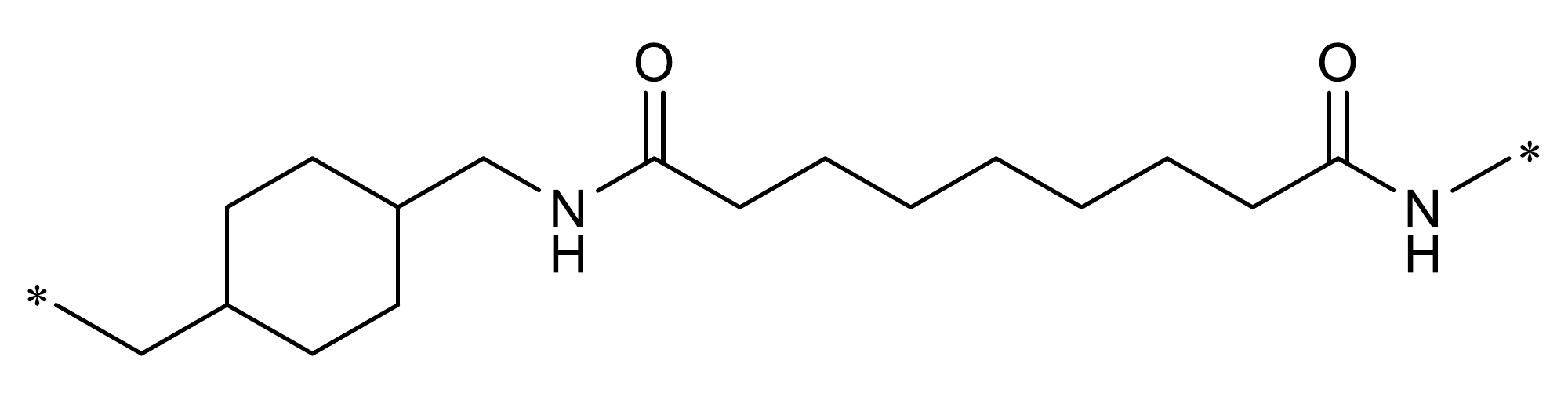}} & 34 \\
        Poly6  &  [*]C=C([*])CO & \raisebox{-0.48\height}{\includegraphics[width=0.12\textwidth, trim=80 50 100 100,clip]{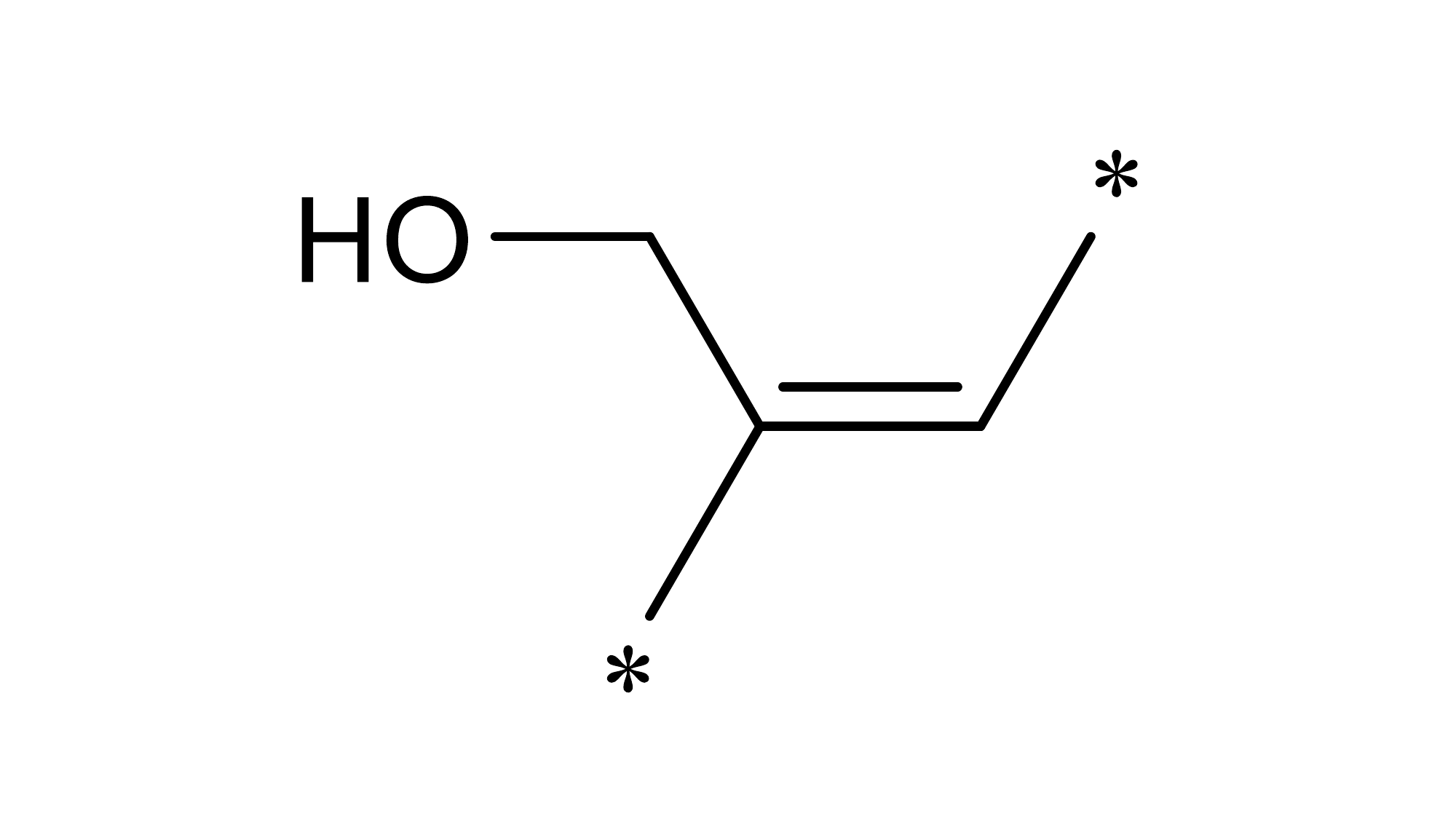}} & 125 \\
        Poly7  & [*]CC([*])N1CCCC1=S & \raisebox{-0.48\height}{\includegraphics[width=0.12\textwidth, trim=80 50 100 80,clip]{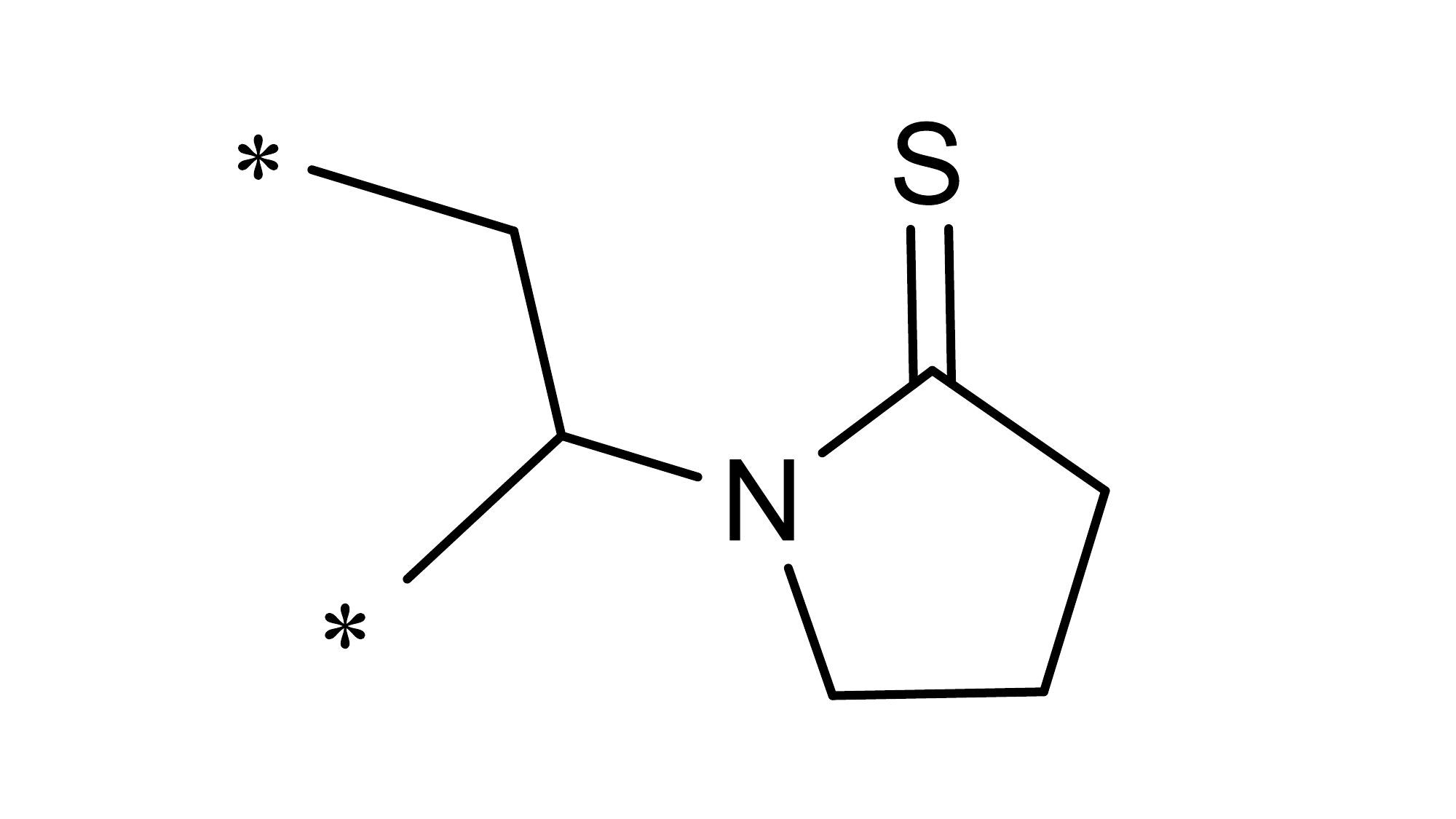}} & 80 \\
        Poly8  & [*]CC(C)(CO[*])COCCC\#N & \raisebox{-0.48\height}{\includegraphics[width=0.15\textwidth, trim=70 60 50 120,clip]{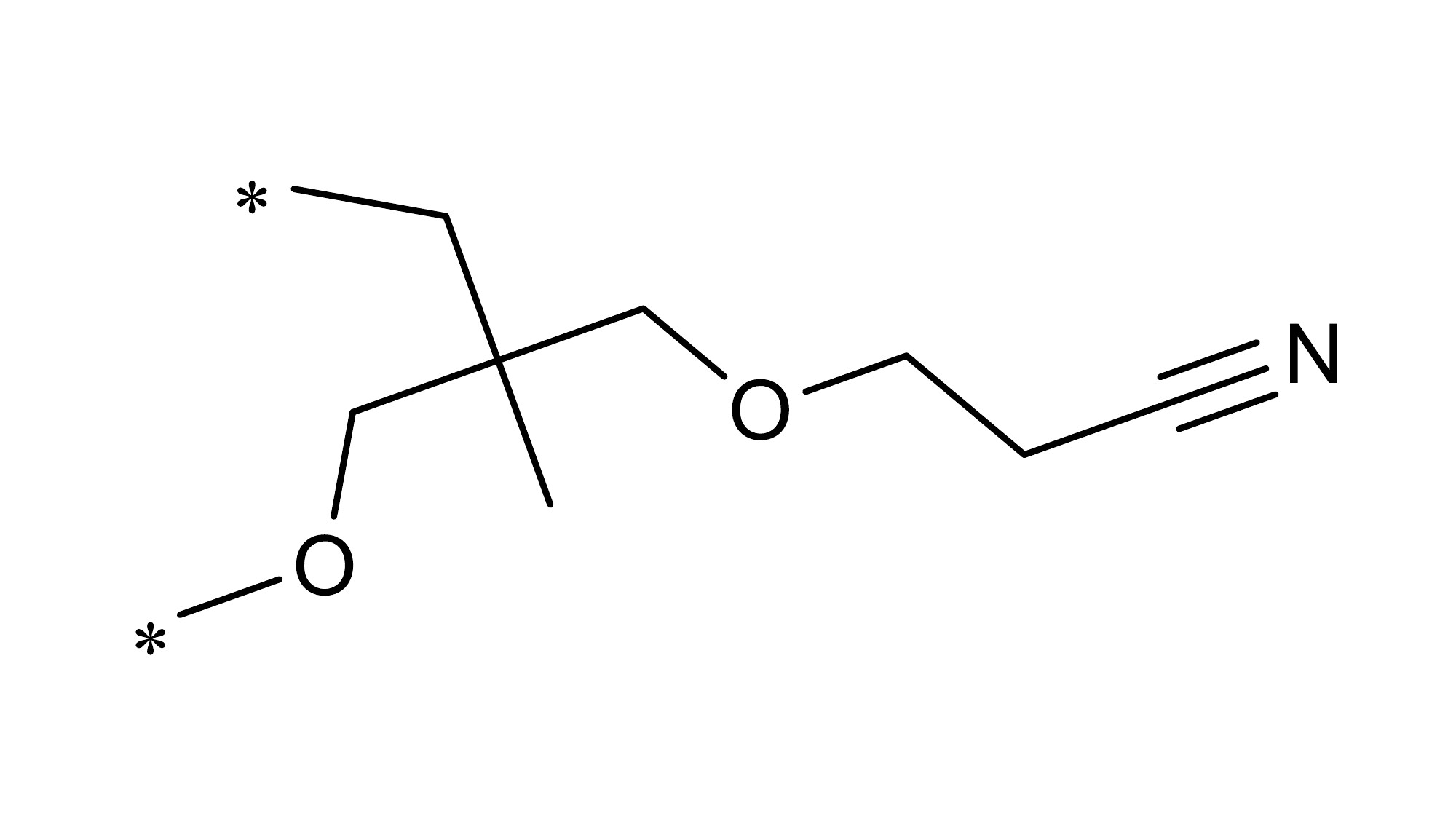}} & 66 \\
        \bottomrule
    \end{tabular}
  \label{tab: selected_polymer}
  \end{footnotesize}
\end{table}

\begin{figure}[t]
    \centering
  \subfigure[Radius of gyration]{
    \includegraphics[width=0.42\textwidth, trim=260 120 290 120,clip]{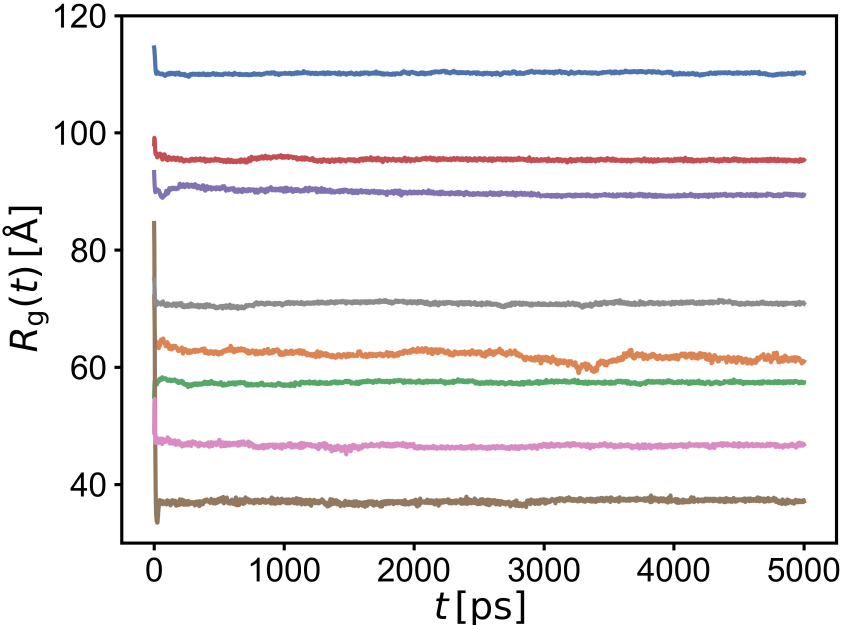}
  }
  \subfigure[Coefficient of variation]{
    \includegraphics[width=0.42\textwidth, trim=265 115 285 120,clip]{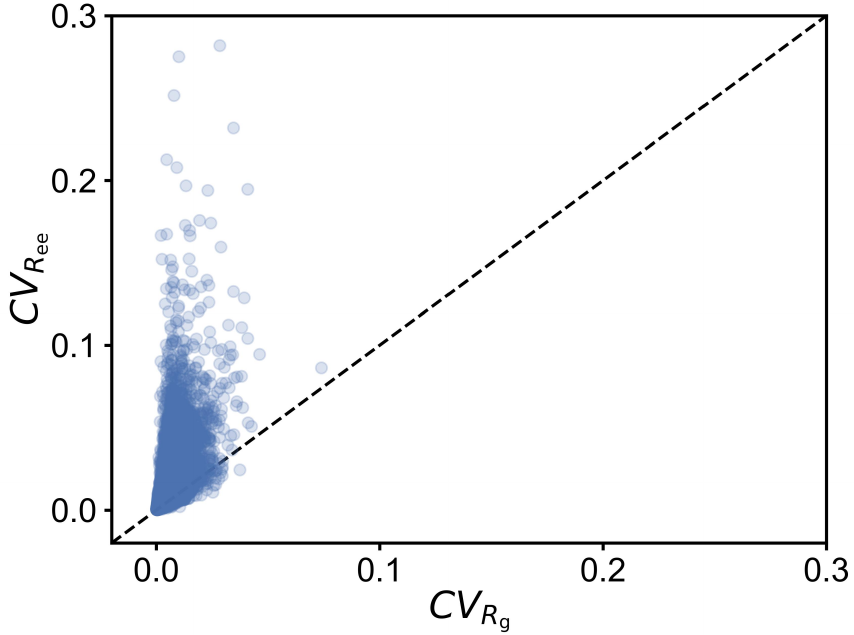}
  }
  \caption{\textbf{\thefigure:} \re{Equilibration and relaxation diagnostics. (a) \rre{Radius of gyration $R_{\rm g}(t)$} for eight representative polymers, where lines in blue, orange, green, red, purple, brown, pink, and gray correspond to Poly1-Poly8 in \ref{tab: selected_polymer}, respectively.} \rre{(b) Coefficients of variation of $R_{\rm g}$ and $R_{\rm ee}$ for all polymers in the full MD dataset.}} 
  \label{fig: md_equil}
\end{figure}

\begin{figure}[t]
    \centering
    \includegraphics[width=1\textwidth, trim=82 125 115 110,clip]{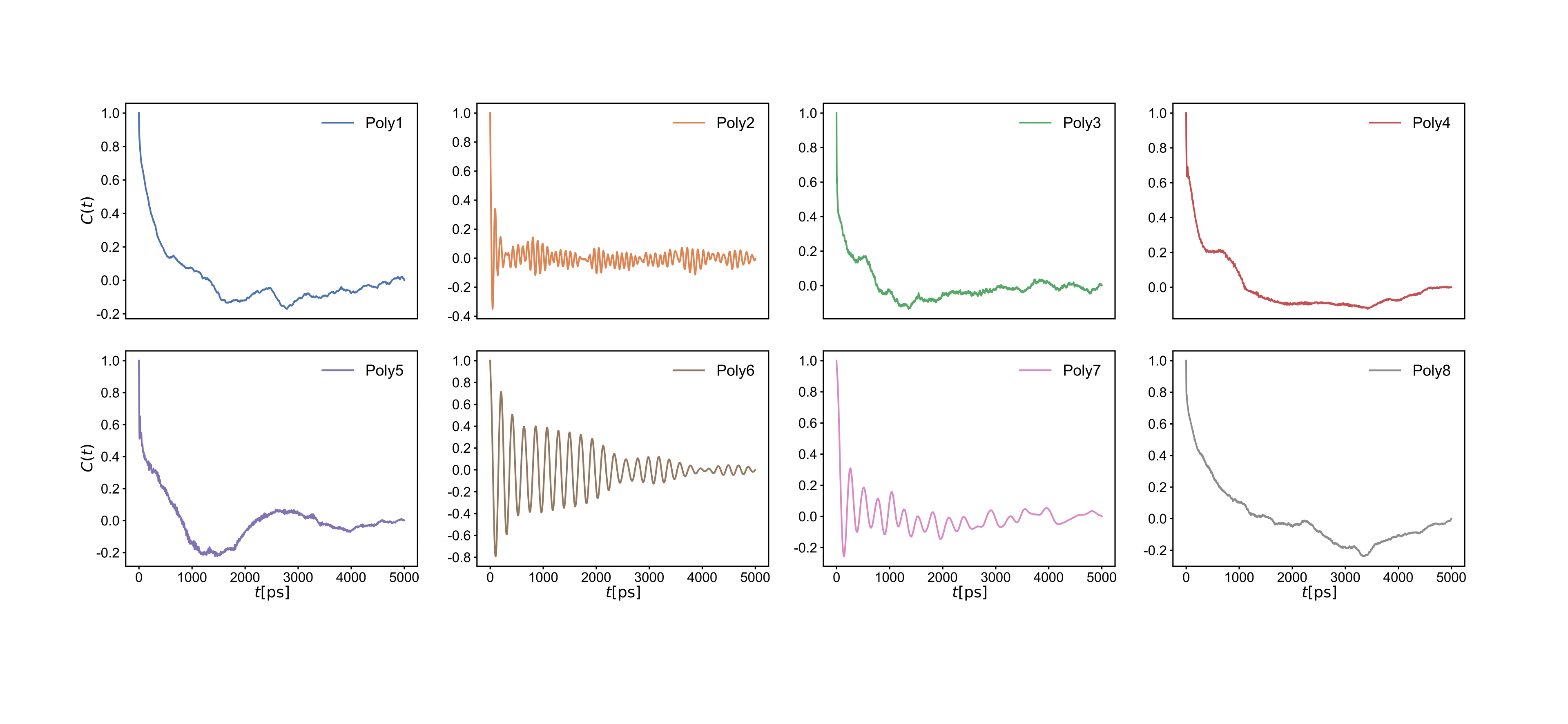}
    \caption{\textbf{\thefigure:} \rre{Autocorrelation function of end-to-end vectors for the eight representative polymers listed in \ref{tab: selected_polymer}.}}
    \label{fig: md_acf}
\end{figure}

We first analyze the trajectory of the representative polymers listed above to examine the time evolution of the \rre{radius of gyration $R_{\rm g}(t)$}. As shown in \ref{fig: md_equil}a, \rre{$R_{\rm g}$} decreases rapidly at the beginning of the simulations and reaches a stable plateau within 1 ns, indicating equilibration of the global chain size. 
In addition, we further compute the time autocorrelation function of the end-to-end vector $C(t)=\langle\mathbf{R_{ee}}(t)\cdot\mathbf{R_{ee}}(0)\rangle/\langle\mathbf{R_{ee}}^2(0)\rangle$ (see \ref{fig: md_acf}), where $\mathbf{R_{ee}}$ denotes the polymer end-to-end vector. 
\rre{The relaxation behavior differs among the polymers, as expected from their different backbone flexibility, side-chain structures, and intramolecular interactions. Nevertheless, the decay of these autocorrelations toward zero indicates the loss of orientational memory and sufficient conformational relaxation over the equilibration timescale.} Together, these analyses support that the polymer conformations used for the dataset are sampled from an equilibrated and relaxed regime.
As a complementary dataset-level diagnostic, we further examined the magnitude of global conformation fluctuations in the final 1 ns sampling window for the full MD dataset. For each polymer, we computed the coefficients of variation for $R_{\rm g}$ and $R_{\rm ee}$ over the last 1 ns of the trajectory, defined as $CV_X=\sigma(X)/\langle X\rangle$, $X \in \{R_{\rm g}, R_{\rm ee}\}$. As shown in \ref{fig: md_equil}b, $CV_{R_{\rm ee}}$ is generally larger than $CV_{R_{\rm g}}$, consistent with the greater sensitivity of the end-to-end distance to global chain reorientation and chain-end motion. By contrast, $R_{\rm g}$ is averaged over the entire chain and is therefore less sensitive to localized conformation changes. Moreover, the data points are concentrated in the lower-left region, indicating finite and bounded fluctuations of these global observables within the sampling window for most polymers. This supports the use of the final 1 ns trajectory segment as a reasonable window for extracting conformations in dataset construction.

\begin{figure}[t]
    \centering
    \includegraphics[width=0.48\textwidth, trim=260 110 280 100,clip]{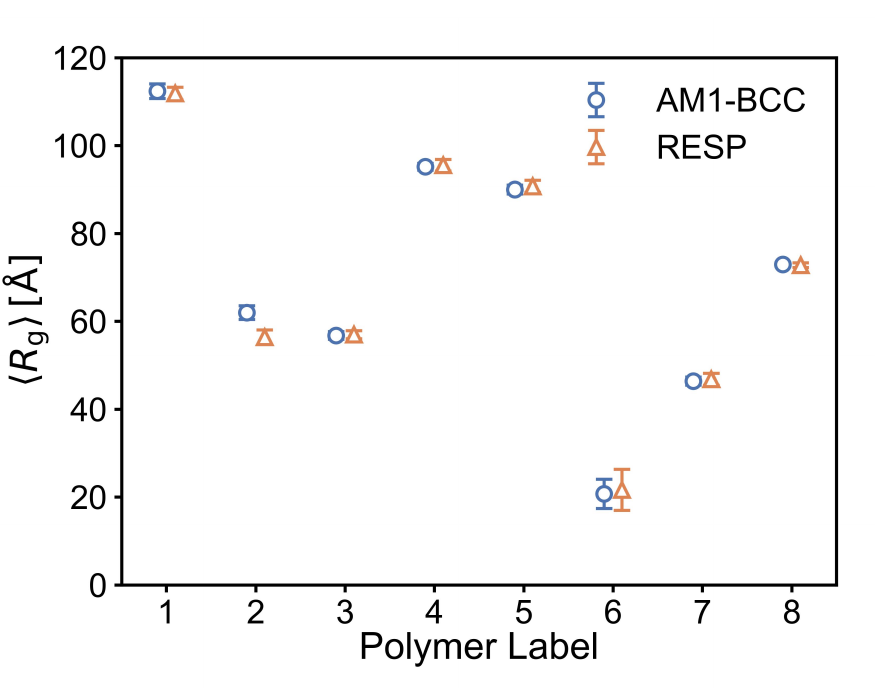}
    \caption{\textbf{\thefigure:} \re{\rre{Ensemble-averaged radius of gyration $\langle R_{\rm g} \rangle$} of the 8 representative polymers from simulations using the AM1-BCC and RESP partial charge atomic assignment schemes. Error bars denote the standard deviation across 5 independent simulations.}} 
  \label{fig: rg_charge}
\end{figure}

\begin{figure}[ht]
    \centering
    \includegraphics[width=1\textwidth, trim=80 125 115 100,clip]{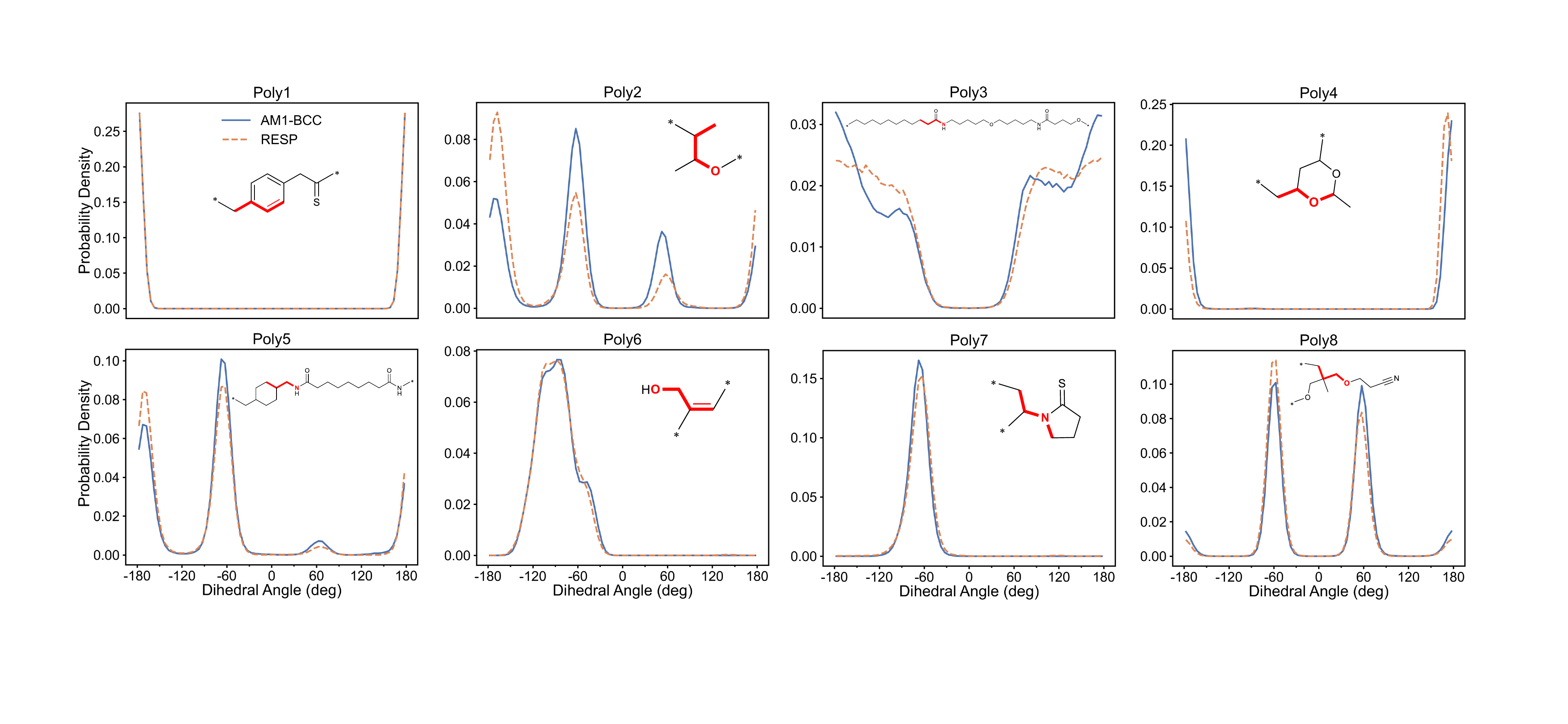}
    \caption{\textbf{\thefigure:} \re{The selected backbone dihedral angle distributions of the 8 representative polymers from simulations using the AM1-BCC and RESP partial charge atomic assignment schemes. The corresponding dihedral is highlighted in red on the 2D repeat-unit structures.}}
    \label{fig: poly_dihedral}
\end{figure}

To evaluate the sensitivity of polymer conformations to the partial atomic charge assignment scheme, we compared polymer conformations obtained from MD simulations using AM1-BCC and RESP~\cite{bayly1993well} charges. RESP charges are derived by fitting to electrostatic potentials computed with Gaussian16~\cite{Gaussian16} at the B3LYP/6-311G(d,p) level using Antechamber. Note that RESP has been widely adopted in atomistic polymer parameterization and simulation studies as a higher-fidelity fixed-charge baseline~\cite{fang2021revised,davel2024parameter}; accordingly, we employ it here as a practical reference for robustness testing. 
For each representative polymer, we perform five paired simulations under the two charge assignment schemes. Within each pair, the two simulations are started from the same initial conformation and employ identical random seeds. After equilibration, an additional 5 ns production process is carried out with a 10 ps sampling interval. As shown in \ref{fig: rg_charge}, the $\langle R_{\rm g} \rangle$ values obtained with AM1-BCC and RESP charges are in close agreement for all 8 polymers. 
We further compare the ensemble-averaged distributions of selected backbone dihedral angles involving heteroatoms between the two charge schemes (see \ref{fig: poly_dihedral}). Despite noticeable differences in peak heights, particularly for polymers with polar-group-containing torsions and more flexible backbones (e.g., Poly2 and Poly3), the overall dihedral distribution trends remain consistent. 
Overall, for the representative neutral polymers and conditions considered here, the conformational statistics relevant to dataset construction are not strongly sensitive to the choice between AM1-BCC and RESP charges.

\textbf{Dataset Statistics.}
\re{We devoted considerable time and resources to constructing a polymer conformation dataset via the pipeline described above.
This dataset comprises over 50,000 linear polymers, each equipped with five conformations uniformly sampled from the final 1 ns of its MD trajectory to ensure the capture of well-equilibrated states.
To rigorously establish the generality and robustness of our model, we systematically characterized the chemical and structural distribution of this dataset.
Regarding the dominant chemistries, this dataset exhibits a high degree of functional group heterogeneity, avoiding monopolization by any single polymer class.
As presented in~\ref{tab: conf_dominant_chemistries}, the three predominant functional groups are ethers (31.97\%), esters (26.94\%), and amides (19.12\%). 
It also demonstrates a robust long-tail distribution covering specialized chemistries such as urethanes (4.77\%), sulfones (3.30\%), and fluorinated groups (3.16\%), thereby underscoring the comprehensive coverage.
In terms of structural diversity, this dataset contains over 11,000 unique Murcko scaffolds, demonstrating exceptional topological diversity.
Furthermore, this dataset demonstrates a highly balanced distribution between aromatic and aliphatic structures, with 64.6\% of repeating units containing aromatic rings and 35.4\% being strictly aliphatic. 
Regarding ionic states, the vast majority of repeating units are neutral (99.77\%), while a small fraction (0.23\%) represents charged species. 
Notably, this distribution is highly consistent with the actual occurrence of polymers, where neutral species predominate in most industrial and commercial applications. 
Collectively, these statistics strictly substantiate that our dataset captures extensive diversity in both chemical composition and geometric complexity.

\begin{table}[t]
    \centering
    \setlength{\tabcolsep}{0.5em}
    \caption{\textbf{\thetable:} \re{The distribution of dominant chemistries (functional groups) within the conformation dataset.}}
    \begin{tabular}{lc|lc}
    \toprule
    \textbf{Functional Group} & \textbf{Percentage} & \textbf{Functional Group} & \textbf{Percentage} \\
    \midrule
    Ether & 31.97\% & Sulfone & 3.30\% \\
    Ester & 26.94\% & Fluorinated & 3.16\% \\
    Amide & 19.12\% & Aliphatic Hydrocarbon & 3.03\% \\
    Urethane & 4.77\% & Carbonate & 2.04\% \\
    Aromatic Hydrocarbon & 3.64\% & Imide & 1.92\% \\
    \bottomrule
    \end{tabular}
    \label{tab: conf_dominant_chemistries}
\end{table}

\begin{figure}[ht]
  \centering
  \subfigure[Training Set]{
    \includegraphics[width=0.313\textwidth]{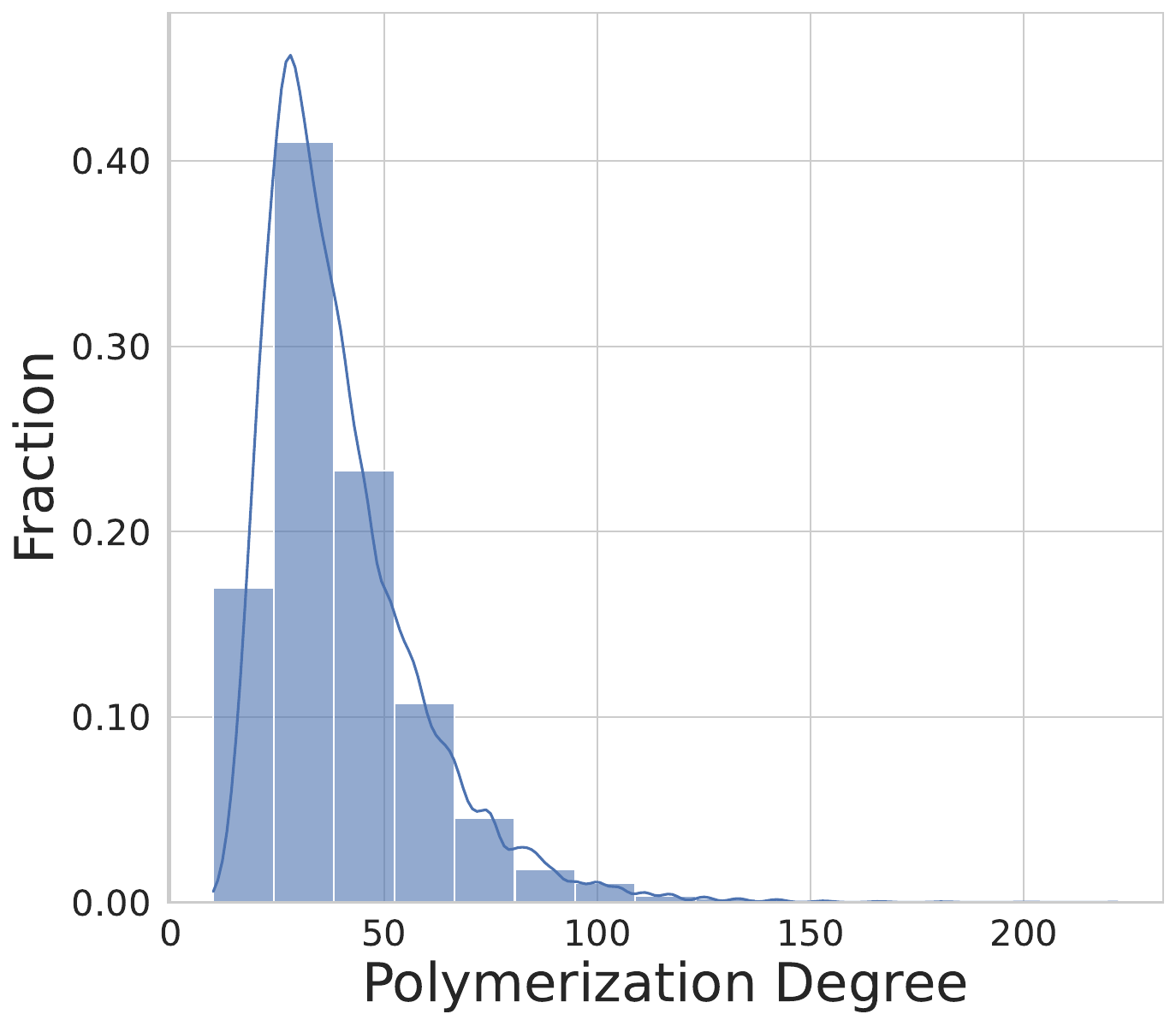}
  }
  \subfigure[Validation Set]{
    \includegraphics[width=0.313\textwidth]{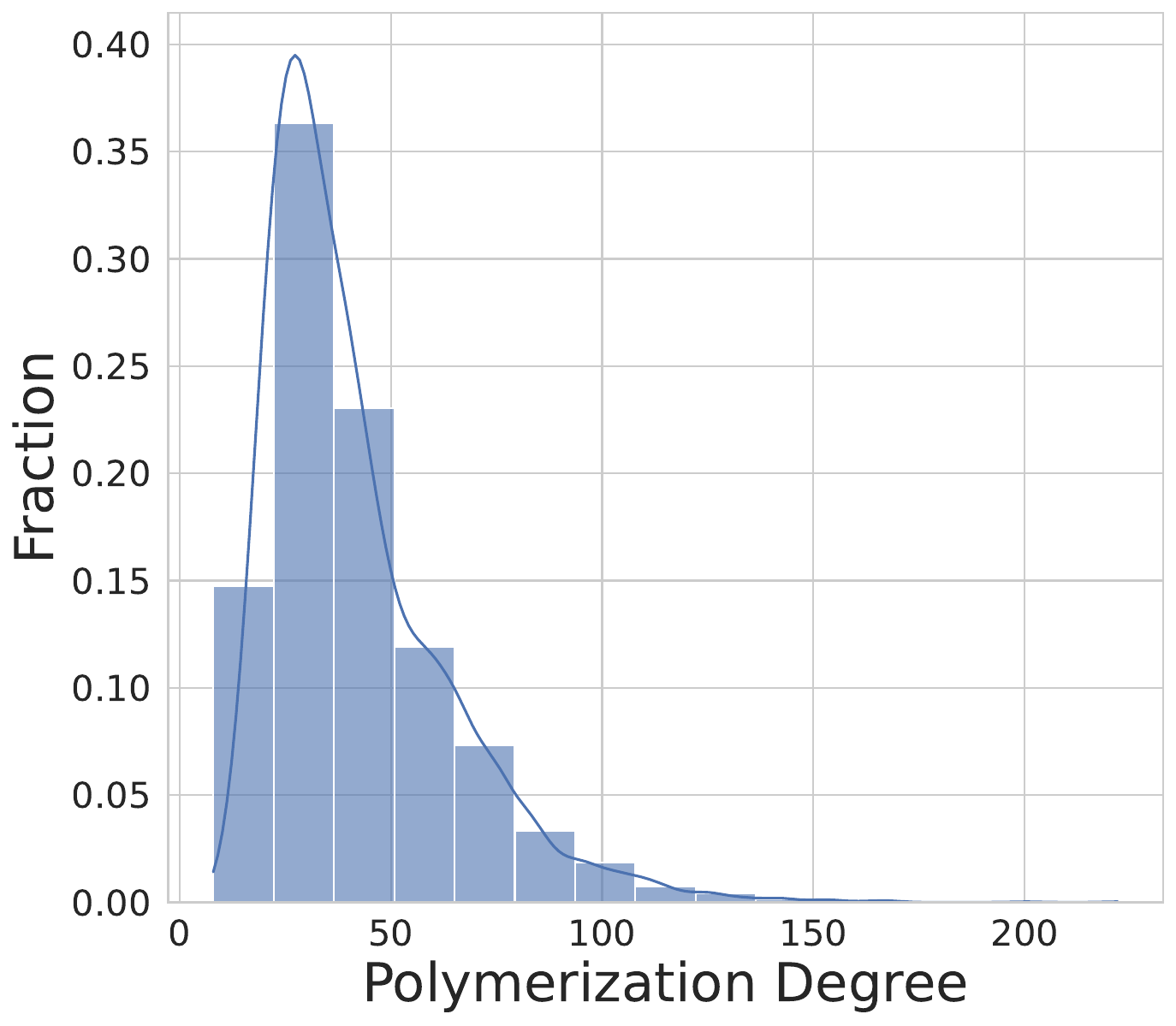}
  }
  \subfigure[Test Set]{
    \includegraphics[width=0.313\textwidth]{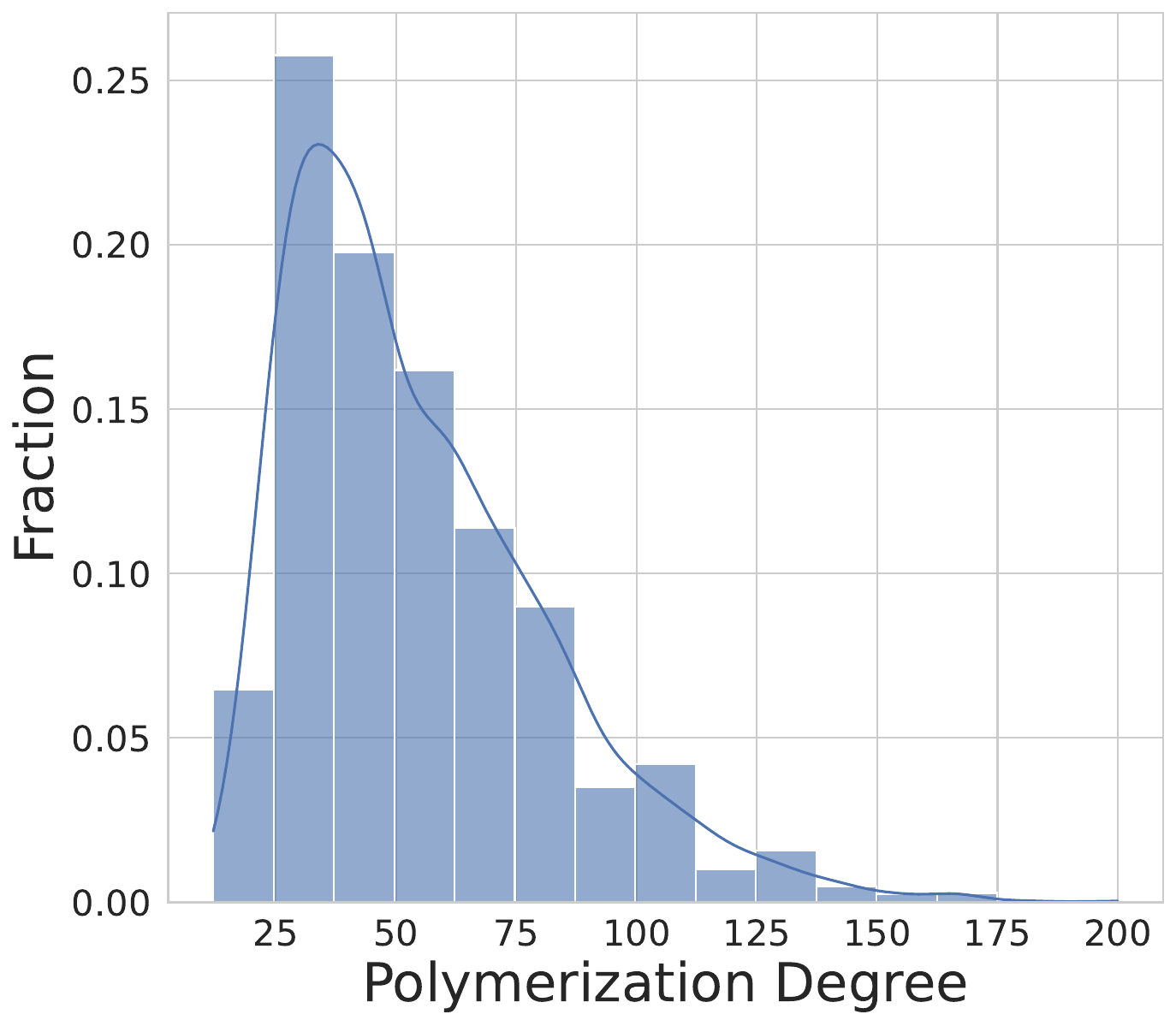}
  }
  \caption{\textbf{\thefigure:} \re{The distribution of polymerization degree (i.e., the number of repeating units per conformation) within the conformation dataset.}} 
  \label{fig: conf_dataset_degree}
\end{figure}

\begin{figure}[ht]
  \centering
  \subfigure[Training Set]{
    \includegraphics[width=0.313\textwidth]{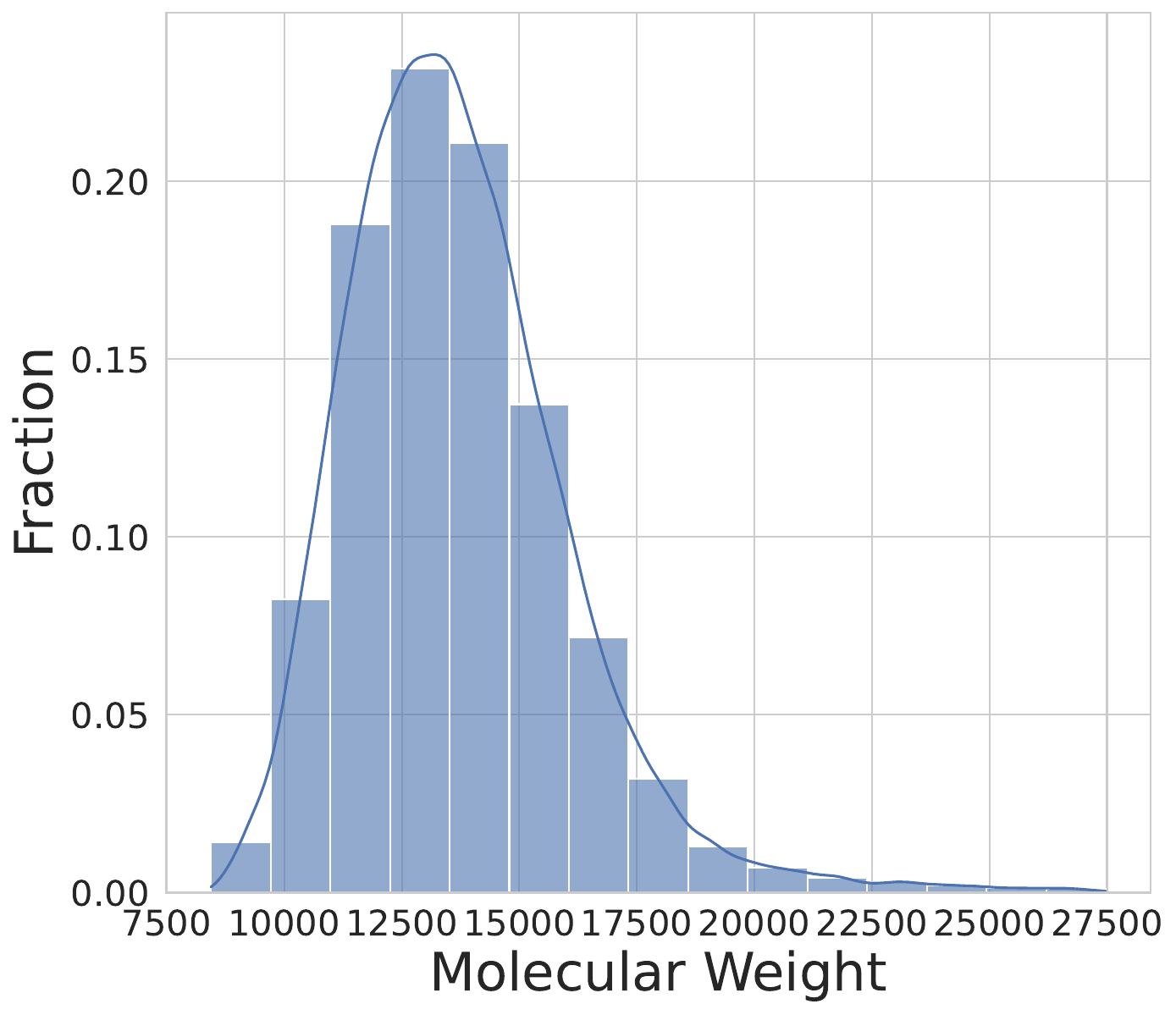}
  }
  \subfigure[Validation Set]{
    \includegraphics[width=0.313\textwidth]{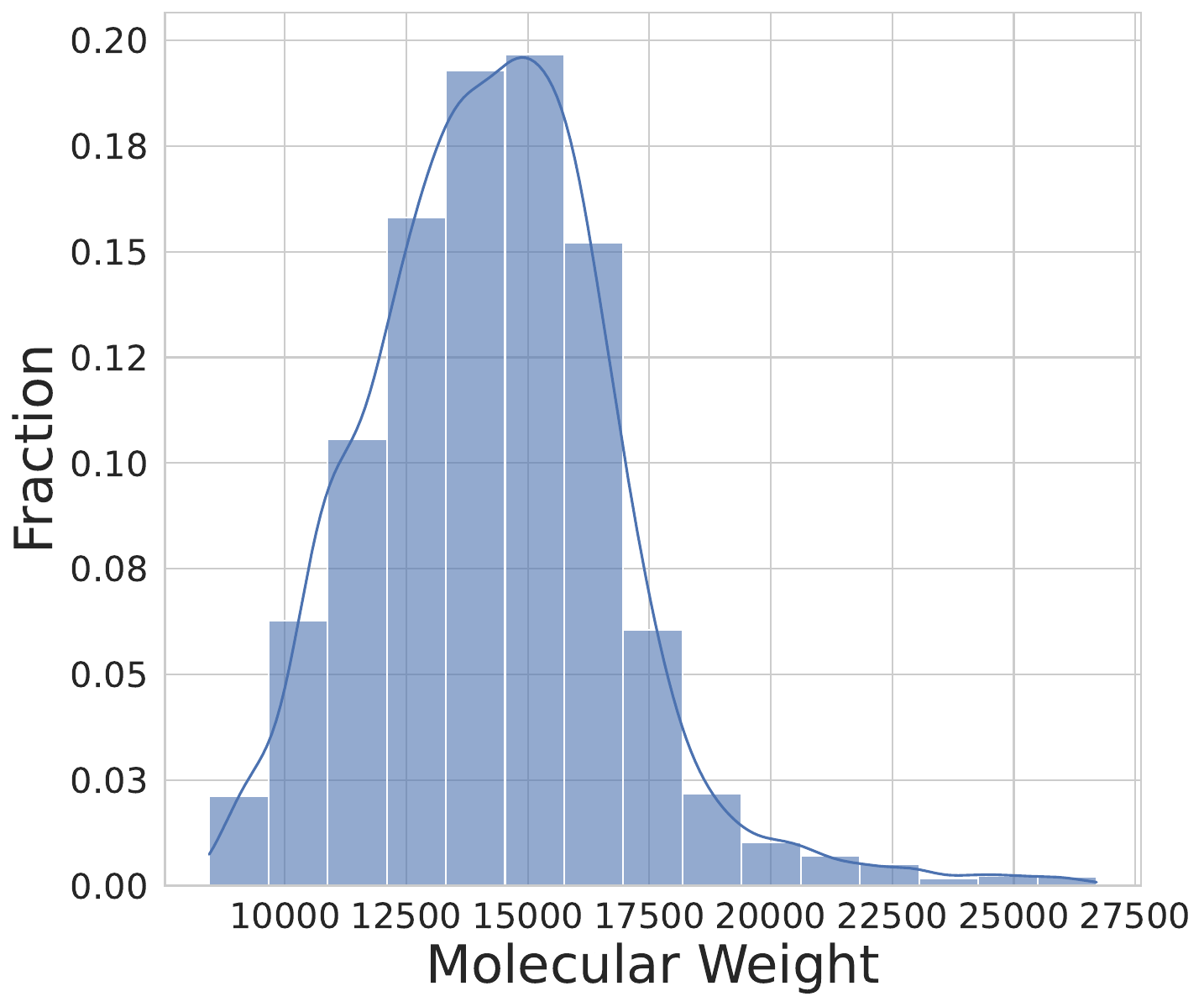}
  }
  \subfigure[Test Set]{
    \includegraphics[width=0.313\textwidth]{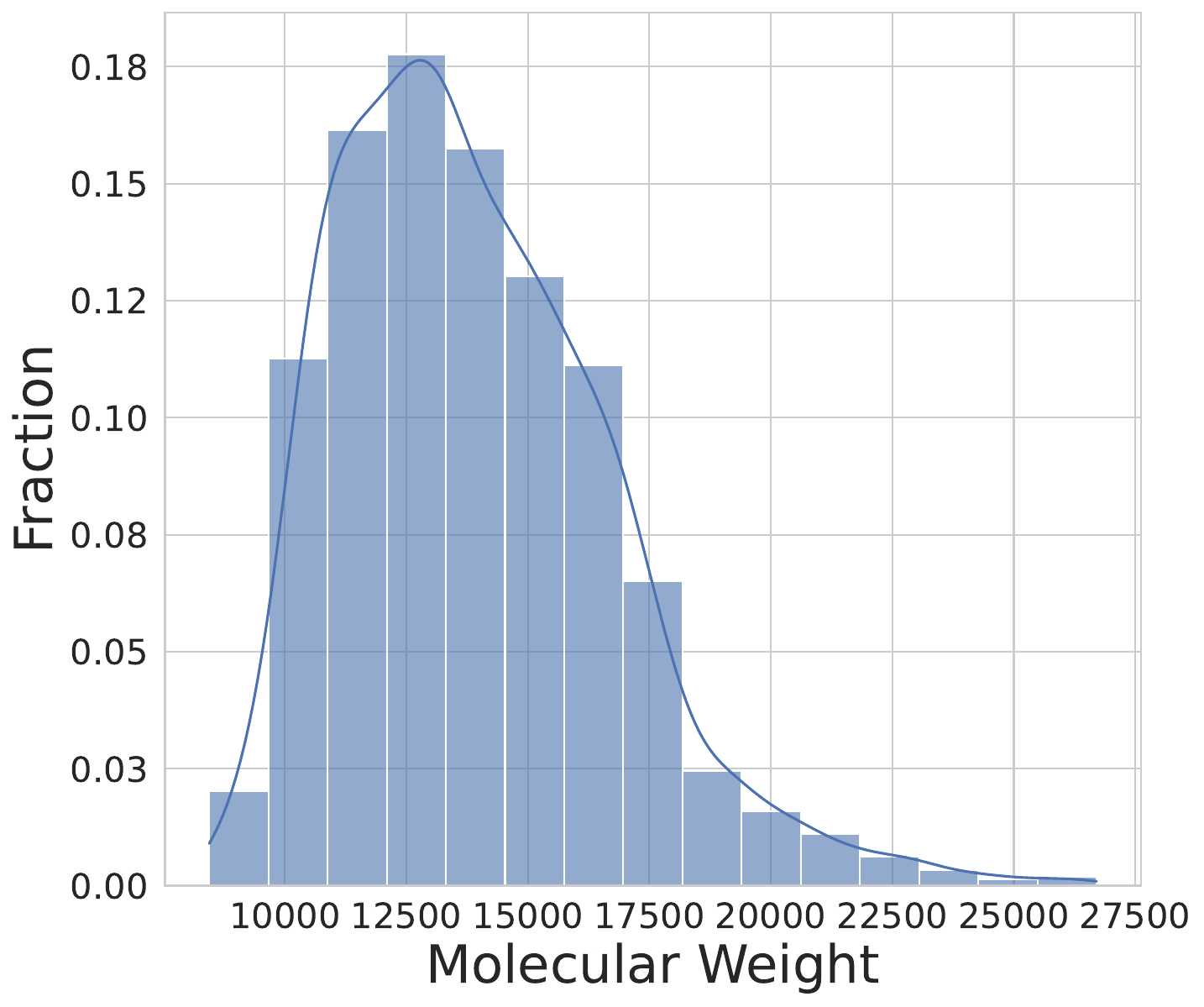}
  }
  \caption{\textbf{\thefigure:} \re{The distribution of molecular weight within the conformation dataset.}} 
  \label{fig: conf_dataset_weight}
\end{figure}

Following standardization, deduplication, and stringent quality control, this dataset is further partitioned into training ($\sim$46k polymers), validation ($\sim$5k), and test ($\sim$2k) sets.
Please note that this test set is used to assess conformation generation capability.
Here, we visualize the distribution of polymerization degree (i.e., the number of repeating units per conformation) and molecular weight to provide insights into the structural complexity and variability of this dataset.
As demonstrated in~\ref{fig: conf_dataset_degree} and~\ref{fig: conf_dataset_weight}, the polymerization degree predominantly ranges from 20 to 100, corresponding to molecular weights of 10,000 to 20,000, with highly consistent distributions across all data splits.
Please note that this specific physical scale is intentionally determined to strike an optimal balance for our single-chain molecular dynamics (MD) pipeline. 
It is sufficiently large to capture fundamental polymer physics, such as intra-chain interactions, steric hindrance, and local folding behaviors, while remaining computationally tractable to ensure that the MD trajectories reliably converge to well-equilibrated states. 
Consequently, by maintaining this rigorous physical fidelity, our dataset successfully captures and covers substantial structural diversity in polymer conformations.
}

\begin{table}[tbp]
  \centering
  \caption{\textbf{\thetable:} The summary of property datasets.}
    \begin{tabular}{lcccc}
        \toprule
        Dataset & Property & Unit & \# Samples & Data Range \\
        \midrule
        Egc   & bandgap (chain) & eV & 3380 & $[0.02, 8.30]$ \\
        Egb   & bandgap (bulk) & eV & 561 & $[0.39, 10.05]$ \\
        Eea   & electron affinity & eV & 368 & $[0.39, 4.61]$ \\
        Ei    & ionization energy & eV & 370 & $[3.55, 9.61]$ \\
        Xc    & crystallization tendency & \% & 432 & $[0.13, 98.41]$ \\
        EPS   & dielectric constant & 1 & 382 & $[2.61, 8.52]$ \\
        Nc    & refractive index & 1 & 382 & $[1.48, 2.58]$ \\
        Eat   & atomization energy & eV/atom & 390 & $[-6.83, -5.02]$ \\
        \bottomrule
    \end{tabular}
  \label{tab: property_dataset}%
\end{table}%

\subsection{Polymer Property Dataset}
While numerous polymer property datasets have been reported, the majority are either not publicly accessible or available only for online querying~\cite{kuenneth2023polybert, otsuka2011polyinfo}.
In this context, following the latest work~\cite{wang2024mmpolymer}, we also use eight polymer property datasets (denoted as Egc, Egb, Eea, Ei, Xc, EPS, Nc, and Eat, respectively) provided in\re{~\cite{kuenneth2021polymer}}.
In particular, these datasets, derived from density functional theory calculations, encompass a broad spectrum of typical properties and are partitioned in line with~\cite{wang2024mmpolymer}, thereby ensuring the reliable assessment of property prediction capability.
More details about these property datasets are summarized in~\ref{tab: property_dataset}.

\subsection{Polymer Design Dataset}
Although the latest work~\cite{liu2024graph} provides one dataset comprising 553 polymers for polymer design, which employs three gas permeability (i.e., O2Perm, CO2Perm, and N2Perm) along with synthesizability as the condition set to balance performance and practical feasibility, some polymers in this dataset are chemically invalid (e.g., lacking polymerization sites).
\re{In this context, we further exclude such chemically inadmissible polymers that failed rigorous checks concerning the presence of essential polymerization sites, linkage viability, chemical sanity, and structural connectivity.
These remaining polymers are then utilized as our design dataset, thereby ensuring data validity and reliable evaluation.}
In particular, this dataset comprises approximately 400 polymers and is partitioned into training, validation, and test sets.

\section{Details on Baselines}\label{SI-sec: baselines}
\subsection{Polymer Conformation Generation Baseline}
For the polymer conformation generation task, given the absence of specialized methods, we have to utilize various molecular conformation generation methods as our baselines, including:  
\begin{itemize}
  \item \textbf{GeoDiff~\cite{xu2022geodiff}} employs the diffusion process directly on the Euclidean coordinate space of atoms, with the SE(3)-equivariant denoising model that preserves roto-translational symmetry.
  \item \textbf{TorsionalDiff~\cite{jing2022torsional}} employs the diffusion process only on the space of torsion angles, with the extrinsic-to-intrinsic score model that satisfies the required symmetries.
  \item \textbf{MCF~\cite{wang2024swallowing}} employs the domain-agnostic diffusion process on the conformer field while making no assumptions about structures, with the non‑equivariant score model that benefits from scale.
  \item \textbf{ET-Flow~\cite{hassan2024flow}} employs flow matching directly on all-atom coordinates while incorporating more informed priors, with the Equivariant Transformer that captures geometric features.
\end{itemize}
Here, we adopt these methods by treating polymers as large molecules containing more atoms, and then train them on the same polymer conformation dataset for performance comparison.
In particular, TorsionalDiff utilizes RDKit~\cite{landrum2013rdkit} by default to generate an initial 3D structure as input.
However, since RDKit does not apply to polymers, we have to utilize the initial 3D structure from the corresponding MD simulation trajectory as its input, unintentionally giving it a biased advantage over other methods.

\subsection{Polymer Property Prediction Baseline}
For the downstream polymer property prediction task, considering methods without pretraining have already been excluded from baselines in previous works~\cite{wang2024mmpolymer}, we directly utilize various state-of-the-art polymer pretraining methods designed for property prediction as our baselines, including: 
\begin{itemize}
  \item \textbf{polyBERT~\cite{kuenneth2023polybert}} and \textbf{Transpolymer~\cite{xu2023transpolymer}} are both BERT-style polymer pretraining frameworks that perform masked language modeling on numerous unlabeled polymers through treating polymers as character sequences (e.g., P-SMILES strings). 
  \item \textbf{MMPolymer~\cite{wang2024mmpolymer}} is the multimodal multitask polymer pretraining framework that not only conducts intra‑modal pretraining within polymer 1D sequential and 3D structural modalities but also leverages cross‑modal contrastive learning to align these two modalities.
\end{itemize}
Meanwhile, to reveal the limitations of directly transplanting molecular methods to polymer-specific tasks, various representative molecular pretraining methods are also utilized as our baselines, including:
\begin{itemize}
  \item \textbf{MolCLR~\cite{wang2022molecular}} is the graph-based molecular pretraining framework that leverages contrastive learning on augmented molecular graphs to maximize the agreement of augmentations from the same molecule while minimizing agreement across different molecules.
  \item \textbf{3D Infomax~\cite{stark20223d}} is the cross-modal molecular pretraining framework that enhances the 2D GNN awareness of 3D geometry through maximizing the mutual information between 2D graph embeddings and the corresponding 3D conformation embeddings.
  \item \textbf{Uni-Mol~\cite{zhou2023unimol}} is the 3D molecular pretraining framework that performs large‑scale self‑supervision to recover masked atoms and denoise 3D coordinates from corrupted molecular conformations.
\end{itemize}

\subsection{Polymer Design Baseline}
For the downstream polymer design task, we directly align with the latest work~\cite{liu2024graph}, utilizing it and its mentioned baselines as our baselines, including:
\begin{itemize}
  \item \textbf{MolGPT~\cite{bagal2021molgpt}} is the sequence-based generative model that represents molecules as sequences and generates them by predicting SMILES tokens autoregressively.
  \item \textbf{GraphGA~\cite{jensen2019graph}} is the graph-based genetic algorithm that evolves graphs via mutation and crossover operators under a validity‑constrained search to optimize target objectives.
  \item \textbf{DiGress~\cite{vignac2023digress}} is the graph-based generative model that generates graphs with categorical node and edge attributes through the discrete denoising diffusion.
  \item \textbf{GDSS~\cite{jo2022score}} is the graph-based generative model that achieves score-based generative modeling of graphs through the system of stochastic differential equations.
  \item \textbf{MOOD~\cite{lee2023exploring}} is the graph-based generative model that incorporates out-of-distribution control into the generative stochastic differential equation to explore the space beyond the training distribution.
  \item \textbf{GraphDiT~\cite{liu2024graph}} is the graph-based generative model that generates graphs through integrating the graph diffusion Transformer with graph-dependent noise.
\end{itemize}

\section{Supplementary Experiments}\label{SI-sec: exp}
\subsection{Polymer Conformation Generation}\label{SI-sec: exp_conf}

\begin{figure*}[t]
    \centering 
        \includegraphics[width=0.995\textwidth]{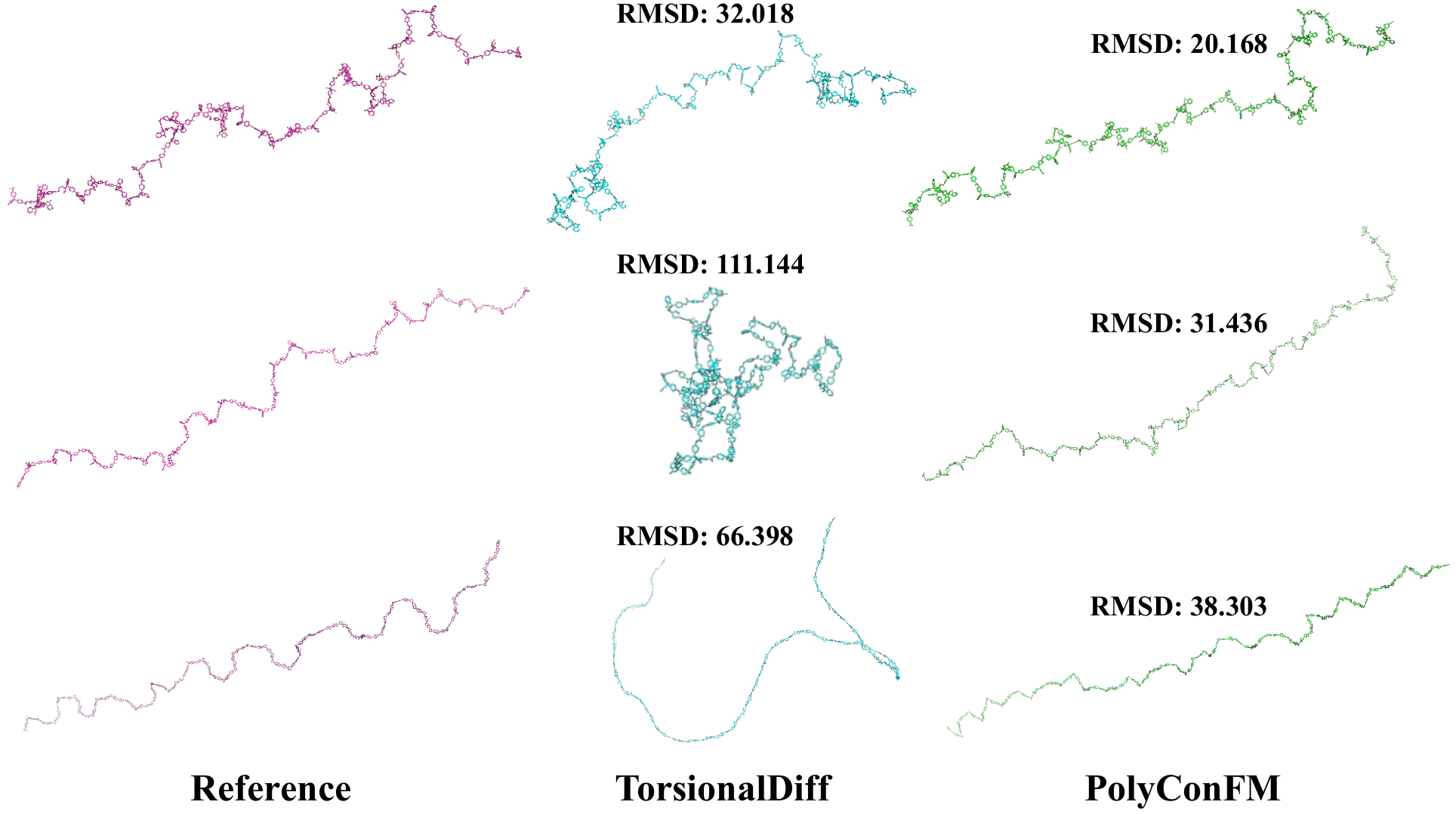} 
    \caption{\textbf{\thefigure:} Several visualization examples of TorsionalDiff (i.e., the best baseline) and PolyConFM on the polymer conformation generation task. \re{In particular, the root-mean-square deviation (RMSD) values are explicitly provided for each generated conformation to quantitatively evaluate geometric fidelity, where lower values indicate better alignment with the reference conformation.}}
    \label{fig: conf_case}
\end{figure*}

\begin{figure*}[t]
    \centering 
        \includegraphics[width=0.995\textwidth]{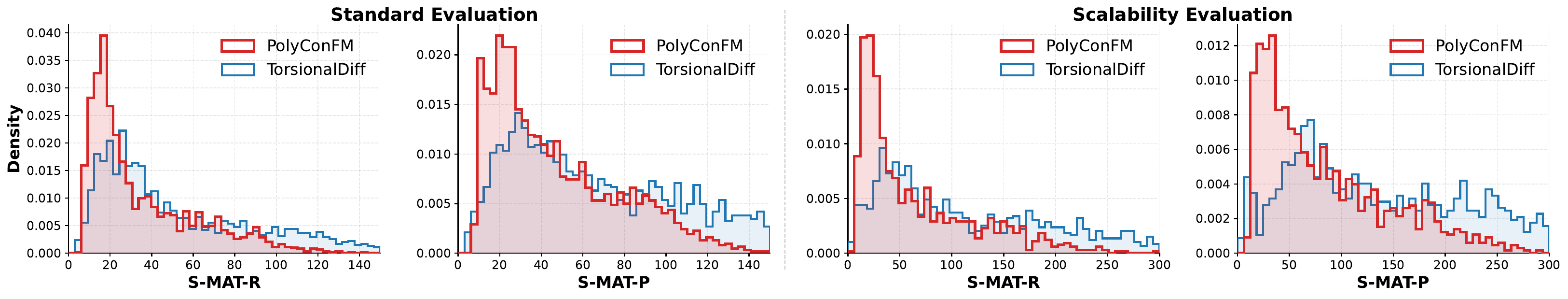} 
    \caption{\textbf{\thefigure:} \re{The density histograms of structure-matching metrics on the polymer conformation generation task, where we compute these metrics corresponding to PolyConFM and TorsionalDiff (i.e., the best baseline) under both standard and scalability evaluations.
    In particular, for the scalability evaluation, we double the number of repeating units per polymer within the test set during inference, thereby forcing models trained on conformations with approximately 2000 atoms to generate larger conformations with approximately 4000 atoms.}
    \rre{All reported results are evaluated on the test set.}}
    \label{fig: conf_histograms}
\end{figure*}

\ref{fig: conf_case} presents several visualization examples comparing PolyConFM with the best baseline (i.e., TorsionalDiff) on the polymer conformation generation task.
In particular, it demonstrates that PolyConFM generates polymer conformations that more closely align with the references, capturing unfolded and relaxed backbones as well as detailed geometry.
By contrast, despite being supplied with biased prior knowledge from the initial polymer structure, TorsionalDiff still yields overly compact or distorted conformations, failing to recover relaxed and extended configurations. 
Taken together, these qualitative comparisons further confirm PolyConFM's superior capability to generate conformations, which is crucial for diverse downstream tasks that depend on accurate structural priors.

Beyond reporting the mean and median values of S-MAT-R and S-MAT-P in Table~\ref{tab: exp_conf}, we further provide their density histograms to reinforce the evidence of the superior structural fidelity achieved by PolyConFM.
As presented in~\ref{fig: conf_histograms}, compared with the best baseline (i.e., TorsionalDiff), PolyConFM exhibits a significantly higher concentration of samples in the low-value regions, which is reflected in its markedly sharper and more left-shifted peaks for both S-MAT-R and S-MAT-P across standard and scalability evaluations, confirming that PolyConFM consistently generates conformations with higher structural accuracy and tighter alignment with the references. 
Notably, while TorsionalDiff exhibits a progressively broadened and right-shifted error distribution in the scalability evaluation, PolyConFM maintains a stable and concentrated profile. 
This statistical consistency further underscores PolyConFM's robustness in maintaining structural coherence across larger conformations, highlighting its superior extrapolation capabilities for multiscale polymer conformation generation.

\begin{figure*}[t]
    \centering
    \begin{subfigure}
        \centering
        \includegraphics[width=0.995\textwidth]{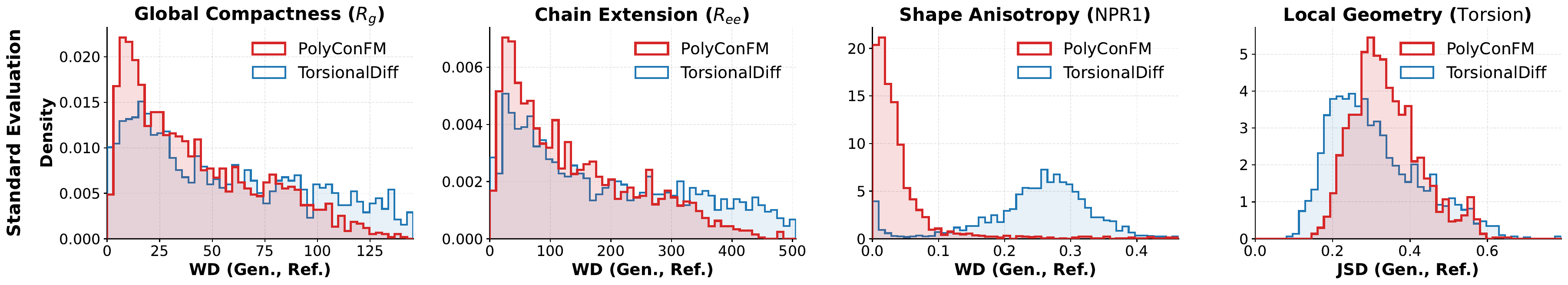}
    \end{subfigure}
    \begin{subfigure}
        \centering
        \includegraphics[width=0.995\textwidth]{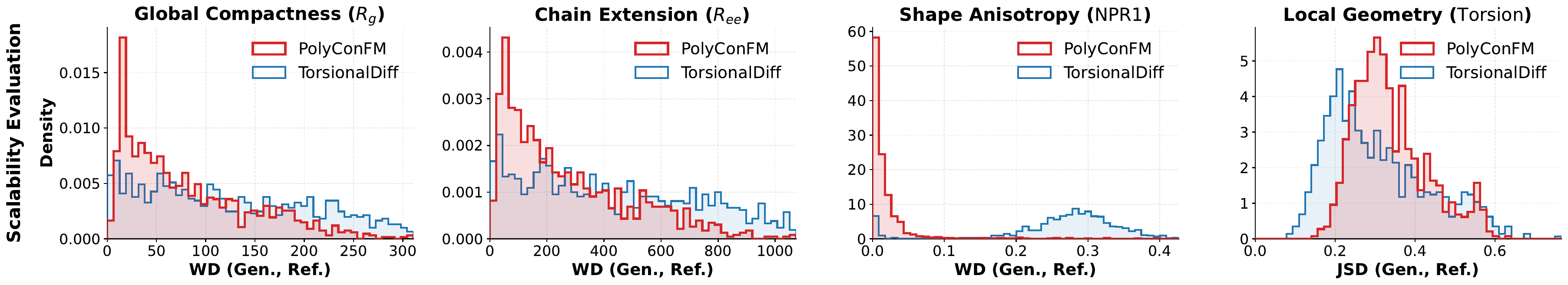}
    \end{subfigure}
   \caption{\textbf{\thefigure:} \re{The density histograms of ensemble-level metrics on the polymer conformation generation task, where we compute distributional divergence between the ensembles generated by PolyConFM (or the best baseline, TorsionalDiff) and reference ensembles for comparison.
   In particular, we employ Wasserstein Distance (WD) to measure distributional divergence across global geometric descriptors ($R_{g}$, $R_{ee}$, and NPR1) and Jensen-Shannon Divergence (JSD) for torsion angle distributions.
   \rre{Here, all reported results are evaluated on the test set.}}
   }
    \label{fig:conf_enesemble_histograms}
\end{figure*}

Considering that both S-MAT-R and S-MAT-P are primarily based on root-mean-square deviation (RMSD), we further extend our evaluation to various ensemble-level physical descriptors, quantifying the distribution divergence between generated and reference ensembles under these physical descriptors.
Specifically, we select four representative physical descriptors to characterize the structural features of the polymer ensemble from global to local scales, including:
\begin{itemize}
    \item \textbf{Global Compactness:} The distribution of the radius of gyration ($R_{\text{g}}$), which measures the overall spatial span and volume occupancy of the polymer.
    \item \textbf{Chain Extension:} The distribution of end-to-end distances ($R_{\text{ee}}$), which captures the long-range correlation and reflects the global scaling behavior of the polymer.
    \item \textbf{Shape Anisotropy:} The distribution of normalized principal moments of the inertia tensor (NPR1), which serves as a geometric proxy for chain stiffness and persistence length to quantify the deviation of the polymer shape from a perfect sphere.
    \item \textbf{Local Geometry:} The distribution of torsion angles, which characterizes local conformational preferences and rotameric states governed by stereochemical constraints.
\end{itemize}
Accordingly, we compute the distributional divergence between the ensembles generated by PolyConFM (or baselines) and the reference ensembles across these descriptors for performance comparison.
Here, we employ the Wasserstein Distance (WD) for the global geometric descriptors ($R_{\text{g}}$, $R_{\text{ee}}$, and NPR1), as it effectively accounts for the metric distance between physical values in Euclidean space. 
For the local geometry, we utilize the Jensen-Shannon Divergence (JSD) to measure corresponding torsion angle distributions. 
This distinction is made because WD is prioritized for continuous observables to capture the magnitude of structural deviations, whereas JSD is better suited for the statistical alignment of angular data.
As presented in~\ref{tab: exp_conf_ensemble} and~\ref{fig:conf_enesemble_histograms}, PolyConFM exhibits superior fidelity in capturing ensemble-level structural features across both global and local scales compared with various baselines.
For global geometric descriptors, PolyConFM achieves a substantial reduction in Wasserstein Distance (WD) for Global Compactness ($R_g$) and Chain Extension ($R_{ee}$), with density histograms revealing error distributions that are significantly more concentrated toward zero than the best baseline (i.e., TorsionalDiff). 
This advantage is particularly pronounced in Shape Anisotropy (NPR1), where PolyConFM maintains a sharp peak near zero with a mean WD of 0.046, and TorsionalDiff exhibits a broad distribution with a mean WD of 0.249.
Regarding local geometry, although TorsionalDiff achieves high distributional similarity in torsion angles because its diffusion process explicitly operates within the torsional space, PolyConFM yields comparable performance while simultaneously delivering a markedly superior distribution alignment of global geometric descriptors.
This comprehensive superiority in capturing both short-range and long-range correlations underscores the reliability of PolyConFM on the polymer conformation generation task.

\begin{table}[t]
  \centering
  \setlength{\tabcolsep}{0.16em}
   \caption{\textbf{\thetable:} \re{The ensemble-level comparison of different methods on the polymer conformation generation task, where we quantify the distribution divergence between generated and reference ensembles under various physical descriptors. In particular, we report the Wasserstein Distance (WD) between generated and reference distributions for global geometric descriptors, including \textbf{Global Compactness} (via radius of gyration, $R_{\text{g}}$), \textbf{Chain Extension} (via end-to-end distance, $R_{\text{ee}}$), and \textbf{Shape Anisotropy} (via normalized principal moments, NPR1), while comparing \textbf{Local Geometry} through Jensen-Shannon Divergence (JSD) of the corresponding torsion angle distributions. Here, these metrics are presented as Mean/Median values, with lower values indicating better performance.}
   \rre{All reported results are evaluated on the test set.}}
    \begin{tabular}{llcccccccc}
    \toprule
    & \multirow{2}[2]{*}{Method} & \multicolumn{2}{c}{$R_{\text{g}}$} & \multicolumn{2}{c}{$R_{\text{ee}}$} & \multicolumn{2}{c}{NPR1} & \multicolumn{2}{c}{Torsion} \\
        \cmidrule(lr){3-4} \cmidrule(lr){5-6} \cmidrule(lr){7-8} \cmidrule(lr){9-10} 
        & & WD Mean  & WD Median & WD Mean  & WD Median & WD Mean  & WD Median & JSD Mean  & JSD Median \\
    \midrule
        \multirow{5}{*}{\begin{sideways}\shortstack{Standard\\Evaluation}\end{sideways}}
            & GeoDiff~\cite{xu2022geodiff} & 86.863  & 82.441  & 289.559  & 270.845  & 0.629  & 0.654  & 0.418  & 0.410  \\
            & TorsionalDiff~\cite{jing2022torsional} & 59.790  & 50.171  & 205.270  & 168.350 & 0.249  & 0.262  & \textbf{0.311}  & \textbf{0.287}  \\
            & MCF~\cite{wang2024swallowing}   & 169.657  & 171.609  & 1280.700 & 1279.374 & 0.488 & 0.506 & 0.499 & 0.496  \\
            & ET-Flow~\cite{hassan2024flow} & 80.165  & 74.927  & 262.895  & 243.561  & 0.569  & 0.590  & 0.399  & 0.386   \\
            & \cellcolor{lightblue}PolyConFM & \cellcolor{lightblue}\textbf{\kern1pt 41.159} & \cellcolor{lightblue}\textbf{\kern1pt 32.400} & \cellcolor{lightblue}\textbf{\kern1pt 135.524} & \cellcolor{lightblue}\textbf{\kern1pt 105.787} & \cellcolor{lightblue}\textbf{\kern1pt 0.046} & \cellcolor{lightblue}\textbf{\kern1pt 0.025} & 
            \cellcolor{lightblue}{\kern0.1pt 0.319} 
            & \cellcolor{lightblue}{\kern0.1pt 0.308} \\
        \midrule
        \multirow{5}{*}{\begin{sideways}\shortstack{Scalability\\Evaluation}\end{sideways}}
            & GeoDiff~\cite{xu2022geodiff} & 175.143 & 163.215 & 593.487 & 555.886 & 0.713 & 0.730 & 0.417 & 0.406  \\
            & TorsionalDiff~\cite{jing2022torsional} & 128.871 & 110.689 & 449.655 & 384.323 & 0.265 & 0.281 & \textbf{0.307} & \textbf{0.275} \\
            & MCF~\cite{wang2024swallowing} & 279.922 & 256.099 & 1384.031 & 1449.293 & 0.558 & 0.575 & 0.534 & 0.504  \\
            & ET-Flow~\cite{hassan2024flow} & 168.308 & 156.184 & 563.914 & 520.762 & 0.654 & 0.665 & 0.395 & 0.382 \\
            & \cellcolor{lightblue}PolyConFM & \cellcolor{lightblue}\textbf{\kern1pt 78.352} & \cellcolor{lightblue}\textbf{\kern1pt 58.681} & \cellcolor{lightblue}\textbf{\kern1pt 263.996} & \cellcolor{lightblue}\textbf{\kern1pt 197.418} & \cellcolor{lightblue}\textbf{\kern1pt 0.022} & \cellcolor{lightblue}\textbf{\kern1pt 0.008} & 
            \cellcolor{lightblue}{\kern0.1pt 0.321} & 
            \cellcolor{lightblue}{\kern0.1pt 0.309} \\
        \bottomrule
    \end{tabular}
  \label{tab: exp_conf_ensemble}
\end{table}

\begin{table}[t]
  \centering
  \caption{\textbf{\thetable:} The structural comparison of different methods on the polymer conformation generation task in terms of Coverage (\%) and Matching (\AA), where we compute Coverage with a threshold of $\delta=25$~\AA\ to distinguish top methods better.
  \rre{Here, all reported results are evaluated on the test set.}}
    \begin{tabular}{lcccccccc}
    \toprule
    \multirow{3}[2]{*}{Method} & \multicolumn{4}{c}{{Recall}}    & \multicolumn{4}{c}{{Precision}} \\
          & \multicolumn{2}{c}{S-COV-R $\uparrow$} & \multicolumn{2}{c}{S-MAT-R $\downarrow$} & \multicolumn{2}{c}{S-COV-P $\uparrow$} & \multicolumn{2}{c}{S-MAT-P $\downarrow$} \\
          \cmidrule(lr){2-3} \cmidrule(lr){4-5} \cmidrule(lr){6-7} \cmidrule(lr){8-9} & Mean  & Median & Mean  & Median & Mean  & Median & Mean  & Median \\
    \midrule
    GeoDiff~\cite{xu2022geodiff} & 0.108  & 0.000  & 93.119  & 89.767  & 0.008  & 0.000  & 95.259  & 91.869  \\
    TorsionalDiff~\cite{jing2022torsional} & {0.172}  & 0.000  & {53.210}  & {38.710}  & {0.100}  & 0.000  & {70.679}  & {60.744}  \\
    MCF~\cite{wang2024swallowing}   & 0.000  & 0.000  & 248.432  & 242.866  & 0.000  & 0.000  & 258.891  & 253.239  \\
    ET-Flow~\cite{hassan2024flow} & 0.089  & 0.000  & 94.057  & 90.475  & 0.064  & 0.000  & 96.896  & 92.877  \\
    \cellcolor{lightblue} \kern-2.5pt PolyConFM & \cellcolor{lightblue} \kern-2.5pt \textbf{0.515} & \cellcolor{lightblue} \kern-2.5pt \textbf{1.000} & \cellcolor{lightblue} \kern-2.5pt \textbf{35.021} & \cellcolor{lightblue} \kern-2.5pt \textbf{24.279} & \cellcolor{lightblue} \kern-2.5pt \textbf{0.336} & \cellcolor{lightblue} \kern-2.5pt \textbf{0.100} & \cellcolor{lightblue} \kern-2.5pt \textbf{46.861} & \cellcolor{lightblue} \kern-2.5pt \textbf{37.996} \\
    \bottomrule
    \end{tabular}
  \label{tab: exp_conf_add}
\end{table}

In addition, as mentioned in Section~\ref{sec: exp_setup_metrics}, although the Coverage metric that relies on a fixed RMSD threshold $\delta$ for structural comparison has been widely adopted in the small molecule domain~\cite{xu2022geodiff}, it's unsuitable for polymer conformation generation since polymers exhibit a far larger conformational space with significant diversity arising from their chain length, flexibility, and repeating units, making a single fixed threshold inadequate for meaningful coverage assessment.
Therefore, we exclude it from our evaluation metrics but report the corresponding performance of various methods under this metric here for reference.
As presented in~\ref{tab: exp_conf_add}, PolyConFM still significantly outperforms all baselines even when incorporating the Coverage metric.
In particular, PolyConFM achieves the highest S-COV-R of 0.515 (mean) and 1.000 (median) for recall while maintaining strong superiority in precision with S-COV-P of 0.336 (mean) and 0.100 (median).
These results demonstrate that PolyConFM generates polymer conformations with superior structural coverage over the reference set, despite the inherent difficulties posed by the flexibility and variability of polymer systems.

\subsection{Polymer Property Prediction}\label{SI-sec: exp_property}
\begin{figure}[t]
  \centering
  \subfigure[Egc]{
    \includegraphics[height=2.8cm]{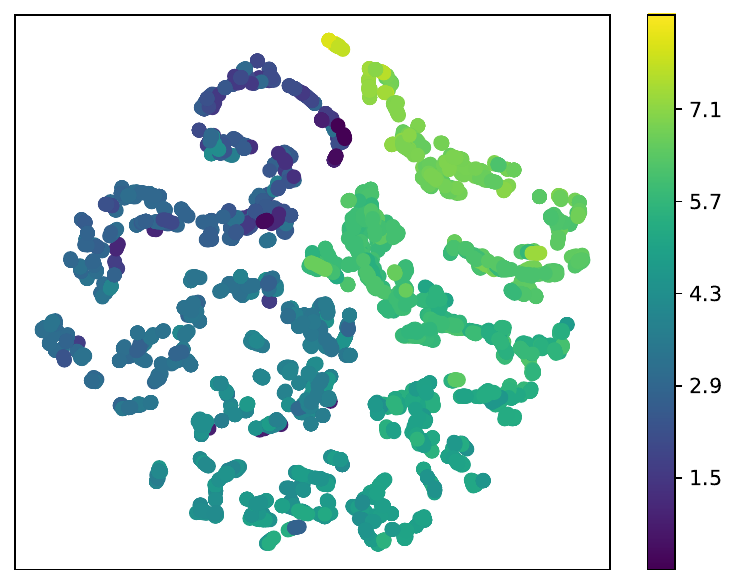}
  }
  \subfigure[Egb]{
    \includegraphics[height=2.8cm]{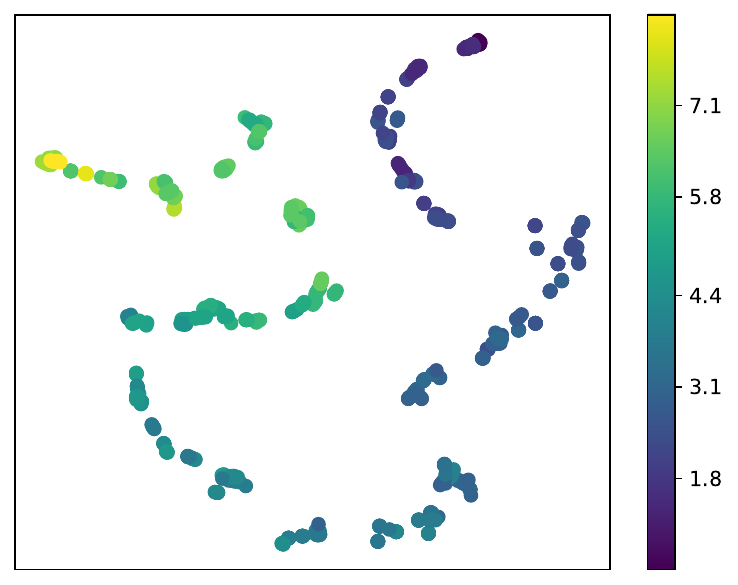}
  }
  \subfigure[Eea]{
    \includegraphics[height=2.8cm]{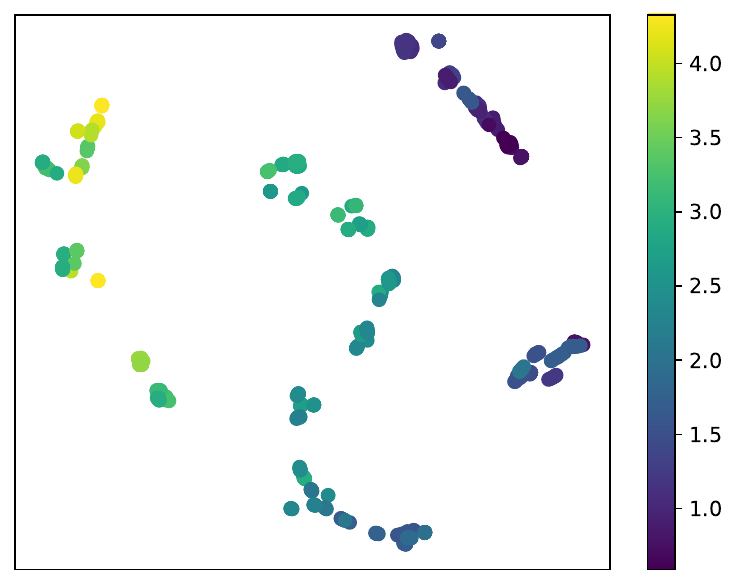}
  }
  \subfigure[Ei]{
    \includegraphics[height=2.8cm]{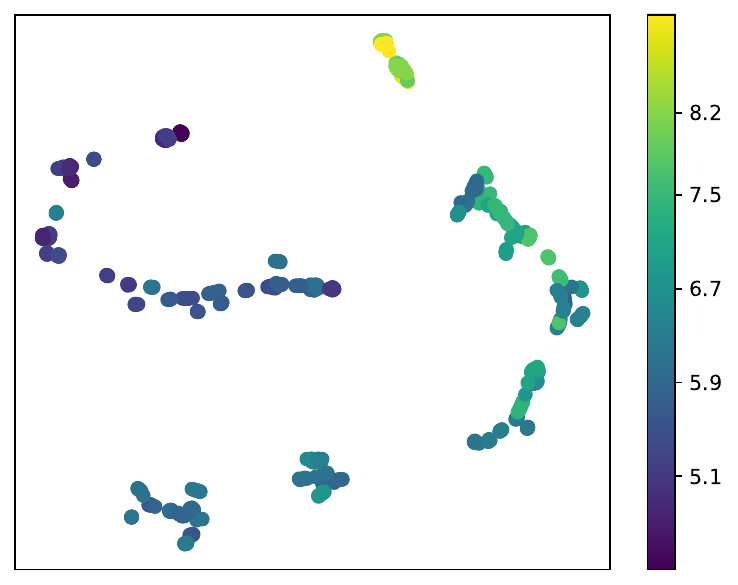}
  }
  \par
  \subfigure[Xc]{
    \includegraphics[height=2.8cm]{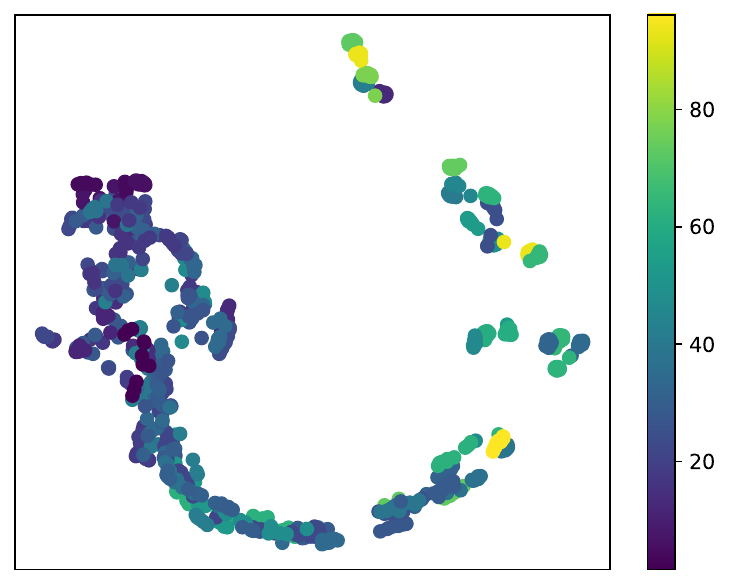}
  }
  \subfigure[EPS]{
    \includegraphics[height=2.8cm]{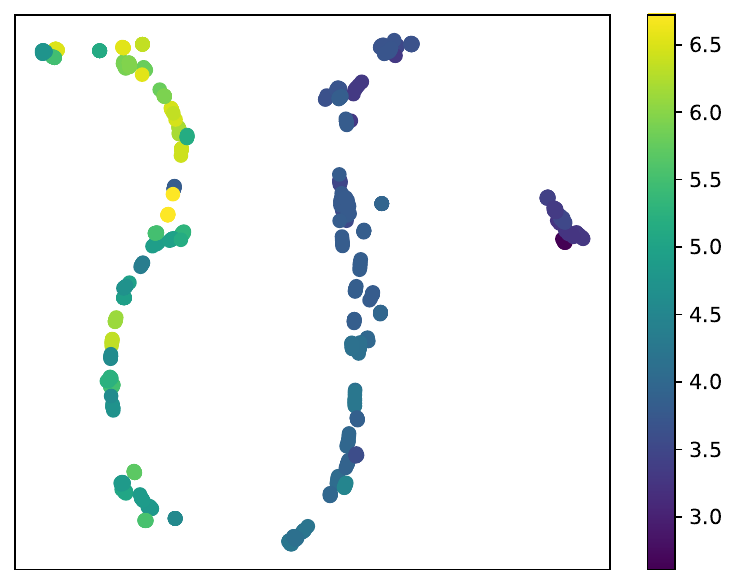}
  }
  \subfigure[Nc]{
    \includegraphics[height=2.8cm]{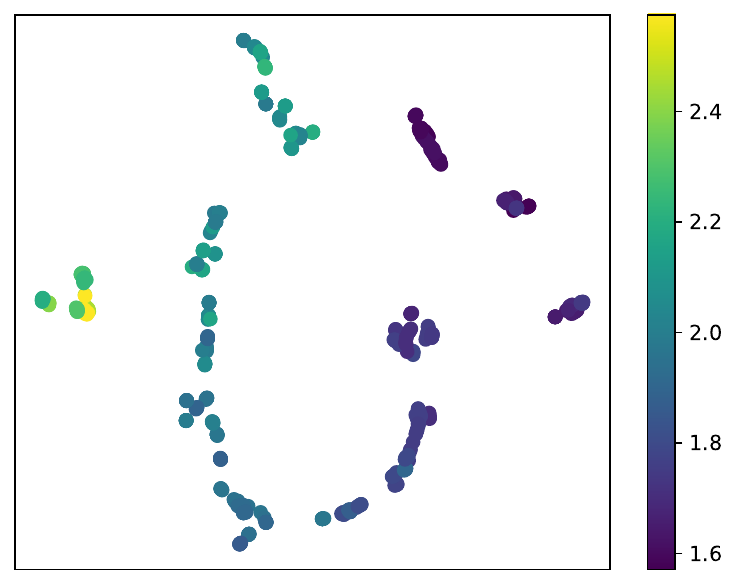}
  }
  \subfigure[Eat]{
    \includegraphics[height=2.8cm]{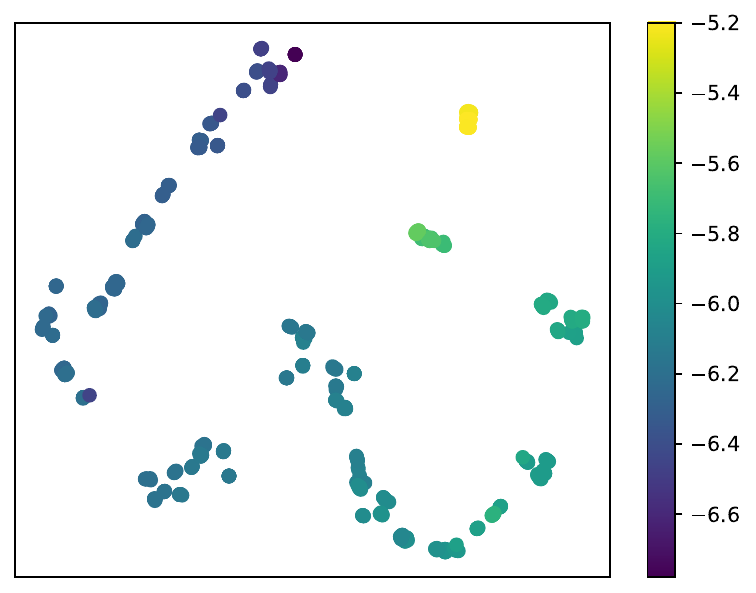}
  }
  \caption{\textbf{\thefigure:} The t-SNE visualization of PolyConFM on the downstream polymer property prediction task, where the ground‑truth property values determine point colors.
  \rre{Here, all reported results are evaluated on the test set.}} 
  \label{fig: exp_property_tsne}
\end{figure}

\ref{fig: exp_property_tsne} presents t-SNE visualization of polymer embeddings learned by PolyConFM on the downstream polymer property prediction task, with point colors indicating ground‑truth property values. 
In particular, these polymer embeddings exhibit coherent manifold structure with smooth value gradients and clear separation between regions of high and low values, evidencing superior property alignment and discriminative capacity.
Moreover, this geometry is consistent across diverse polymer properties, suggesting that PolyConFM indeed captures transferable and property‑relevant factors of variation and thus reliably distinguishes polymers with differing property levels.
Taken together, these observations demonstrate that PolyConFM embodies strong inductive biases toward structure–property relationships, thereby underpinning its leading performance in polymer property prediction.

\begin{figure}[t]
  \centering
  \subfigure[Egc]{
    \includegraphics[height=2.8cm]{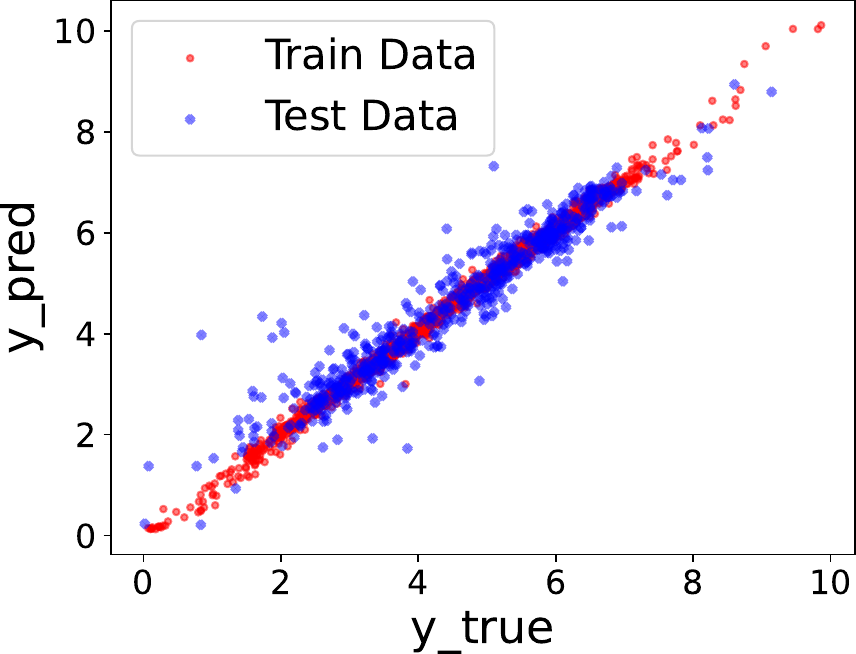}
  }
  \subfigure[Egb]{
    \includegraphics[height=2.8cm]{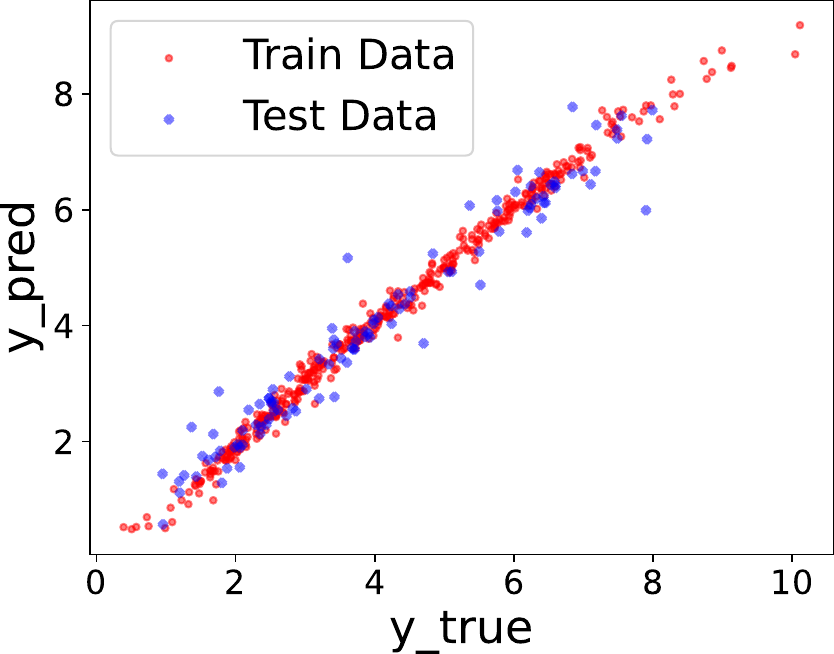}
  }
  \subfigure[Eea]{
    \includegraphics[height=2.8cm]{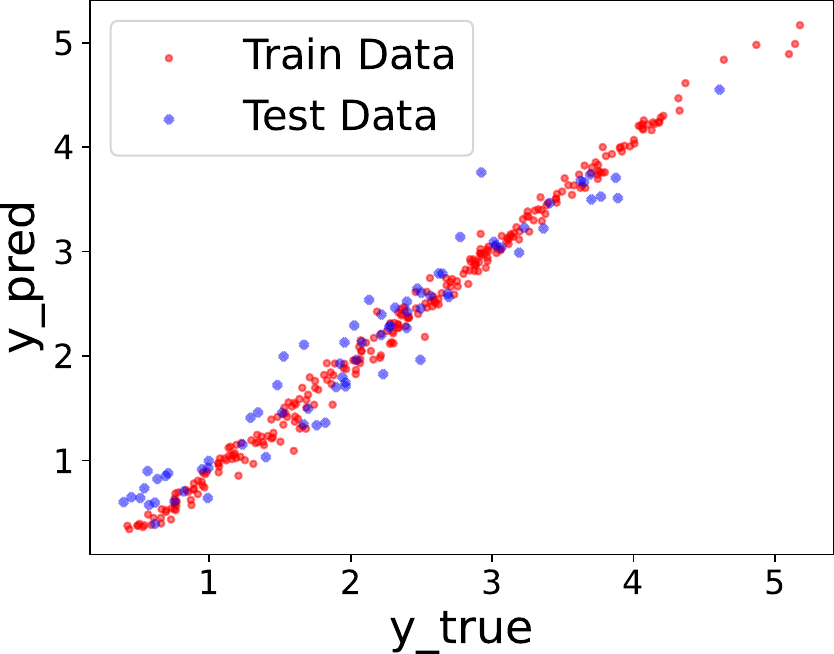}
  }
  \subfigure[Ei]{
    \includegraphics[height=2.8cm]{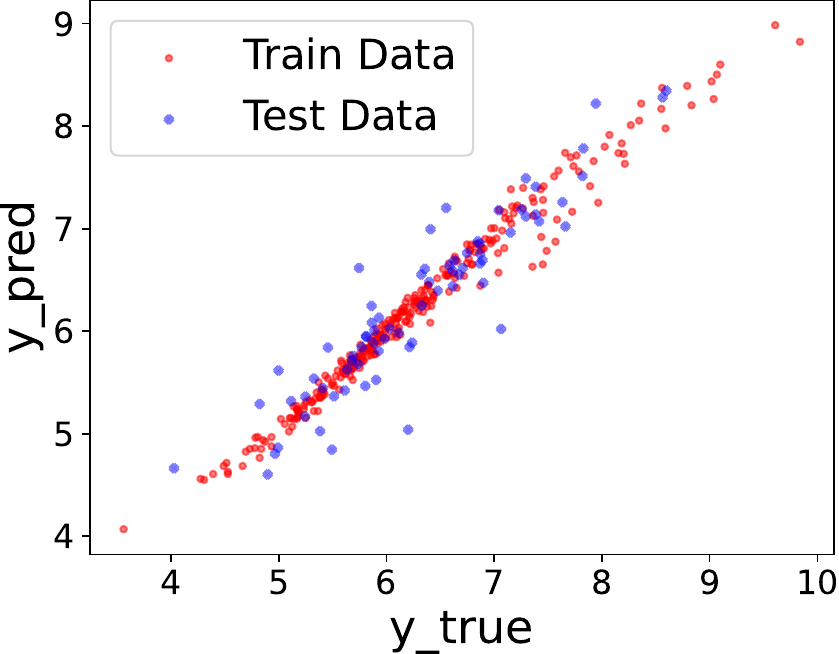}
  }
  \subfigure[Xc]{
    \includegraphics[height=2.75cm]{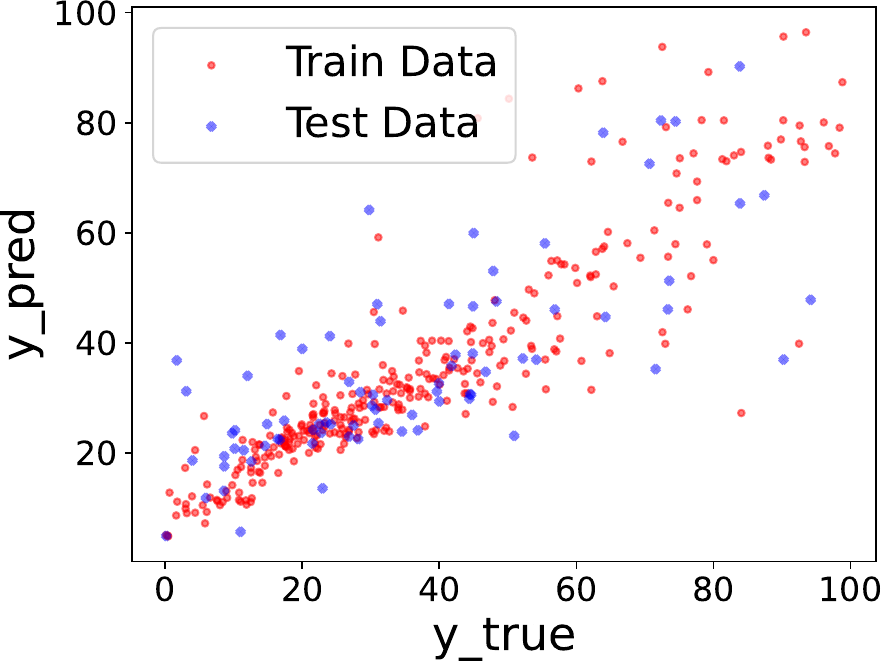}
  }
  \subfigure[EPS]{
    \includegraphics[height=2.75cm]{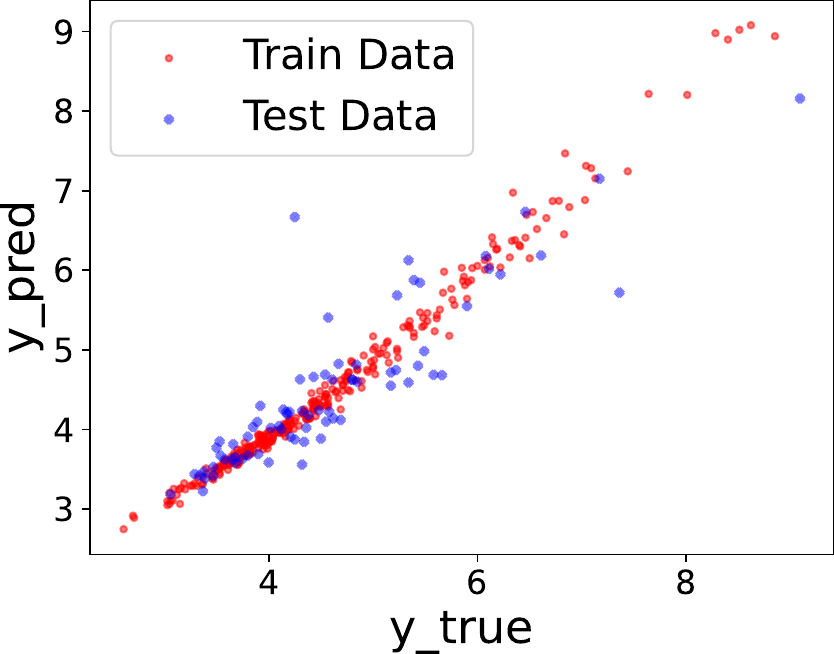}
  }
  \subfigure[Nc]{
    \includegraphics[height=2.75cm]{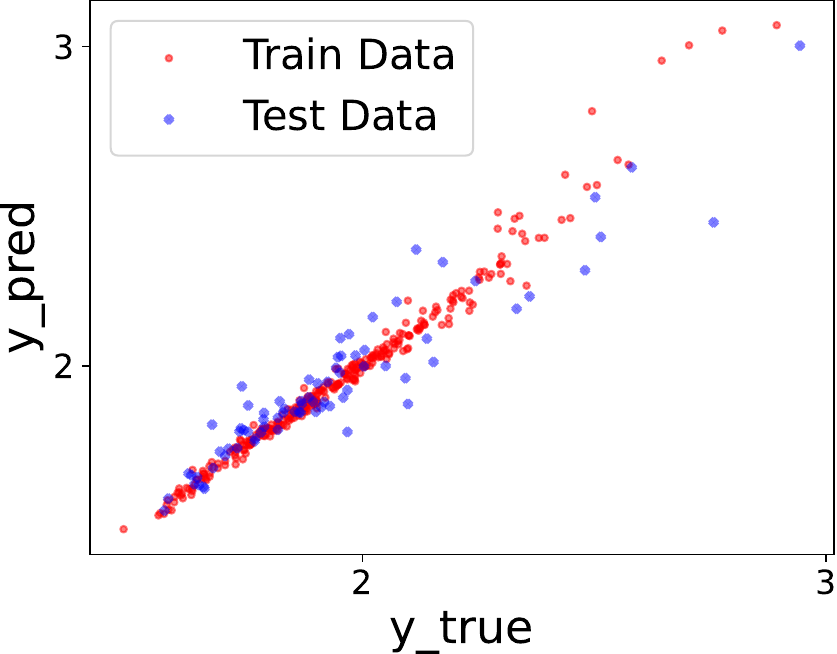}
  }
  \subfigure[Eat]{
    \includegraphics[height=2.75cm]{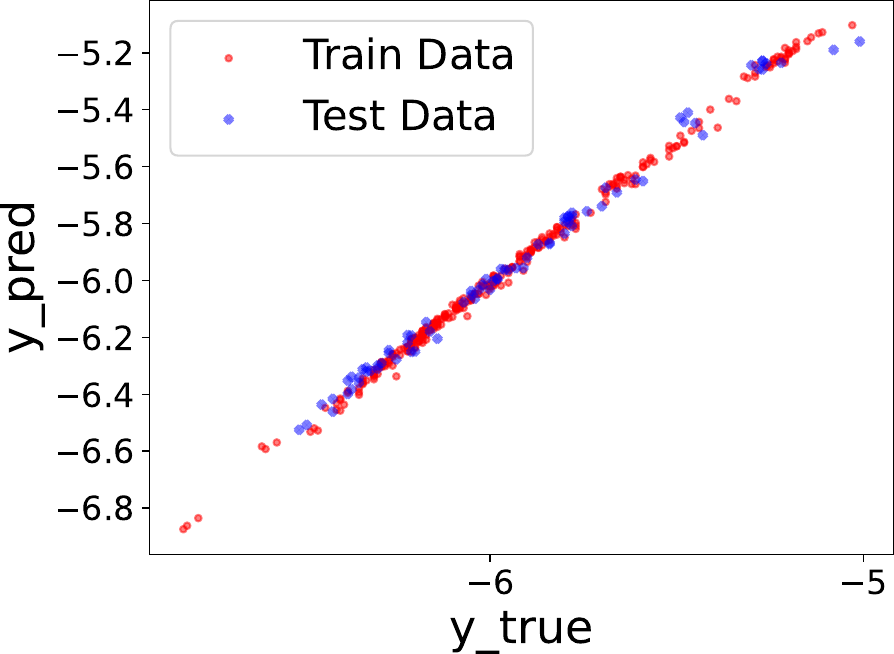}
  }
  \caption{\textbf{\thefigure:} The scatter plots of PolyConFM on the downstream polymer property prediction task, covering eight typical polymer property datasets.} 
  \label{fig: exp_property_scatter}
\end{figure}

In addition, \ref{fig: exp_property_scatter} presents scatter plots comparing PolyConFM’s predicted property values with ground truth on the downstream polymer property prediction task.
Here, points cluster tightly around the identity line across diverse polymer properties, with train and test samples substantially overlapping, indicating consistent generalization and minimal distribution shift between splits.
In particular, properties with narrow dynamic ranges (e.g., Eat) adhere very closely to the identity line, and those with broader ranges (e.g., Egc) also exhibit an approximate linear trend with only a few extreme outliers, suggesting limited heteroscedastic error.
Collectively, these results corroborate PolyConFM’s stable calibration across properties and robust generalization under varying value scales, thereby underpinning its reliable performance in polymer property prediction.

\begin{table*}[t]
    \setlength{\tabcolsep}{0.15em}
    \caption{\textbf{\thetable:} \re{The performance comparison of different methods on the downstream polymer property prediction task, where we utilized the \textbf{cluster-based dataset splitting strategy} and report the mean $\pm$ standard deviation over five independent runs with different random seeds.}
    \rre{Here, all reported results are evaluated on the test set.}}
    \centering
    \begin{footnotesize}
    \scalebox{1}{
    \begin{tabular}{llcccccccc}
    \toprule
     & Method & Egc   & Egb   & Eea   & Ei    & Xc    & EPS   & Nc   & Eat \\
    \midrule
    \multirow{4}{*}{\begin{sideways}RMSE $\downarrow$\end{sideways}}
        
        & Uni-Mol~\cite{zhou2023unimol} & 0.715$_{\pm\text{0.025}}$ & 0.965$_{\pm\text{0.164}}$  & 0.512$_{\pm\text{0.030}}$ & 0.515$_{\pm\text{0.010}}$ & 24.343$_{\pm\text{0.956}}$ & 0.446$_{\pm\text{0.038}}$ & 0.081$_{\pm\text{0.013}}$ & 0.146$_{\pm\text{0.040}}$  \\
        
        & Transpolymer~\cite{xu2023transpolymer} & 0.691$_{\pm\text{0.024}}$ & 0.927$_{\pm\text{0.051}}$ & 0.453$_{\pm\text{0.049}}$ & 0.438$_{\pm\text{0.066}}$ & 21.897$_{\pm\text{0.804}}$ & 0.436$_{\pm\text{0.038}}$ & 0.083$_{\pm\text{0.010}}$ & 0.134$_{\pm\text{0.011}}$ \\
        
        & MMPolymer~\cite{wang2024mmpolymer} & 0.709$_{\pm\text{0.043}}$ & 
        0.921$_{\pm\text{0.063}}$  & 0.412$_{\pm\text{0.076}}$  & 0.465$_{\pm\text{0.123}}$ & 21.281$_{\pm\text{0.301}}$ & 0.375$_{\pm\text{0.056}}$ & 
        0.070$_{\pm\text{0.007}}$ & 0.099$_{\pm\text{0.018}}$ \\

        & \cellcolor{lightblue} \kern-2.5pt PolyConFM & 
        {\cellcolor{lightblue}\textbf{0.648}$_{\pm\text{0.015}}$} & {\cellcolor{lightblue}\textbf{0.816}$_{\pm\text{0.186}}$} & {\cellcolor{lightblue}\textbf{0.324}$_{\pm\text{0.033}}$} & {\cellcolor{lightblue}\textbf{0.359}$_{\pm\text{0.026}}$} & {\cellcolor{lightblue}\textbf{19.569}$_{\pm\text{1.151}}$} & {\cellcolor{lightblue}\textbf{0.277}$_{\pm\text{0.018}}$} & 
        {\cellcolor{lightblue}\textbf{0.054}$_{\pm\text{0.006}}$} & {\cellcolor{lightblue}\textbf{0.083}}$_{\pm\text{0.006}}$ \\
    \midrule
    \multirow{4}{*}{\begin{sideways}$R^2$ $\uparrow$ \end{sideways}}
          
      & Uni-Mol~\cite{zhou2023unimol} & 0.804$_{\pm\text{0.014}}$ & 0.731$_{\pm\text{0.096}}$  & 0.721$_{\pm\text{0.032}}$ & 0.454$_{\pm\text{0.022}}$ & 0.112$_{\pm\text{0.068}}$ & 0.502$_{\pm\text{0.085}}$ & 0.695$_{\pm\text{0.100}}$ & 0.830$_{\pm\text{0.095}}$  \\
      
      & Transpolymer~\cite{xu2023transpolymer} &
     0.817$_{\pm\text{0.013}}$ & 0.757$_{\pm\text{0.027}}$ & 0.780$_{\pm\text{0.048}}$ & 0.598$_{\pm\text{0.117}}$ & 0.282$_{\pm\text{0.053}}$ & 0.524$_{\pm\text{0.084}}$ & 0.685$_{\pm\text{0.075}}$ & 0.863$_{\pm\text{0.022}}$ \\
      
      & MMPolymer~\cite{wang2024mmpolymer} & 
      0.807$_{\pm\text{0.023}}$ & 0.760$_{\pm\text{0.034}}$ & 0.814$_{\pm\text{0.071}}$ & 0.530$_{\pm\text{0.247}}$ & 
      0.322$_{\pm\text{0.019}}$ & 0.644$_{\pm\text{0.105}}$ & 0.775$_{\pm\text{0.045}}$ & 0.924$_{\pm\text{0.029}}$ \\

      & \cellcolor{lightblue} \kern-2.5pt PolyConFM & 
      {\cellcolor{lightblue}\textbf{0.839}$_{\pm\text{0.008}}$} & {\cellcolor{lightblue}\textbf{0.804}$_{\pm\text{0.096}}$} & {\cellcolor{lightblue}\textbf{0.888}$_{\pm\text{0.023}}$} & {\cellcolor{lightblue}\textbf{0.734}$_{\pm\text{0.039}}$} & 
      {\cellcolor{lightblue}\textbf{0.426}$_{\pm\text{0.069}}$} & {\cellcolor{lightblue}\textbf{0.808}$_{\pm\text{0.025}}$} & {\cellcolor{lightblue}\textbf{0.865}$_{\pm\text{0.028}}$} & {\cellcolor{lightblue}\textbf{0.948}}$_{\pm\text{0.008}}$ \\
    \bottomrule
    \end{tabular}}
    \end{footnotesize}
  \label{tab:cluster_results}
\end{table*}

\begin{figure*}[ht]
    \centering
    \begin{subfigure}
        \centering
        \includegraphics[width=0.999\textwidth]{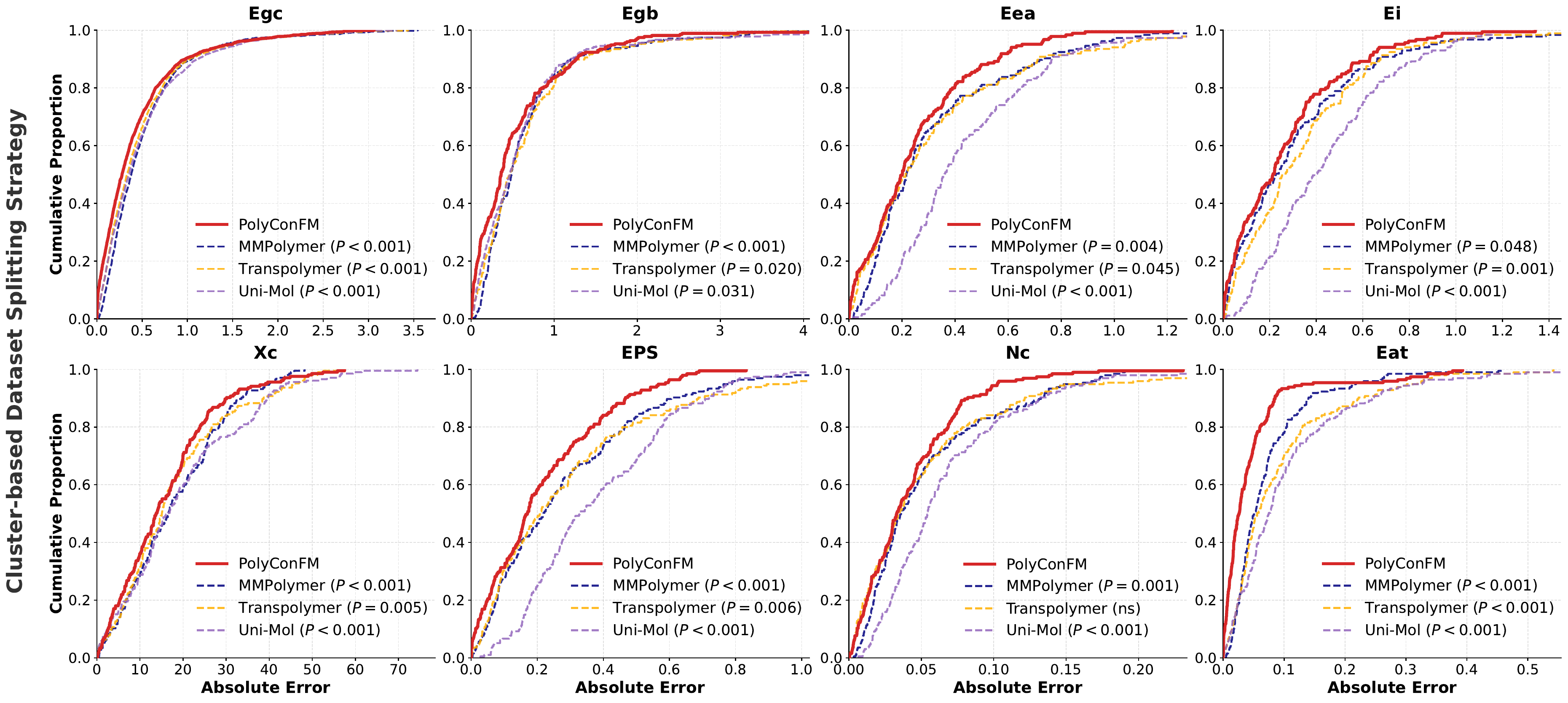}
    \end{subfigure}
    \begin{subfigure}
        \centering
        \includegraphics[width=0.999\textwidth]{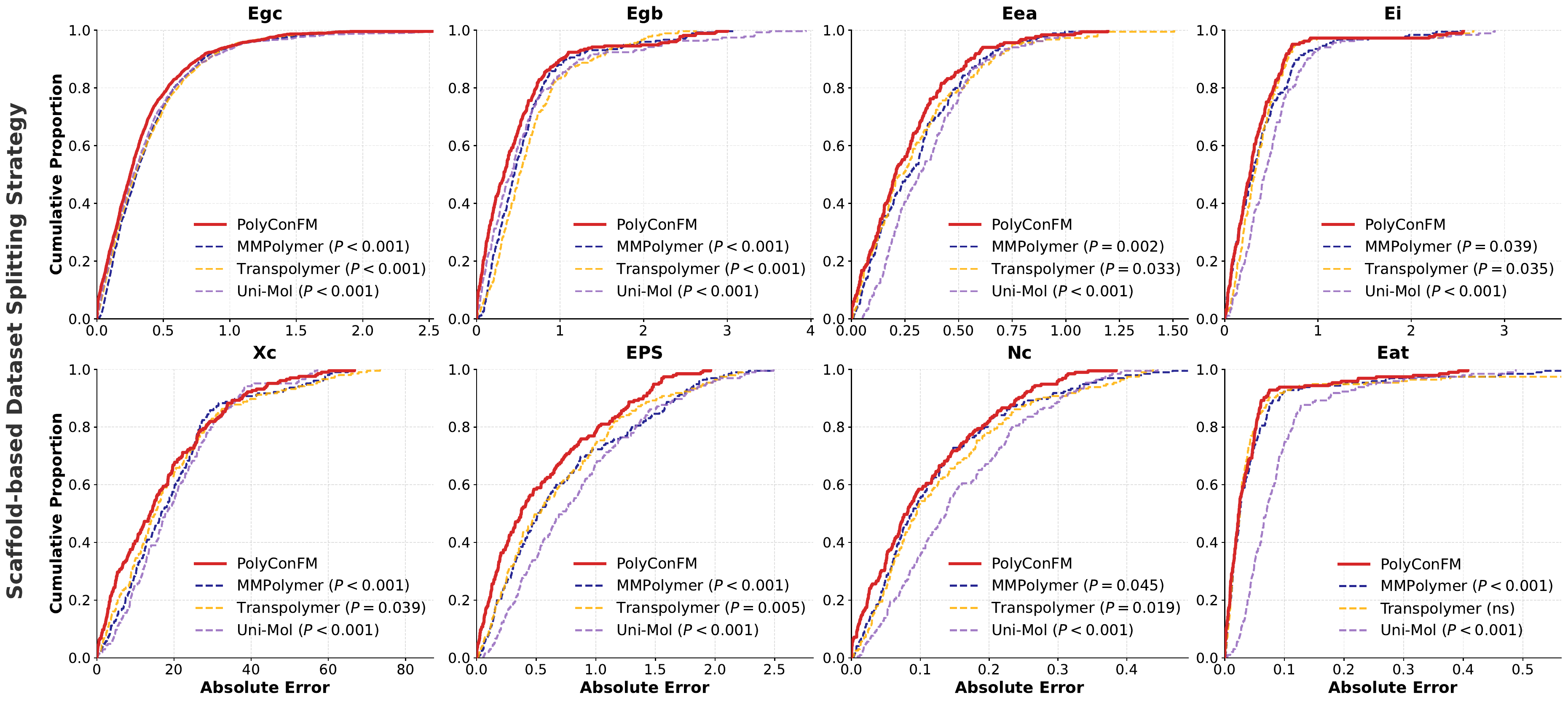}
    \end{subfigure}
   \caption{\textbf{\thefigure:} \re{The empirical cumulative distribution functions
   of prediction errors on the downstream polymer property prediction task, where we compare PolyConFM against baselines under both cluster-based dataset splitting strategy (top two rows) and scaffold-based dataset splitting strategy (bottom two rows) across eight representative polymer property datasets. 
   In each panel, a curve positioned closer to the upper-left corner indicates a higher proportion of test samples with lower absolute errors, signifying superior predictive accuracy.
   We report $P$-values from one-sided Wilcoxon signed-rank tests in the legends to assess the statistical significance of PolyConFM’s improvements over each baseline (ns: not significant, i.e., $P \ge 0.05$), demonstrating that PolyConFM achieves statistically meaningful gains ($P < 0.05$ in most cases) under these challenging structural generalization scenarios.}
   \rre{Here, all reported results are evaluated on the test set.}
   }
    \label{fig:property_cdf}
\end{figure*}

Since models are often tasked with predicting properties for polymers that differ significantly from those in the training set when deployed in practice, we further incorporate two more challenging dataset splitting strategies to strictly benchmark generalization capabilities:
\begin{itemize}
    \item \textbf{Cluster-based Dataset Splitting Strategy:} To rigorously prevent data leakage from homologous series, we utilize Butina clustering based on Morgan fingerprints (radius 2, 1024 bits, Tanimoto cutoff 0.4). With the target splitting ratio of approximately 8:1:1, the training set is populated primarily with the largest clusters to cover the dominant chemical space, and the remaining clusters, which are structurally distinct from the training samples, are allocated to the validation and test sets, thereby ensuring that the model is evaluated on unseen polymer structures.
    \item \textbf{Scaffold-based Dataset Splitting Strategy:} To rigorously test the model's capacity for chemical extrapolation, we partition the datasets based on Bemis-Murcko scaffolds. With the target splitting ratio of approximately 8:1:1, the training set is populated primarily with the most frequent scaffolds to cover the common structural frameworks, and the remaining data, composed of unique scaffolds absent from the training set, is partitioned into validation and test sets, thereby ensuring that the model is evaluated on unseen polymer frameworks.
\end{itemize}
Here, we directly compare PolyConFM against three most competitive baselines (i.e., Uni-Mol, Transpolymer, and MMPolymer) identified in Table~\ref{tab: exp_proprety} under both challenging splitting strategies.
As presented in~\ref{tab:cluster_results} and~\ref{tab:scaffold_results}, PolyConFM still consistently outperforms these baselines across all evaluation metrics on all property datasets.
In particular, under the cluster-based dataset splitting strategy, PolyConFM exhibits superior robustness against structural homology, yielding substantial gains across various properties.
This performance indicates that PolyConFM effectively mitigates the performance inflation caused by homologous data leakage, demonstrating a genuine understanding of diverse chemical spaces.
Similarly, under the scaffold-based dataset splitting strategy, while the performance of all baselines significantly declines when tasked with predicting properties for unseen polymer frameworks, PolyConFM preserves a substantial performance margin across various properties, underscoring its exceptional ability to capture structure–property relationships.
Meanwhile, considering the relatively small scale of these property datasets, reporting only the mean and standard deviation may be insufficient to guarantee the robustness of the observed performance improvements.
To provide a more rigorous evaluation, we further perform significance testing by aggregating the absolute prediction errors from five independent runs for each method.
As presented by the empirical cumulative distribution functions of prediction errors in~\ref{fig:property_cdf}, PolyConFM consistently maintains a higher proportion of test samples with lower absolute errors compared to all baselines, as indicated by its curves being positioned closest to the upper-left corner.
One-sided Wilcoxon signed-rank tests between the error distributions of PolyConFM and each baseline also confirm that improvements achieved by PolyConFM are statistically significant ($P < 0.05$ in most cases) under both splitting strategies.
These comprehensive evaluations consistently validate the superior performance of PolyConFM, thereby establishing its reliability as a powerful tool for property prediction.

\begin{table*}[t]
    \setlength{\tabcolsep}{0.15em}
    \caption{\textbf{\thetable:} \re{The performance comparison of different methods on the downstream polymer property prediction task, where we utilized the \textbf{scaffold-based dataset splitting strategy} and report the mean $\pm$ standard deviation over five independent runs with different random seeds.}
    \rre{Here, all reported results are evaluated on the test set.}}
    \centering
    \begin{footnotesize}
    \scalebox{1}{
    \begin{tabular}{llcccccccc}
    \toprule
     & Method & Egc   & Egb   & Eea   & Ei    & Xc    & EPS   & Nc   & Eat \\
    \midrule
    \multirow{4}{*}{\begin{sideways}RMSE $\downarrow$\end{sideways}}
        
        & Uni-Mol~\cite{zhou2023unimol} & 0.612$_{\pm\text{0.015}}$ & 0.923$_{\pm\text{0.054}}$  & 0.418$_{\pm\text{0.014}}$ & 0.686$_{\pm\text{0.068}}$ & 23.716$_{\pm\text{1.000}}$ & 1.001$_{\pm\text{0.131}}$ & 0.184$_{\pm\text{0.009}}$ & 0.121$_{\pm\text{0.013}}$  \\
        
        & Transpolymer~\cite{xu2023transpolymer} & 0.544$_{\pm\text{0.009}}$ & 0.817$_{\pm\text{0.034}}$ & 0.401$_{\pm\text{0.061}}$ & 0.567$_{\pm\text{0.010}}$ & 24.488$_{\pm\text{1.319}}$ & 0.891$_{\pm\text{0.041}}$ & 0.167$_{\pm\text{0.012}}$ & 0.131$_{\pm\text{0.005}}$ \\
        
        & MMPolymer~\cite{wang2024mmpolymer} & 0.547$_{\pm\text{0.025}}$ & 
        0.794$_{\pm\text{0.064}}$  & 0.379$_{\pm\text{0.021}}$  & 0.575$_{\pm\text{0.019}}$ & 24.069$_{\pm\text{0.680}}$ & 0.927$_{\pm\text{0.139}}$ & 
        0.158$_{\pm\text{0.010}}$ & 0.102$_{\pm\text{0.009}}$ \\

        & \cellcolor{lightblue} \kern-2.5pt PolyConFM & 
        {\cellcolor{lightblue}\textbf{0.498}$_{\pm\text{0.017}}$} & {\cellcolor{lightblue}\textbf{0.752}$_{\pm\text{0.045}}$} & {\cellcolor{lightblue}\textbf{0.344}$_{\pm\text{0.035}}$} & {\cellcolor{lightblue}\textbf{0.532}$_{\pm\text{0.042}}$} & {\cellcolor{lightblue}\textbf{22.055}$_{\pm\text{1.783}}$} & {\cellcolor{lightblue}\textbf{0.742}$_{\pm\text{0.068}}$} & 
        {\cellcolor{lightblue}\textbf{0.138}$_{\pm\text{0.010}}$} & {\cellcolor{lightblue}\textbf{0.078}}$_{\pm\text{0.005}}$ \\
    \midrule
    \multirow{4}{*}{\begin{sideways}$R^2$ $\uparrow$\end{sideways}}
          
      & Uni-Mol~\cite{zhou2023unimol} & 0.710$_{\pm\text{0.015}}$ & 0.721$_{\pm\text{0.032}}$  & 0.807$_{\pm\text{0.013}}$ & 0.234$_{\pm\text{0.155}}$ & 0.076$_{\pm\text{0.078}}$ & 0.302$_{\pm\text{0.184}}$ & 0.387$_{\pm\text{0.062}}$ & 0.457$_{\pm\text{0.123}}$  \\
      
      & Transpolymer~\cite{xu2023transpolymer} &
     0.771$_{\pm\text{0.008}}$ & 0.782$_{\pm\text{0.018}}$ & 0.820$_{\pm\text{0.054}}$ & 0.479$_{\pm\text{0.018}}$ & 0.014$_{\pm\text{0.108}}$ & 0.453$_{\pm\text{0.050}}$ & 0.494$_{\pm\text{0.071}}$ & 0.363$_{\pm\text{0.045}}$ \\
      
      & MMPolymer~\cite{wang2024mmpolymer} & 
      0.768$_{\pm\text{0.021}}$ & 0.793$_{\pm\text{0.033}}$ & 0.842$_{\pm\text{0.018}}$ & 0.465$_{\pm\text{0.035}}$ & 
      0.049$_{\pm\text{0.054}}$ & 0.399$_{\pm\text{0.174}}$ & 0.545$_{\pm\text{0.054}}$ & 0.616$_{\pm\text{0.067}}$ \\

      & \cellcolor{lightblue} \kern-2.5pt PolyConFM & 
      {\cellcolor{lightblue}\textbf{0.808}$_{\pm\text{0.013}}$} & {\cellcolor{lightblue}\textbf{0.815}$_{\pm\text{0.022}}$} & {\cellcolor{lightblue}\textbf{0.869}$_{\pm\text{0.027}}$} & {\cellcolor{lightblue}\textbf{0.540}$_{\pm\text{0.071}}$} & 
      {\cellcolor{lightblue}\textbf{0.198}$_{\pm\text{0.128}}$} & {\cellcolor{lightblue}\textbf{0.620}$_{\pm\text{0.070}}$} & {\cellcolor{lightblue}\textbf{0.651}$_{\pm\text{0.051}}$} & {\cellcolor{lightblue}\textbf{0.776}}$_{\pm\text{0.027}}$ \\
    \bottomrule
    \end{tabular}}
    \end{footnotesize}
  \label{tab:scaffold_results}
\end{table*}

\begin{figure*}[t]
   \centering
   \includegraphics[width=0.999\textwidth]{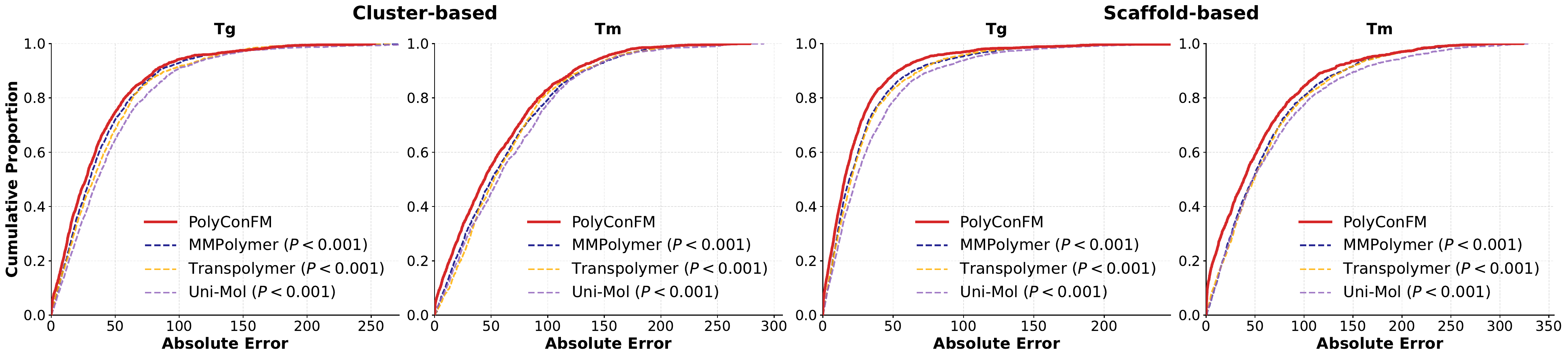}
   \caption{\textbf{\thefigure:} \re{The empirical cumulative distribution functions
   of prediction errors on the downstream polymer property prediction task, where we compare PolyConFM against baselines under both cluster-based dataset splitting strategy (left two panels) and scaffold-based dataset splitting strategy (right two panels) across two critical thermal properties: \textbf{Glass Transition Temperature} ($T_g$) and \textbf{Melting Temperature} ($T_m$).
   In each panel, a curve positioned closer to the upper-left corner indicates a higher proportion of test samples with lower absolute errors, signifying superior predictive accuracy.
   We report $P$-values from one-sided Wilcoxon signed-rank tests in the legends to assess the statistical significance of PolyConFM’s improvements over each baseline, demonstrating that PolyConFM achieves highly significant performance gains ($P < 0.001$ in all cases) under these challenging structural generalization scenarios.} 
   \rre{Here, all reported results are evaluated on the test set.}
   }
   \label{fig:property_Tg_Tm_cdf}
\end{figure*}

To rigorously validate the necessity and superiority of our conformation-centric modeling paradigm, we further extend property prediction experiments to two critical thermal properties: \textbf{Glass Transition Temperature} ($T_g$) and \textbf{Melting Temperature} ($T_m$).
Here, given their intrinsic dependence on global structural features, these thermal properties are highly conformation-sensitive and thus uniquely suited to demonstrate the powerful modeling capability of PolyConFM.
As presented in Table~\ref{tab:main_tg_tm_results}, PolyConFM consistently achieves state-of-the-art performance for both thermal properties across the two splitting strategies. 
In particular, under the scaffold-based dataset splitting strategy, PolyConFM reduces the RMSE by almost 12\% for $T_g$ and 9\% for $T_m$, while improving the $R^2$ by almost 8\% for $T_g$ and 15\% for $T_m$ compared to the best baseline.
Furthermore, we also illustrate the empirical cumulative distribution functions of prediction errors for both thermal properties in~\ref{fig:property_Tg_Tm_cdf} to provide a more rigorous evaluation.
Here, as indicated by the curves positioned closest to the upper-left corner, PolyConFM consistently maintains a significantly higher proportion of test samples with lower absolute errors.
Meanwhile, one-sided Wilcoxon signed-rank tests confirm that these performance gains are statistically substantial ($P < 0.001$ in all cases against all baselines), as detailed in the legends of each panel.
These comprehensive evaluations validate that our conformation-centric modeling paradigm indeed effectively captures the underlying physical dependencies governing complex polymer systems, highlighting its necessity and superiority for high-fidelity property prediction.
Notably, while traditional models mentioned in some existing works~\cite{averochkin2026machine, he2026bridging} could report better results (e.g., higher $R^2$ or lower RMSE) for $T_g$ and $T_m$, such outcomes are typically obtained under random-based dataset splitting, which fails to prevent significant structural similarities between training and test sets, thereby allowing models to perform simple interpolation within the familiar chemical space.
In contrast, this work compares PolyConFM and baselines under more rigorous dataset splitting (i.e., cluster-based and scaffold-based), which ensures that training and test sets inhabit distinct chemical regions, thereby forcing models to generalize to entirely unseen chemical families rather than performing simple interpolation.
Given that $T_g$ and $T_m$ are fundamentally governed by long-range conformations rather than local motifs, these results under these rigorous out-of-distribution (OOD) settings more faithfully reflect PolyConFM's predictive power.

Following the guidelines established in~\cite{ash2025practically}, we further incorporate 5$\times$5 cross-validation to facilitate more rigorous and robust comparisons.
Specifically, we integrate 5$\times$5 cross-validation with two advanced splitting strategies suggested in~\cite{ash2025practically}, serving the dual purpose of preventing structural data leakage and benchmarking generalization capabilities:
\begin{itemize}
    \item \textbf{Cluster-based 5$\times$5 Cross-Validation}: We first group structurally similar samples into the same fundamental unit using the Butina clustering algorithm based on Morgan fingerprints (radius 2, 1024 bits, Tanimoto cutoff 0.4). 
    By treating these fundamental units as indivisible blocks, we ensure that structurally similar samples are never split across different folds, thereby satisfying the requirement for generalization evaluation.
    Following this, we execute the standard $5\times5$ cross-validation procedure (i.e., 5 iterations of 5-fold cross-validation): In each iteration, these fundamental units are shuffled and greedily allocated into 5 folds to maintain balanced fold sizes, then the resulting folds are iteratively rotated to serve as the training, validation, and test sets.
    Notably, the entire pipeline strictly adheres to the guidelines established in~\cite{ash2025practically}, effectively preventing structural data leakage and rigorously evaluating out-of-distribution (OOD) generalization.
    \item \textbf{Scaffold-based 5$\times$5 Cross-Validation}: We first group samples sharing the same Bemis-Murcko scaffold into the same fundamental unit. 
    By treating these fundamental units as indivisible blocks, we ensure that structurally similar samples are never split across different folds, thereby satisfying the requirement for generalization evaluation.
    Following this, we execute the standard $5\times5$ cross-validation procedure (i.e., 5 iterations of 5-fold cross-validation): In each iteration, these fundamental units are shuffled and greedily allocated into 5 folds to maintain balanced fold sizes, then the resulting folds are iteratively rotated to serve as the training, validation, and test sets.
    Notably, the entire pipeline strictly adheres to the guidelines established in~\cite{ash2025practically}, effectively preventing structural data leakage and rigorously evaluating out-of-distribution (OOD) generalization.
\end{itemize}

\begin{figure}[H]
  \centering
  \subfigure[Egc]{
    \includegraphics[height=4.5cm]{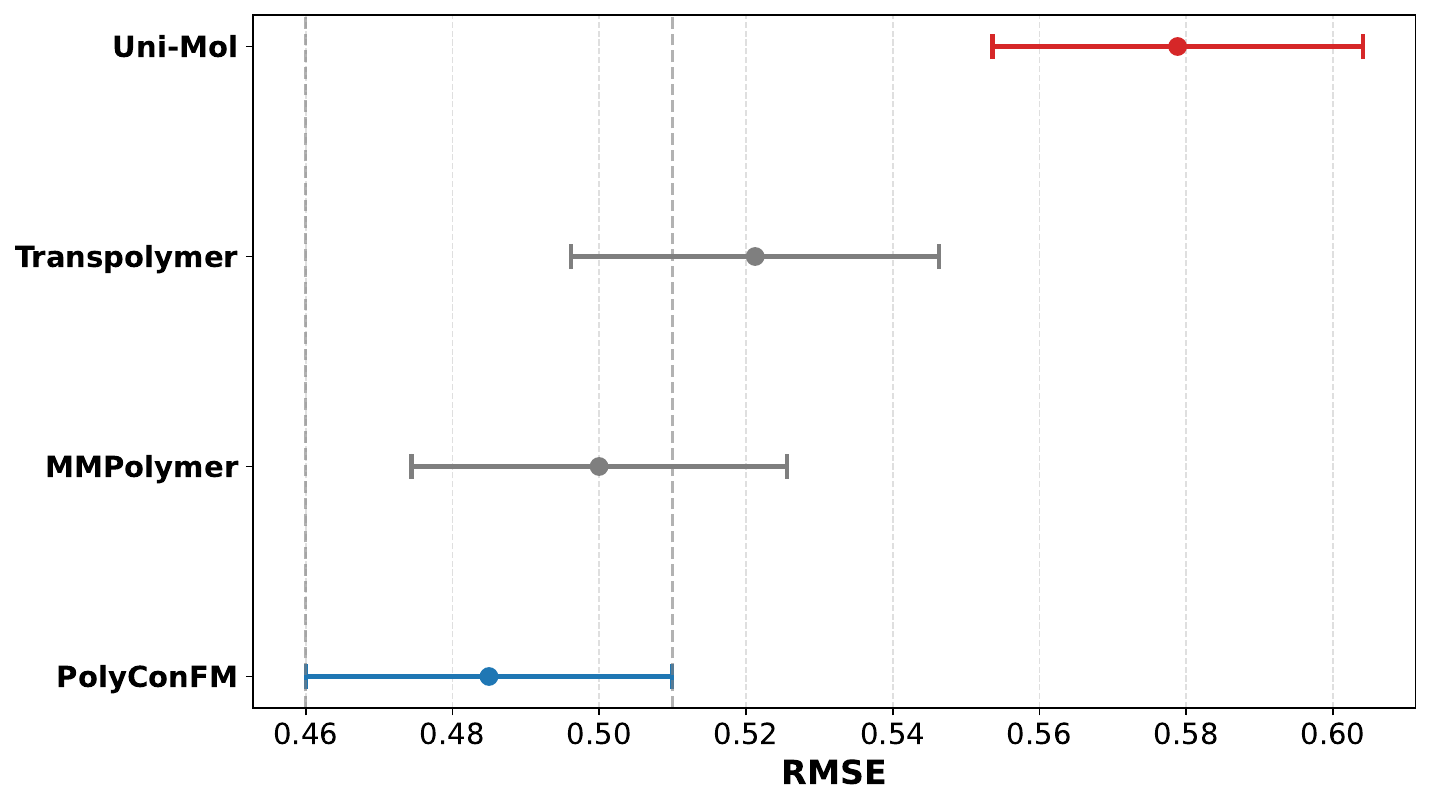}
  }
  \subfigure[Egb]{
    \includegraphics[height=4.5cm]{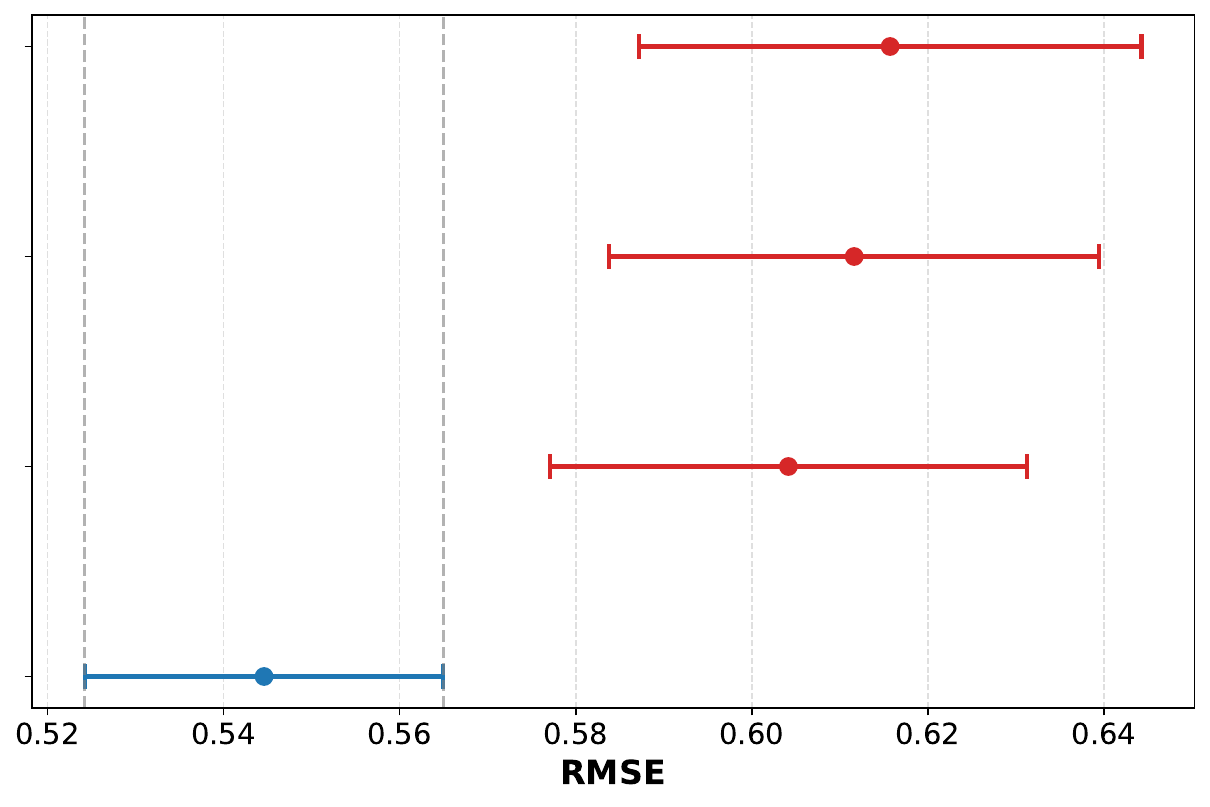}
  }
  \subfigure[Eea]{
    \includegraphics[height=4.5cm]{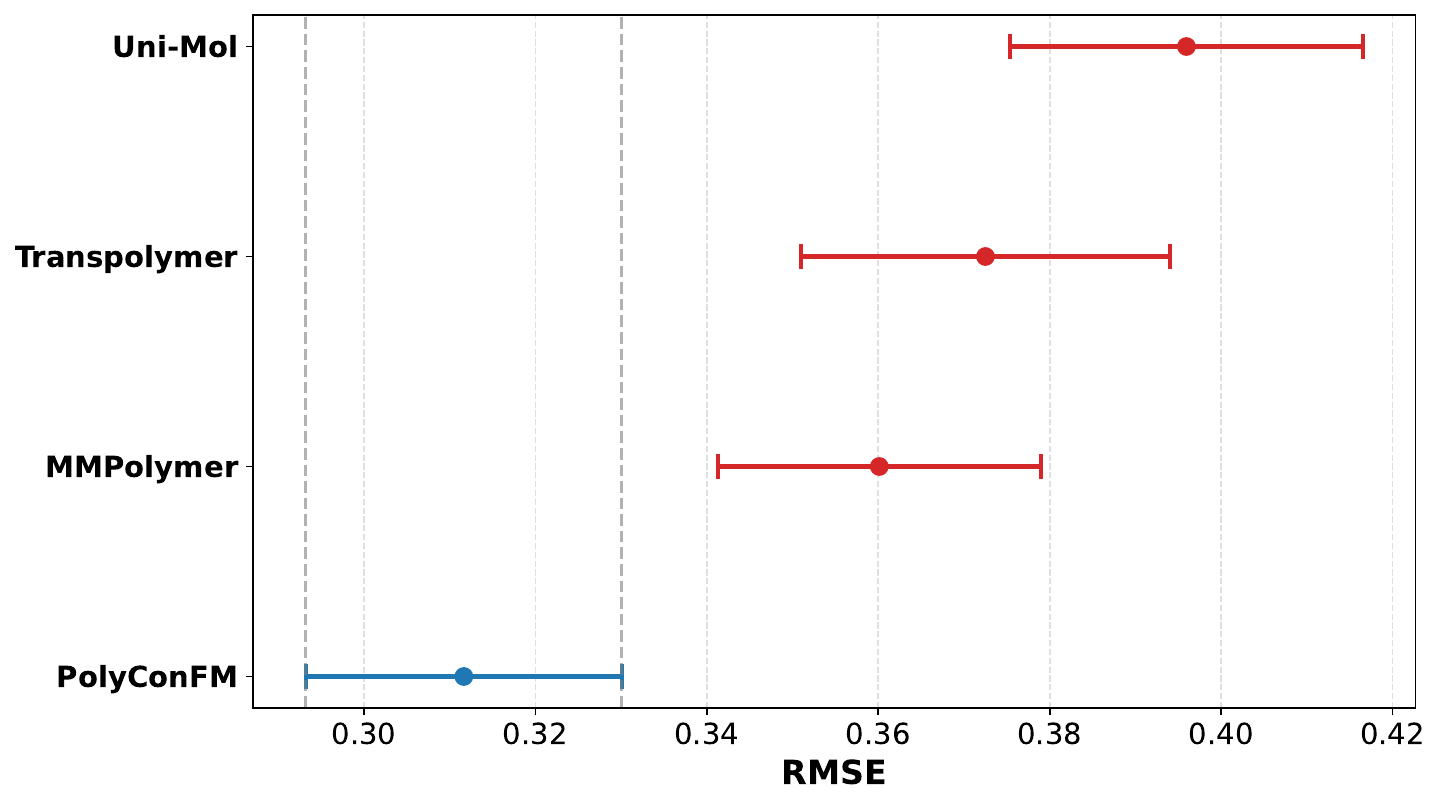}
  }
  \subfigure[Ei]{
    \includegraphics[height=4.5cm]{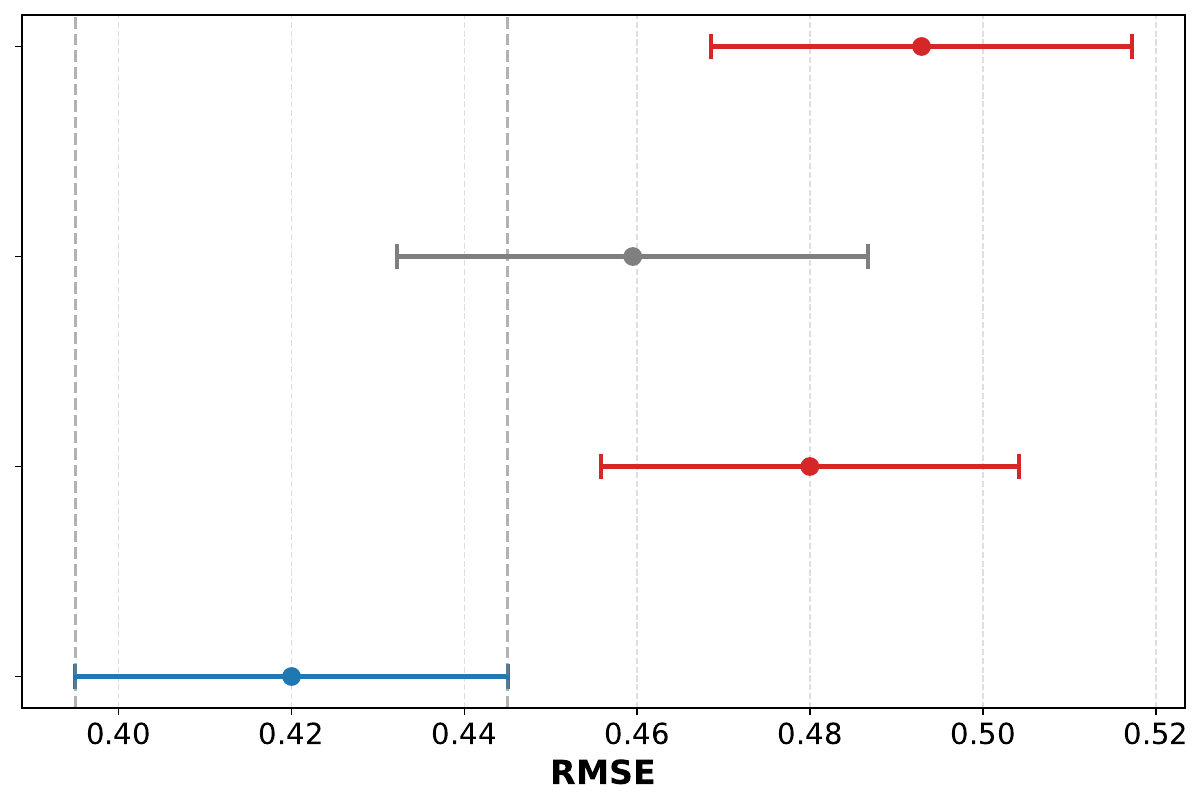}
  }
  \subfigure[Xc]{
    \includegraphics[height=4.5cm]{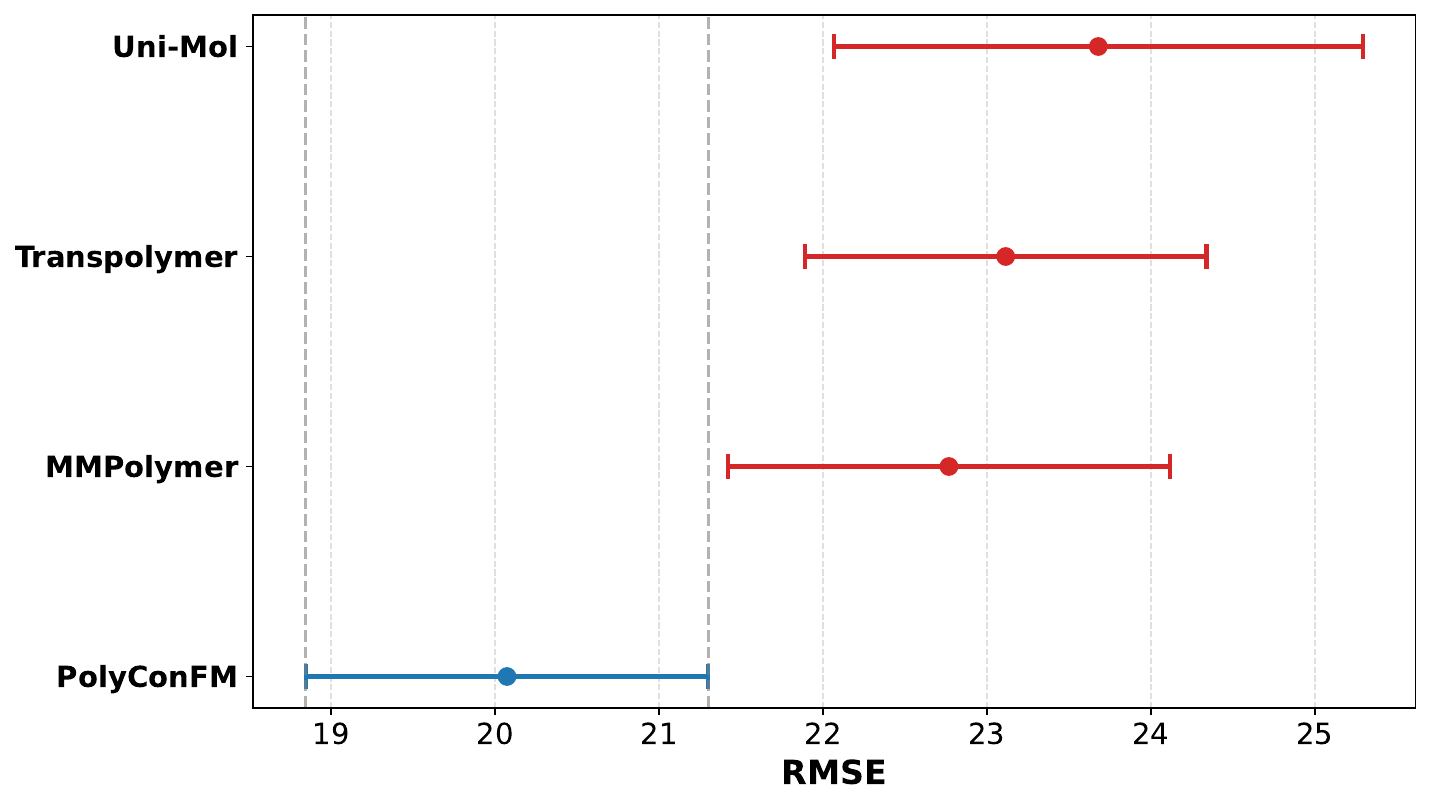}
  }
  \subfigure[EPS]{
    \includegraphics[height=4.5cm]{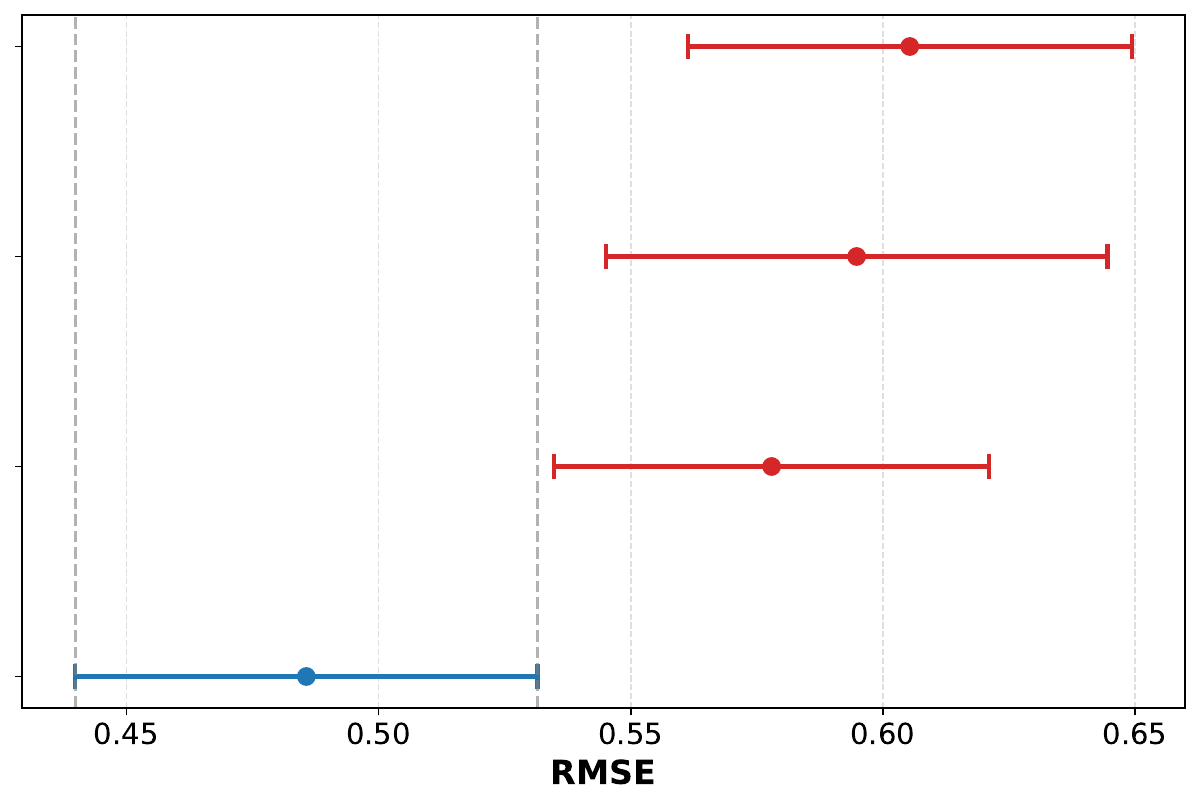}
  }
  \subfigure[Nc]{
    \includegraphics[height=4.5cm]{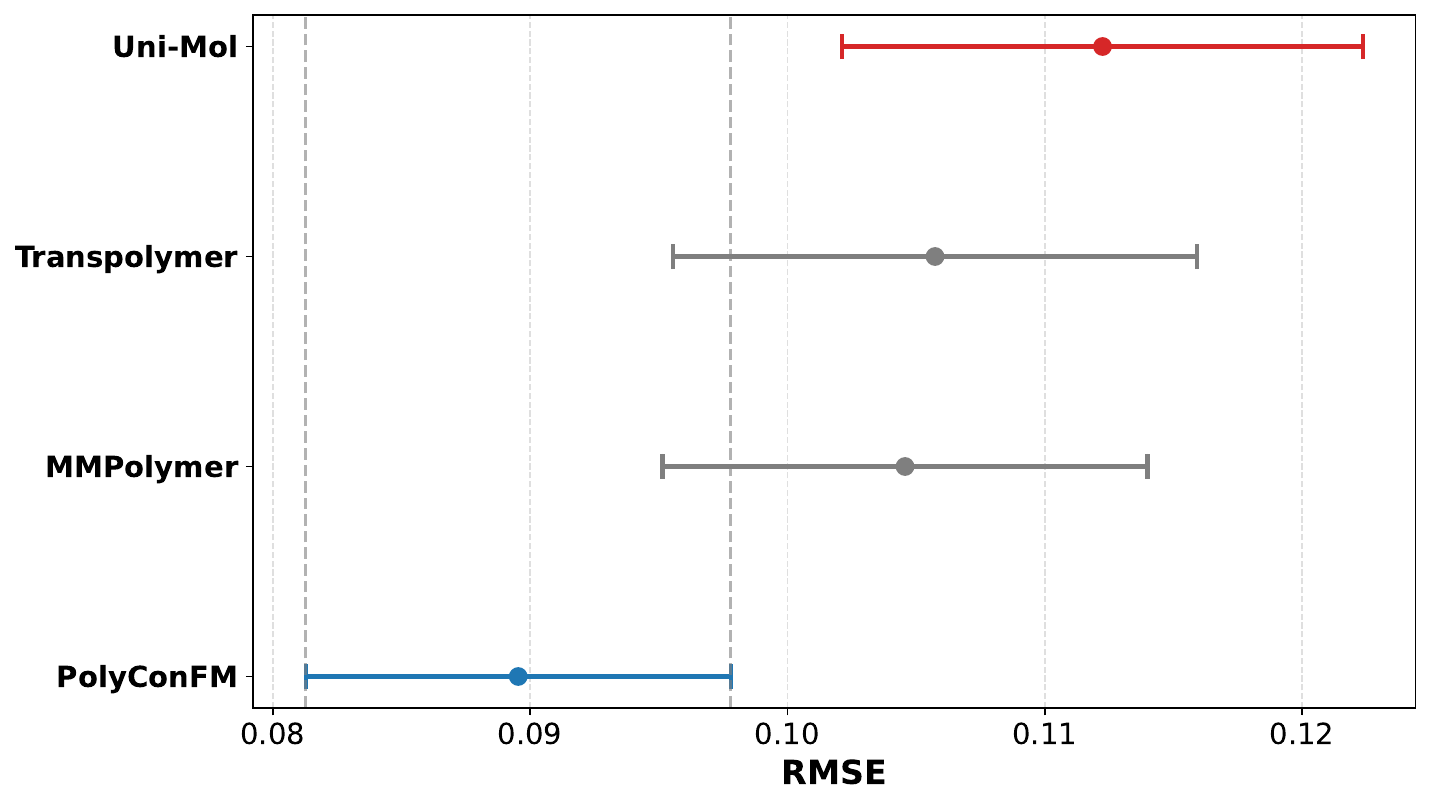}
  }
  \subfigure[Eat]{
    \includegraphics[height=4.5cm]{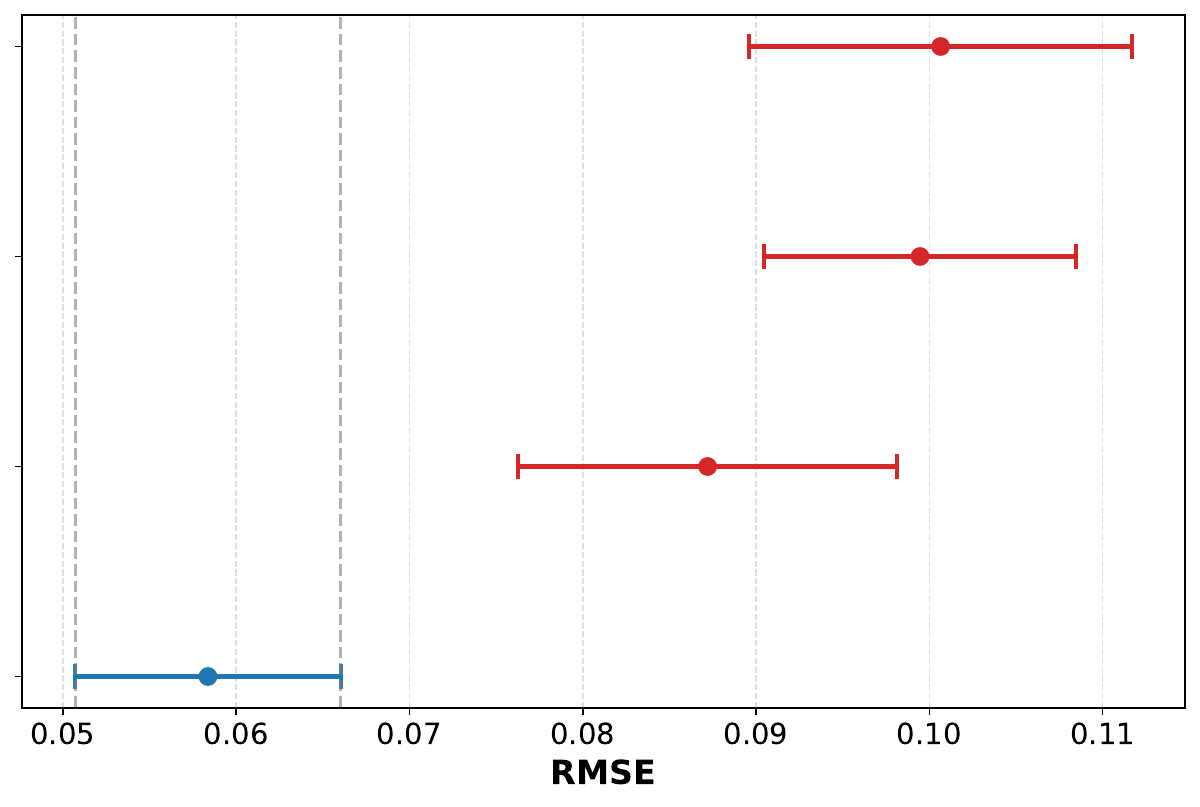}
  }
  \caption{\textbf{\thefigure:} \rre{The simultaneous confidence interval plots of the test-set \textbf{RMSE} (i.e., root mean squared error) on the downstream polymer property prediction task, where we compare PolyConFM against baselines under the \textbf{cluster-based 5$\times$5 cross-validation}.
  Using the Tukey HSD testing procedure with a significance level of 0.05, the method with the best performance is displayed in blue, methods equivalent to the best model are represented in gray, and methods that show statistically significant differences from the best model are indicated in red. 
  The entire evaluation pipeline strictly adheres to the guidelines established in~\cite{ash2025practically}.
  }} 
  \label{fig: exp_cluster_5x5_cv_CI_RMSE}
\end{figure}

\begin{figure}[H]
  \centering
  \subfigure[Egc]{
    \includegraphics[height=4.5cm]{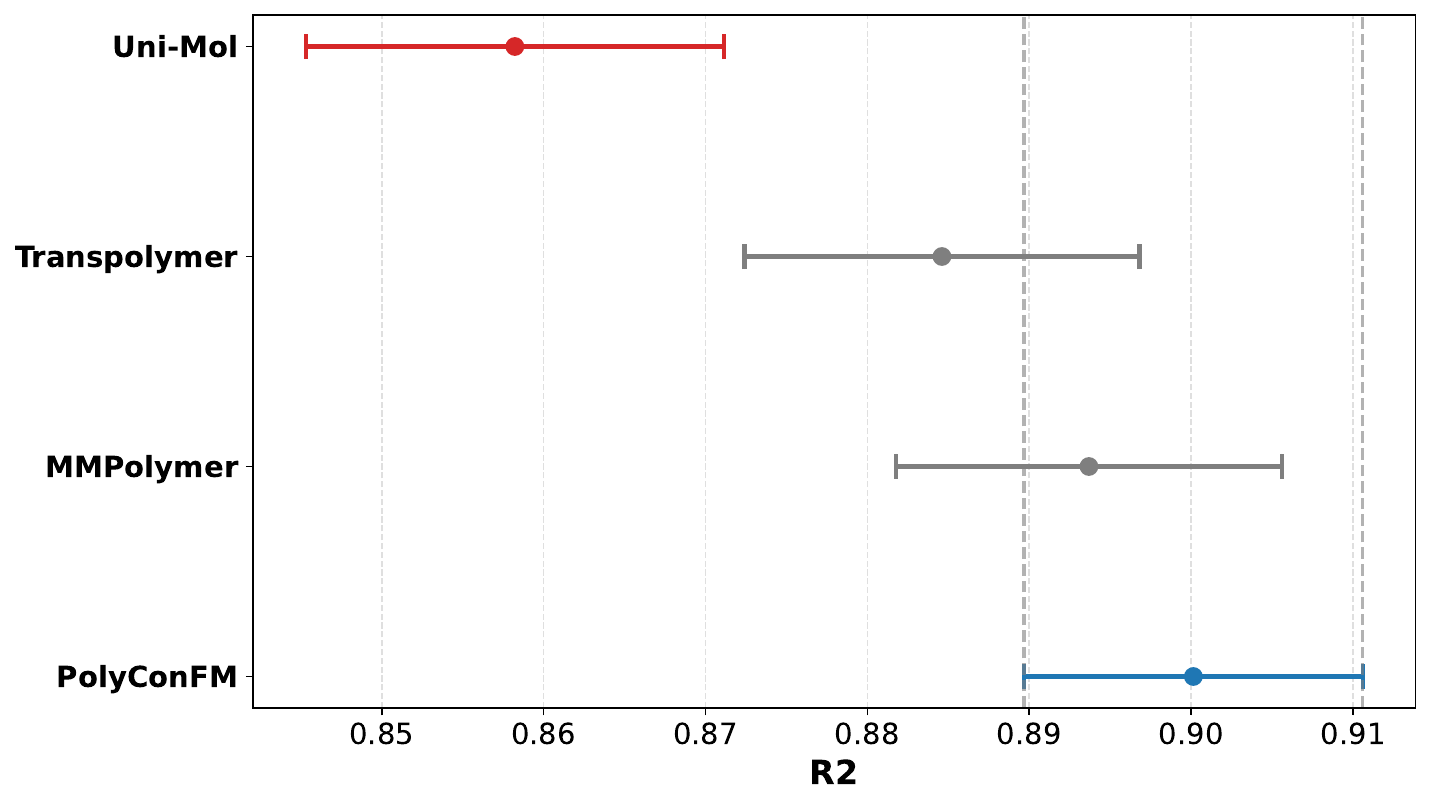}
  }
  \subfigure[Egb]{
    \includegraphics[height=4.5cm]{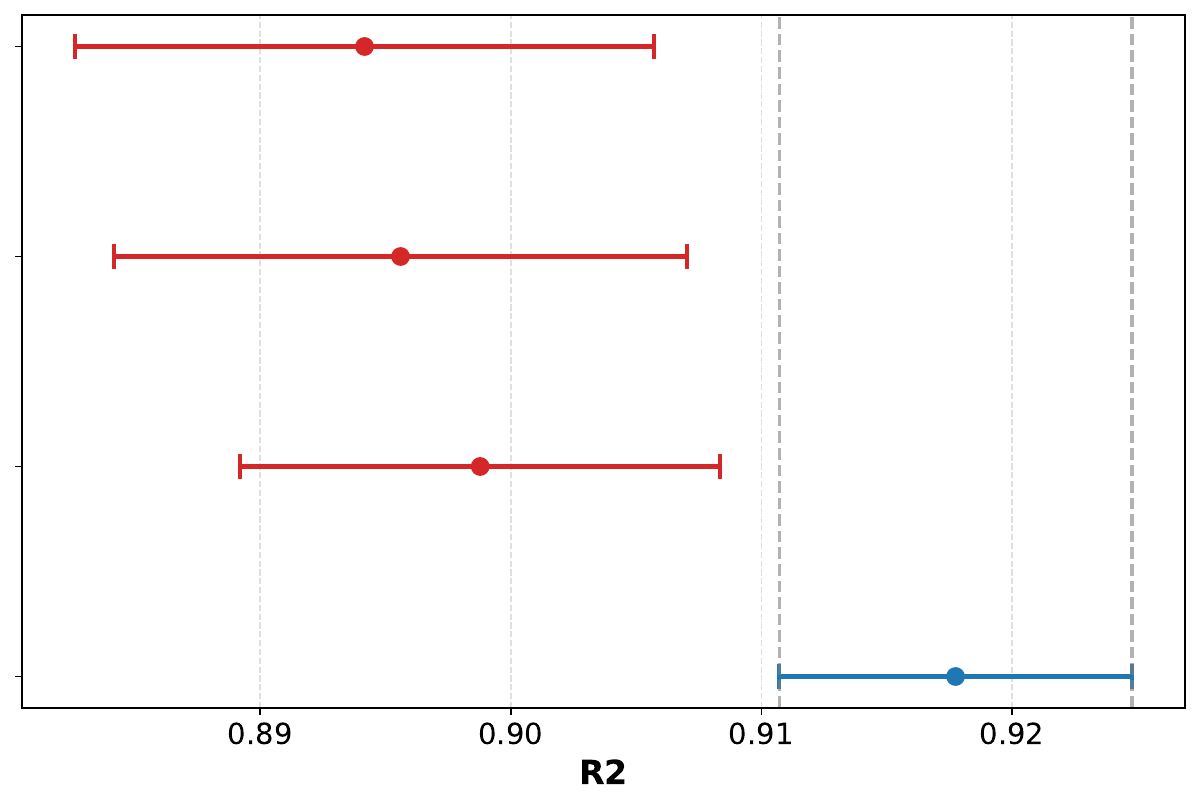}
  }
  \subfigure[Eea]{
    \includegraphics[height=4.5cm]{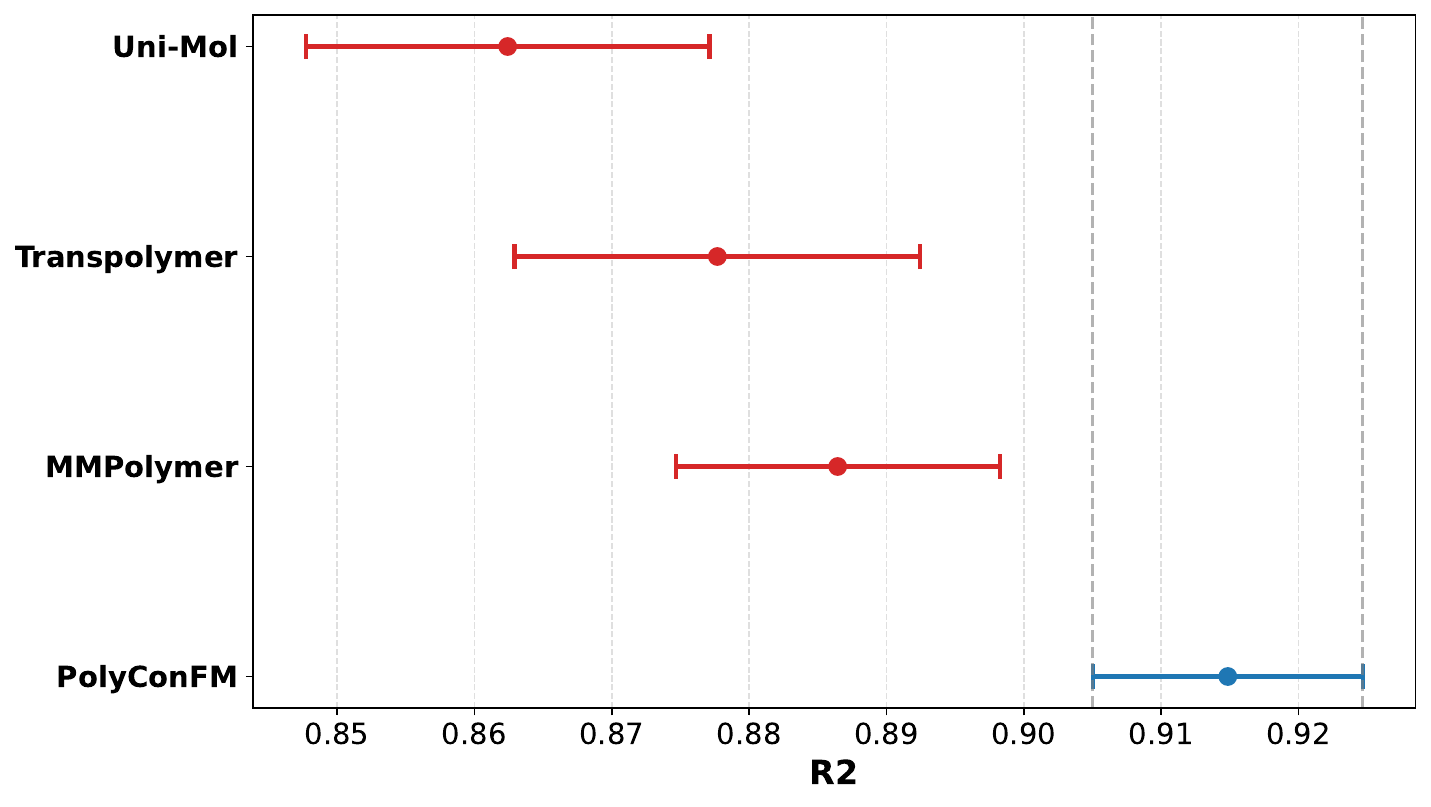}
  }
  \subfigure[Ei]{
    \includegraphics[height=4.5cm]{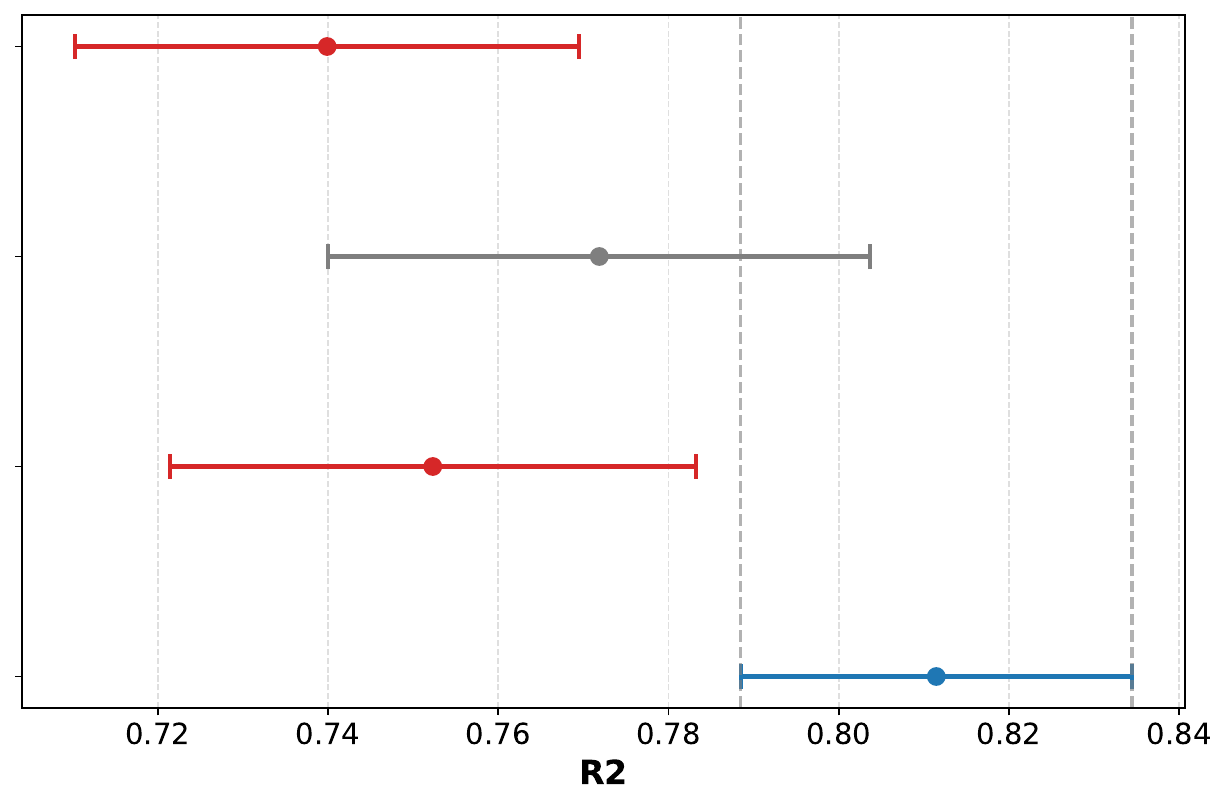}
  }
  \subfigure[Xc]{
    \includegraphics[height=4.5cm]{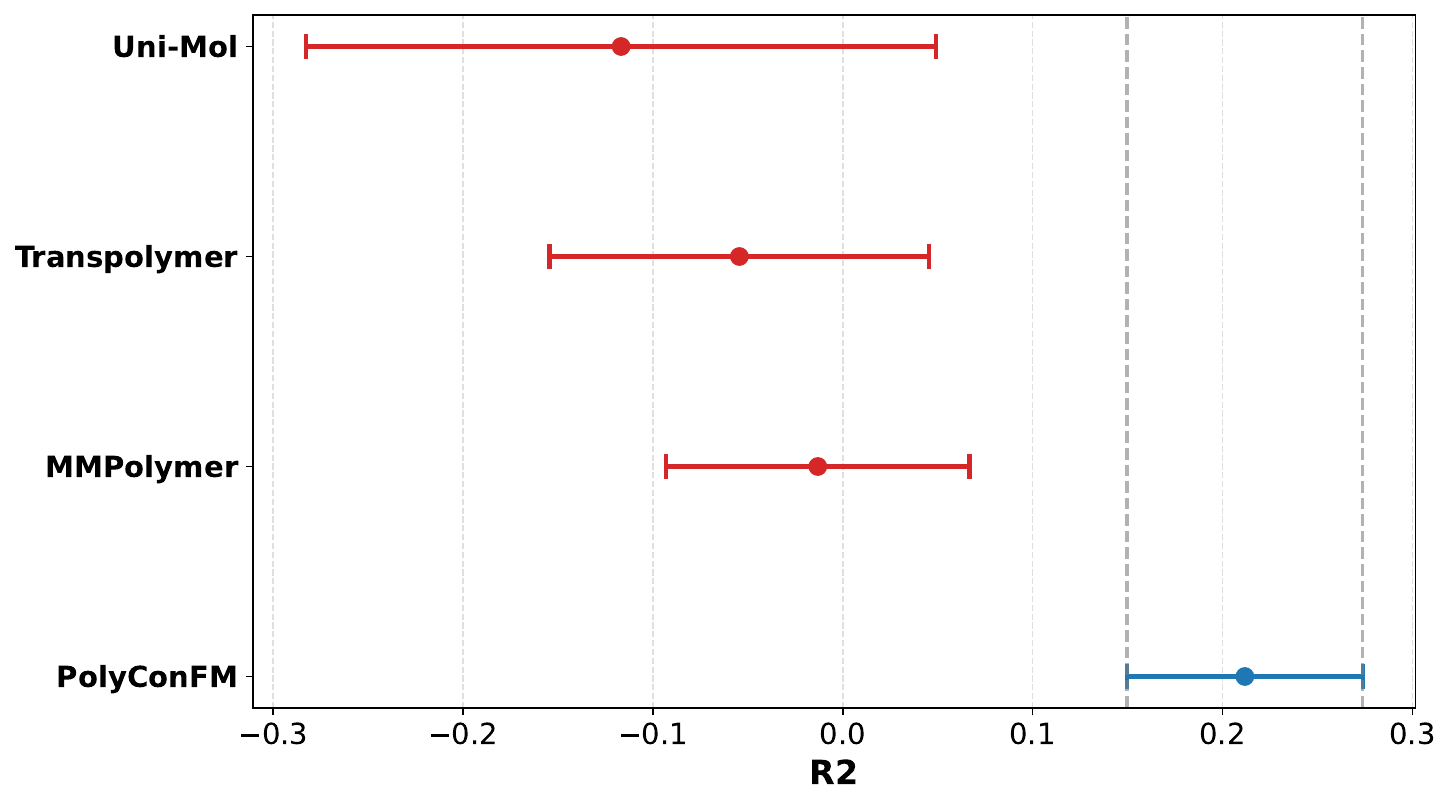}
  }
  \subfigure[EPS]{
    \includegraphics[height=4.5cm]{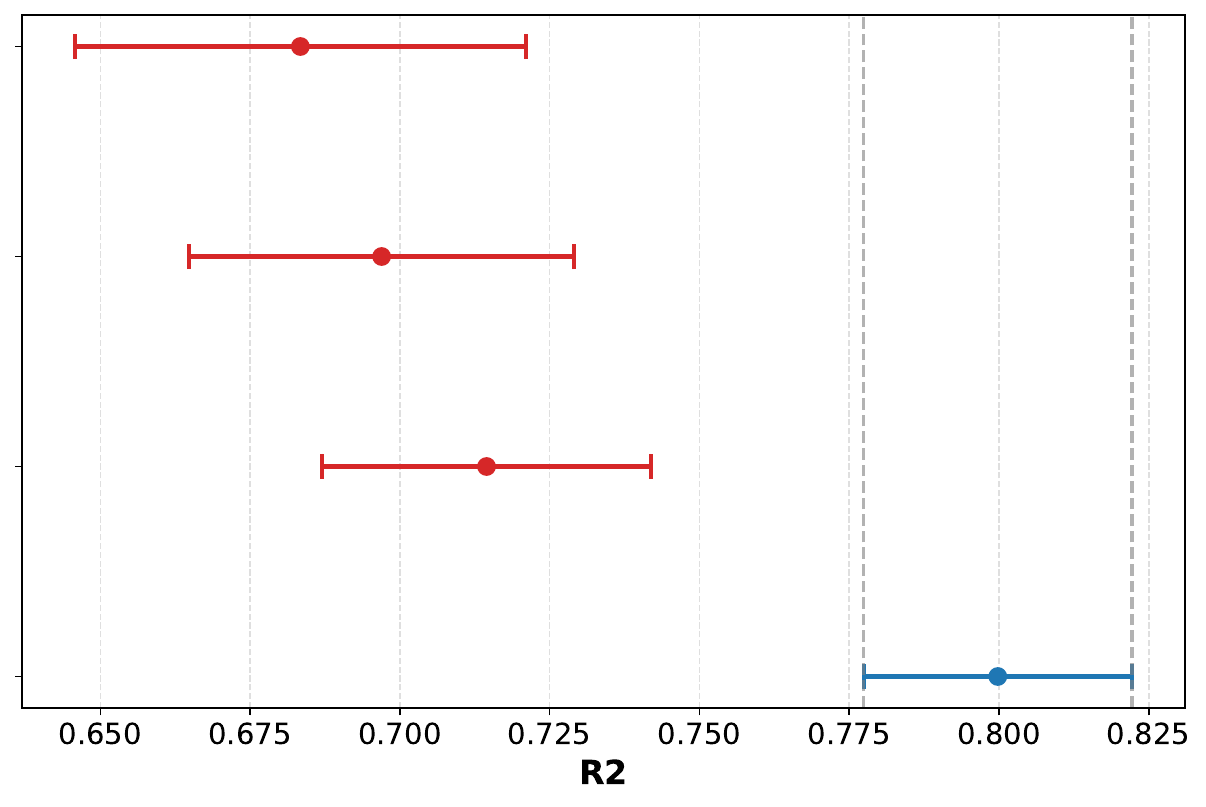}
  }
  \subfigure[Nc]{
    \includegraphics[height=4.5cm]{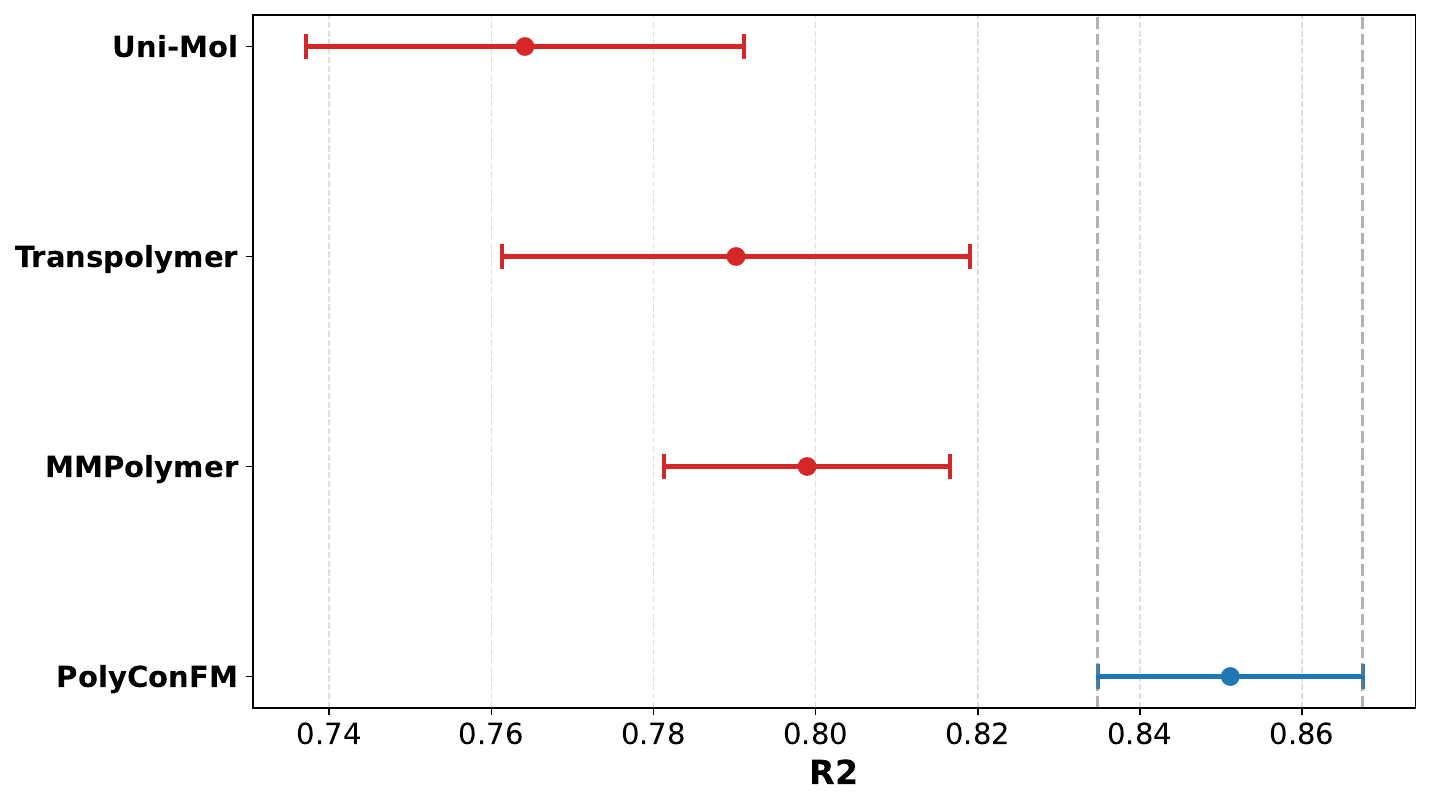}
  }
  \subfigure[Eat]{
    \includegraphics[height=4.5cm]{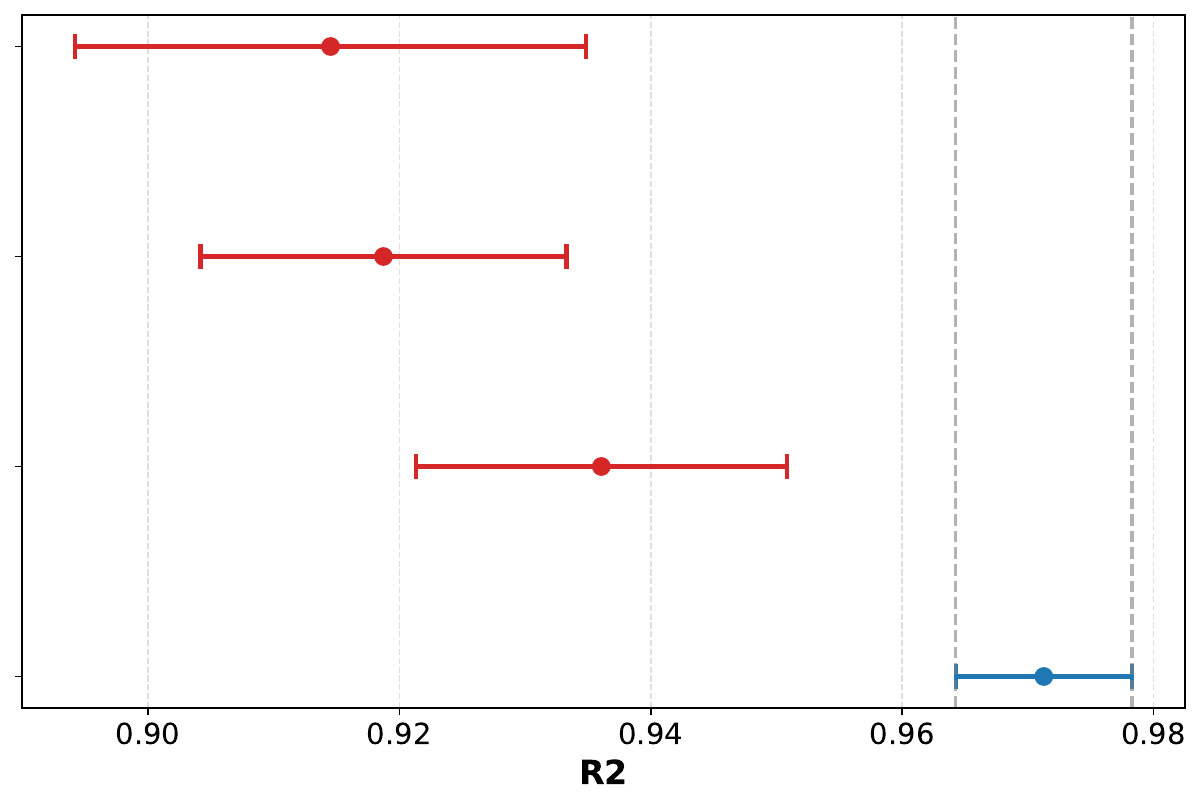}
  }
  \caption{\textbf{\thefigure:} \rre{The simultaneous confidence interval plots of the test-set $\bm {R^2}$ (i.e., coefficient of determination) on the downstream polymer property prediction task, where we compare PolyConFM against baselines under the \textbf{cluster-based 5$\times$5 cross-validation}.
  Using the Tukey HSD testing procedure with a significance level of 0.05, the method with the best performance is displayed in blue, methods equivalent to the best model are represented in gray, and methods that show statistically significant differences from the best model are indicated in red. 
  The entire evaluation pipeline strictly adheres to the guidelines established in~\cite{ash2025practically}.
  }}
  \label{fig: exp_cluster_5x5_cv_CI_R2}
\end{figure}

\begin{figure}[H]
  \centering
  \subfigure[Egc]{
    \includegraphics[height=4.8cm]{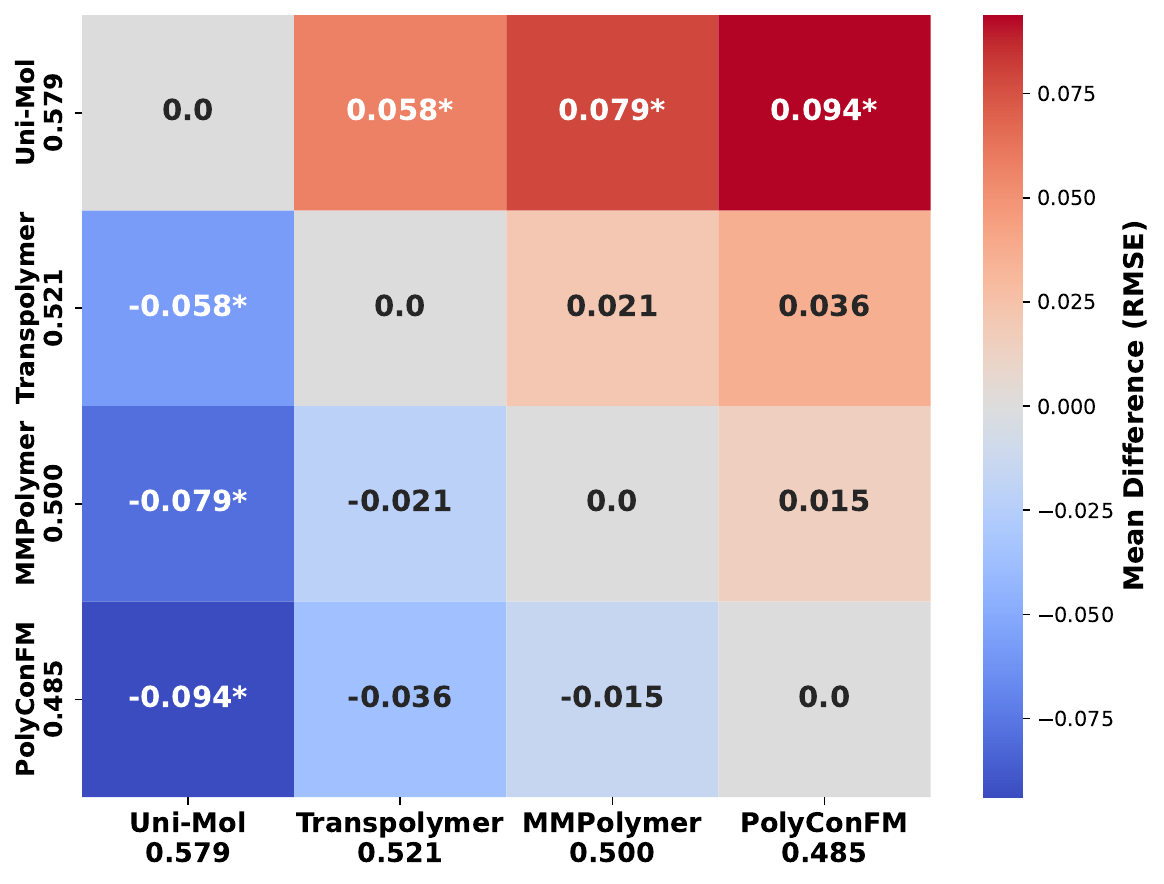}
  }
  \subfigure[Egb]{
    \includegraphics[height=4.8cm]{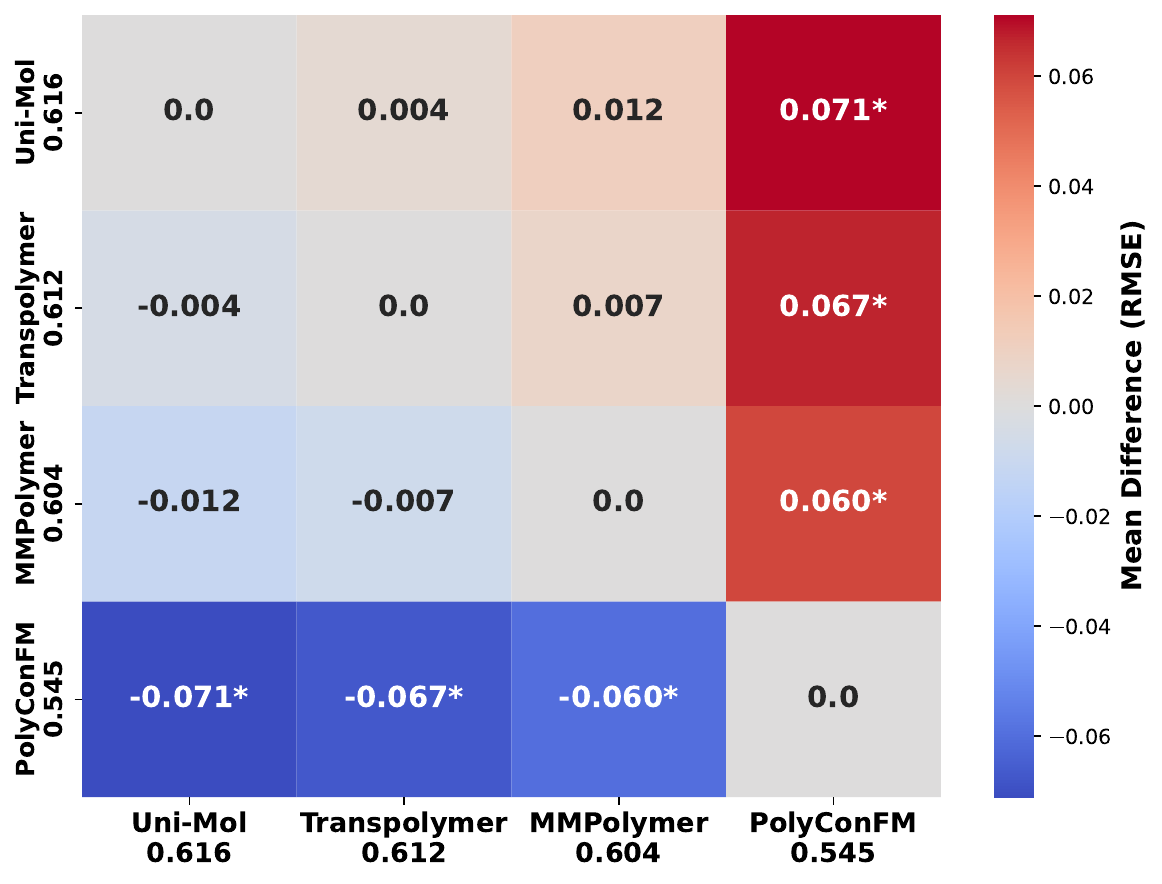}
  }
  \subfigure[Eea]{
    \includegraphics[height=4.8cm]{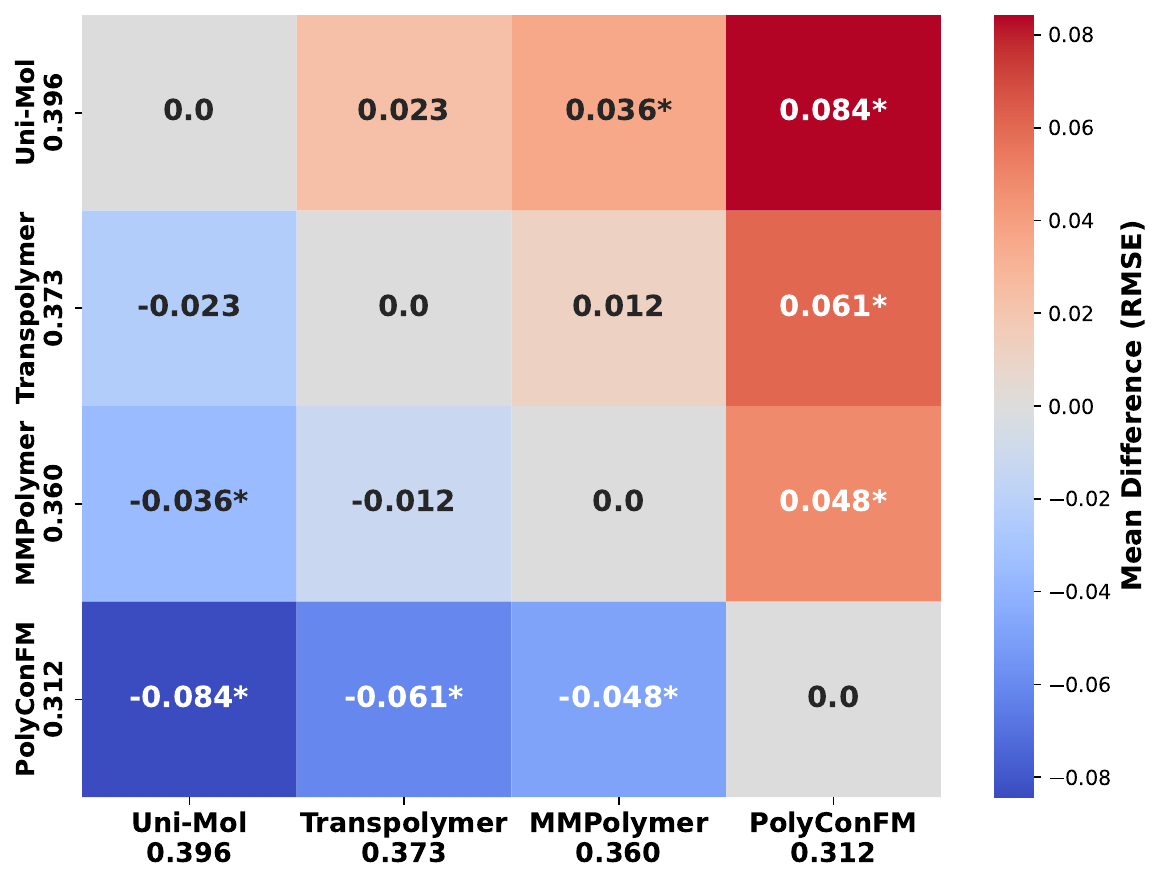}
  }
  \subfigure[Ei]{
    \includegraphics[height=4.8cm]{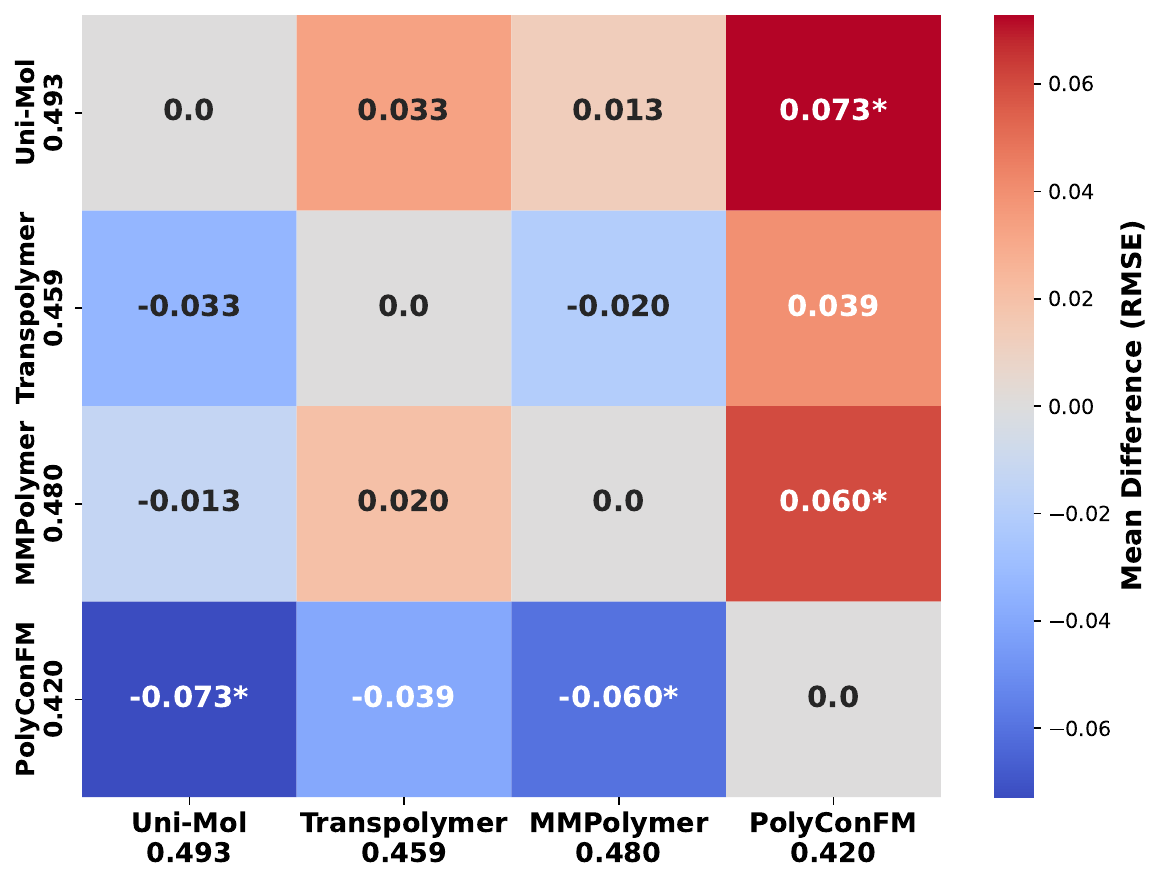}
  }
  \subfigure[Xc]{
    \includegraphics[height=4.8cm]{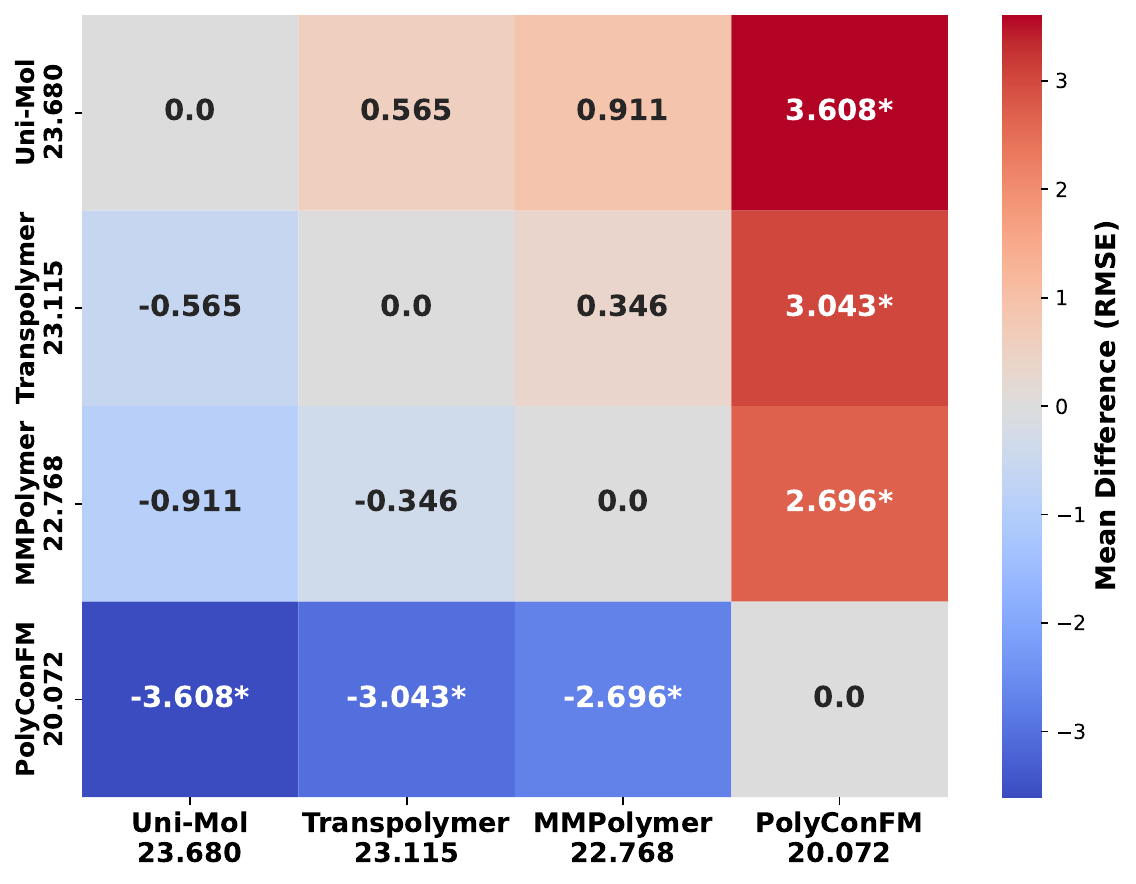}
  }
  \subfigure[EPS]{
    \includegraphics[height=4.8cm]{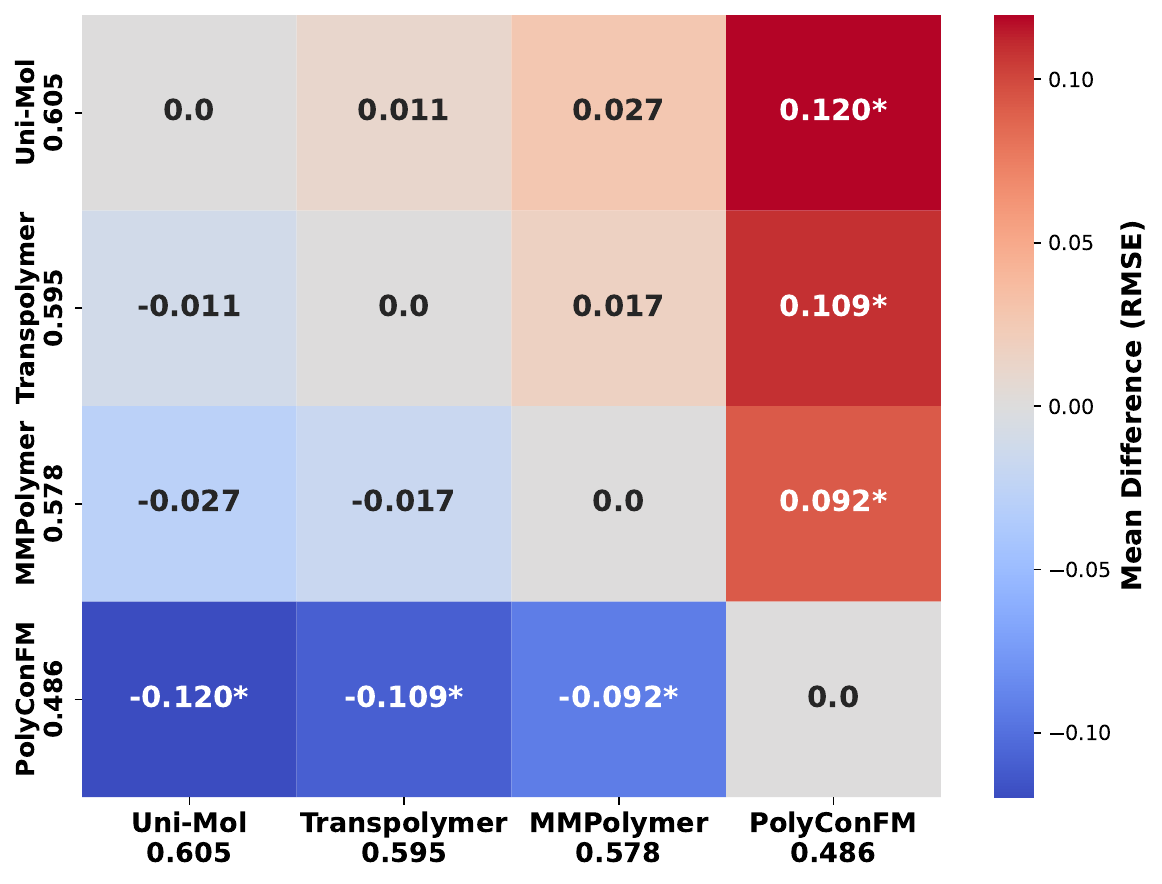}
  }
  \subfigure[Nc]{
    \includegraphics[height=4.8cm]{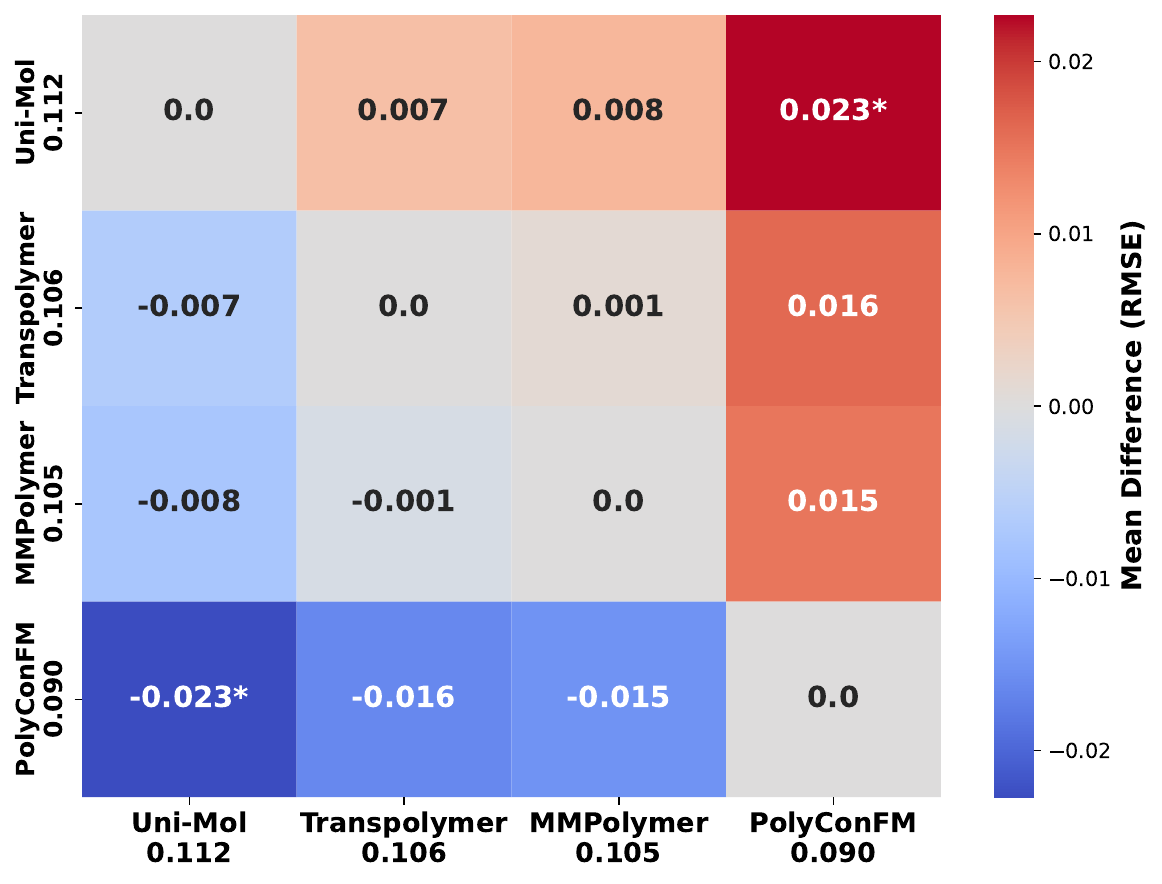}
  }
  \subfigure[Eat]{
    \includegraphics[height=4.8cm]{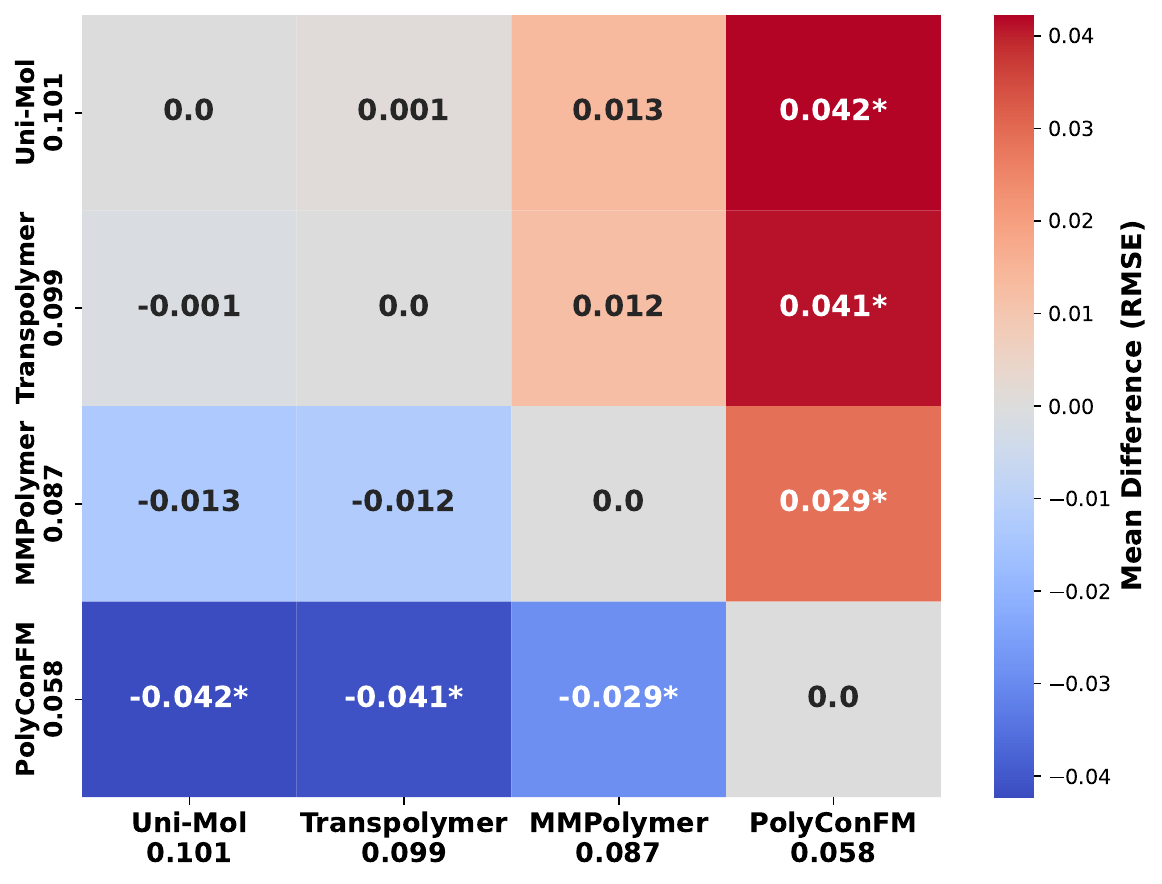}
  }
  \caption{\textbf{\thefigure:} \rre{The multiple comparisons similarity plots of the test-set \textbf{RMSE} (i.e., root mean squared error) on the downstream polymer property prediction task, where we compare PolyConFM against baselines under the \textbf{cluster-based 5$\times$5 cross-validation}.
  Using the Tukey HSD testing procedure with a significance level of 0.05, statistically significant differences are annotated by star symbols.
  The entire evaluation pipeline strictly adheres to the guidelines established in~\cite{ash2025practically}.
  }} 
  \label{fig: exp_cluster_5x5_cv_MCSim_RMSE}
\end{figure}

\begin{figure}[H]
  \centering
  \subfigure[Egc]{
    \includegraphics[height=4.8cm]{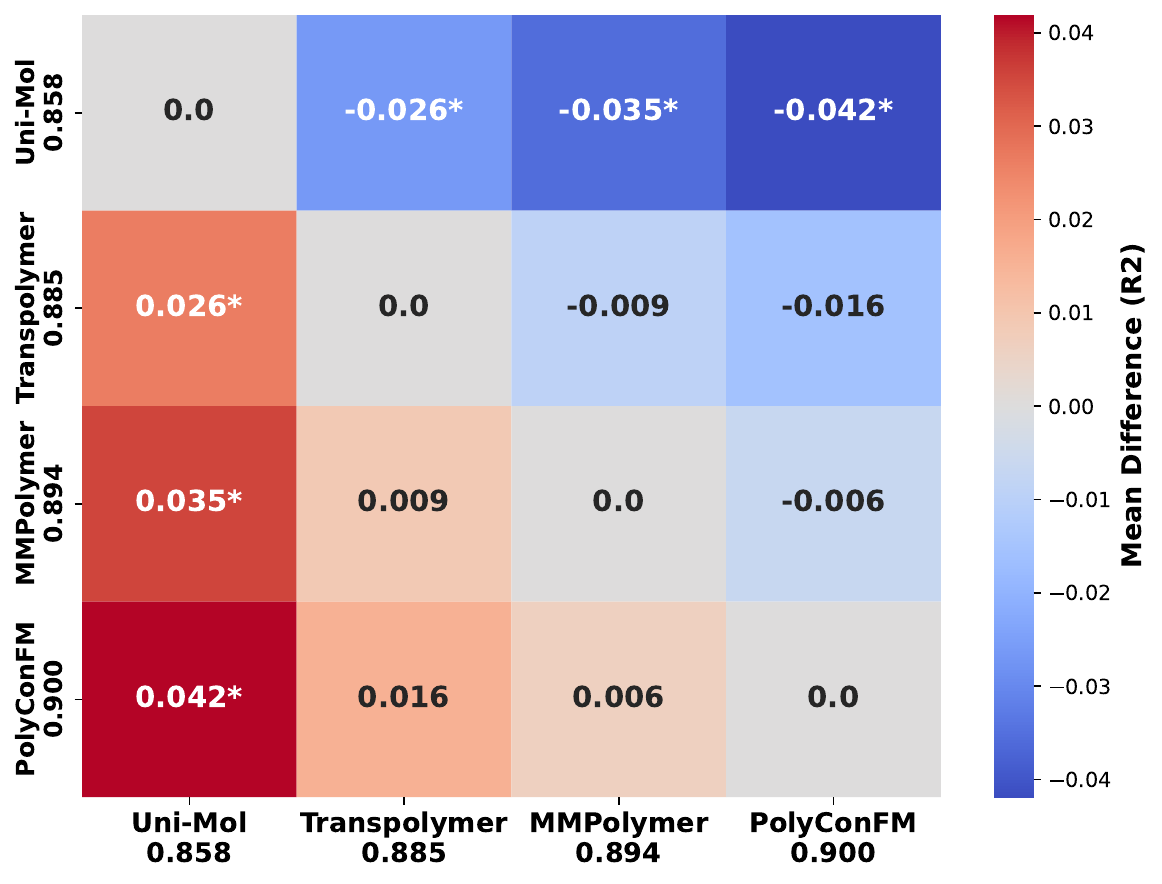}
  }
  \subfigure[Egb]{
    \includegraphics[height=4.8cm]{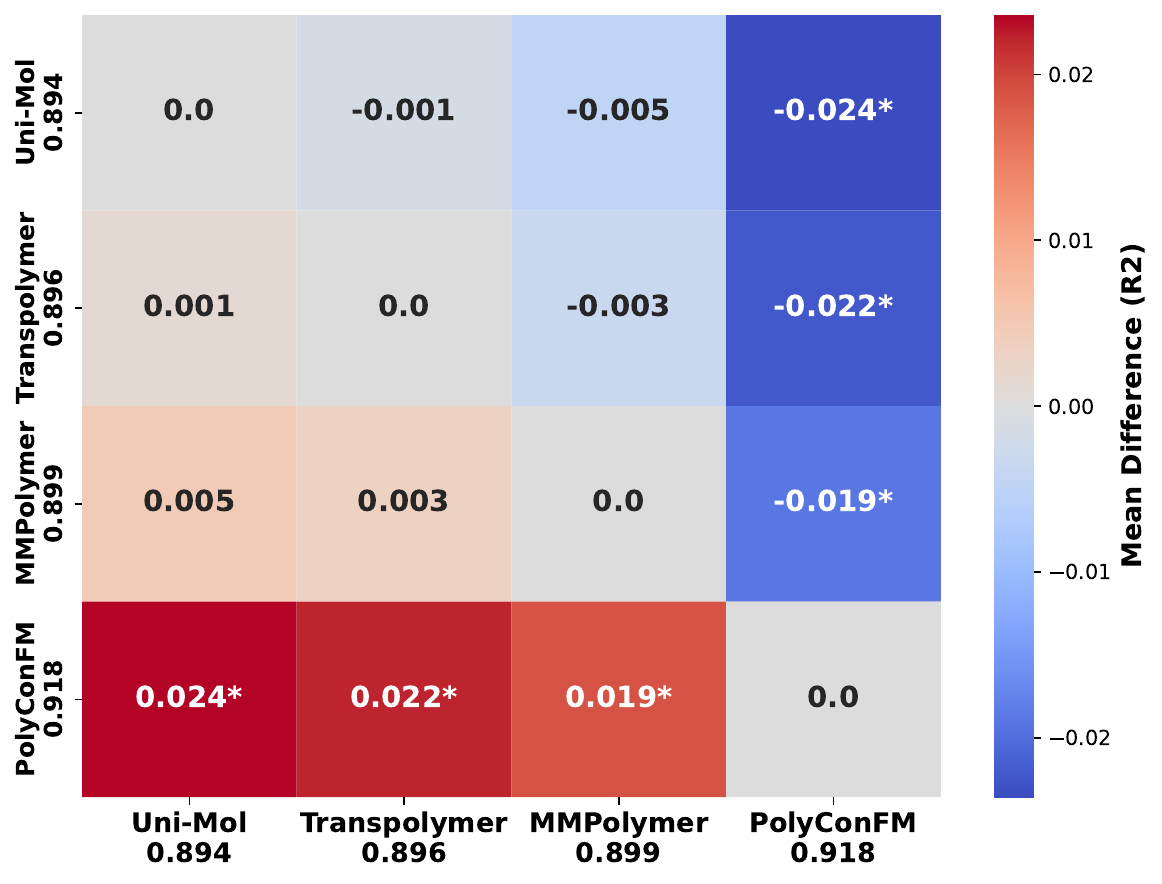}
  }
  \subfigure[Eea]{
    \includegraphics[height=4.8cm]{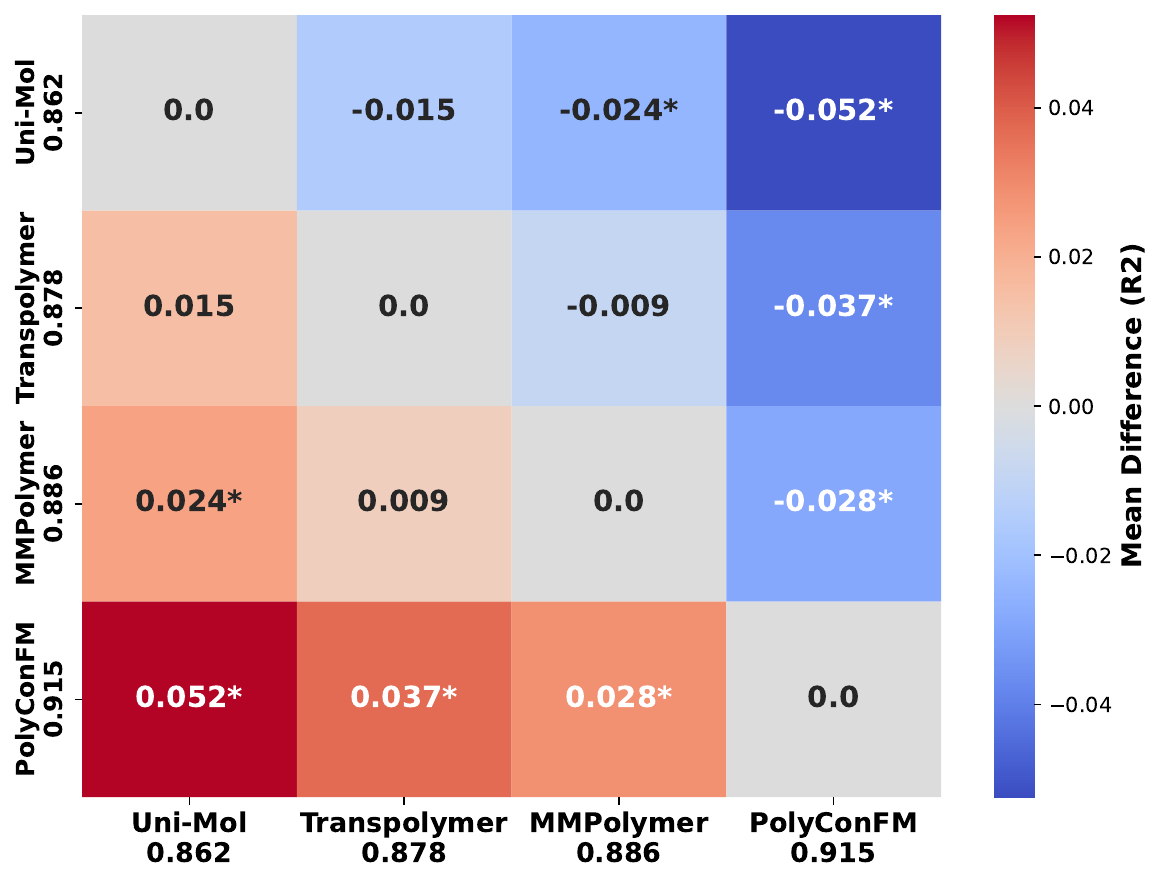}
  }
  \subfigure[Ei]{
    \includegraphics[height=4.8cm]{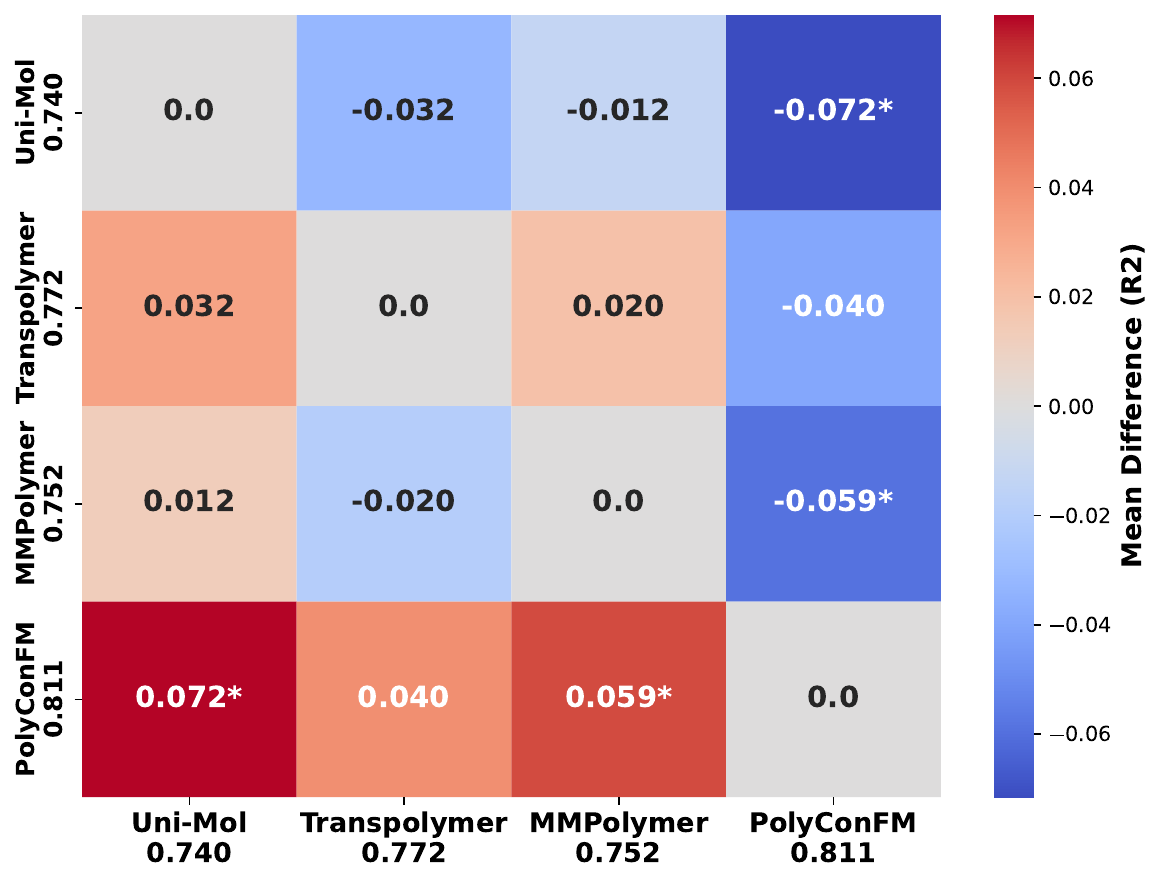}
  }
  \subfigure[Xc]{
    \includegraphics[height=4.8cm]{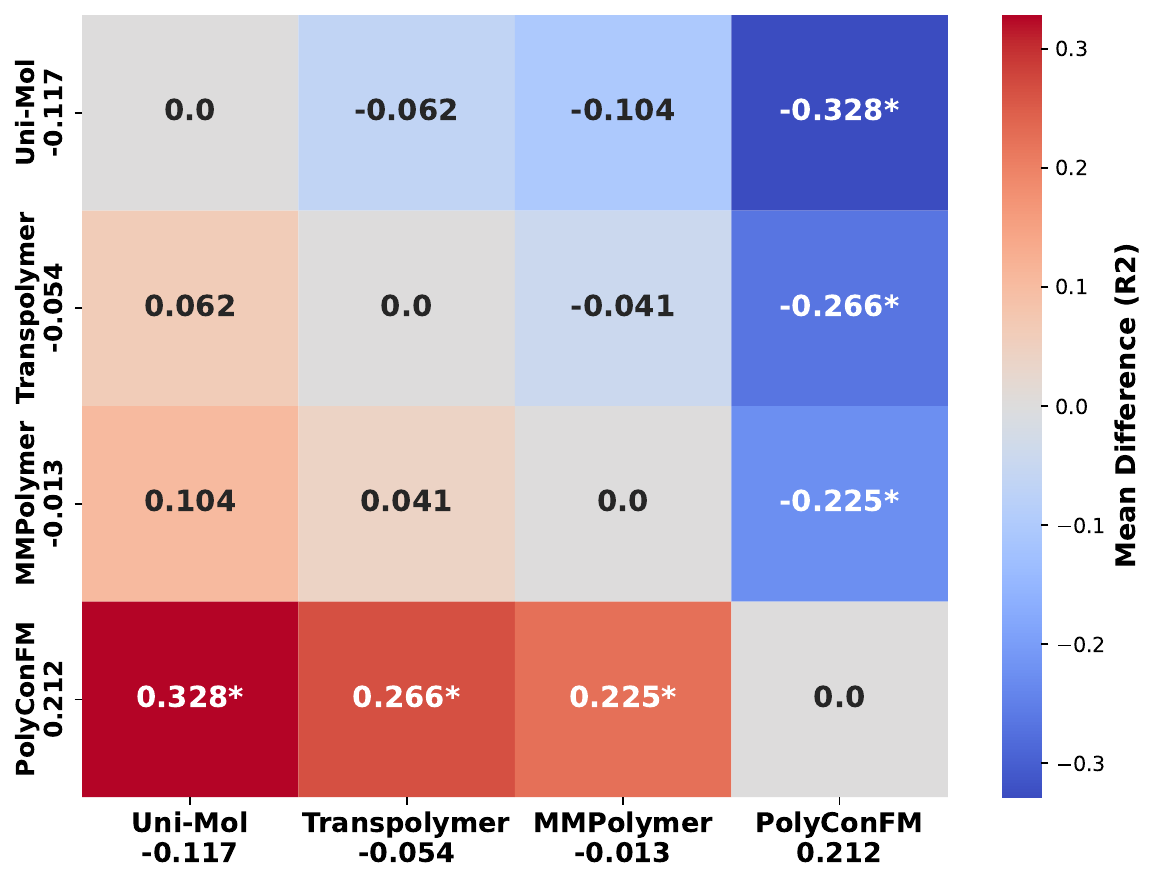}
  }
  \subfigure[EPS]{
    \includegraphics[height=4.8cm]{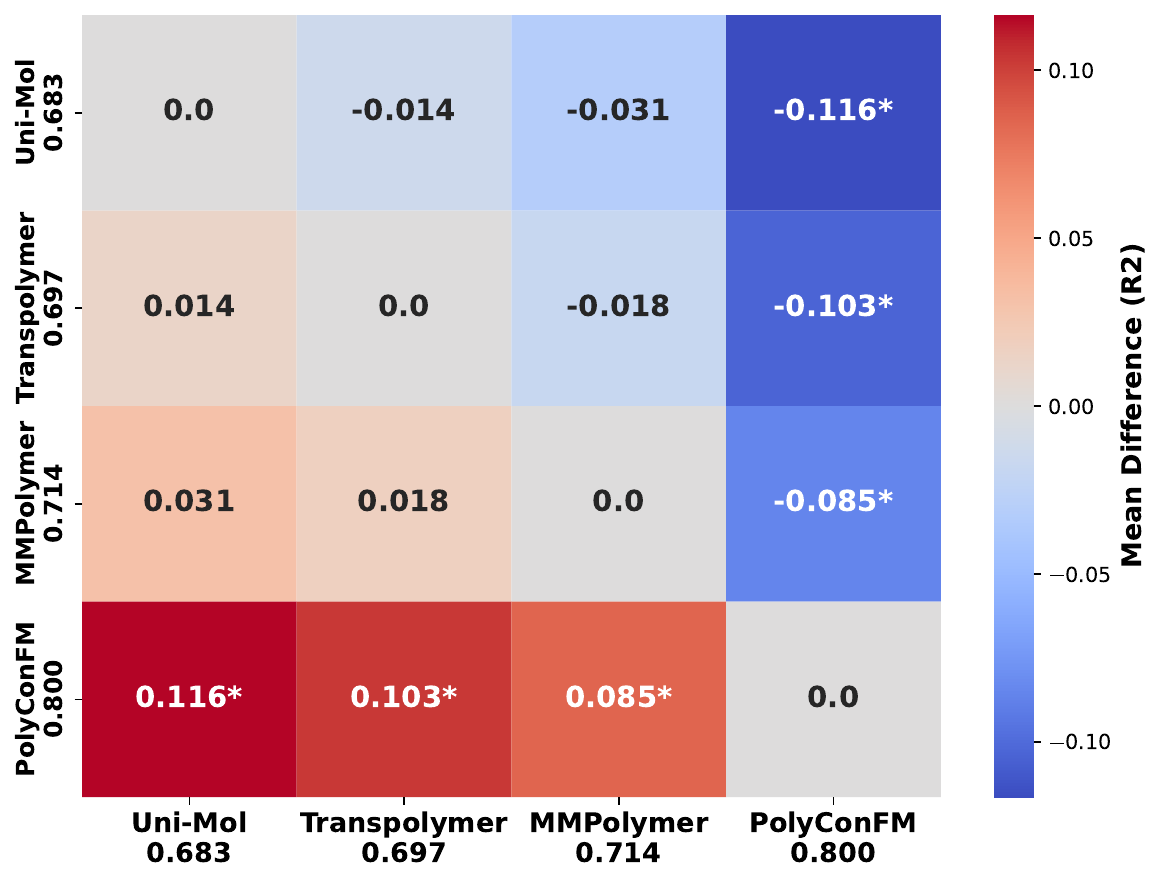}
  }
  \subfigure[Nc]{
    \includegraphics[height=4.8cm]{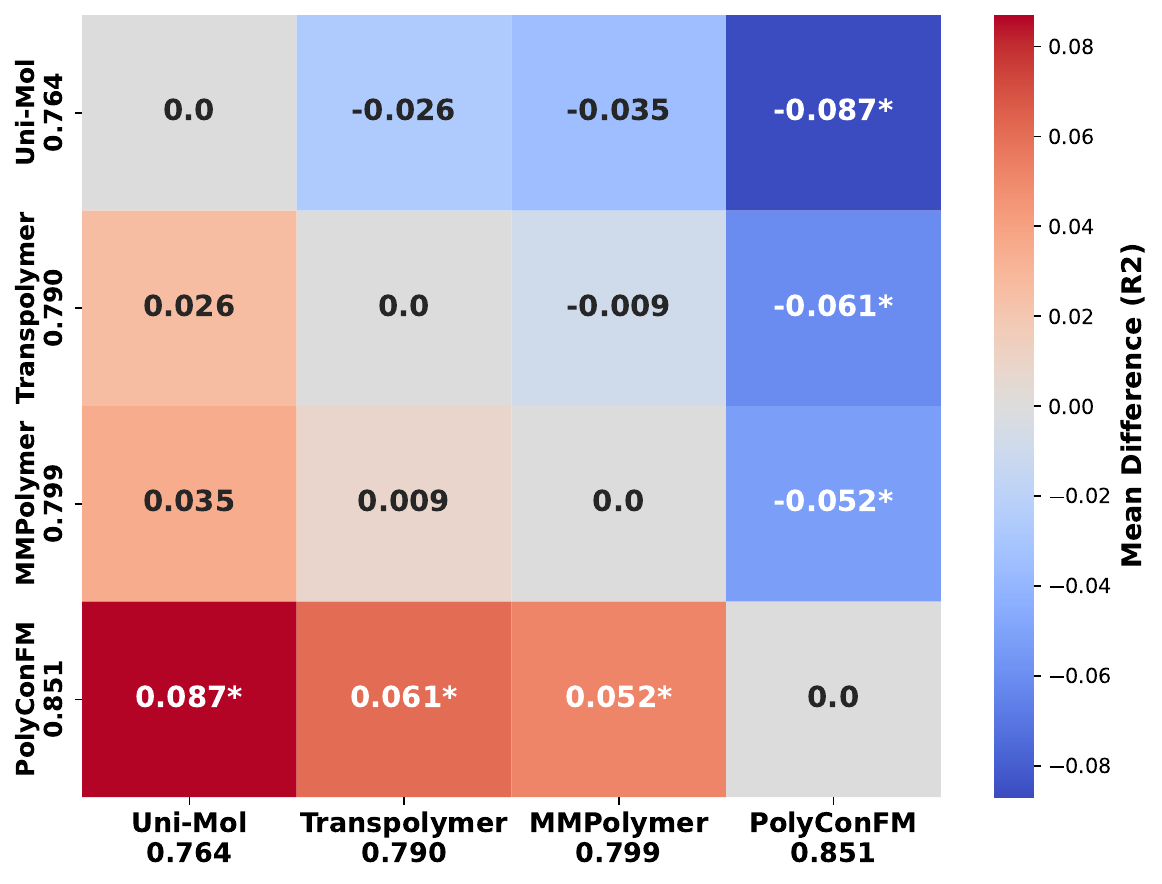}
  }
  \subfigure[Eat]{
    \includegraphics[height=4.8cm]{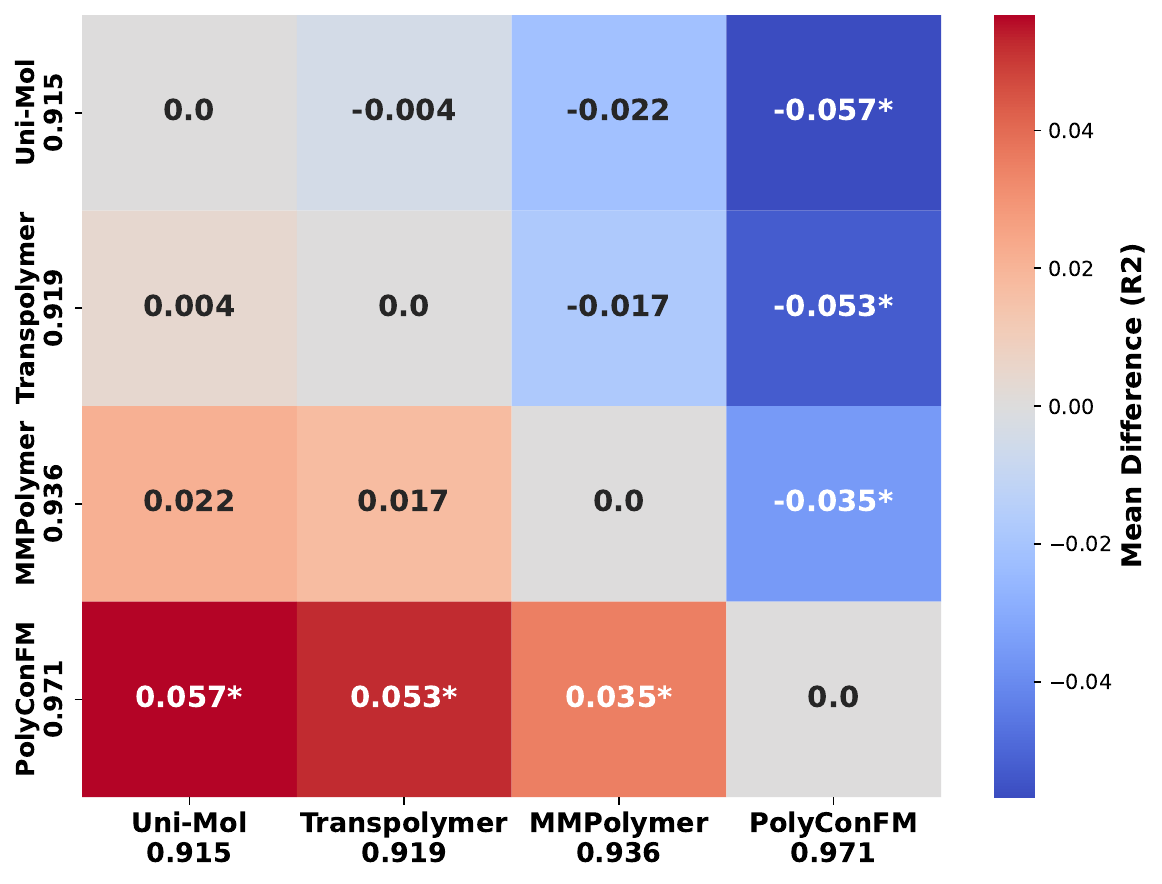}
  }
  \caption{\textbf{\thefigure:} \rre{The multiple comparisons similarity plots of the test-set $\bm {R^2}$ (i.e., coefficient of determination) on the downstream polymer property prediction task, where we compare PolyConFM against baselines under the \textbf{cluster-based 5$\times$5 cross-validation}.
  Using the Tukey HSD testing procedure with a significance level of 0.05, statistically significant differences are annotated by star symbols.
  The entire evaluation pipeline strictly adheres to the guidelines established in~\cite{ash2025practically}.
  }} 
  \label{fig: exp_cluster_5x5_cv_MCSim_R2}
\end{figure}

\begin{figure}[H]
  \centering
  \subfigure[Egc]{
    \includegraphics[height=4.5cm]{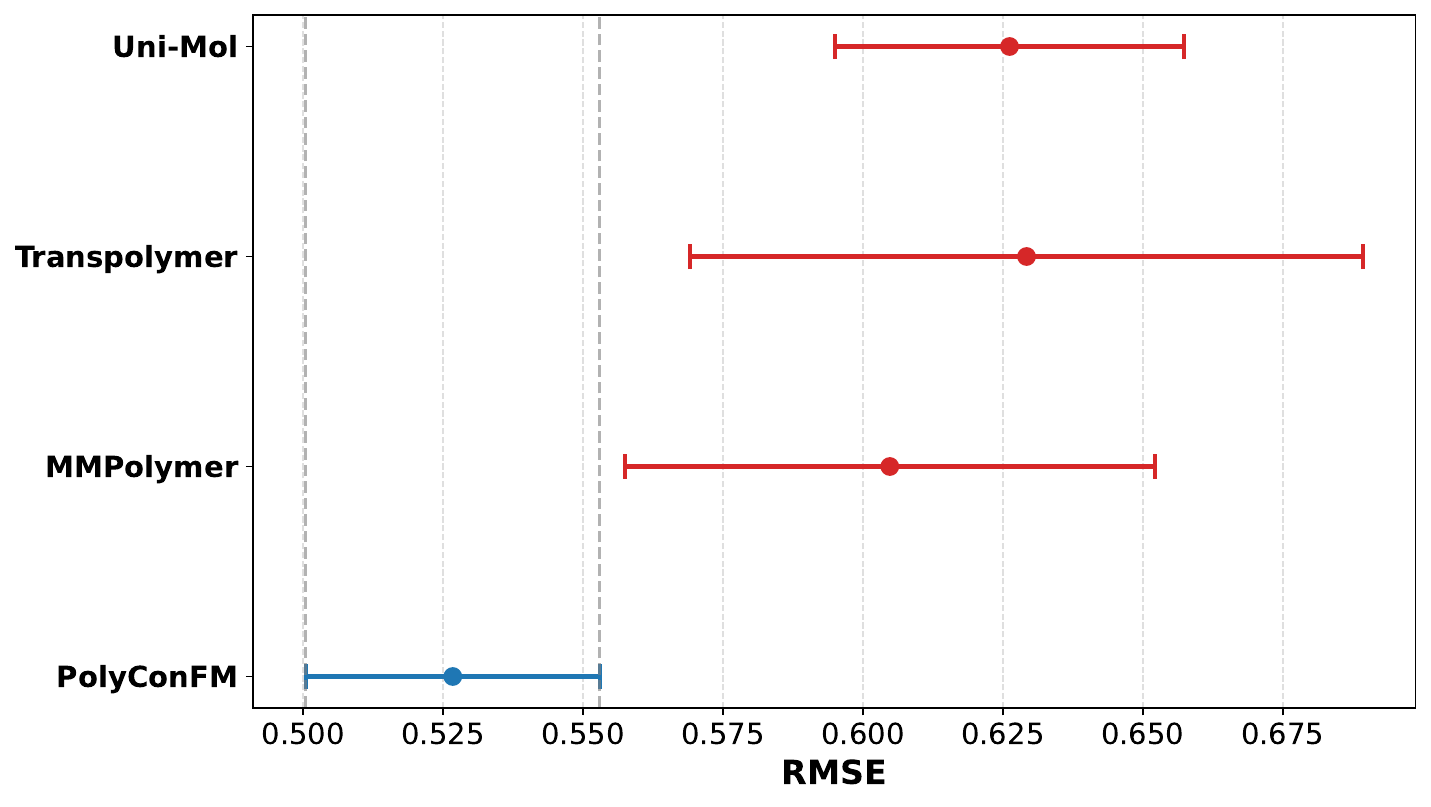}
  }
  \subfigure[Egb]{
    \includegraphics[height=4.5cm]{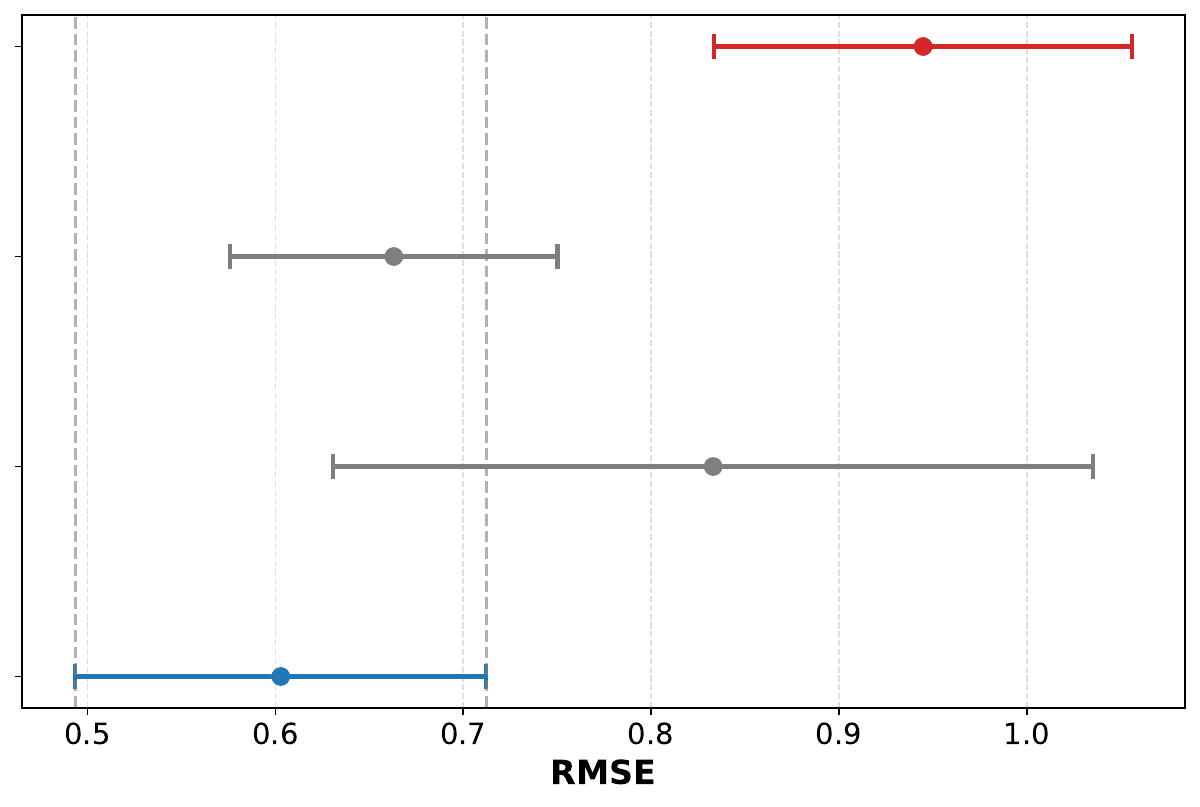}
  }
  \subfigure[Eea]{
    \includegraphics[height=4.5cm]{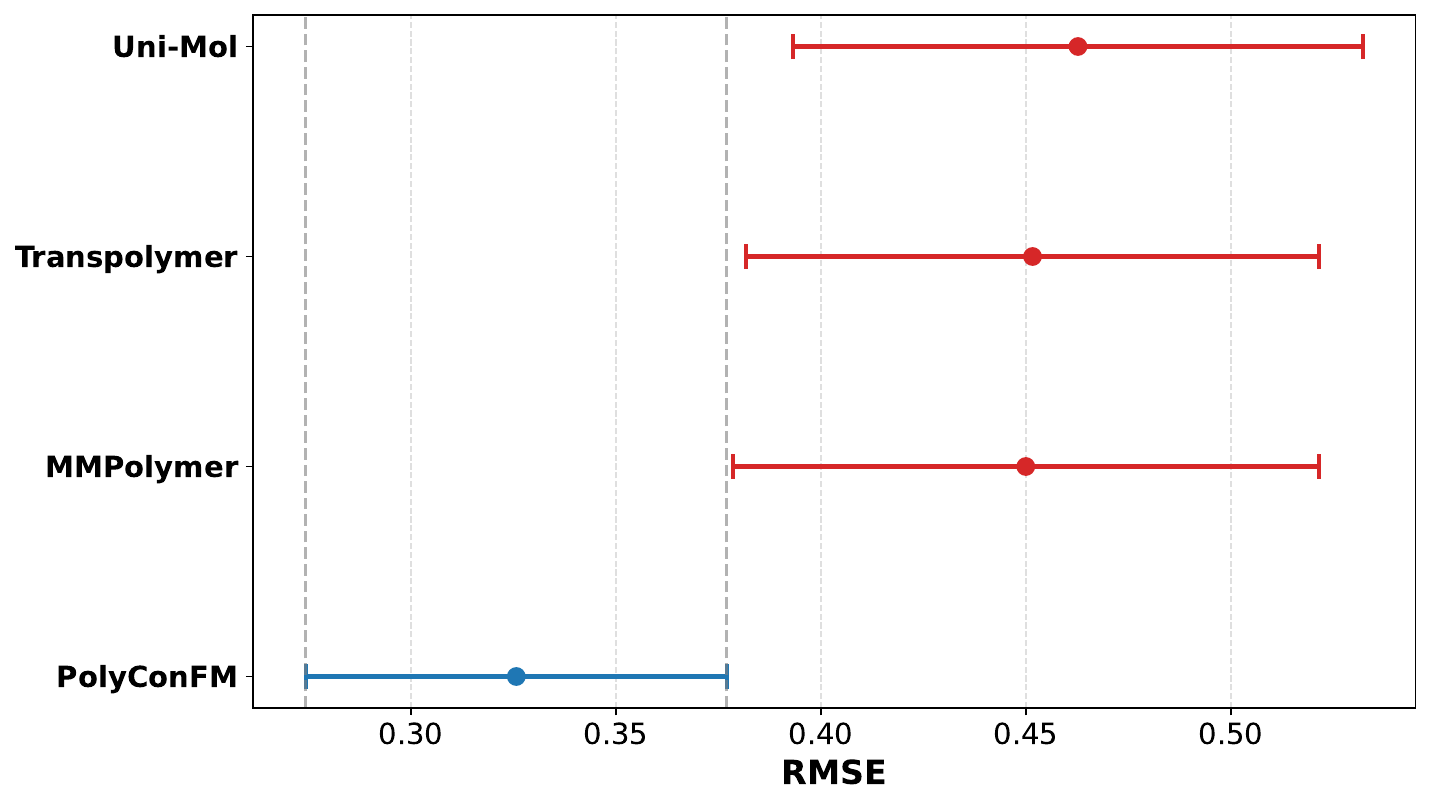}
  }
  \subfigure[Ei]{
    \includegraphics[height=4.5cm]{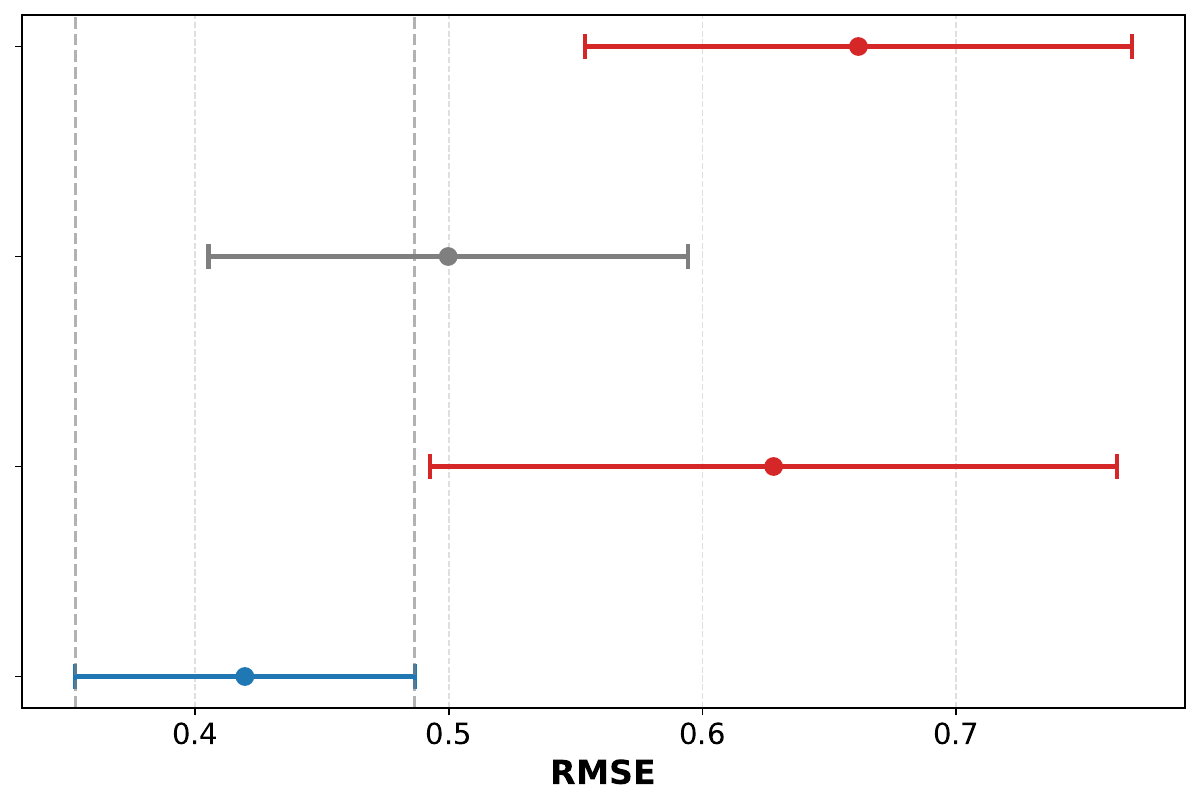}
  }
  \subfigure[Xc]{
    \includegraphics[height=4.5cm]{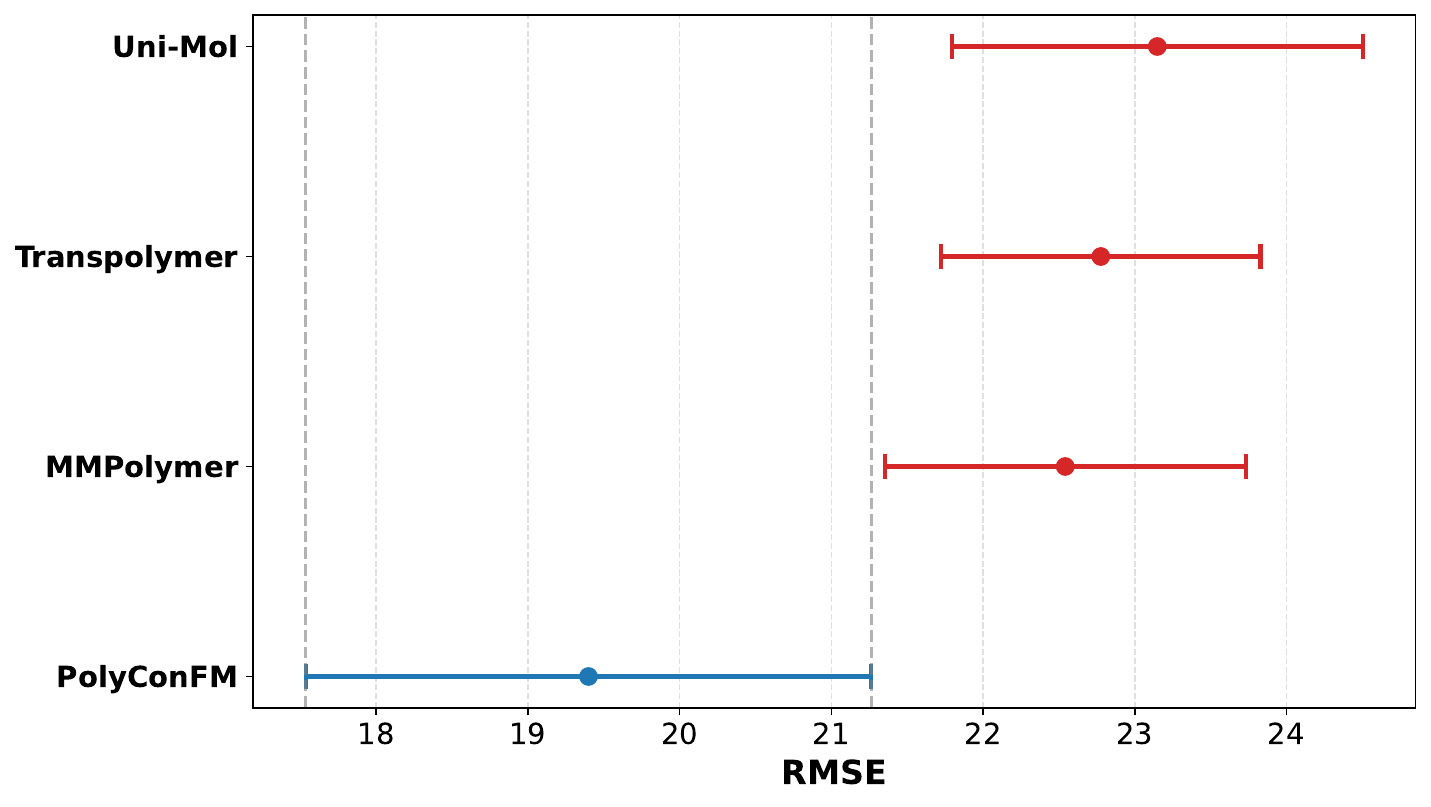}
  }
  \subfigure[EPS]{
    \includegraphics[height=4.5cm]{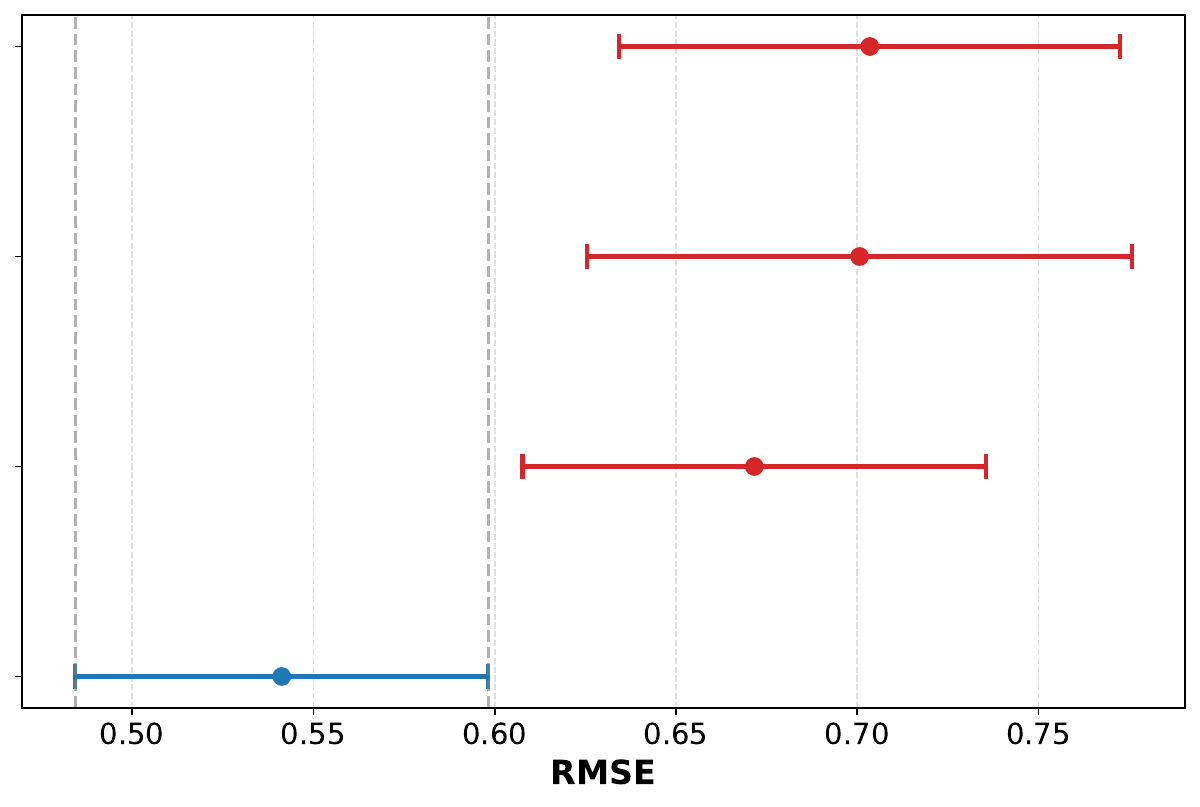}
  }
  \subfigure[Nc]{
    \includegraphics[height=4.5cm]{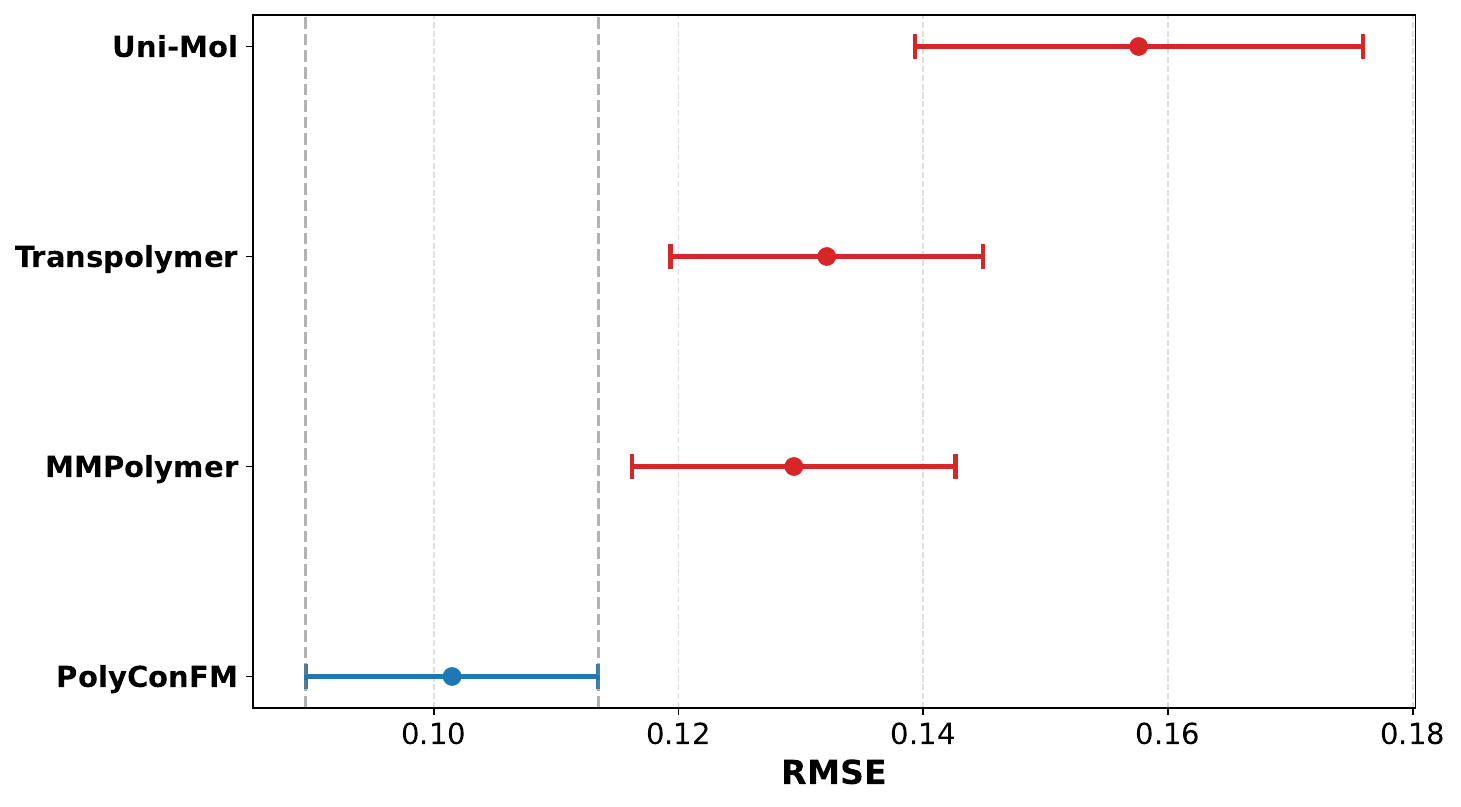}
  }
  \subfigure[Eat]{
    \includegraphics[height=4.5cm]{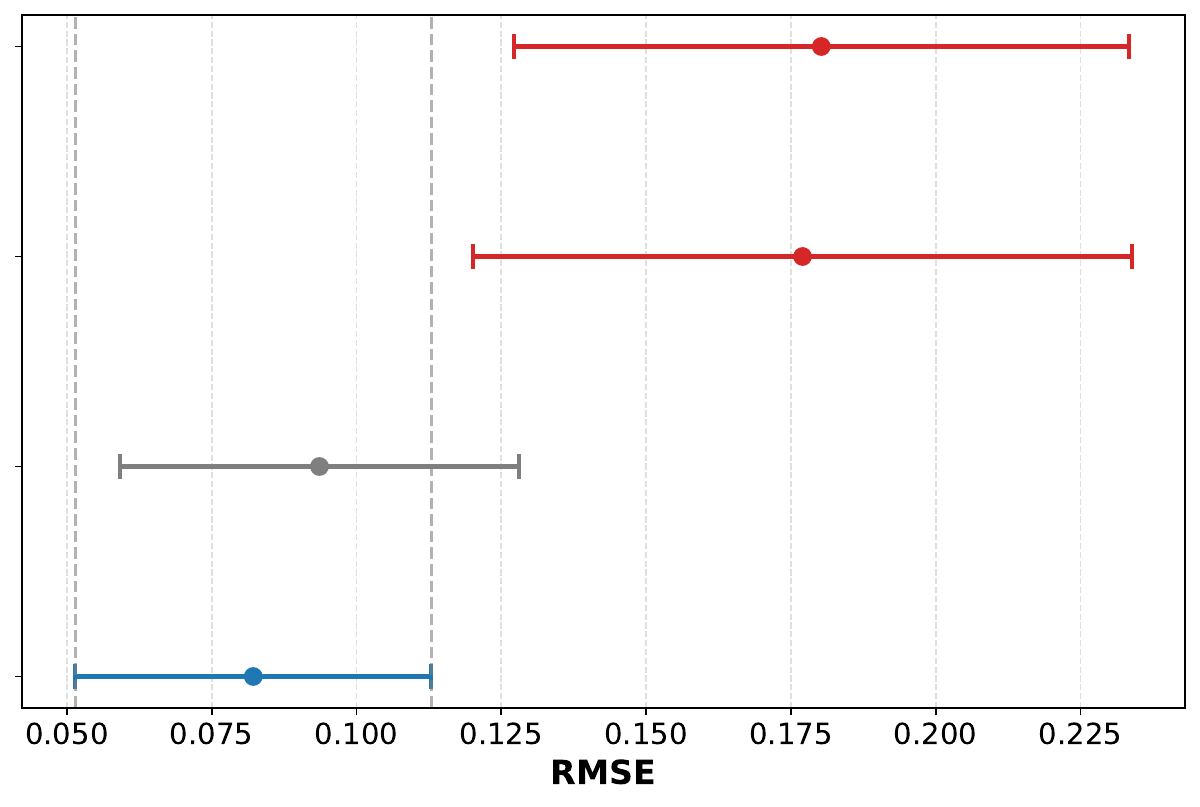}
  }
  \caption{\textbf{\thefigure:} \rre{The simultaneous confidence interval plots of the test-set \textbf{RMSE} (i.e., root mean squared error) on the downstream polymer property prediction task, where we compare PolyConFM against baselines under the \textbf{scaffold-based 5$\times$5 cross-validation}.
  Using the Tukey HSD testing procedure with a significance level of 0.05, the method with the best performance is displayed in blue, methods equivalent to the best model are represented in gray, and methods that show statistically significant differences from the best model are indicated in red. 
  The entire evaluation pipeline strictly adheres to the guidelines established in~\cite{ash2025practically}.
  }} 
  \label{fig: exp_scaffold_5x5_cv_CI_RMSE}
\end{figure}

\begin{figure}[H]
  \centering
  \subfigure[Egc]{
    \includegraphics[height=4.5cm]{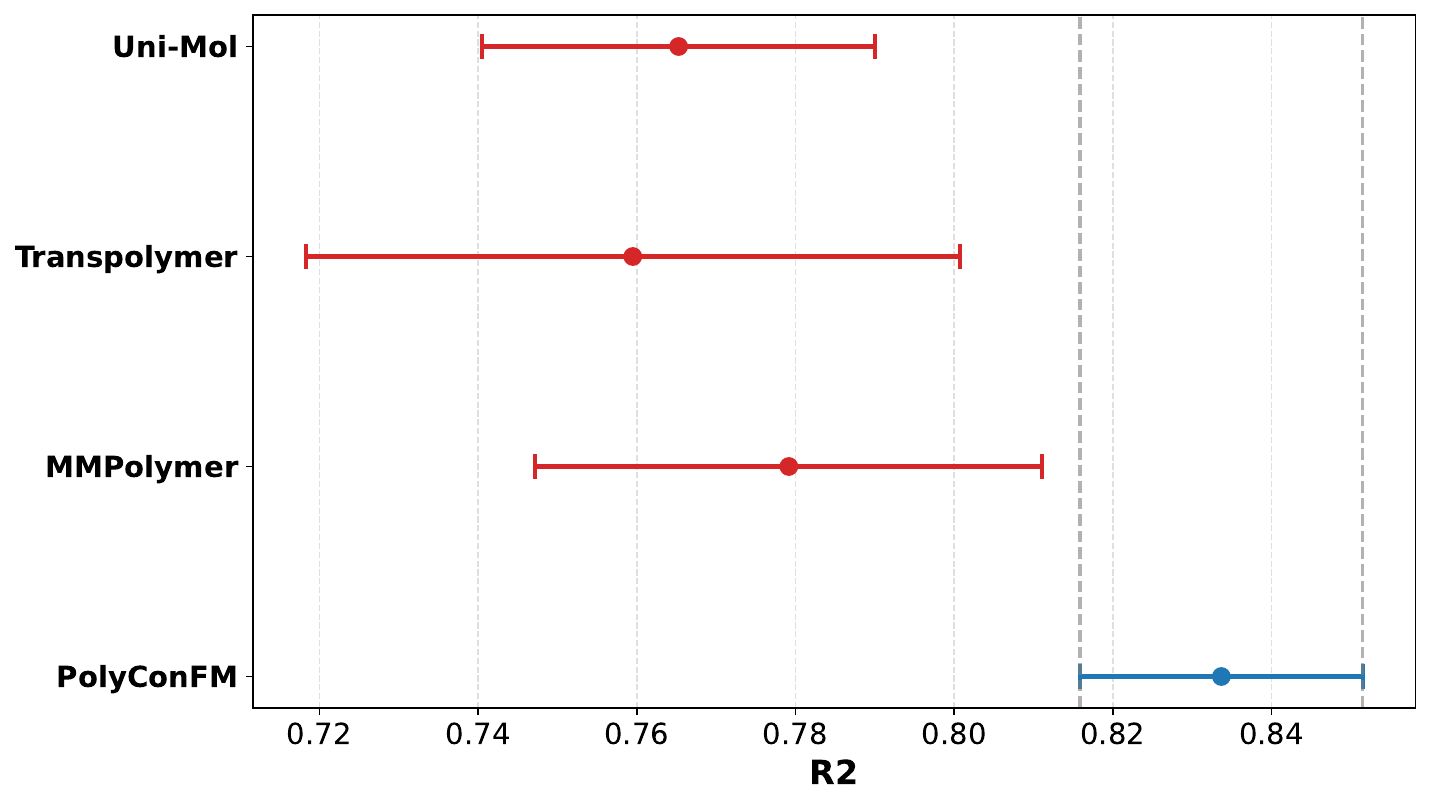}
  }
  \subfigure[Egb]{
    \includegraphics[height=4.5cm]{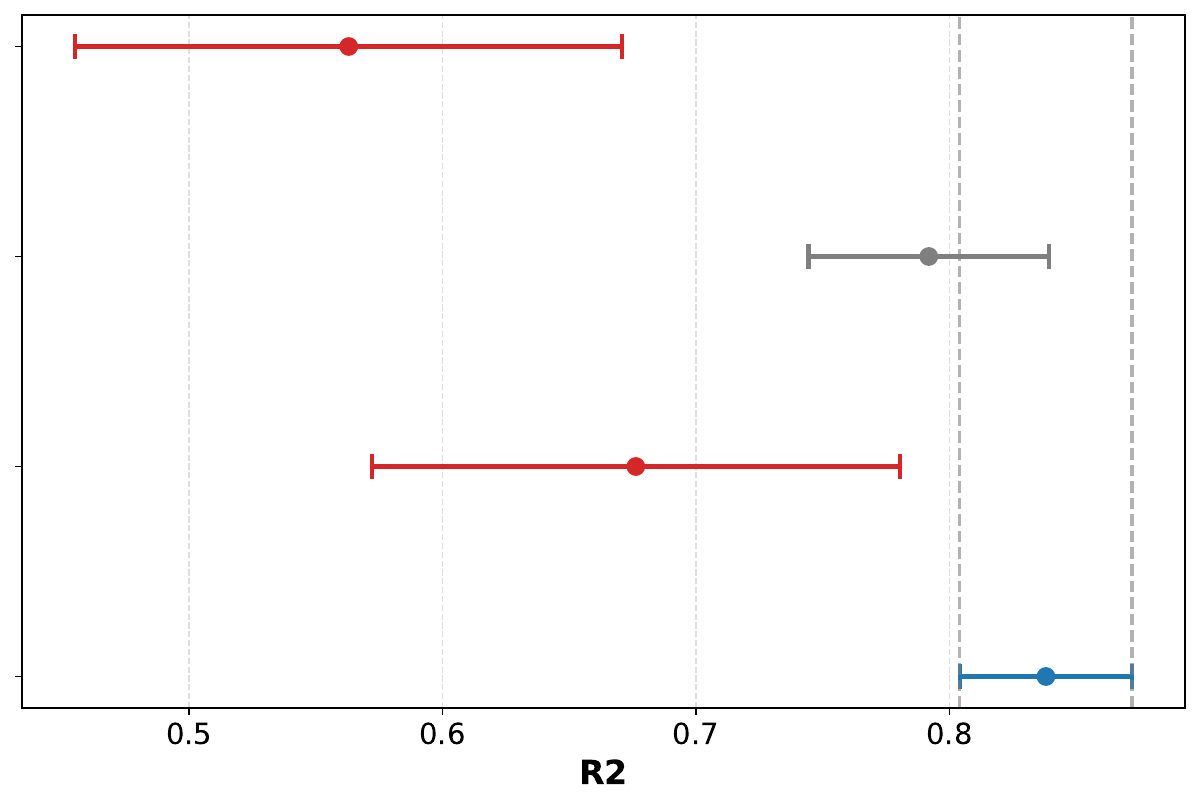}
  }
  \subfigure[Eea]{
    \includegraphics[height=4.5cm]{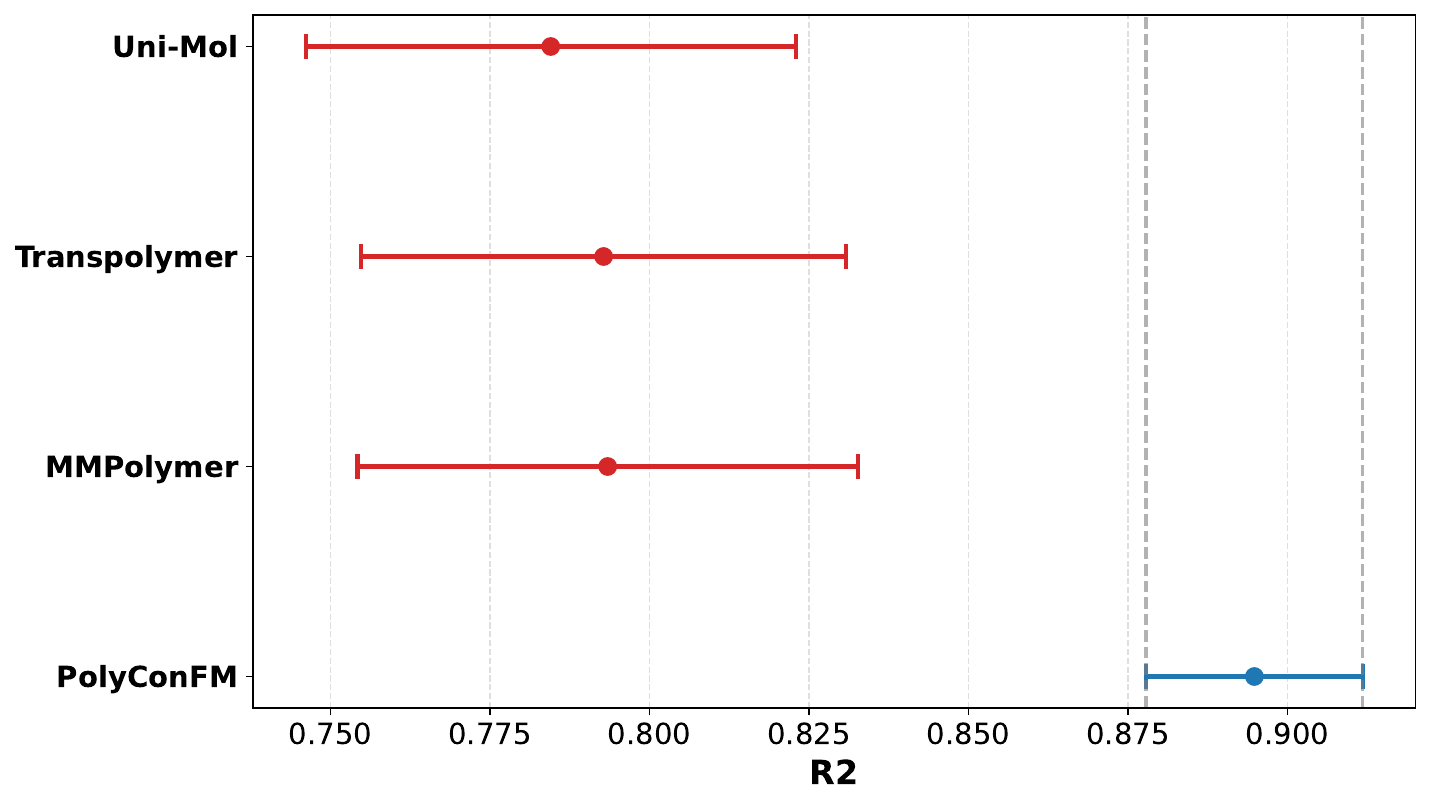}
  }
  \subfigure[Ei]{
    \includegraphics[height=4.5cm]{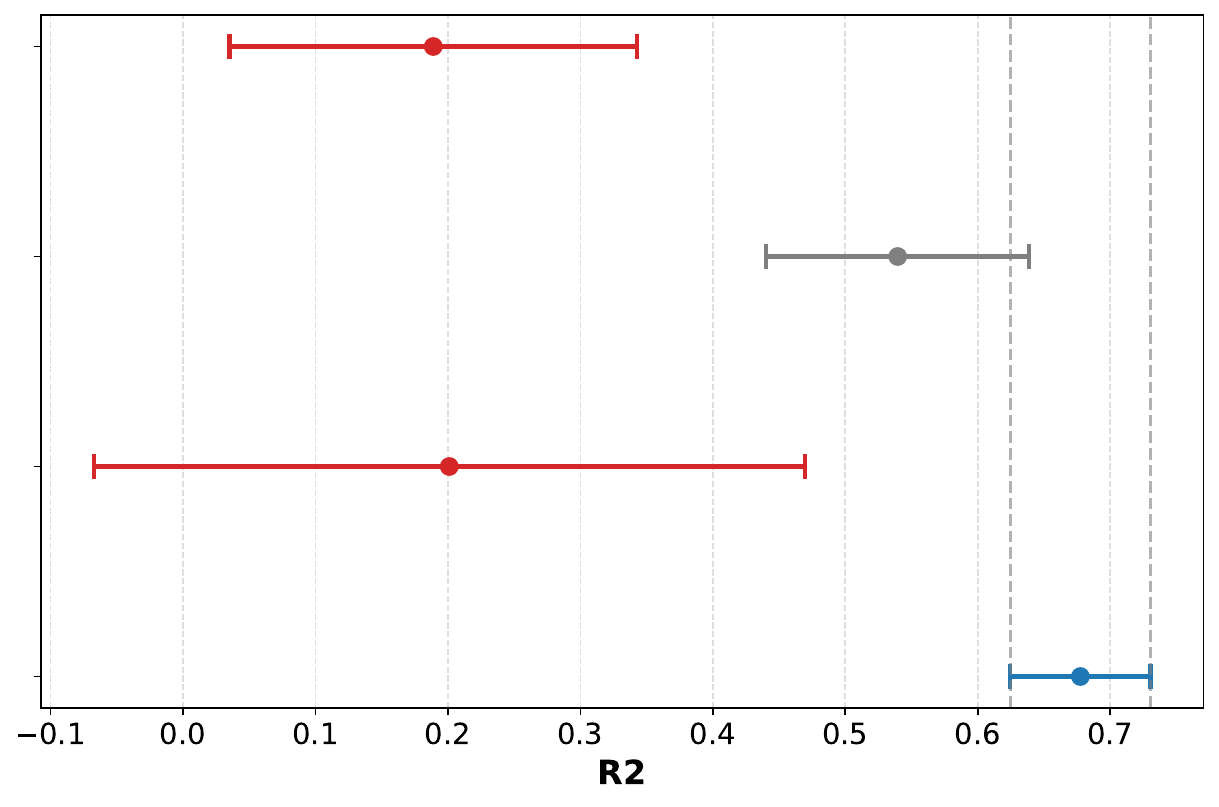}
  }
  \subfigure[Xc]{
    \includegraphics[height=4.5cm]{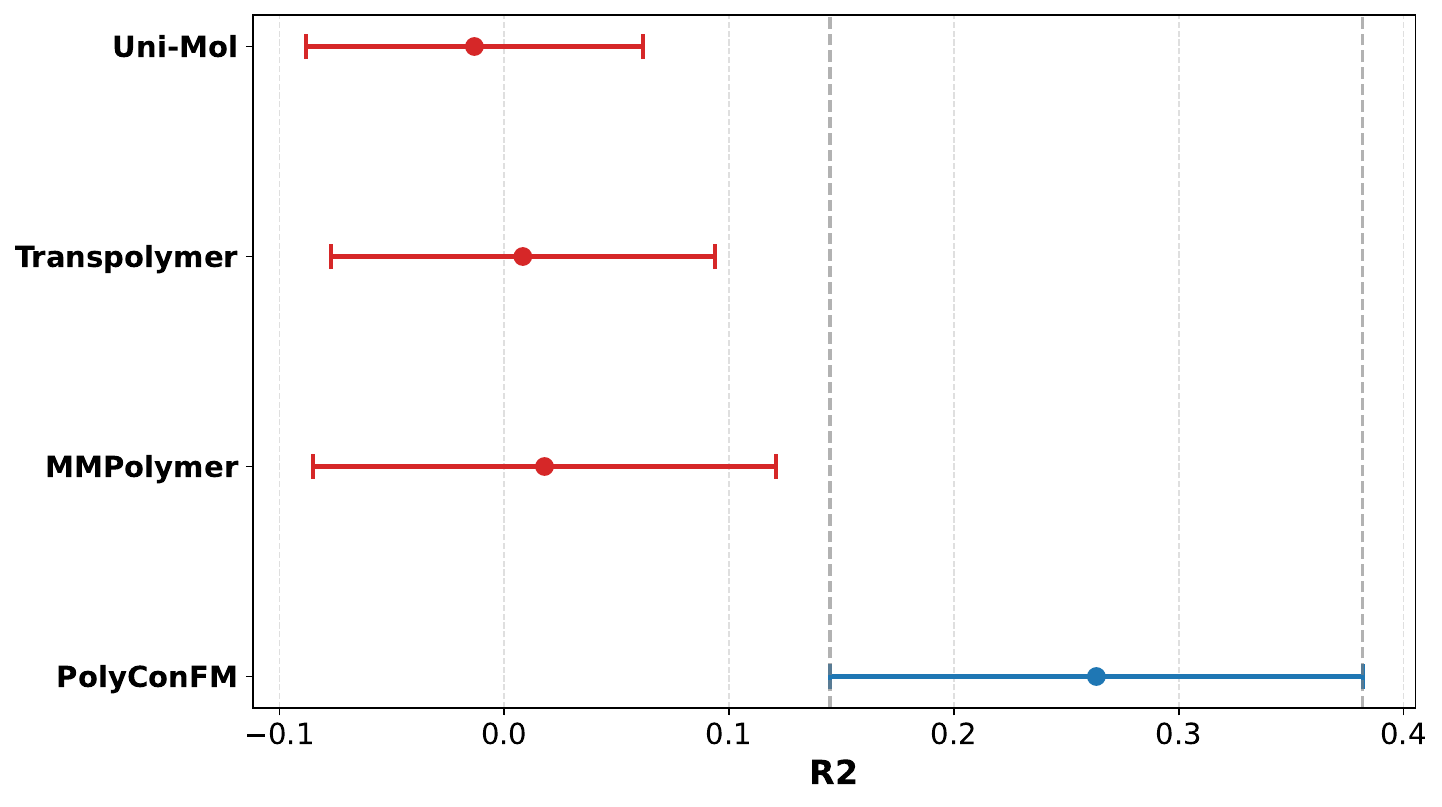}
  }
  \subfigure[EPS]{
    \includegraphics[height=4.5cm]{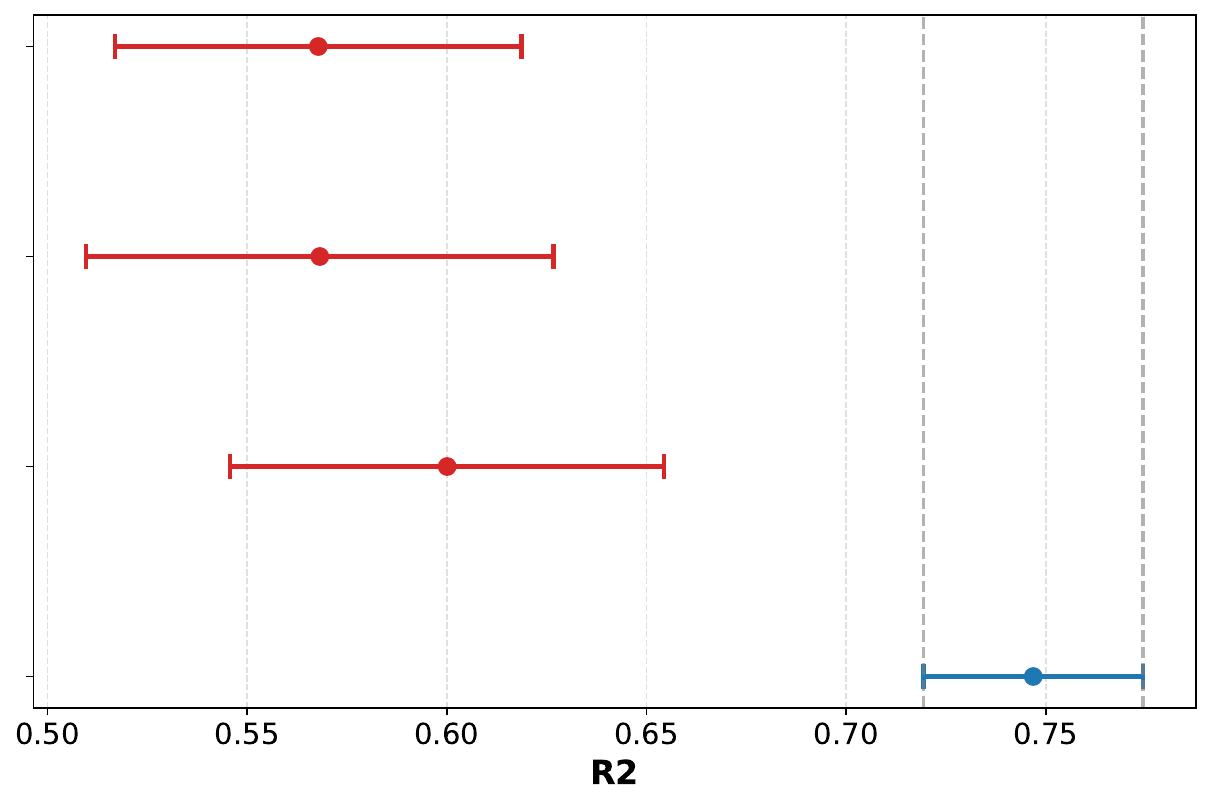}
  }
  \subfigure[Nc]{
    \includegraphics[height=4.5cm]{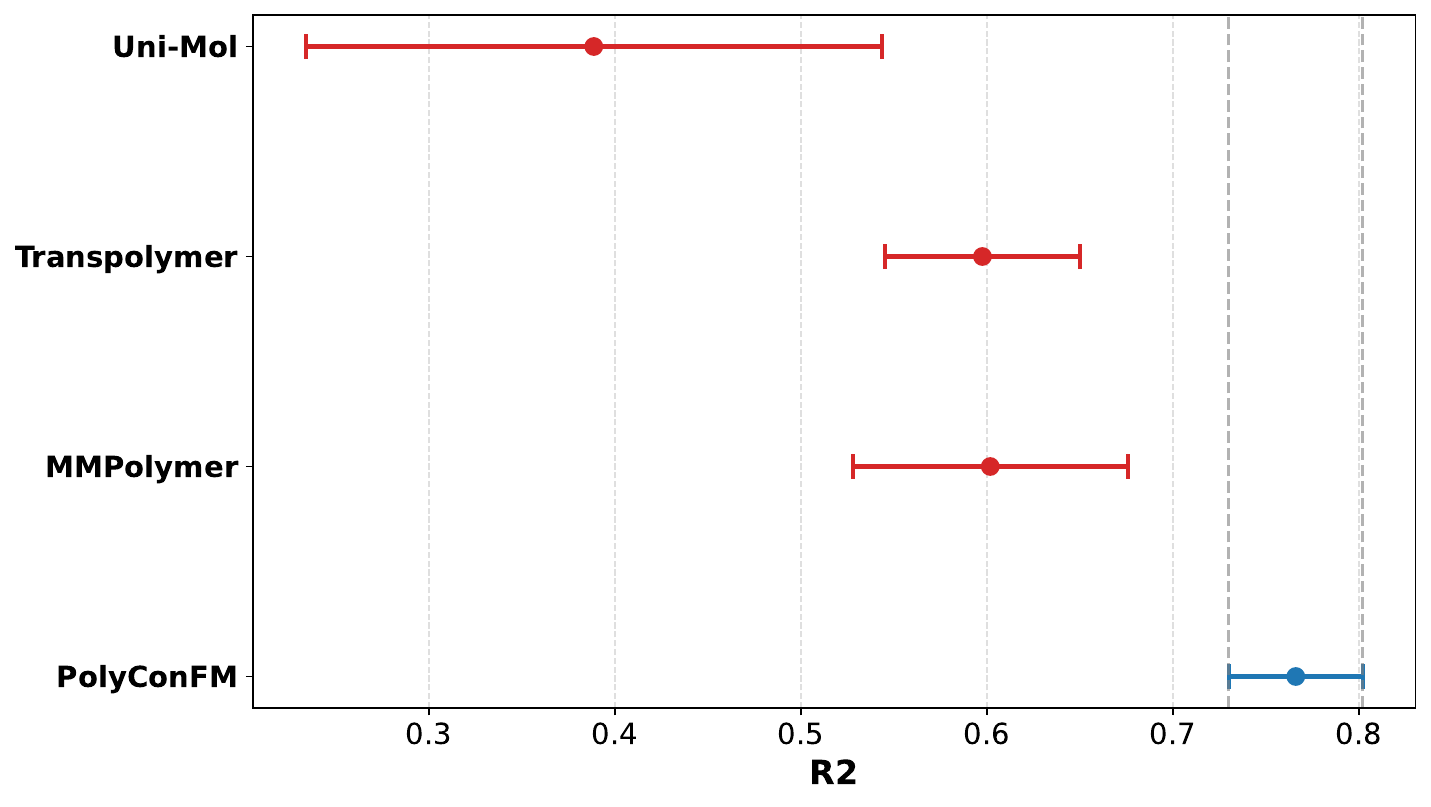}
  }
  \subfigure[Eat]{
    \includegraphics[height=4.5cm]{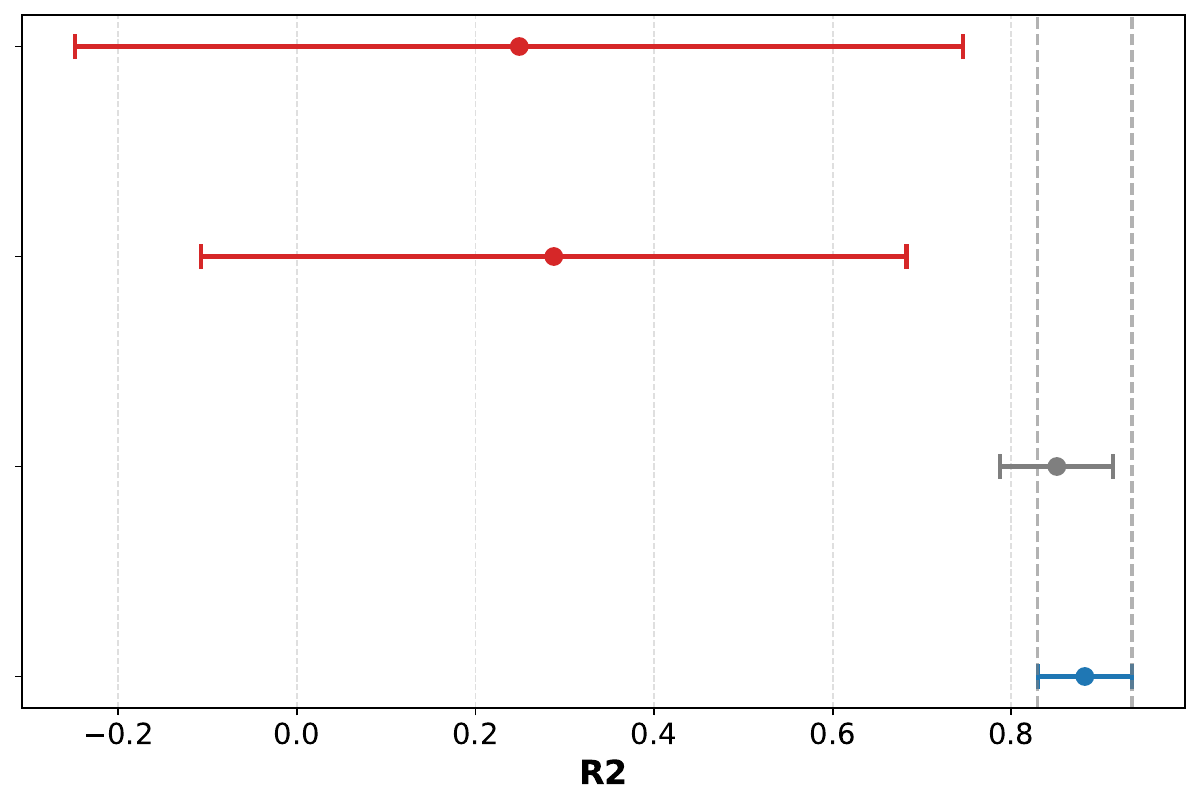}
  }
  \caption{\textbf{\thefigure:} \rre{The simultaneous confidence interval plots of the test-set $\bm {R^2}$ (i.e., coefficient of determination) on the downstream polymer property prediction task, where we compare PolyConFM against baselines under the \textbf{scaffold-based 5$\times$5 cross-validation}.
  Using the Tukey HSD testing procedure with a significance level of 0.05, the method with the best performance is displayed in blue, methods equivalent to the best model are represented in gray, and methods that show statistically significant differences from the best model are indicated in red. 
  The entire evaluation pipeline strictly adheres to the guidelines established in~\cite{ash2025practically}.
  }} 
  \label{fig: exp_scaffold_5x5_cv_CI_R2}
\end{figure}

\begin{figure}[H]
  \centering
  \subfigure[Egc]{
    \includegraphics[height=4.8cm]{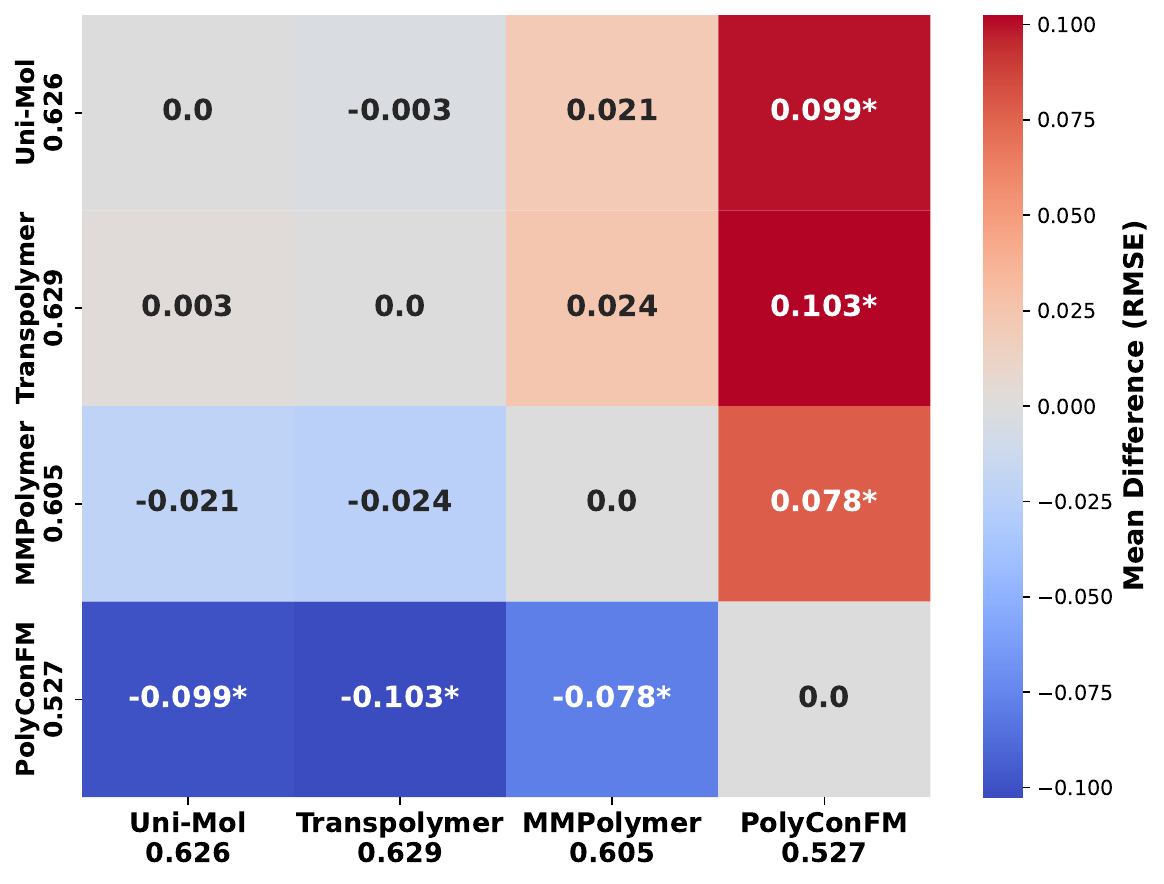}
  }
  \subfigure[Egb]{
    \includegraphics[height=4.8cm]{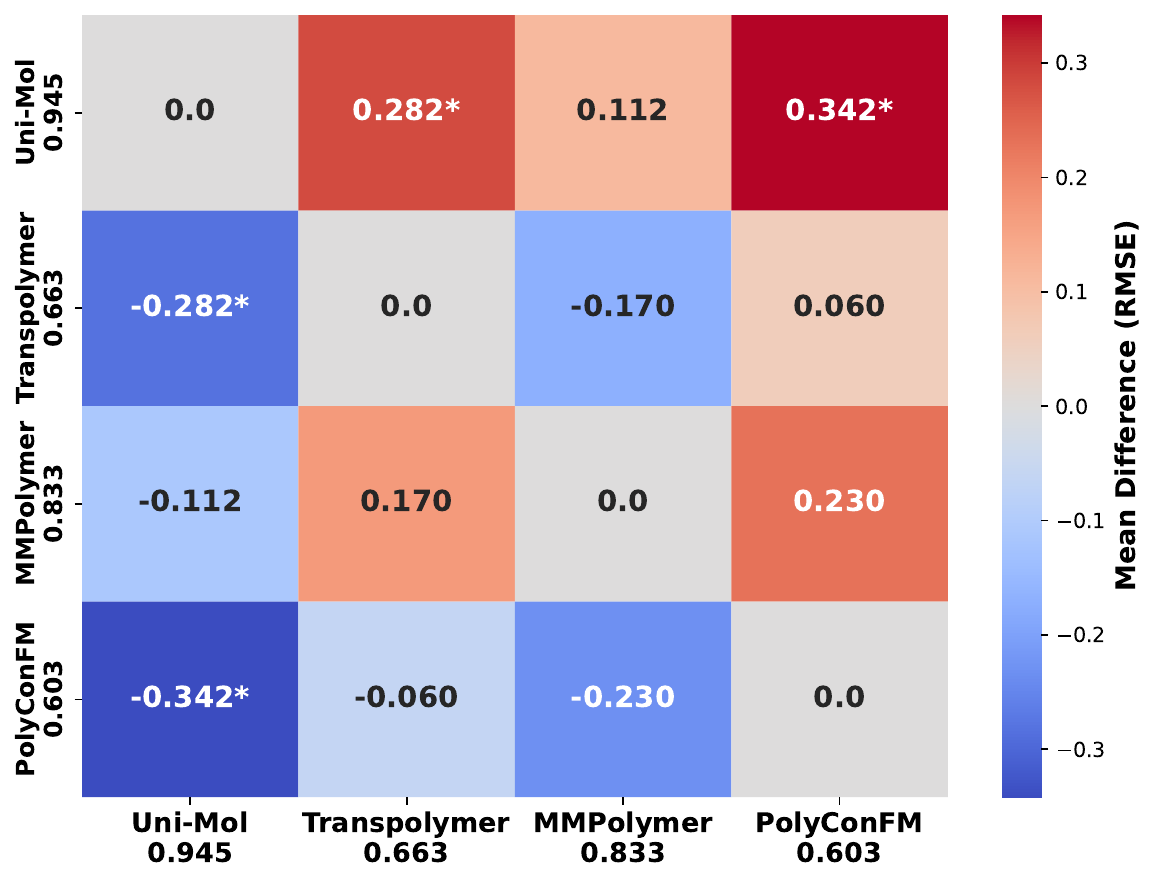}
  }
  \subfigure[Eea]{
    \includegraphics[height=4.8cm]{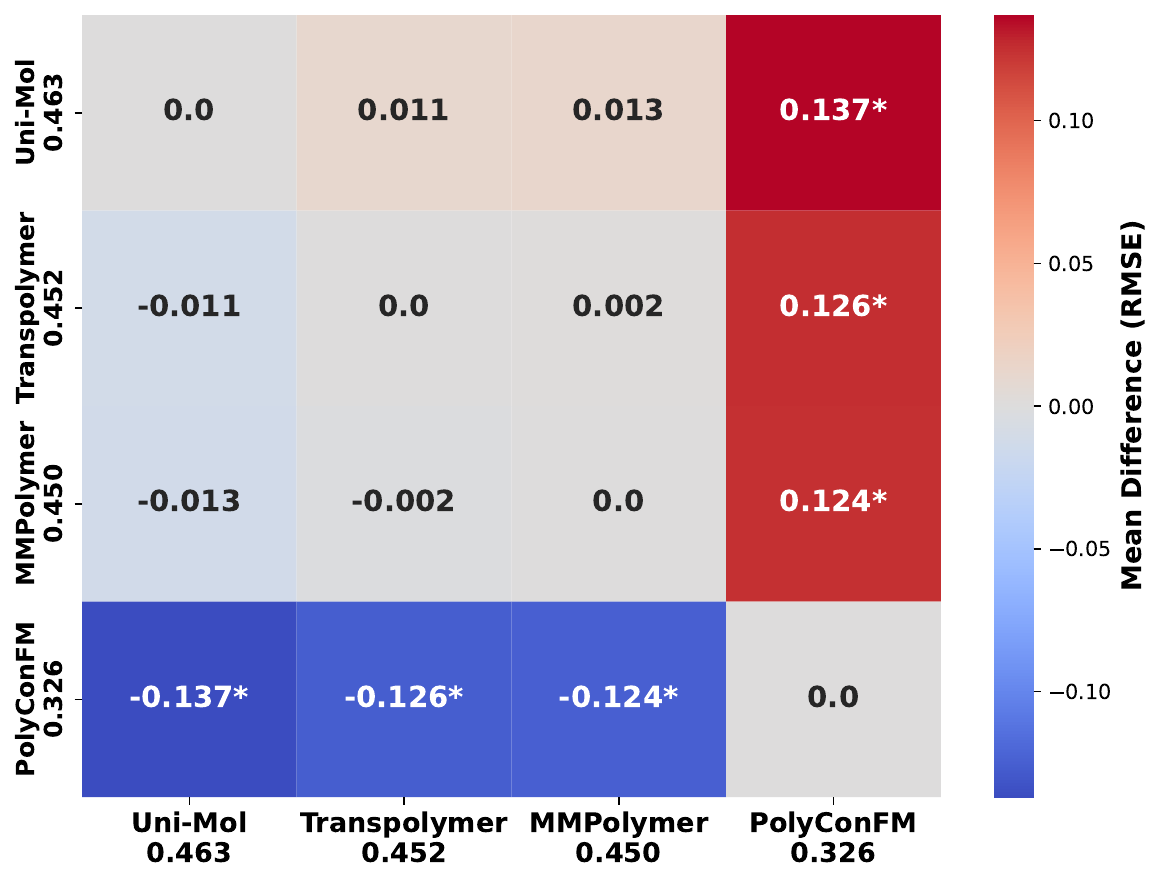}
  }
  \subfigure[Ei]{
    \includegraphics[height=4.8cm]{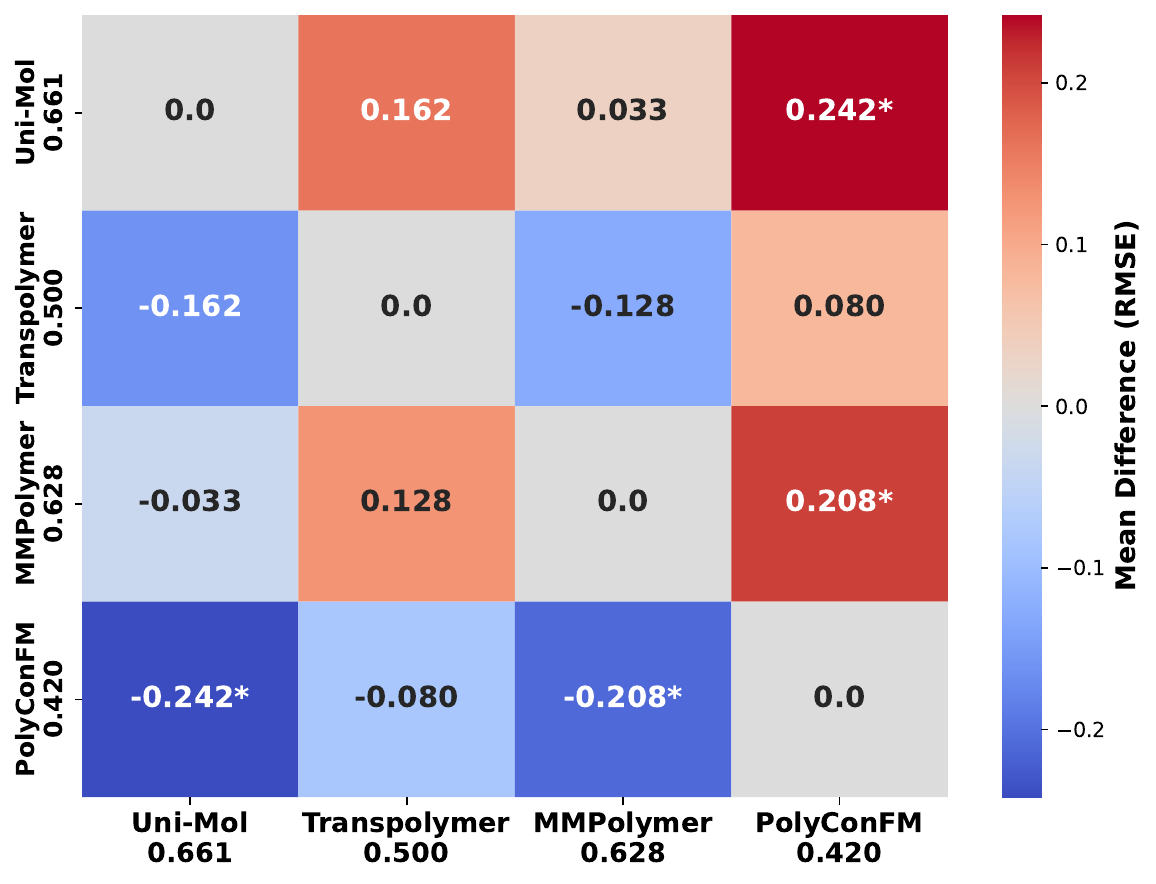}
  }
  \subfigure[Xc]{
    \includegraphics[height=4.8cm]{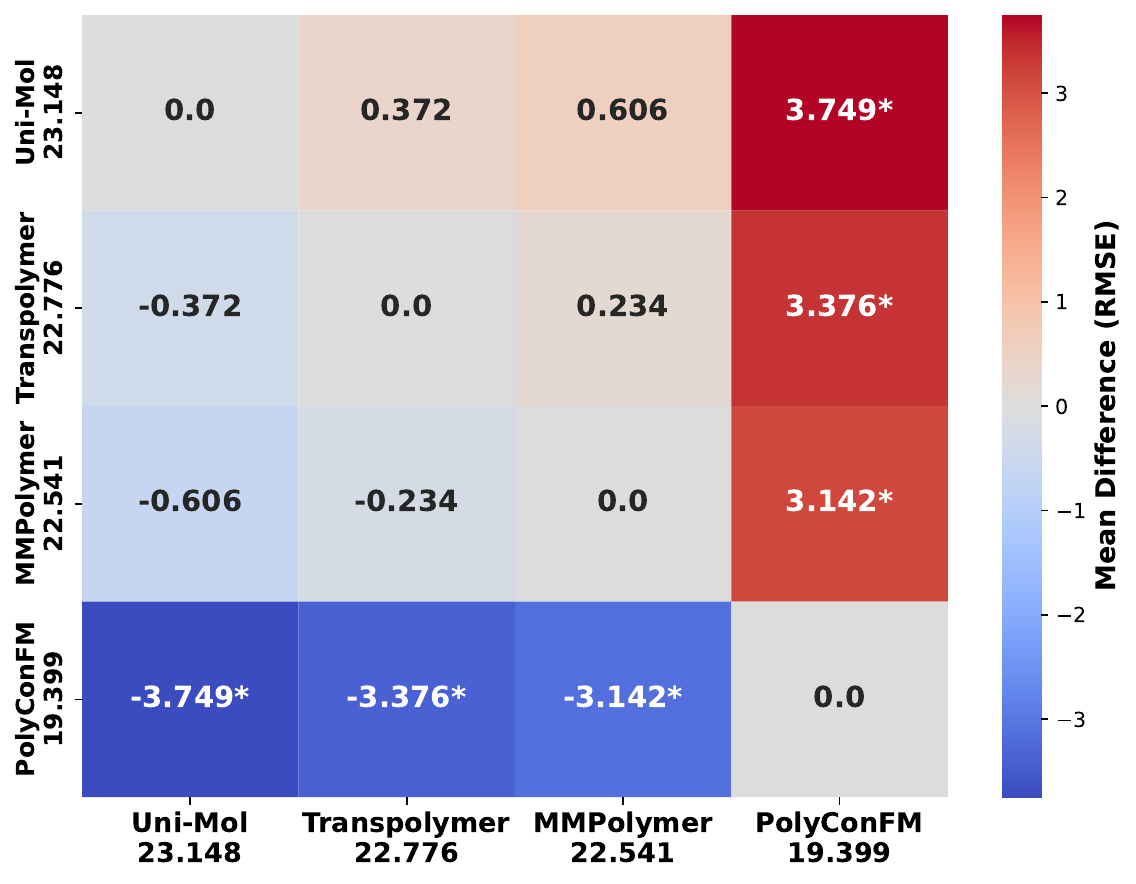}
  }
  \subfigure[EPS]{
    \includegraphics[height=4.8cm]{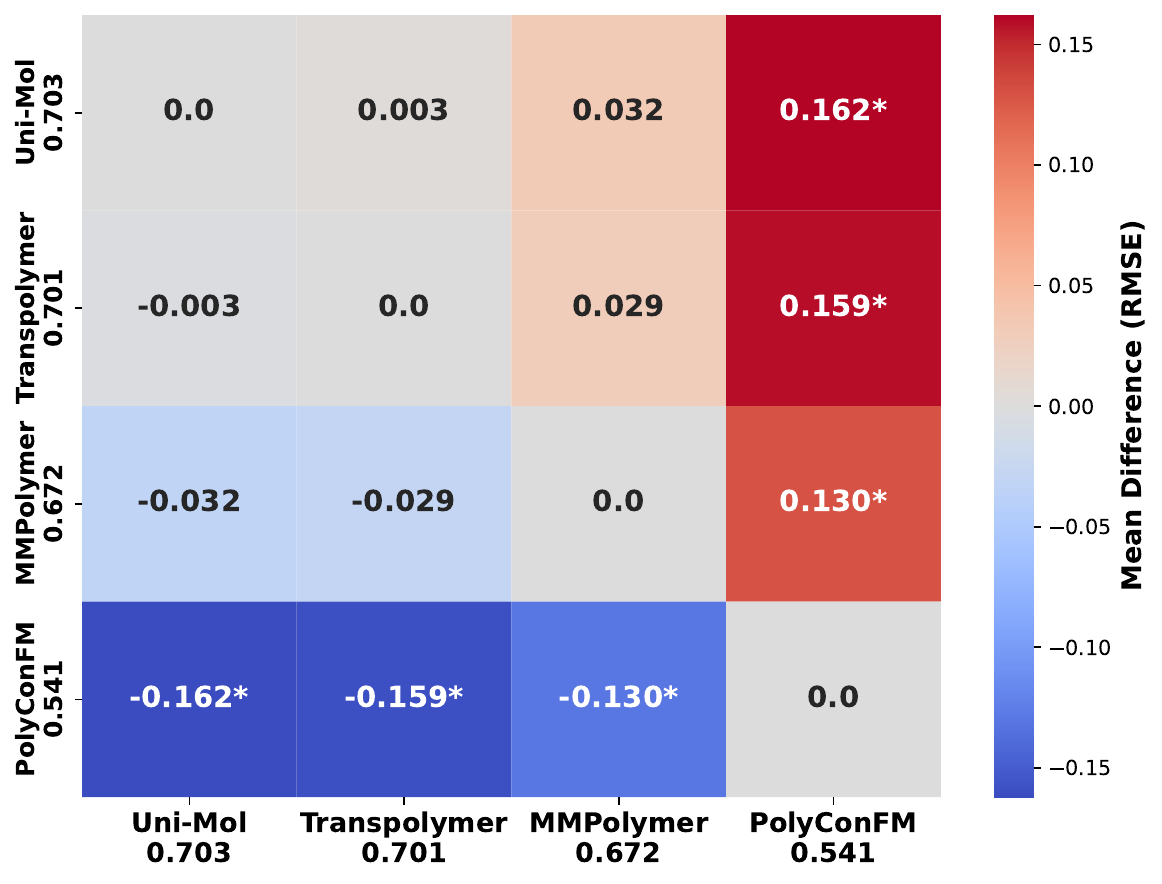}
  }
  \subfigure[Nc]{
    \includegraphics[height=4.8cm]{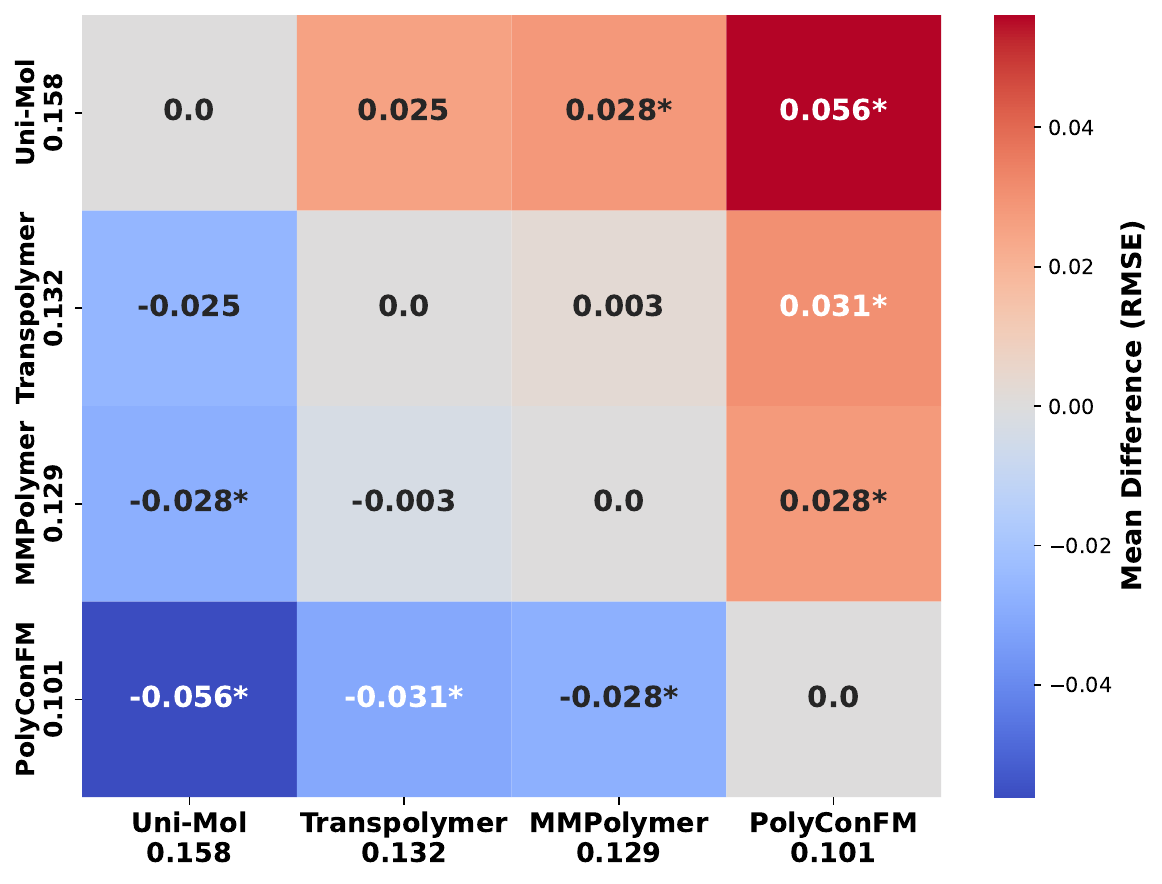}
  }
  \subfigure[Eat]{
    \includegraphics[height=4.8cm]{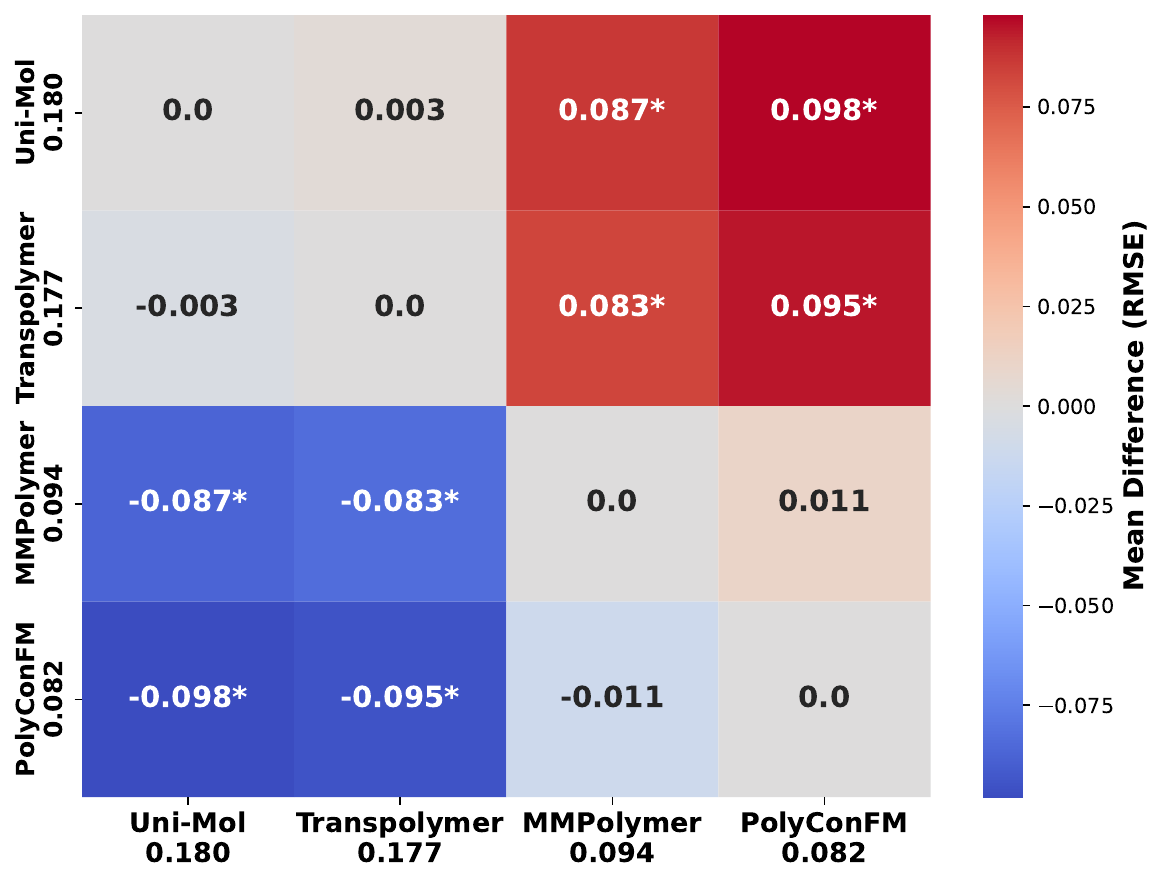}
  }
  \caption{\textbf{\thefigure:} \rre{The multiple comparisons similarity plots of the test-set \textbf{RMSE} (i.e., root mean squared error) on the downstream polymer property prediction task, where we compare PolyConFM against baselines under the \textbf{scaffold-based 5$\times$5 cross-validation}.
  Using the Tukey HSD testing procedure with a significance level of 0.05, statistically significant differences are annotated by star symbols.
  The entire evaluation pipeline strictly adheres to the guidelines established in~\cite{ash2025practically}.
  }} 
  \label{fig: exp_scaffold_5x5_cv_MCSim_RMSE}
\end{figure}

\begin{figure}[H]
  \centering
  \subfigure[Egc]{
    \includegraphics[height=4.8cm]{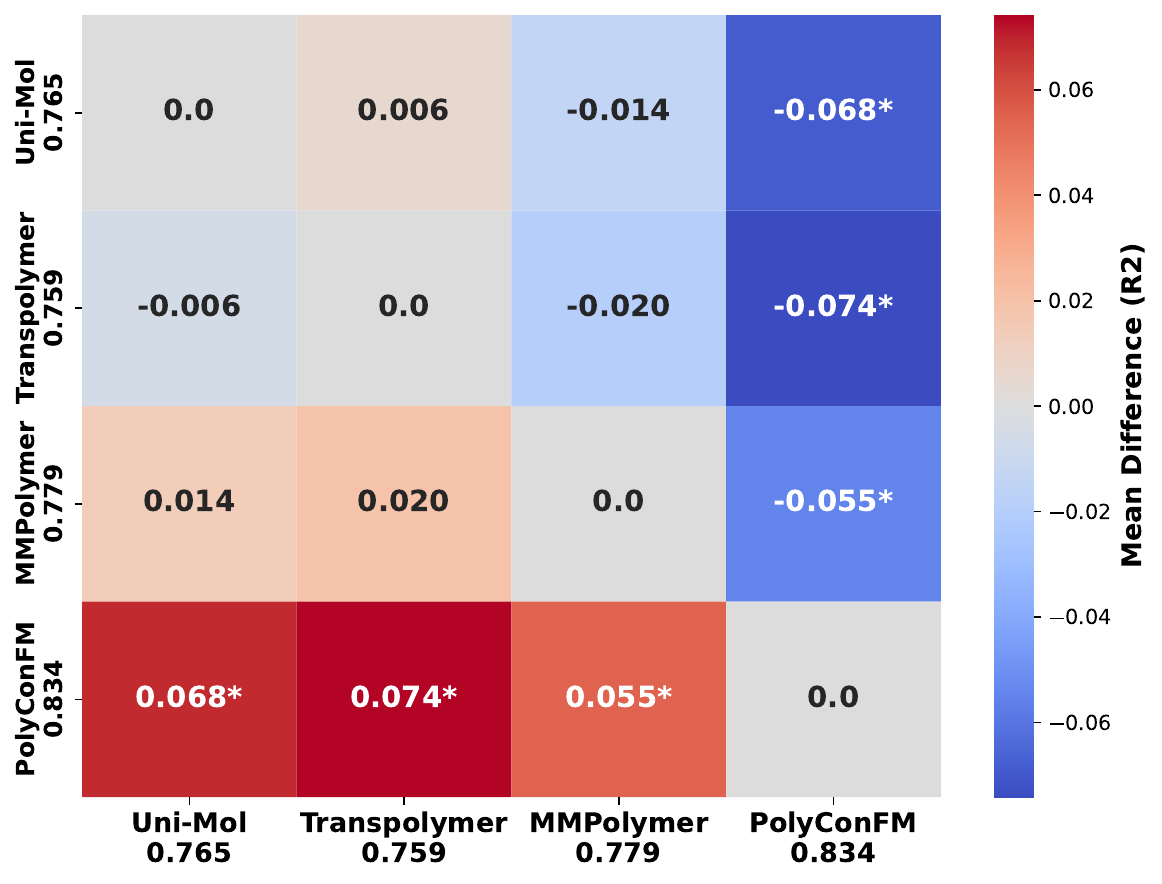}
  }
  \subfigure[Egb]{
    \includegraphics[height=4.8cm]{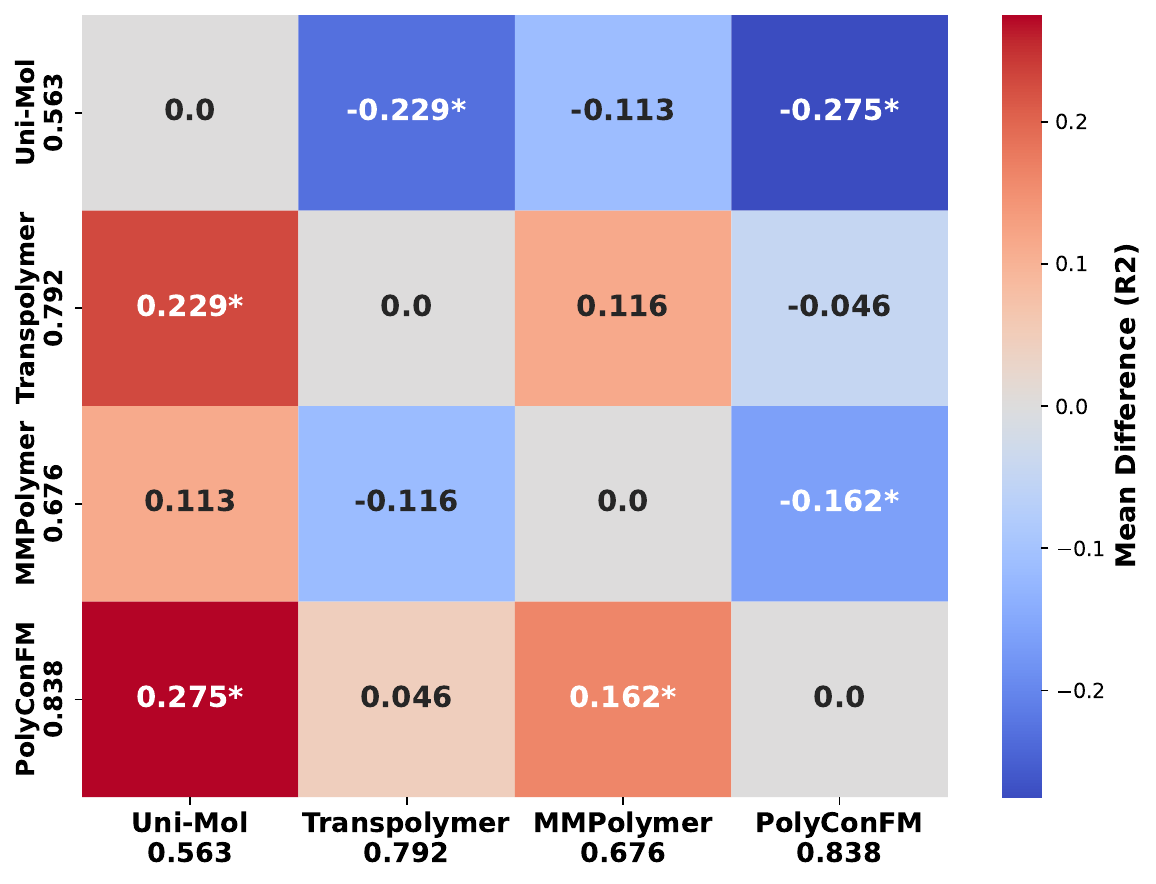}
  }
  \subfigure[Eea]{
    \includegraphics[height=4.8cm]{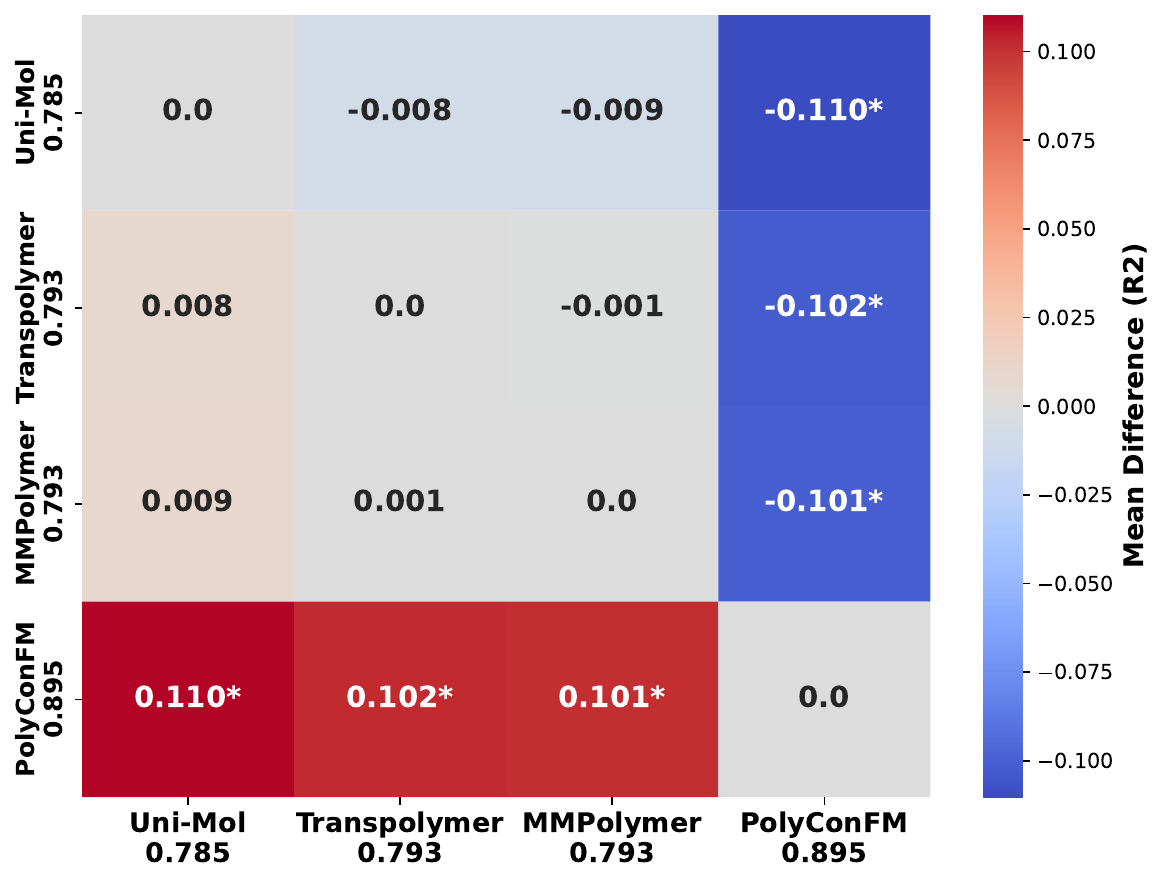}
  }
  \subfigure[Ei]{
    \includegraphics[height=4.8cm]{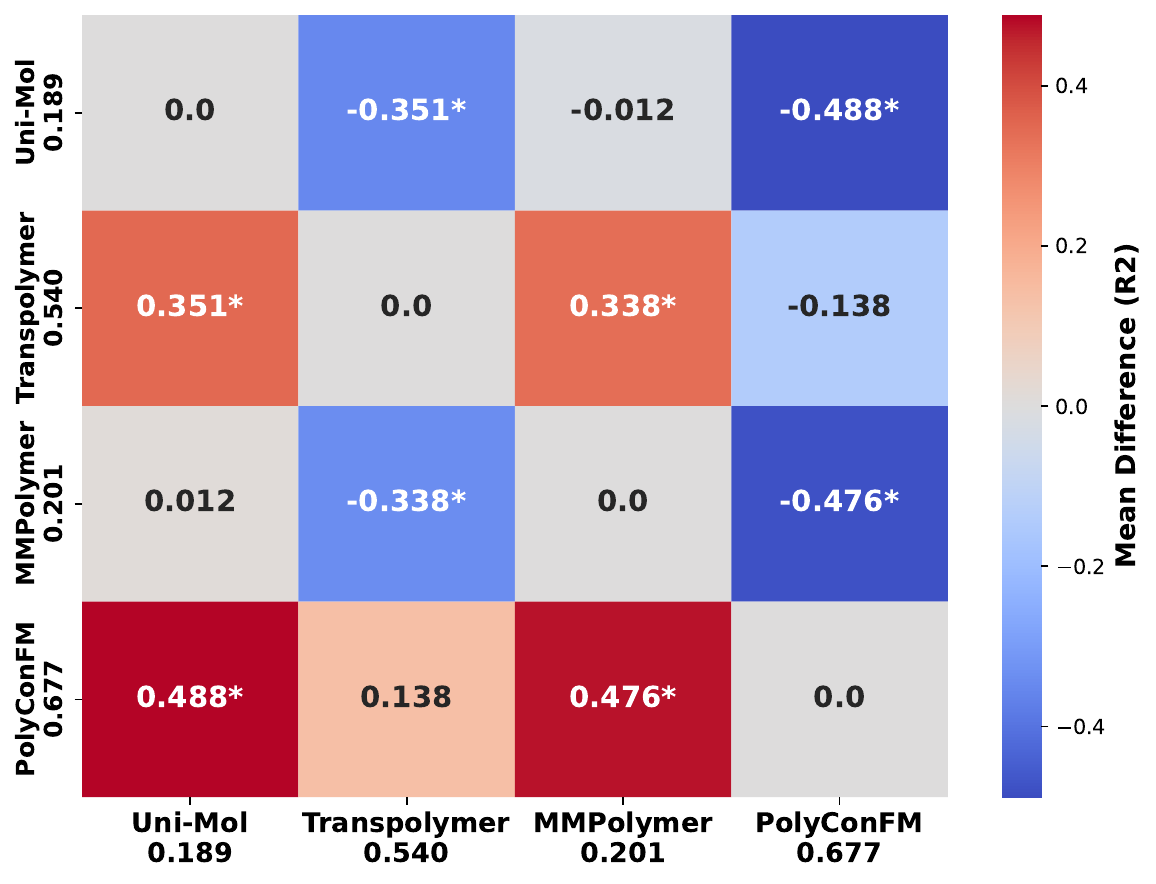}
  }
  \subfigure[Xc]{
    \includegraphics[height=4.8cm]{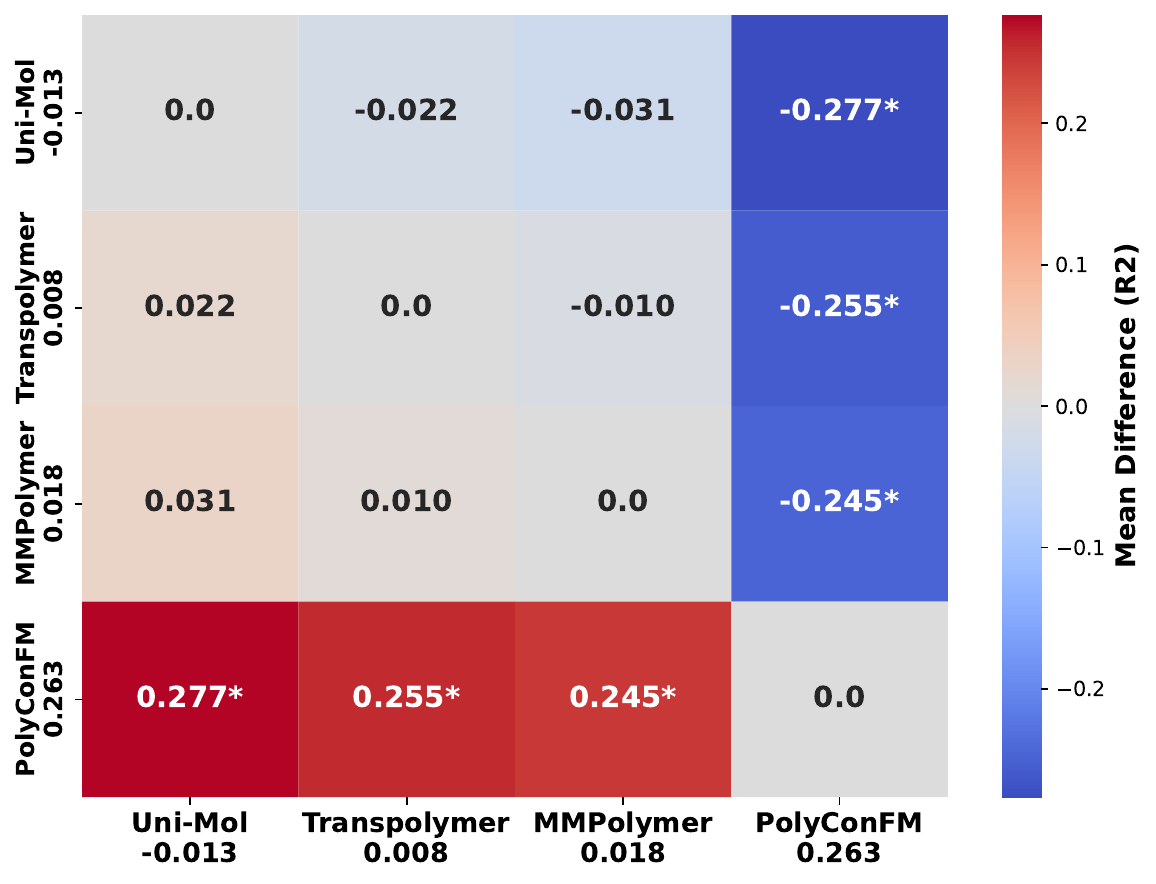}
  }
  \subfigure[EPS]{
    \includegraphics[height=4.8cm]{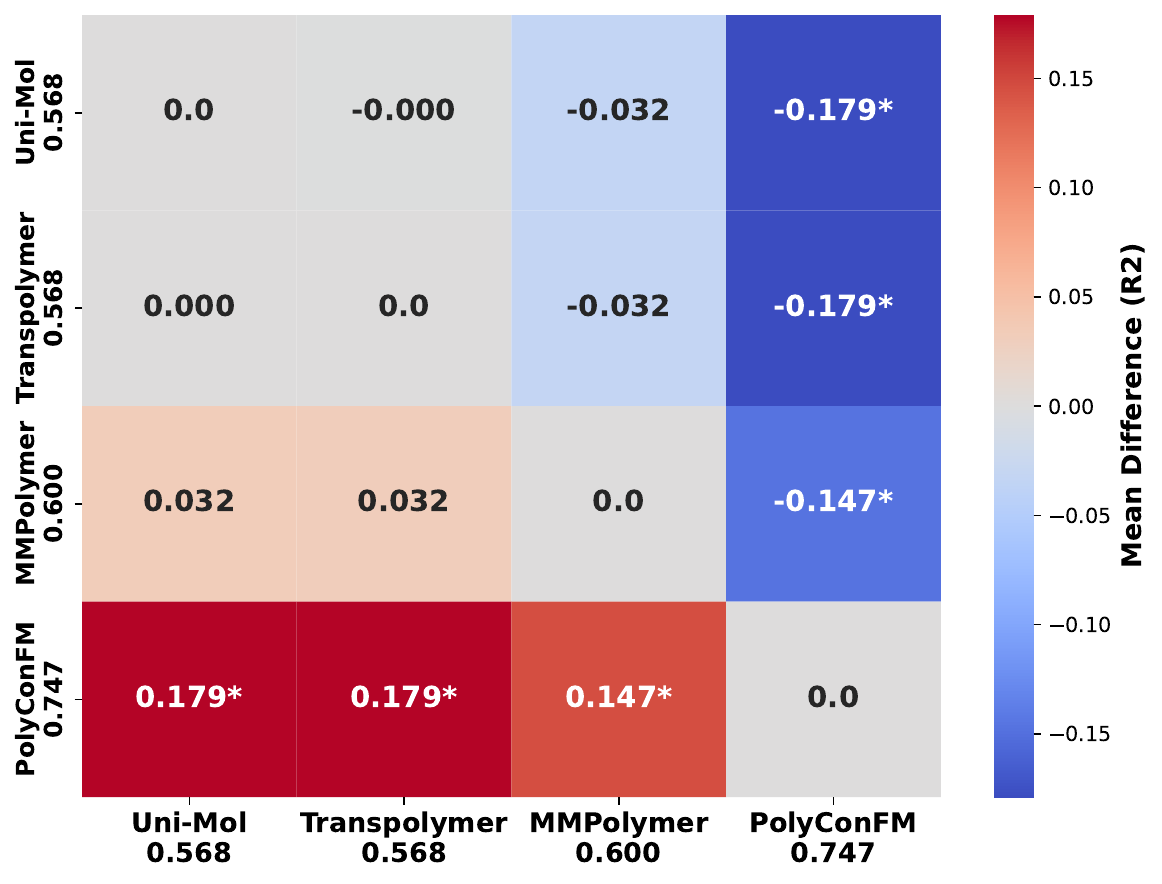}
  }
  \subfigure[Nc]{
    \includegraphics[height=4.8cm]{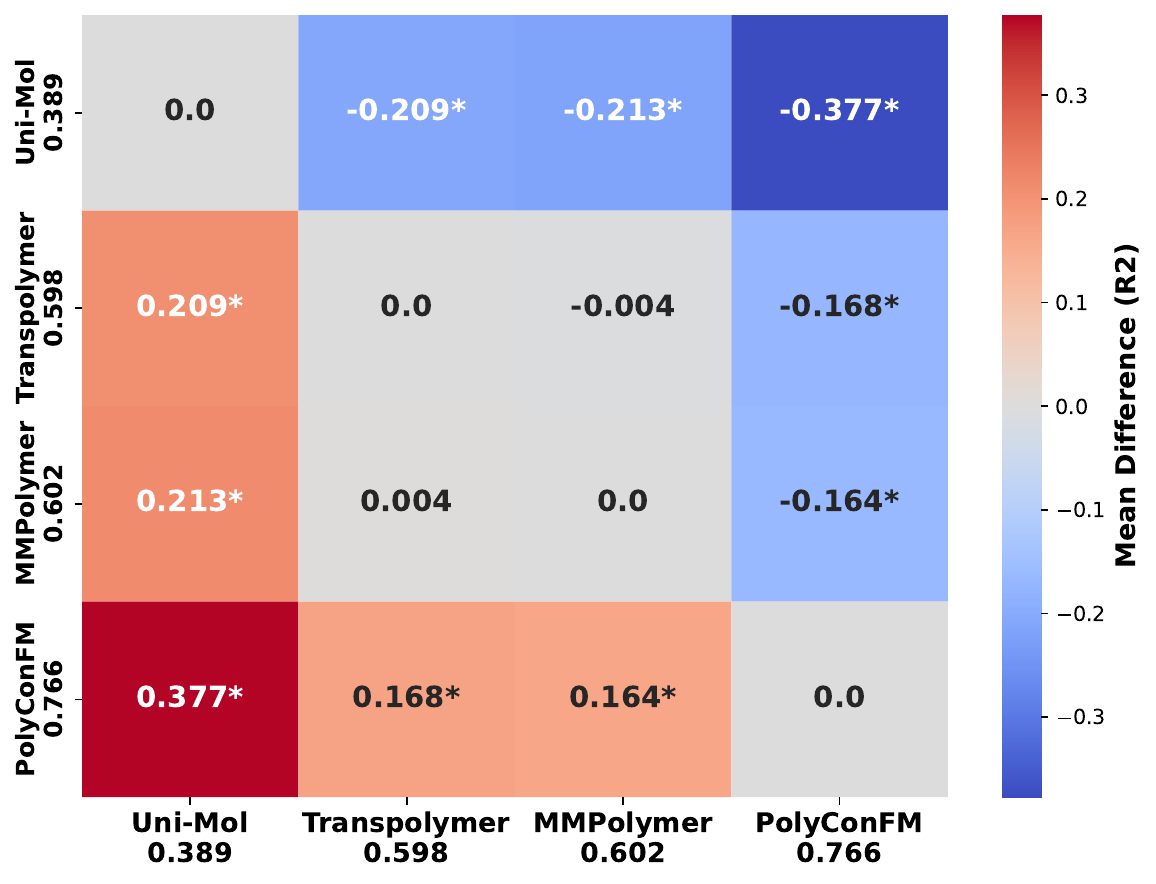}
  }
  \subfigure[Eat]{
    \includegraphics[height=4.8cm]{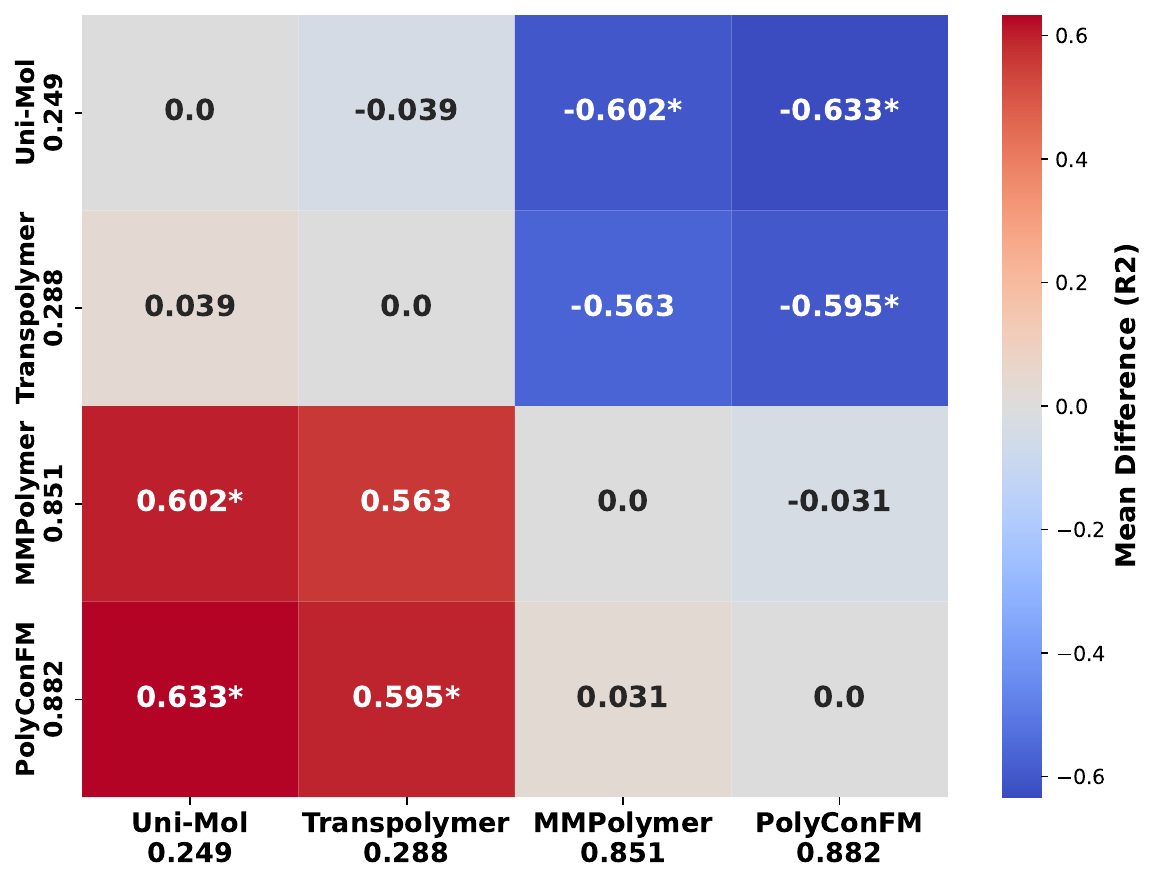}
  }
  \caption{\textbf{\thefigure:} \rre{The multiple comparisons similarity plots of the test-set $\bm {R^2}$ (i.e., coefficient of determination) on the downstream polymer property prediction task, where we compare PolyConFM against baselines under the \textbf{scaffold-based 5$\times$5 cross-validation}.
  Using the Tukey HSD testing procedure with a significance level of 0.05, statistically significant differences are annotated by star symbols.
  The entire evaluation pipeline strictly adheres to the guidelines established in~\cite{ash2025practically}.
  }} 
  \label{fig: exp_scaffold_5x5_cv_MCSim_R2}
\end{figure}

Here, we directly compare PolyConFM against three most competitive baselines (i.e., Uni-Mol, Transpolymer, and MMPolymer) identified in Table~\ref{tab: exp_proprety} under both cluster-based 5$\times$5 cross-validation and scaffold-based 5$\times$5 cross-validation.
Besides, instead of traditional performance-metric tables, we adopt the simultaneous confidence interval plot and multiple comparisons similarity plot from~\cite{ash2025practically} to present our results, while the Tukey HSD test from~\cite{ash2025practically} is also employed for pairwise comparisons to assess statistical significance.
As presented in~\ref{fig: exp_cluster_5x5_cv_CI_RMSE}--\ref{fig: exp_scaffold_5x5_cv_MCSim_R2}, PolyConFM consistently achieves superior performance across all eight property datasets, significantly outperforming these most competitive baselines under both cluster-based 5$\times$5 cross-validation and scaffold-based 5$\times$5 cross-validation.
More importantly, statistical analysis using the Tukey HSD test further confirms that the superior performance of PolyConFM is statistically significant in most cases.
As evidenced by the simultaneous confidence interval plots (i.e., \ref{fig: exp_cluster_5x5_cv_CI_RMSE}, \ref{fig: exp_cluster_5x5_cv_CI_R2}, \ref{fig: exp_scaffold_5x5_cv_CI_RMSE}, \ref{fig: exp_scaffold_5x5_cv_CI_R2}), PolyConFM is consistently represented in blue, indicating it is the best-performing model, while the competing baselines (i.e., Uni-Mol, Transpolymer, and MMPolymer) are frequently marked in red, signifying a statistically significant performance gap.
In addition, we observe that in a few specific cases (e.g., the Nc dataset under the cluster-based 5$\times$5 cross-validation), PolyConFM’s RMSE improvement over baselines is not statistically significant, whereas its $R^2$ advantage remains highly significant. 
This discrepancy stems from the fact that the 25 independent test sets obtained via the 5$\times$5 cross-validation have different label variances (i.e., $Var(y)$), thereby preventing the constant mapping between RMSE and $R^2$.
Given that $R^2 = 1 - \text{(RMSE)}^2 / Var(y)$ (Eq.~\ref{eq: rmse_r2}), one certain baseline might achieve a statistically comparable RMSE by performing well on specific test sets with low label variance, but its lower $R^2$ reveals that it actually fails to capture the relative data distribution.
In contrast, PolyConFM consistently maintains high $R^2$, demonstrating superior robustness regardless of data partitioning variations.
All these results underscore that, even under the most stringent 5$\times$5 cross-validation, PolyConFM consistently exhibits superior predictive performance and exceptional reliability, establishing it as a powerful tool for predicting polymer properties. 

\newpage

\newpage
\subsection{Polymer Design}\label{SI-sec: exp_design}

Following established benchmarks in this field~\cite{liu2024graph, gao2022sample}, we also utilize the Random Forests (RF) trained on task-related polymers as the evaluation oracle for the downstream polymer design task. 
Specifically, recent studies such as GraphDiT~\cite{liu2024graph} have already systematically compared RF with other typical surrogate models, including Gaussian Processes (GP) and Support Vector Machines (SVM), to validate the rationality of this oracle choice.
Their analysis demonstrates that RF not only yields highly consistent relative rankings but also delivers superior predictive accuracy for task-related polymer properties (including three specific gas permeabilities of interest this work).
Besides, while perfectly approximating the ground-truth properties via any surrogate model is inherently difficult, it is crucial to note that both PolyConFM and all baselines are evaluated using the same evaluation oracle.
It means that any inherent biases of the chosen surrogate model apply equally to all methods, thereby effectively neutralizing potential evaluation disparities and ensuring a strictly fair comparison for their relative performance.

Beyond ensuring the calibration of our chosen oracle through the low MAE values validated in the previous work~\cite{liu2024graph}, we further provide comprehensive uncertainty estimates to verify that PolyConFM does not exploit potential oracle quirks (a phenomenon also known as ``reward hacking'').
Specifically, we leverage the intrinsic ensemble architecture of the Random Forest to quantify the predictive uncertainty, which is defined as the standard deviation of predictions across all constituent decision trees for each candidate polymer.
Notably, in the context of ensemble learning, high variance among base learners typically indicates that a sample resides in the out-of-distribution (OOD) region, a classic hallmark of reward hacking where models generate unrealistic candidates to exploit the surrogate's blind spots.
As presented in~\ref{tab: exp_design_oracle_uncertainty}, we report the mean and median of the uncertainty values calculated across all generated polymers for each method, demonstrating that PolyConFM consistently achieves the lowest predictive uncertainty among all methods across all targeted gas permeabilities.
Most importantly, the uncertainty profiles of PolyConFM closely align with the Reference values derived from real polymers.
These comprehensive results confirm that PolyConFM successfully generates physically plausible and in-distribution polymers that remain within the well-characterized chemical space, rather than drifting into unreliable OOD regions to inflate performance scores.

\begin{table}[t]
\centering
\setlength{\tabcolsep}{4.3pt}
\caption{\textbf{\thetable:} \re{The statistical analysis of predictive uncertainty across different methods.
For each candidate polymer, the predictive uncertainty is defined as the standard deviation of predictions across all constituent decision trees within the Random Forest oracle.
We report the \textbf{Mean} and \textbf{Median} of these uncertainty values calculated across all generated polymers.
In particular, the same statistical metrics corresponding to \textbf{Reference} are calculated using real polymers from the test set to reflect the intrinsic variance of the underlying physical distribution.
}}
\begin{tabular}{lcccccccc}
\toprule
\multirow{2}[2]{*}{Method} & \multicolumn{4}{c}{Mean} & \multicolumn{4}{c}{Median} \\
\cmidrule(lr){2-5} \cmidrule(lr){6-9}
 & O2Perm $\downarrow$ & N2Perm $\downarrow$ & CO2Perm $\downarrow$ & Average $\downarrow$ & O2Perm $\downarrow$ & N2Perm $\downarrow$ & CO2Perm $\downarrow$ & Average $\downarrow$ \\
\midrule
Reference & 0.381 & 0.409 & 0.408 & {0.399} & 0.315 & 0.341 & 0.322 & {0.326} \\
\midrule
MolGPT~\cite{bagal2021molgpt}    & 0.670 & 0.700 & 0.711 & {0.694} & 0.571 & 0.598 & 0.613 & {0.594} \\
GraphGA~\cite{jensen2019graph}   & 0.611 & 0.611 & 0.620 & {0.614} & 0.565 & 0.576 & 0.568 & {0.570} \\
DiGress~\cite{vignac2023digress}   & 1.184 & 1.234 & 1.192 & {1.203} & 1.147 & 1.226 & 1.142 & {1.172} \\
GDSS~\cite{jo2022score}      & 1.297 & 1.046 & 1.562 & {1.302} & 1.129 & 1.049 & 1.639 & {1.272} \\
MOOD~\cite{lee2023exploring}      & 1.307 & 1.016 & 1.445 & {1.256} & 1.233 & 0.967 & 1.439 & {1.213} \\
GraphDiT~\cite{liu2024graph} & 0.657 & 0.683 & 0.670 & {0.670} & 0.521 & 0.570 & 0.540 & {0.544} \\
\cellcolor{lightblue} \kern-2.8pt PolyConFM & \cellcolor{lightblue} \kern-2.8pt \textbf{0.577} & \cellcolor{lightblue} \kern-2.8pt \textbf{0.590} & \cellcolor{lightblue} \kern-2.8pt \textbf{0.596} & 
\cellcolor{lightblue} \kern-2.8pt \textbf{0.588} & \cellcolor{lightblue} \kern-2.8pt \textbf{0.460} & 
\cellcolor{lightblue} \kern-2.8pt \textbf{0.464} & \cellcolor{lightblue} \kern-2.8pt \textbf{0.419} & 
\cellcolor{lightblue} \kern-2.8pt \textbf{0.448} \\
\bottomrule
\end{tabular}
\label{tab: exp_design_oracle_uncertainty}
\end{table}

\begin{figure*}[t]
    \centering 
        \includegraphics[width=0.999\textwidth]{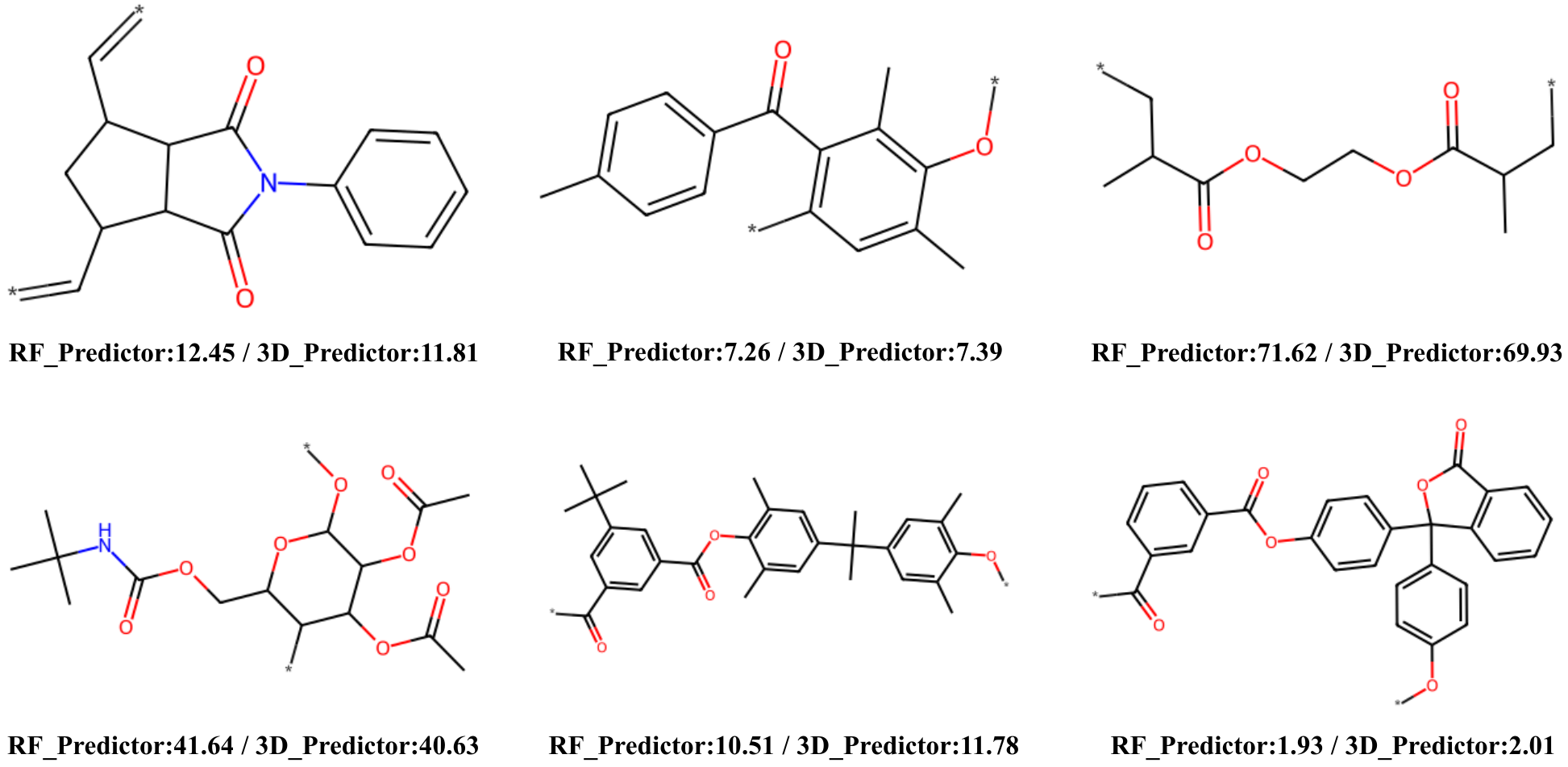} 
    \caption{\textbf{\thefigure:} \re{High-fidelity validation of the Random Forest (RF) oracle through the 3D-aware predictor, where we report their predicted property values on several representative polymers.}}
    \label{fig: design_oracle_validation}
\end{figure*}

Moreover, given that PolyConFM also achieves state-of-the-art performance on the polymer property prediction task (as detailed in Section~\ref{sec: results_property}), we further employ this 3D-aware model as a higher-fidelity predictor to verify the consistency and reliability of our RF-based evaluation results.
In particular, we select several representative polymers and re-predict their properties using this higher-fidelity predictor.
As presented in~\ref{fig: design_oracle_validation}, the property values predicted by our RF oracle demonstrate a remarkable consensus with those from the high-fidelity predictor across a wide numerical range.
This strong consensus underscores the credibility of our RF oracle in evaluating polymers, thereby ensuring that PolyConFM optimizes physically grounded properties rather than exploiting topological artifacts.

\begin{figure*}[t]
    \centering 
        \includegraphics[width=0.995\textwidth]{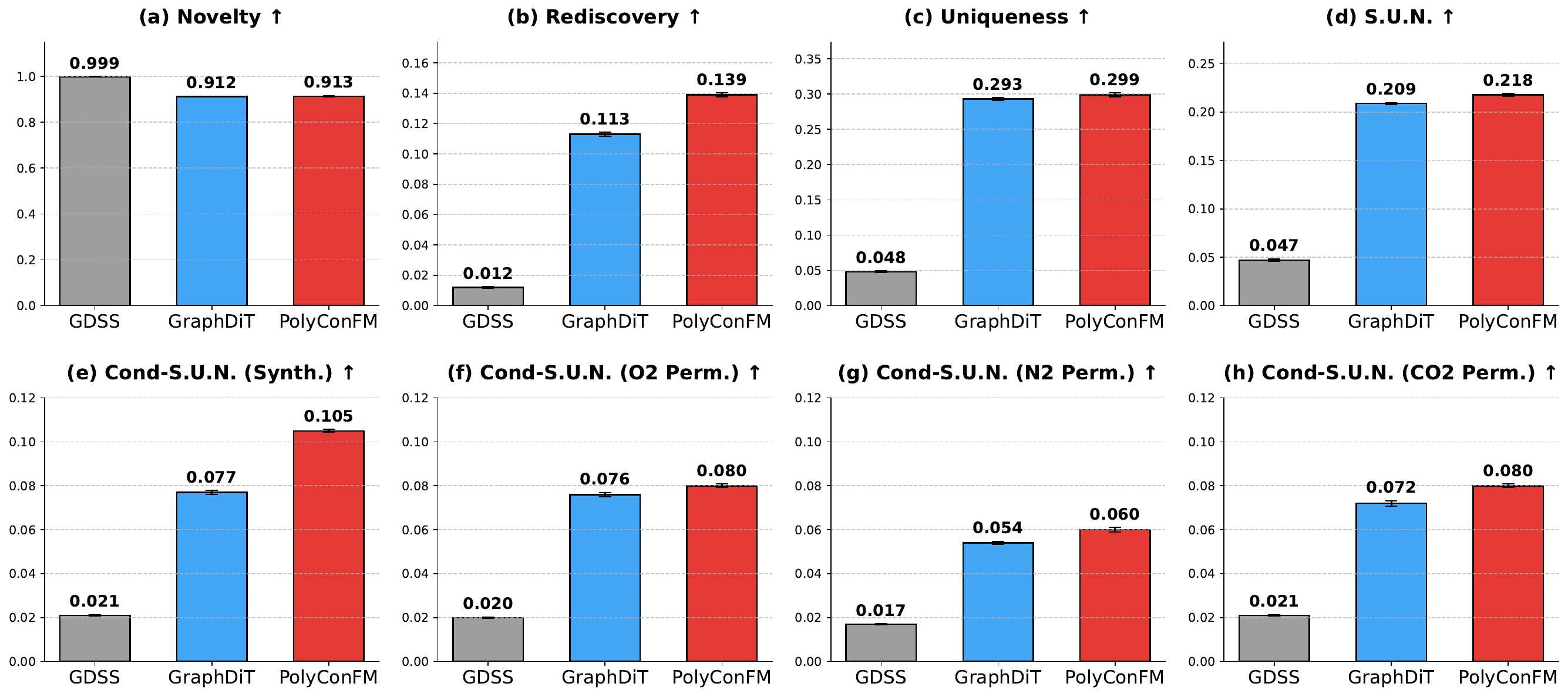} 
    \caption{\textbf{\thefigure:} \rre{Further comparison of PolyConFM and top-2 baselines (i.e., GraphDiT and GDSS) on the downstream polymer design task under the following evaluation metrics: (a) \textbf{Novelty} measures the proportion of generated polymers that are absent from the training set; (b) \textbf{Rediscovery} measures the proportion of reference polymers from the unseen test set that are successfully regenerated; (c) \textbf{Uniqueness} measures the proportion of generated polymers that are distinct; (d) \textbf{S.U.N.} measures the proportion of generated polymers that are simultaneously \textbf{S}table, \textbf{U}nique, and \textbf{N}ovel; and (e)-(h) \textbf{Conditional S.U.N.} measures the proportion of generated polymers that satisfy S.U.N. criteria while achieving an absolute error $\le 1.0$ against the targeted condition.}
    \rre{Here, all reported results are evaluated on the test set.}}
    \label{fig: design_abl_metrics}
\end{figure*}

To demonstrate the superiority and effectiveness of PolyConFM on the downstream polymer design task, we further compare it against its top-2 baselines (i.e., GraphDiT and GDSS) using an augmented set of evaluation metrics, including Novelty, Rediscovery, Uniqueness, the S.U.N. rate, and the Conditional S.U.N. rate.
This comprehensive evaluation rigorously verifies whether the model truly captures the complex chemical distribution while maintaining high fidelity in satisfying targeted conditions.
As presented in~\ref{fig: design_abl_metrics}, PolyConFM consistently maintains state-of-the-art performance under these evaluation metrics. 
In terms of distribution learning (a-d), PolyConFM achieves the highest S.U.N. rate (0.218) and Rediscovery rate (0.139). While GDSS shows near-perfect Novelty (0.999), its negligible Rediscovery (0.012) and S.U.N. (0.047) rates reveal that it primarily generates physically meaningless noise. 
In contrast, PolyConFM's balance between Novelty (0.913) and Rediscovery confirms its superior ability to capture the underlying chemical distribution.
The advantages are even more striking in the context of condition control (e-h). 
Specifically, PolyConFM significantly improves the Conditional S.U.N. rate for synthetic accessibility by~40\% over the best baseline (i.e., GraphDiT), while maintaining the highest Conditional S.U.N. rates across all gas permeability targets (O2, N2, CO2). 
In general, these results collectively validate that PolyConFM is an effective and reliable model for practical polymer design, providing a significant performance leap over existing methods.

\begin{table*}[t]
    \centering
    \setlength{\tabcolsep}{2pt}
    \caption{\textbf{\thetable:} \rre{The performance comparison of different methods on the downstream polymer design task, and the best result for each metric has been bolded. In particular, the conditioning set only comprises the synthetic score (Synth.) and a single gas permeability property.}
    \rre{Here, all reported results are evaluated on the test set.}} 
    \begin{tabular}{llccccccc}
    \toprule
    & \multirow{2}[2]{*}{Method} & \multicolumn{4}{c}{Distribution Learning} & \multicolumn{2}{c}{Condition Control} \\
    \cmidrule(lr){3-6} \cmidrule(lr){7-8} 
    & & Coverage $\uparrow$ & Diversity $\uparrow$ & Similarity $\uparrow$ & Distance $\downarrow$ & Synth. $\downarrow$ & Property $\downarrow$ \\ 
    \midrule
    & GDSS~\cite{jo2022score}      & $0.667_{\pm0.000}$ & $0.786_{\pm0.004}$ & $0.001_{\pm0.000}$ & $35.597_{\pm0.013}$ & $1.304_{\pm0.006}$ & $0.981_{\pm0.005}$ \\
    & GraphDiT~\cite{liu2024graph}  & $1.000_{\pm0.000}$ & $0.845_{\pm0.001}$ & $0.977_{\pm0.001}$ & $\phantom{0}6.910_{\pm0.032}$  & $1.138_{\pm0.005}$ & $0.811_{\pm0.003}$ \\
    \rowcolor{lightblue}
    \cellcolor{white}\multirow{-3}{*}{O2Perm}  
    & PolyConFM & $1.000_{\pm0.000}$ & $\textbf{0.847}_{\pm0.002}$ & $\textbf{0.980}_{\pm0.001}$ & $\phantom{0}\textbf{6.605}_{\pm0.013}$  & $\textbf{0.782}_{\pm0.006}$ & $\textbf{0.806}_{\pm0.002}$ \\
    \midrule
    & GDSS~\cite{jo2022score}      & $0.667_{\pm0.000}$ & $\textbf{0.875}_{\pm0.003}$ & $0.012_{\pm0.002}$ & $38.944_{\pm0.013}$ & $1.450_{\pm0.006}$ & $3.272_{\pm0.017}$ \\
    & GraphDiT~\cite{liu2024graph}  & $1.000_{\pm0.000}$ & $0.847_{\pm0.004}$ & $0.974_{\pm0.002}$ & $\phantom{0}7.117_{\pm0.022}$  & $1.035_{\pm0.002}$ & $\textbf{1.011}_{\pm0.007}$ \\
    \rowcolor{lightblue}
    \cellcolor{white}\multirow{-3}{*}{N2Perm}  
    & PolyConFM & $1.000_{\pm0.000}$ & $0.855_{\pm0.001}$ & $\textbf{0.977}_{\pm0.000}$ & $\phantom{0}\textbf{6.709}_{\pm0.025}$  & $\textbf{0.817}_{\pm0.005}$ & $1.012_{\pm0.004}$ \\
    \midrule
    \multirow{3}{*}{CO2Perm} & GDSS~\cite{jo2022score}      & $0.667_{\pm0.000}$ & $\textbf{0.864}_{\pm0.003}$ & $0.015_{\pm0.000}$ & $39.013_{\pm0.040}$ & $1.202_{\pm0.004}$ & $1.191_{\pm0.003}$ \\
    & GraphDiT~\cite{liu2024graph}  & $1.000_{\pm0.000}$ & $0.849_{\pm0.001}$ & $0.976_{\pm0.001}$ & $\phantom{0}6.773_{\pm0.024}$  & $1.075_{\pm0.003}$ & $0.828_{\pm0.003}$ \\
    \rowcolor{lightblue}
    \cellcolor{white}\multirow{-3}{*}{CO2Perm}  
    & PolyConFM & $1.000_{\pm0.000}$ & $0.846_{\pm0.002}$ & $\textbf{0.978}_{\pm0.001}$ & $\phantom{0}\textbf{6.479}_{\pm0.023}$  & $\textbf{0.846}_{\pm0.003}$ & $\textbf{0.806}_{\pm0.002}$ \\
    \bottomrule
    \end{tabular}
    \label{tab: exp_design_add}
\end{table*}

\ref{tab: exp_design_add} summarizes the performance of various methods on the downstream polymer design task, where only taking the synthetic score (Synth.) and a single gas permeability property as the conditioning set.
Here, we still compare their capability along both distribution learning and condition control in tandem, ensuring a comprehensive and balanced evaluation.
As presented in this Table, PolyConFM preserves a favorable trade‑off between distributional fidelity and conditional satisfaction, significantly outperforming all baselines on all conditioning sets (Synth.\&O2Perm, Synth.\&N2Perm, Synth.\&CO2Perm). 
In particular, for each conditioning set, it consistently secures complete heavy‑atom type coverage, the highest fragment‑level similarity, and the lowest Fréchet ChemNet Distance, while maintaining competitive diversity.
Besides, compared with the best baseline, it reduces MAE on the synthetic score by at least 20\%, demonstrating consistently enhanced condition control across all conditioning sets.
Taken together, these results further confirm PolyConFM's superior capability, establishing it as a powerful and reliable tool for polymer design.

\end{appendix}
\end{document}